\documentclass[11pt, logo, a4paper]{deepmind}

\pdftrailerid{redacted}
\usepackage{kantlipsum, lipsum}
\usepackage{dsfont}
\usepackage{dm-colors}
\correspondingauthor{Mike Johanson (\texttt{mjohanson@deepmind.com}) or Joel Leibo (\texttt{jzl@deepmind.com})}

\usepackage[authoryear, sort&compress, round]{natbib}

\usepackage[utf8]{inputenc} 
\usepackage[T1]{fontenc}    
\usepackage{hyperref}       
\usepackage{url}            
\usepackage{booktabs}       
\usepackage{amsfonts}       
\usepackage{nicefrac}       
\usepackage{microtype}      
\usepackage{lipsum}
\usepackage{fancyhdr}       
\usepackage{graphicx}       
\usepackage[dvipsnames]{xcolor}
\usepackage{subcaption}
\usepackage{xspace}
\usepackage[export]{adjustbox}
\usepackage{makecell}
\usepackage{multirow}
\usepackage{appendix}
\usepackage{siunitx}
\usepackage{float}
\usepackage{boldline}
\usepackage{setspace}

\newcommand{\BibTeX}{\rm B\kern-.05em{\sc i\kern-.025em b}\kern-.08em\TeX}
\makeatletter
\DeclareRobustCommand\onedot{\futurelet\@let@token\@onedot}
\def\@onedot{\ifx\@let@token.\else.\null\fi\xspace}
\def\eg{\emph{e.g}\onedot}
\def\ie{\emph{i.e}\onedot}
\def\etc{\emph{etc}\onedot}
\definecolor{darkgreen}{RGB}{0,125,0}

\newcounter{mjNoteCounter}

\newcounter{jlNoteCounter}

\newcounter{ftNoteCounter}

\newcounter{ehNoteCounter}

\title{Emergent Bartering Behaviour in Multi-Agent Reinforcement Learning}

\author[1]{Michael Bradley Johanson}
\author[1]{Edward Hughes}
\author[1]{Finbarr Timbers}
\author[1]{Joel Z. Leibo}

\affil[1]{DeepMind}

\begin{abstract}
Advances in artificial intelligence often stem from the development of new environments that abstract real-world situations into a form where research can be done conveniently. This paper contributes such an environment based on ideas inspired by elementary Microeconomics. Agents learn to produce resources in a spatially complex world, trade them with one another, and consume those that they prefer. We show that the emergent production, consumption, and pricing behaviors respond to environmental conditions in the directions predicted by supply and demand shifts in Microeconomics. We also demonstrate settings where the agents' emergent prices for goods vary over space, reflecting the local abundance of goods. After the price disparities emerge, some agents then discover a niche of transporting goods between regions with different prevailing prices---a profitable strategy because they can buy goods where they are cheap and sell them where they are expensive. Finally, in a series of ablation experiments, we investigate how choices in the environmental rewards, bartering actions, agent architecture, and ability to consume tradable goods can either aid or inhibit the emergence of this economic behavior. This work is part of the environment development branch of a research program that aims to build human-like artificial general intelligence through multi-agent interactions in simulated societies. By exploring which environment features are needed for the basic phenomena of elementary microeconomics to emerge automatically from learning, we arrive at an environment that differs from those studied in prior multi-agent reinforcement learning work along several dimensions. For example, the model incorporates heterogeneous tastes and physical abilities, and agents negotiate with one another as a grounded form of communication. To facilitate further work in this vein we will release an open-source implementation of the environment as part of the Melting Pot suite~\citep{leibo2021scalable}. 
\end{abstract}

\begin{document}

\maketitle

\begin{spacing}{0.5}
\tableofcontents
\end{spacing}

\section{Introduction}

We would like to build artificial agents capable of innovating as humans do. We believe that theoretical and algorithmic frameworks such as decision theory and reinforcement learning (or \textbf{RL}) are relevant to this goal. However, one reason we have not yet succeeded in building such agents stems from a fundamental tension between what these theories are good at---mainly strengthening weak and unlikely behaviors so they become more prevalent and refined---and what we actually want them to do: discover truly novel and innovative behaviors that do not occur with any frequency at all at the start of their learning.

As anyone who ever tried to train a pigeon to bowl will attest, if you must wait to provide the first reward until the pigeon naturally emits an approximation of the desired behavior, you will have to wait a very long time indeed~\citep{peterson2004day}. In terms of Bayesian decision theory, the prior distribution must contain the novel behavior within its support. Otherwise, no amount of evidence for its superiority will suffice to nudge its probability off zero~\citep{kalai1993rational}. L. J. Savage illustrated the problem with a pair of proverbs. A small world is one where you can always ``look before you leap''. A large world is one where you must sometimes ``cross that bridge when you come to it''~\citep{binmore2007rational, savage1951foundations}. Bayesian decision theory is only valid in small worlds. Yet the real world is large. 

In large worlds, RL becomes a theory of how to strengthen weak behaviors, not a theory of how to generate wholly new ones. This is because an innovative behavior cannot be reinforced until a close-enough approximation to it has been emitted for the first time. RL researchers employ a variety of methods to encourage agents to continually emit novel behaviors. Many amount to injecting randomness in to action selection, as in $\epsilon\text{-greedy}$ and entropy regularized Boltzmann exploration~\citep{sutton2018reinforcement}. \cite{osband2019deep} calls this approach random \emph{dithering}. Dithering provides an agent with opportunities to experience the rewards that may be obtained with behaviors it never tried before. However, dithering is an inefficient way to traverse a large world. Consider: an $\epsilon\text{-greedy}$ agent selects the action it currently thinks is best with probability $1 - \epsilon$ and otherwise selects uniformly at random from all available actions. The probability of it emitting any particular novel behavioral sequence---\ie a sequence containing no actions initially thought to be valuable---decays exponentially as the length of the sequence increases~\citep{kakade2003sample, osband2019deep}. Optimistic initialization is another approach to exploration, which positively biases the agent's initial reward estimates for every state and action~\citep{sutton2018reinforcement}. However, absent prior knowledge concerning where to place one's optimism, it reduces to a general drive to explore all states and actions, an impossible task and an inefficient and distracting bias in large worlds.

There are many other more sophisticated approaches to exploration in RL but all struggle in large worlds for the same reason: the farther the target behavior is from the current best known behavior, the more the agent must experiment with actions it believes to be unattractive in order to find it. In large worlds the target behavior may be very far away indeed. More sophisticated approaches to exploration seek to choose experimental actions more judiciously than random dithering, but there are limits to how much efficiency can be gained this way without incorporating prior knowledge~\citep{osband2019deep}. As a result, RL is often bad at discovering beneficial behaviours (\ie action sequences) when they never occur by chance before learning and a random agent never emits them. Yet these are exactly the behaviors that we are most interested in discovering. We know that humans can innovate: examples include composing Beethoven's 5th symphony, designing spacecraft to take astronauts to the moon, and devising agricultural technology to feed billions of people. We want to develop algorithms that can innovate as humans do.

So far we have assumed that we cannot rely on the agent's learning environment having any particular structure. The logic underlying this assumption is clear: since we want our RL agents to succeed in \textit{any} environment it follows we must prefer exploration techniques such as $\epsilon\text{-greedy}$ that eventually converge in \textit{all} environments, particularly those where no prior knowledge can be leveraged. This starting point led us to a pessimistic conclusion about RL's ability to explain innovation. On the other hand, allowing prior knowledge would completely change the picture. There are many RL algorithms that explore efficiently when provided with correct prior knowledge in one form or another, even in large worlds~(e.g~\cite{gupta2018meta}). Perhaps by restricting our attention to natural environments like those in which humans evolved we can uncover the roots of innovation. After all, natural environments do not cover more than a vanishingly tiny slice of the space of all possible environments, and they may have useful properties relevant to promoting exploration by the agents that inhabit them. Moreover, the RL environments we typically use for research were never intended as models of situations conducive to the evolution or development of natural intelligence. They could systematically fail to capture the important properties of natural environments.

Indeed, two independent strands of biological evidence and reasoning suggest that most of our RL environments miss something important that is present in natural environments. In laboratory animals, manipulations of the rearing environment produce profound effects on both brain structure and behavior. For example, laboratory rodent environments may be enriched by using larger cages which contain larger groups of other individuals---creating more opportunities for social interaction, variable toys and feeding locations, and a wheel to allow for voluntary exercise. Rearing animals in such enriched environments improves their learning and memory, increases synaptic arborization, and increases total brain weight~\citep{van2000neural}. The second strand of biological reasoning concerns the ``social brain hypothesis'' for the evolutionary emergence of intelligence in the primate order~\citep{dunbar1998social}. It is based on the observation that a species' brain size correlates well with its typical social group size (adjusted for overall body size). The correlation holds over the entire primate order, which spans three orders of magnitude in brain size~\citep{dunbar2017there}. As a general rule, primates who live in larger groups have larger brains. The social brain hypothesis suggests that larger social groups, which were necessary for reasons such as mitigating predation risk, gave rise to myriad new problems of social origin. The need to solve these was the driver for the evolution of greater and greater intelligence in the primate order. In RL terms, the hypothesis is that the socially enriched environments of the ``brainier'' species contained the right mix of problems to encourage agents to devote effort toward finding intelligent solutions.

Moreover, many critical innovations concern the coordinated behavior of more than one agent. The RL community's standard method of addressing the problem of exploration, intrinsic motivation (e.g.~\cite{pathak2017curiosity}), is uniquely unsuited for the finding of equilibria involving extensive coordination between agents since intrinsic motivations are, by definition, intrinsic---depending only on self-generated signals, which cannot easily be correlated to those in others. 

Fortunately there is a subfield of RL that considers social enriched environments: \textbf{multi-agent reinforcement learning (MARL)}. Multi-agent environments are inherently non-stationary, as each agent's stream of experience and optimal behaviour changes as the other agents learn and change their behaviour. As the population learns, new niches may be created that an agent can fill, or other agents may start to contest an agent's current niche. This provides agents with \emph{extrinsic} motivation to continually explore new behaviors as the population adapts \citep{leibo2019autocurricula, balduzzi2019open, baker2019emergent, wang2019paired}. In theory such multi-agent systems may continue to explore forever. In practice they often reach an equilibrium point and stop exploring. Efforts to understand how these systems work create tension with the dominant ``single-agent paradigm'' of artificial intelligence and cognitive science. In short, all the representations that matter are no longer inside the agent's ``head'', but rather are distributed in some fashion between the agent, the population, the environment, and the training protocol itself.

Multi-agent environments feature causal forces that provide extrinsic motivation to explore. For example: consider the supply and demand forces in Microeconomics. They are created by the aggregate behavior of individual agents. They constitute real motivational forces for the utility maximizing agents who populate economic theory in the sense that changes in supply and demand cause systematic incentives for agents to change their behavior in specific directions. For instance, an increase in the price at which I can sell a widget incentivizes me to produce more widgets. Likewise, a decrease in the number of agents competing with one another to buy my widgets incentivizes me to lower the price at which I sell them, or to decrease production. These same forces can sometimes motivate innovation. If demand for the widgets made in my factory increases strongly enough---and I cannot simply raise the price---then I become incentivized to find ways to make more of them or to make them more efficiently, perhaps by improving a manufacturing process. The example illustrates that there is not necessarily any explore-exploit trade-off. In fact, real world environments often have properties that make it possible for agents to \textbf{explore by exploitation} (see also~\cite{leibo2019malthusian}).

So far we have argued that the social environment---that is, the set of other agents in the world---shape the reward landscape and thereby provide extrinsic motivation for agents to explore and innovate. The underlying environment, or \textbf{substrate}, that these agents inhabit also plays a role in motivating agents to innovate. For instance, a very simple substrate consisting of just an empty room devoid of objects admits no innovation, no matter how large or complex the agent population is that inhabits it. We can extend this kind of argument much further. Consider what would happen if you connected a large language model such as GPT-3~\citep{brown2020language} to the latest RL-based agent that solves complex problems in a 3D world (to be concrete, perhaps consider~\cite{parisotto2020stabilizing} or any other state of the art single-agent RL algorithm applicable to 3D simulated worlds). Also, for this thought experiment, assume that we have somehow solved the problem of language grounding (described in for example~\cite{harnad1990symbol}). Choose as the substrate a 3D simulation with realistic physics such as the one underlying the DMLab-30 suite of environments~\citep{espeholt2018impala}. Then connect 100 of these state-of-the-art RL agents to it. Give them the ability to talk to one another by querying the large language model and sending one another streams of text. Since we have assumed that the language grounding problem has already been solved, the agents are thus able to refer to all the objects in their world by name, and that knowledge is integrated into their broader language understanding provided by the language model. Let all 100 agents live simultaneously in the same simulated world. Now, what would happen? The ``social-is-all-you-need'' hypothesis appears to suggest that this would be enough to set off a cumulative cultural innovation explosion ratchet. But would it really? Surely the specific properties of the substrate matter too. We do not believe that \textit{any} environment containing sufficient complexity can generate innovation, not even for a multi-agent system with deeply complex individual agents having a lot of cognitive capacity\footnote{For instance, some environments are just too simple or afford too little means of communication and too little interdependence for a ``social-is-all-you-need'' intelligence ratchet to get off the ground.}. This raises the question: which properties of the environment matter and which do not? Are there necessary and sufficient conditions for an environment to ``allow for'' substantial innovation? How do we even study such questions? The present paper concerns a particular hypothesis in this realm: that properties that are important in microeconomics will also be important for motivating agents to explore and innovate. The reason is that economics is a science of incentives. The environmental properties highlighted in economics are those that create incentives for agents to interact. Meanwhile in AI, incentives are also the motive force for agents to explore. Incentives are what we think are lacking in the social-only thought experiment. Without any incentive to innovate, agents simply won't.

Exploration by exploitation depends on the environment to furnish incentives for exploration. Incentives induce gradients in value over policy space. For instance, competitive incentives provide an intrinsic drive for exploration. As soon as one agent starts habitually exploiting any particular solution it creates an incentive for other agents to invest time and energy into learning how to adapt in response. In strictly competitive two-player settings such as the games of go or heads-up poker~\citep{silver2017mastering, bowling2015hulhe}, we would describe the agents as adapting to exploit each other, and (if using appropriate algorithms) would eventually converge to a stalemate: a Nash equilibrium. However, that point may be arbitrarily far away from their initial behavior. In a large world, it is possible for agents to continually refine their behaviours and innovate new ones to better adapt to each other for a very long time.

Economic behaviors like production, consumption, and trade are enacted by individuals, but weaved together in a complex whole composed of the interaction of many individuals and the environments they inhabit. A mutually supporting system of such behavior exerts incentives on individuals to learn specific kinds of things, such as how to improve the efficiency of a production process. This is true for human economies and ought also to be true in artificial economies. \cite{read1958pencil} illustrates the idea with a short story told from the perspective of a pencil proudly describing its heritage. The pencil's wood came from a straight grain cedar tree in Northern California, cut down by a team of loggers bearing saws, trucks, and rope. The pencil's graphite was mined in Sri Lanka and its mining involved a range of other tools, and had to be shipped by sea to the pencil factory. Numerous dockworkers, sailors, and lighthouse keepers all take action to ensure its safe passage. The story goes on like this for several pages, until the reader is left with real sense of awe at the complexity of the globe-spanning cooperative machine that gave birth to the pencil. \cite{read1958pencil} points out that it is not even necessary that all the people involved in the pencil's construction ever lay eyes on the final product or have any interest in it themselves. But the pencil-producing system as a whole hangs together anyway. Indeed, it does more than that that: it thrives. The individuals involved need not care about pencils, or be aware of upstream or downstream steps in the creation of pencils, but are all linked nonetheless through the system of incentives provided by the market economy. All act toward their local ends, and the {``invisible~hand''} of the market coordinates their activities. Furthermore, consider what happens if demand for pencils increases, something that could occur for instance if the overall population size increased. All else being equal, greater demand for pencils creates greater demand for graphite and cedar wood. It incentivizes all of the many different individuals involved in the pencil production supply chain to make local efficiency improvements, so they may ultimately sell more of their product to obtain greater profits, taking advantage of the increased demand. In terms of reinforcement learning, the incentives created by the market economy are experienced by agents as gradients in value over policy space. Imposing such a gradient has a strong effect on the policies that agents may ultimately learn.

The emergence of economic behavior in MARL engages somewhat different logic than in economics. Consider the following illustrative example resembling what can be found in any introductory economics textbook. Farmer Alice---who has corn---and farmer Bob---who has chickens---have utility functions such that they can benefit from trade. Alice wants some chickens and Bob wants some corn. A simple economic analysis proceeds from there. One result is a model that predicts their supply and demand behavior: the classic intersecting supply and demand curves model~(e.g.~\cite{samuelson1995economics}). However, this derivation assumed from the outset that Alice and Bob already know \textit{how to trade}. For example: Alice must know that she would gain more utility if she had chickens, that Bob has chickens, that Bob wants corn, that there exists a sequence of actions they could each take to exchange these goods, and that Bob is also aware of these facts. In economic models that assume rational agents, this knowledge and behaviour is taken for granted. This assumption makes sense: the human agents being modelled perform these exchanges many times every day without much thought. However, when the agents are reinforcement learning agents that learn through their own stream of experience, with no \textit{a priori} knowledge of what their observations mean, what effect their actions have, or that other agents are in fact goal-seeking and adapting entities and not just non-adaptive parts of the environment that happen to move around, it is not at all clear that the agents will learn to trade with each other. Further, even if they do discover the sequences of actions required to produce, trade, and consume goods, it is similarly unclear that their behaviour will evolve in the directions we predict of theoretical rational agents under the pressure of abstract supply and demand forces.

In this paper, we will study exactly this emergence of microeconomic behaviour---production, trading, and consumption---in populations of state-of-the-art reinforcement learning agents. In our environment called \textbf{Fruit Market}, deep reinforcement learning agents learn from scratch how to produce, trade, and consume resources in order to maximize their individual reward. When the environment is changed in ways familiar to a Microeconomics 101 student, by adjusting environmental features related to supply or demand, the population's equilibrium production, consumption, and pricing behaviour largely shifts in the directions we expect from Microeconomics. Our environment includes dimensions of space and time, allowing for the emergence of phenomenon such as local prices reflecting the nearby abundance of resources, and arbitrage behaviour by agents who learn to exploit those price differences. However, our work is not really about applying state-of-the-art artificial intelligence to economic modeling (for that, see instead~\cite{zheng2020ai}). Instead, our goal is to explore this emergence of microeconomic behaviour just as we would study any other social behaviour within the broad project of creating \textbf{artificial general intelligence (AGI)}.

A key element of this work is our exploration of what microeconomic knowledge, if any, must be built into the environment in order for current state-of-the-art agents to discover and refine production, consumption, barter, and arbitrage behaviors. We restrict ourselves to only adjusting the environment. The agents we use are generic deep reinforcement learning agents which have been widely used in other MARL research. They start training from a randomly initialized state and have no domain-specific prior knowledge, parameter tuning, or code. In this setting, we find numerous ways to manipulate the environment that can radically change the final behavior that a population converges to, including whether trade flourishes between agents, or does not emerge at all. In addition to our empirical results demonstrating successful learning by the agents, we also include a large analysis of these environmental choices to show why they were made, and how alternative choices perform. For example, we will demonstrate that current agents do not learn to trade if their actions for doing so are overly generic, such as ``drop an item on the ground'' or ``give an item to another agent''. These actions could be used to trade goods, but it is difficult to learn to use them appropriately: why give an item to another agent if they have not yet learned to give something else in return? However, if the environment includes a mechanism that facilitates trading by making the exchange atomic---simultaneously swapping items between agents that have agreed---then the agents do consistently learn how to trade. In the real world, there would be a whole system of conventions, norms, and institutions to support this, such as concepts of private property ownership that serve to coordinate everyone's expectations so that trades can proceed more-or-less atomically~\citep{segal2013property}. When we assume an automatic trade facilitation mechanism we sidestep the critical question of how all that structure could emerge. This move turns out to be essential for the present work. We would not have been able to make progress using today's state-of-the-art generic agents otherwise. However, if our agents cannot learn without such mechanisms in the future, it will ultimately have negative implications for our MARL agents' generality since there are surely many important economic behaviors and phenomena that follow from properties of the underlying market-inducing conventions, norms, and institutions~\citep{coase1988firm}. Without dismissing the importance of such market-inducing structures, we set them aside for now. By focusing our attention on the case featuring the automatic trade resolution mechanism we are able to make progress on downstream questions like how environmental changes (\eg supply and demand shifts) affect emergent production, consumption, barter, and arbitrage behaviors. Learning these behaviors is still a complex feat for MARL agents as they involve interleaved decisions of where and what to harvest and where to travel to find others to trade with, as well as what offers to make and accept.

AI research relies on simulation environments that capture important cognitive and social challenges. This is because agents that learn in such environments face incentives that push them to develop habits of cognition and discover essential concepts that describe their world and how they can effectively behave in it~\citep{silver2021reward}. One implication of this research method is that, towards the goal of constructing generally capable agents, researchers must continually grow the set of environments under consideration. Eventually it should reflect a full accounting for all conceptually distinct principles of intelligence. For the domain of social intelligence, so rich in real-life, its mirror in MARL research remains woefully incomplete. Our goal in this work is to add these themes of trading, negotiation, specialization, and adaptation to a changing population to the areas examined in MARL research. To facilitate further research in this direction, we have prepared an open-source version of our environment which we will incorporate into the next release of the Melting Pot environment suite~\citep{leibo2021scalable}\footnote{Melting Pot is available at \url{https://github.com/deepmind/meltingpot}.}.

\section{Related Work}
\label{sec:related}

Many AGI researchers take an approach to building advanced learning systems based on the idea of reverse engineering human intelligence. Part of the reverse engineering approach involves trying to elucidate the correct list of cognitive abilities that one would need to establish a putative AGI possesses in order for us to declare victory and consider it a human-like artificial agent. The most obvious such abilities are the ``classic'' cognitive abilities like perception, attention, and memory. However, numerous other schemes exist (e.g.~\cite{spelke2007core, schneider2018cattell}), and the problem of how to measure whether or not a putative AGI displays a particular ability is not entirely resolved~\citep{hernandez2017measure}. However, recent developments that adapt to RL the relentless focus on measuring generalization long advocated by the supervised machine learning community are widely seen as a positive methodological development (e.g.~\cite{zhang2018study, machado2018revisiting, fortunato2019generalization, cobbe2019quantifying, juliani2019obstacle, crosby2020animal, leibo2021scalable}).

\subsection{Exploration and the road to AGI}

In one paradigm, the cognitive abilities are themselves regarded as potentially emergent from generic experiential learning under a simple reward function~\citep{silver2021reward}; it is called the \emph{reward is enough} hypothesis. It contrasts with the hypothesis that specialized inductive biases will be needed for each ability (as advocated for instance by~\cite{lake2017building} and~\cite{marcus2018innateness}). From our vantage point, the important thing about the reward is enough hypothesis is that it casts the problem of cognitive ability discovery as one of extremely deep exploration. It requires agents to emit behaviors very far from those they would emit randomly, ``sculpting'' subsystems like long-term memory out of a pluripotent initial neural machinery\footnote{In cognitive science, Cecilia Heyes has articulated a theory of cognitive ability discovery that is broadly compatible with ours~\citep{heyes2019precis}. In her view, some cognitive abilities such as natural linguistic proficiency are built up by generic learning processes operating within the context of cultural evolution. Her term for these is \textbf{cognitive gadget}. For instance, the ability to read is clearly a cognitive gadget since written language is no more than 6000 years old, too recent for genetic evolution to have produced a specialized reading mechanism. The reward is enough account of \cite{silver2021reward} similarly holds that cognitive abilities may emerge from generic learning mechanisms and adds the unique hypothesis that for AI, one specific such mechanism, reinforcement learning, is sufficient to originate all the other cognitive abilities.}. The distance in behavior space that such exploration much traverse is truly gargantuan.

For years the prototypical exploration problem in RL has been the Atari game Montezuma's Revenge~\citep{bellemare2013arcade}. In this 2D ``flip-screen'' game, the player must guide a character through a series of rooms where only a rather precise sequence of movements can get them through safely and non-zero rewards are very rare. Thus policy and value gradients are often near zero and reinforcement learning is very inefficient. Exploration research motivated by Montezuma's Revenge addresses the hypothesis that sparse rewards are the main difficulty in exploration. The idea is that if only the rewards were instead dense---\ie frequent---then, even if those rewards were not the ``real'' reward, they would still induce a strong gradient capable of guiding policy learning around the space, where it would eventually discover the solution. The sparse reward hypothesis led researchers to propose a great variety of intrinsic motivation models: modified reward functions that push agents to seek novelty or empowerment (e.g.~\cite{bellemare2016unifying, gregor2016variational, burda2018exploration, eysenbach2018diversity, karl2019unsupervised}). These methods have been successful in Montezuma's Revenge. But, on their own, it is unclear how they could be made to scale up to the truly gargantuan amount of exploration required to discover novel cognitive abilities demanded by~\cite{silver2021reward}.

Many researchers have sought to build off the observation that correct prior knowledge of the environment, when presented in an accessible fashion, can be used to structure exploration (e.g.~\cite{schwartz2019language, goyal2019using, mirchandani2021ella, mu2022improving} and \cite{tam2022semantic} all explored natural language-based representations of prior knowledge). Much of the emphasis in AGI-oriented single-agent RL research is accordingly based on approaches that seek to learn rich models and representations of the world that can then be deployed to support such long-term goal directedness~\citep{hung2019optimizing, schrittwieser2020mastering, hessel2021muesli}. For instance, starting with some amount of common sense, agents could represent that they do not know about a certain area, make a concerted plan to investigate it, and then sequence all their actions over a long period of time in order to follow through on that plan, integrating the knowledge thus acquired into their general world model, and then repeat the process again from its new stronger starting point (e.g.~\cite{botvinick2017building, lampinen2020transforming, shanahan2022abstraction}). One version of this approach makes the goal directedness explicit via generalized value functions~\citep{sutton2011horde} with the idea that, ideally, the (sub)-goals themselves would come from something like the agent's abstract understanding of its world~\citep{vezhnevets2017feudal, veeriah2021discovery}.

An alternative, and really quite different, approach starts by essentially giving up on solving the sparse reward problem within the simulation itself. Instead, this approach relies on human trainers to provide the diverse data needed to train the artificial agent using imitation learning~\citep{ziebart2010modeling, ho2016generative, finn2016connection, osa2018algorithmic}, offline RL~\citep{zolna2020offline}, or by abstracting RL away to treat the problem as one of data-driven sequence modeling~\cite{chen2021decision}. Using human data in this way it is possible to resolve the chicken-and-egg problem of RL. You need not first emit a behavior before it can be reinforced if real humans provide the stream of experience ~\citep{abramson2020imitating}.

In their own ways, all the aforementioned approaches have been motivated via the sparse reward hypothesis concerning the difficulty of engendering exploration deep enough to build cognitive abilities. The central hypothesis motivating much of the AGI-oriented MARL research on the other hand is quite different. In MARL it is natural to cast the problem not as sparse reward, but rather as premature convergence to local optima that are not good enough. Even bad local optima, if isolated from the better regions of policy space by vast intervening regions that are even worse, create basins of attraction that  are difficult to escape once entered. In MARL these local optima are also associated with equilibria. Other players will be acting in some fashion that creates a local part of policy space with an incentive structure from which gradient-guided learning cannot escape. The problem of exploring deeply enough to build cognitive abilities is thus recast for MARL as a problem of equilibrium selection~\citep{harsanyi1988general, gintis2009game}. A similar perspective prevails in the study of ecosystem evolution (e.g.~\cite{swenson2000artificial}) and game theoretically informed political philosophy~\citep{sugden1986economics, binmore1994game, skyrms1996evolution, gintis2014bounds}. Successful origination of significant new innovations entails adaptive radiation in niche space~\citep{boyd2011cultural, leibo2019malthusian}. 

The critical factors controlling multi-agent joint exploration are thus which equilibria exist in joint policy space and how smooth and traversable are the non-zero gradient paths that link them to one another. Both are determined by the interaction of physical and social properties of the simulated system. Physical properties include the simulated landscape. Social properties include the numbers of other agents as well as their tastes and preferences. In light of this, MARL researchers have considered a variety of different basic models of the multi-agent learning problem~\citep{shoham2007if}, often stressing just one component (e.g. physical or social) at a time. One basic distinction is between algorithms that assume the ``rules of the game'' are \emph{not} given (incomplete information, large worlds), in which case they must explore to discover them; versus those that assume access to a perfect simulator of the game (or world) dynamics. The latter is often called the \emph{planning} setting and it includes recent work on games like poker and go~\citep{silver2017mastering, bowling2015hulhe, moravcik2017deepstack, brown2018superhuman, bard2020hanabi}. The present paper is concerned with the former case: settings where exploration is needed to discover the dynamics of the world.

Most prior multi-agent reinforcement learning research on complex social situations where it is necessary to explore falls into one of the following three categories:
\begin{enumerate}
    \item Pure conflicting interests (zero sum games)
    \item Pure common interest (pure coordination games)
    \item Mixed motivation settings such as sequential social dilemmas and bargaining problems
\end{enumerate}
Lately there has been a push to consolidate all these disparate strands of research into a single combined benchmark called Melting Pot, the idea being that MARL algorithms ought to be generic enough to work across all three categories~\citep{leibo2021scalable}.

\subsection{Pure conflicting interests}

Exploration is often facilitated in zero sum games as a result of ``arms race'' type learning dynamics---called \textbf{exogenous autocurricula} in the terminology of~\cite{leibo2019autocurricula}. For instance, in one project agents were trained to play a team-based first-person shooter computer game based on Quake~3~\citep{jaderberg2019human}. The game was 2v2 Capture the Flag. Doing well in this game requires agents to develop ``martial'' skills such as aiming, chasing, shooting, dodging behind cover, and competitive strategizing, as well as navigation and memory skills like exploring to find the opposing team's flag and remembering a quick path to bring it back to the goal, and cooperation skills to work effectively with a teammate. Learning of these skills was driven by the need to continually outperform opponent teams. Since all teams trained simultaneously, there would usually be, for any given agent, another in the population at an appropriate skill level such that learning to defeat them would convey valuable lessons. When an agent's performance is weak or overfit to a particular situation or opponent, other agents in the co-training population learn to exploit them, thereby incentivizing them to unlearn the aspects of their behavior that cause it to perform poorly. In this way, the training process becomes self-correcting and may accumulate new innovations over time. \cite{bansal2018emergent} applied a similar approach to a 3D sumo wrestling game with simulated physics. In practice, successful co-adaptation algorithms for pure conflict settings generally play not just against the latest (and strongest) policy, but also against as large and diverse as possible a set of older policies \citep{Lanctot17PSRO, balduzzi2019open, czarnecki2020real}. This is the same insight underpinning the successful Nash league approach to Starcraft II~\cite{vinyals2019grandmaster}. Among poker researchers and professionals (who similarly learn from experience), such interactions are described with the phrase: ``When you exploit your opponent, you are also teaching them.''.

Some of the most impressive examples of innovation arising from competitive multi-agent co-adaptation are in cases with asymmetric agent roles. For instance, the interactions of adversarial setter and solver agents can be used to motivate substantial exploration~\citep{sukhbaatar2018intrinsic}, an idea that was also applied impressively to robotics applications~\citep{openai2021asymmetric}. Also, a team from Open AI showed that a population of co-adapting agents playing the game of hide-and-seek were driven to discover tools and how to use them strategically in their environment~\citep{baker2019emergent}.

\subsection{Pure common interest}

Unlike pure conflict situations where co-adaptation effects are often helpful to exploration and generalization, in pure common interest settings co-adaptation usually has the opposite effect. When your task is to cooperate with a partner, finding a good response to their current policy \emph{disincentivises} them from further exploration. Thus partner agents overfit to each other's quirks, which causes them to generalize poorly and not explore enough. Accordingly, recent research in this area has been directed toward generating agents capable of adapting on the fly, \eg by metalearning~\citep{duan2016rl, wang2016learning}, so that they can then coordinate with a diverse set of other teammates, who might even be human in some cases~\citep{carroll2019utility, wu2020too, strouse2021collaborating}.

Another line of work on situations of pure common interest is concerned with emergent communication and signalling systems. For instance, \cite{foerster2016learning} studied the benefits of using a differentiable communication channel to pass information between agents and \cite{lazaridou2017multi} studied how emergent communication patterns in referential games can be grounded in natural language by co-training networks on both an interactive task (a multi-agent referential game) and a passive task (supervised image-labeling task with natural language image labels). More recent work in this area considered zero-shot coordination protocols where agents must adapt to new partners who may have learned different ``languages''~\citep{bullard2020exploring, zhu2021few, hu2020other}. 

One especially relevant strand of this literature concerns communication that is intrinsically grounded in the semantics of the underlying game and thus is not ``cheap talk''~\citep{lewis2017deal, cao2018emergent}. In this line of work, communication is regarded as negotiation over how to divide a set of items. It is a temporally extended interaction because agents make a sequence of proposals, continuing until acceptance, whereupon rewards are provided to both agents according to their preferences for each item and their agreed split. Agents have different preferences over the items from one another, and do not know each other's preferences. Thus it is possible to negotiate cleverly and thereby capture a larger share of reward for oneself. Initially~\cite{cao2018emergent} found that it was necessary to modify the reward function to force agents to be prosocial by having them optimize the collective return (sum of both players' rewards). However, more recently~\cite{noukhovitch2021emergent} showed how that restriction could be lifted, turning the environment into a mixed motivation setting, the class of environments featuring both competitive and cooperative motivations to which we now turn.

\subsection{Mixed motivation settings}

Several new concepts are important for describing how physical and social properties of multi-agent systems jointly cause the emergence of mixed motivation incentive structures. On the physical side, the concept of a \textbf{resource} is useful for explaining how environments differ from one another. Following \cite{ostrom2005understanding}'s schema, we classify resources along two dimensions \textbf{excludability} and \textbf{subtractability} (Table~\ref{table:goods}). Excludability refers to how efficiently users of the resource may be excluded from accessing it. For example, I can exclude others from accessing resources in my home by locking the door, but I cannot exclude others from accessing the fish in the harbor without some extraordinary intervention like sending naval vessels to blockade it. Subtractability refers to the degree to which one user obtaining a benefit from the resource depletes the amount of resource remaining. For instance, a hamburger is a subtractable resource because once I have eaten it it is gone.

\begin{table}
    \centering
    \begin{tabular}{l|l|l}
                   & \cellcolor{gray!25}\textbf{Exclusion is Feasible} & \cellcolor{gray!25}\textbf{Exclusion is Not Feasible} \\ 
        \hlineB{4}
        \cellcolor{gray!25}\textbf{Subtractable}  & private goods & common-pool resources \\
        \hline
        \cellcolor{gray!25}\textbf{Non-Subtractable} & club goods  & public goods  \\
    \end{tabular}
    \vspace{0.5em}
    \caption{Classification of resources by excludability and subtractability properties (adapted from~\cite{ostrom2005understanding}). A resource is excludable when it is possible to exclude another agent from accessing it. A resource is subtractable when consumption by one agent reduces the amount available for consumption by others. Standard examples of different categories of resources are: public safety (public good), clean drinking water (common-pool resource), cable TV channel (club good), and housing (private good).}
    \label{table:goods}
\end{table}

Non-excludability makes individuals interdependent. Actions to consume or produce non-excludable resources typically have externalities. That is, the choices taken by any one individual have an effect on many others. Non-excludable resources are thus often associated with social dilemmas~\citep{ostrom2005understanding}. Social dilemmas are situations where there is a tension between individual and collective rationality~\citep{kollock1998social}. They have been studied in game theory for decades and more recently generalized to \textbf{sequential social dilemmas} to enable their study in more complex environments containing spatial and temporal structure and dynamics using MARL~\citep{leibo2017ssd}.

Resources that are both non-excludable and non-subtractable are called \textbf{public goods}~\citep{ostrom2005understanding}, see Table~\ref{table:goods}. Real life examples include clean air and national defense. Groups often face situations where they must invest in order to provide a public good. In such situations it is possible for individuals to free ride: benefiting from the work of the others without working oneself. Thus situations calling for public good provision are generally rife with social dilemmas that, when they go unresolved, produce incentives---\ie individual policy gradients---pointing toward under-investment relative to the socially optimal amount of investment. \cite{mckee2021deep} studied a sequential social dilemma game called Clean Up that models an irrigation dilemma in which food only grows if an aquifer is clean, so at least some individuals must expend time and effort to clean it, an activity that contributes to the public good. However, all face incentives to free ride by standing away from the aquifer, near where the food will grow, waiting for others to do the hard work of cleaning so that they may enjoy its benefits without contributing any of their own work. \cite{mckee2021deep} found that humans are only able to find cooperative solutions to Clean Up when they can easily track one another's identity and reputation for contributing to the public good. Generic self-interested deep RL agents  failed to learn to cooperate regardless of whether they were in anonymous or identifiable conditions. The authors then showed that by endowing the agents with an inductive bias encoding the concept of competitive altruism~\citep{hardy2006nice}, their results then resembled the human results: failure to cooperate in the case of anonymous players but success in the case of identifiable players with salient reputations.

Any resource that is non-excludable and subtractable is called a \textbf{common pool resource}~\citep{ostrom2005understanding}, see Table~\ref{table:goods}. In this case, when individuals act selfishly, they may destroy a surplus that would otherwise accrue to all~\citep{ostrom1990governing}, for instance by over-consuming and thereby destroying resources on which all would otherwise benefit. Examples include common grazing pastures, fisheries, and forests. In each it is difficult or impossible for individuals to exclude one another's access. But whenever an individual obtains a benefit from such a common-pool resource, the remaining amount available for appropriation by others is at least somewhat diminished. For example, the overall fish stock is reduced whenever you remove fish from the sea; the trick is to appropriate \emph{sustainably} by removing fish more slowly than their natural birth rate replaces them given the size of the existing stock. If each individual agent’s marginal benefit of appropriation exceeds their share of the cost of further depletion, then they are predicted to continue their appropriation until the resource becomes degraded. This inexorable logic is called the \textbf{tragedy of the commons}~\citep{hardin1968tragedy, ostrom1990governing}. It is typically impossible for an individual acting unilaterally to escape this fate; since even if one were to restrain their appropriation, the effect would be too small to make a difference (assuming the group is large). Thus individual-level innovation is not sufficient to evade the tragedy of the commons. Any group-level innovation that resolves such a social dilemma must involve changing the behavior of a critical fraction of the participants~\citep{schelling1973hockey}.

Common pool resources give rise to situations where agents face incentives that guide their exploration in a systematically self-defeating direction. For the sequential social dilemma game ``Commons Harvest'', first introduced in~\cite{perolat2017multi}, agents that emit random actions typically get higher scores than agents who have begun to learn. However, there are endogenous ways by which agents can, through their learning, come to alleviate the effect of these perverse incentives. Suppose that, by building a fence around the resource or some other means, access to it can be made exclusive to just one agent. That agent is then called the owner and the resource is called a private good~\citep{ostrom2003types}, see Table~\ref{table:goods}. The owner is incentivized to avoid over-appropriation so as to safeguard the value of future flow of benefits from the resource from which they and they alone will profit. Such effects are recapitulated in these models. Agents learn strategies wherein they exclude others from a portion of the resource. Then, in accord with predictions from economics~\citep{ostrom1990governing, acheson2005spatial, janssen2008turfs, turner2013territoriality}, sustainable appropriation strategies emerge more readily in the ``privatized'' zones than they do elsewhere.

The different incentive structures of Clean Up and Commons Harvest have myriad implications. For instance, they respond differently to sanctioning motivations. Inducing agents to police one another's behavior by making (some of) them averse to disadvantageous inequity leads to cooperation in Commons Harvest but not Clean Up. In Commons Harvest, agents must learn to associate their overconsumption with negative consequences. Disadvantageous inequity averse agents feel incentivized to punish the agents who consume the most resources the most rapidly. The punishment pattern thus provided is sufficient to teach restraint, which is cooperation in Commons Harvest. On the other hand, disadvantageous inequity aversion is ineffective for Clean Up because it doesn't actually signal what the agent needs to positively do. It merely communicates what \emph{not} to do rather than what \emph{to} do. There is not enough information in the disadvantageous-inequity-aversion-induced pattern of punishment to build a whole new behavior this way~\citep{hughes2018inequity}.

Beyond inequity aversion, there are a range of other agents and learning mechanisms capable of resolving sequential social dilemmas. Several of them are algorithmically similar to the inequity aversion mechanism; they work by replacing purely self-interested individual reward functions with reward functions that take the rewards of other agents into account~\citep{gemp2020d3c, mckee2020social, baker2020emergent}. Other work in this vein extended the reward function modification method by coupling it to population-based training~\citep{jaderberg2019human} so that the reward functions themselves can evolve over the course of training~\citep{wang2019evolving}. The authors used this model to study the conditions under which the more altruism-promoting reward functions could evolve and found their best results in a case resembling group selection, in accord with the expectation from evolutionary theory~\citep{nowak2006five}. Another prominent and rather different approach is based on the concept of reciprocity. Famously in iterated prisoners dilemma, agents are incentivized to cooperate if their partner always punishes defection by defecting themselves (tit-for-tat~\cite{axelrod1984evolution}). Both \cite{kleiman2016coordinate} and \cite{lerer2017maintaining} describe hierarchical MARL agents where a hard-coded high-level controller implementing tit-for-tat decides whether to play a cooperating policy, trained with joint reward, or a defecting policy, trained using the default self-interested rewards. \cite{eccles2019learning} explored a different way of achieving cooperation via reciprocity: learn to recognize the ``niceness level'' of other agents' policies and then imitate their niceness level back to them. This approach performed well on both Clean Up and Commons Harvest.

The concept of heterogeneous preferences is important for MARL in mixed motivation settings, and will be especially important in the present paper. We regard an individual's taste as their own private information concerning the utilities they may place over the set of possible outcomes. Taste differences between agents give rise to \textbf{bargaining} problems. One recent paper considers normative disagreement as a bargaining problem~\citep{stastny2021normative}. Other work considers the case of multiple mutually exclusive public goods with complex spatial and temporal dynamics using MARL~\citep{koster2020model, vinitsky2021learning}. These interlinked collective action problems admit more than one way to cooperate, but agents have divergent preferences over which way is best. In this case, uncoordinated cooperation may be no better than mutual defection. Heterogeneous preferences make these social-dilemma-like bargaining situations more difficult. However, in the more straightforward bargaining setting of the present work, heterogeneous preferences will have the opposite effect. Agents can gain surplus utility by trading with one another precisely because they have different tastes for the various items in their world. 

In the environment considered in the present work, Fruit Market, the underlying resources are both excludable and subtractable; thus they are classified as \textbf{private goods}~\citep{ostrom2005understanding}. Also, agents have heterogeneous preferences. Thus Fruit Market falls into a part of the mixed motivation design space that has not been explored much in prior MARL work. For instance, Fruit Market is not a social dilemma. However, it is closer to the incentives structures often studied in agent-based computational economics---the field to which we now turn.

\subsection{Agent-Based Computational Economics}
\label{sec:related:ace}

In economics, \textbf{Agent-based Modeling (ABM or AB)} or \textbf{Agent-based Computational Economics (ACE)} are approaches to predicting and understanding the behaviour of economies through the interaction of individual agents~\citep{tesfatsion2021, tesfatsion2006, richiardi2017future, richiardi2014missing}. Unlike statistical approaches that model real data, or theoretical approaches that assume a population's convergence to equilibrium, agent-based approaches model the behaviours and incentives of independent entities (\eg firms, families, or individuals). The goal of this approach is to model and understand the emergent population-level economic phenomenon from the interaction of smaller components, without the rationality and consistency assumptions needed by approaches that assume equilibrium behaviour.

While the architecture of an ABM or ACE computational experiment appears to have much in common with MARL (one environment process which many individual agent processes connect to), their different objectives lead to different choices in the environments and agents. For example, for agent-based economic models to be useful, their agents' individual behavior should be simple. Hand-crafted and non-adaptive rules-based agents are commonly used. This is suitable for purpose because the focus is on generating surprising population-level results. Complex population-level effects often arise from simple individual behaviors. When adaptive or learning agents are used, their behaviour may appear constrained (from a MARL point of view) to only choose somewhat reasonable actions. ABM and ACE models emphasize the interaction between agents more so than the architecture or learning abilities of the agents themselves. In contrast, in MARL research, it is the agents and their learning dynamics that we are most concerned with, and we usually do not require our agents to accurately model any real-life or rational behaviour, so long as they earn reward. With that distinction in objective noted, however, we found the ACE modelling principles proposed by Tesfatsion to be well aligned with MARL objectives~\cite[Section 2]{tesfatsion2021}, and also note the aspiration to environments complex enough to support the emergence of generally intelligent agents~\cite[Section 6]{tesfatsion2021}.

Several decades of prior work in economics used agent-based models to study trading between agents, as we do in this work. However, as with any scientific model, including ours, the problem is abstracted from the real life setting to create a simpler and more tractable model that excludes extraneous details. For example, models in ABM, ACE, and economics-themed MARL environments may make simplifying choices in the following aspects, each representing a spectrum of options:

\begin{itemize}

    \item \textbf{Production}. Do agents decide what and how much to produce? Alternatively, are they automatically granted goods at the start of each episode, or are production decisions only represented at a very high level to focus attention on trading?

    \item \textbf{Consumption}. Do agents make choices about what and how much to consume, or are all goods automatically consumed as a ``bundle'' at the end of each episode? Are goods only consumed for immediate reward, or can agents also use them in other ways, such as saving them for future consumption, or selling or trading them to earn more reward?

    \item \textbf{Trade Prices}. Do agents choose what prices to buy and sell goods at, or what offers to make when negotiating with other agents? Alternatively, are the agent's prices determined automatically (\eg at the midpoint of two trading agents' marginal utility functions), or does the environment calculate one optimal price (\eg a Walrasian equilibrium price) and enforce its use by all agents?

    \item \textbf{Trade Partners and Interaction}. Do agents choose which specific other agents to negotiate and trade with? Are agents restricted to only trade with other predetermined ``nearby'' agents according to an adjacency graph, or can they trade with anyone in the population? Do agents interact directly with each other to exchange goods by using a sequence of actions, or does the environment facilitate trading by swapping their goods in one atomic step, or do agents submit bids and asks to a local or global ``order book'' in the environment which pairs them up and exchanges goods automatically, without direct interaction?
    
    \item \textbf{Trade Quantity}. Do agents decide how many goods to buy and sell at a chosen time and price, thus allowing strategic decisions (\eg hold some back to consume for reward, or to trade in the future if prices improve)? Alternatively, is the quantity decision automated, or do agents always sell all of their goods (if possible) after selecting a price, or are the goods being sold valueless to the agent such that they should always be sold at any available price?
    
    \item \textbf{Spatial granularity}. Does the environment have spatial dimensions, such that some agents, resources, or other entities are closer together than others? Are agents fixed in one location, or can they move through space, requiring decisions about where to produce and trade goods? Are spatial locations represented as discrete states representing nations, cities, square kilometers, square meters (about a person-sized area), or smaller (\eg does the environment represent the difference between a player holding an apple in an outstretched hand suggesting a gift, versus the player carrying that same apple in a backpack)? Alternatively, is space essentially continuous (\eg modelled at an extremely fine granularity) but with the relevant players and entities represented at one of those larger resolutions? Alternatively, is space removed as a consideration, and all agents are ``in the same location'' (\eg in one room, or online) while they interact?
    
    \item \textbf{Temporal granularity}. Is each episode represented as a single timestep, where players take one action and then see a result? Does the interaction take place over a number of discrete rounds where players take specific types of actions (\eg, all players commit to production levels, then all players choose prices to sell goods at, then all goods are sold and players receive their payouts)? Alternatively, are episodes a long series of discrete (potentially fine-grained) timesteps where players choose arbitrary actions? Does each timestep represent a time interval on the order of months, days, minutes, seconds, or less (approaching continuous time)?
    
    \item \textbf{Observation richness}. Do all agents observe the true state of the environment, or does each agent have a private observation (\eg as in an imperfect information game such as poker) that provides each agent with their own perspective? Does the observation contain only a small number of highly relevant state variables, or is the observation a rich and multimodal set of sensory inputs, such as egocentric (the world appears to move around the player, whether first-person or top-down) or allocentric (the player moves through the world) visual data represented by pixels?
    
    \item \textbf{Model provided}. Are agents given a perfect model of part or all of the environment to use for planning, or must they learn their behaviour (or build their own internal model) from experience? A common case is the agent's knowledge of their own utility function: is reward granted through some known function that can be tractably optimized (\eg a Cobb-Douglas utility function\footnote{A Cobb-Douglas function is of the form $f(\vec{x}) = \prod_i x_i^{\lambda_i}$, where $x_i$ is an expenditure on good $i$ and $\lambda_i$ is an elasticity constant for that good: whether there are increasing or diminishing returns for having more of it. Cobb-Douglas functions can be used to model an agent's production or utility objectives; both involve choosing a bundle of goods to consume in order to maximize some value. As a utility function, an agent might need to divide their budget between expenditures on food and housing, where spending zero on either good gives zero utility, and the optimal allocation can be computed tractably.}), or does the agent have an unknown utility function and must learn what behaviours grant reward through trial and error?
    
    \item \textbf{Agent quality}. Are the agents simple programs written by hand to produce a desired behaviour, such as decision trees or state machines? Do they optimize a small number of internal parameters to adjust to recent conditions, without deviating too far from a predetermined policy? Alternatively, are they reinforcement learning agents that learn arbitrary behaviours from their own experience, and so might perform randomly, poorly, or unrealistically early or even late in training, but might also discover effective behaviours on their own, or innovate new behaviours that would have been too difficult to program in manually?
    
\end{itemize}

The best choices for each of these aspects depends on the research goal being pursued. If the goal is to model a specific real-world scenario with agents, it makes sense to manually design the agents to encode or stay close to the desired behavior reflecting how the real-world agents behave, and to simplify the environment to expose only the most salient factors and decisions. There is no need to use computationally expensive state-of-the-art deep reinforcement learning agents that learn their own policy from experience, if you can already program in the specific (and economically rational) behaviour that you want them to use, and only intend to study the emergent population-level phenomenon arising from those simple agents. Much of the related work we have encountered uses simple environments and agents in order to study such emergent population-level behaviours. We will briefly note several examples here, focusing on abstract environments that attempt to investigate trading behaviour itself, and not on work that attempts to model real-world economies or behaviours. After this, we will describe in more detail the recent ``AI Economist'' work~\cite{zheng2020ai} that we believe is the closest comparison to our work because both use deep reinforcement learning agents.

In an introduction to ACE,~\cite{tesfatsion2006} describes an environment called ``The ACE Trading World''. The environment has two resources (hash and beans) and money, and three types of agents (hash-producing firms, beans-producing firms, and consumers). The environment has no spatial component and takes place over a series of discrete time periods, where agents make specific types of decisions. The firms start by choosing how much of their resource to produce and what price to charge, and then consumers attempt to buy bundles of hash and beans to maximize their utility, purchasing the lowest-priced goods first. The process then repeats, with agents updating their internal models of the predicted availability and price of each resource. The environment is used in two settings: one that is not agent-based and instead uses a ``Walrasian Auctioneer'' to calculate the equilibrium quantities and prices, and another that instead uses agents that learn to make these production and pricing decisions independently. In this work, Tesfatsion stresses the importance of agent survival in ACE models: in this environment, firms that become insolvent and consumers that fail to meet their subsistence needs are removed from the simulation, so that only successful agents remain. Where standard economic models focus on the behavior of economies operating at equilibrium and ``survival is assured as a modeling assumption'', Tesfatsion sees ACE models as stressing the agents' ability to both survive and prosper~\citep[Section 4.2]{tesfatsion2006}.

In a set of related papers, \cite{alvin1992decentralized}, \cite{wilhite2001}, and \cite{venkat2010emergence} studied the impact of spatial layouts of agents on trade. These works considered large sets of agents and adjacency graphs describing which pairs of agents were close enough to trade, and in \cite{venkat2010emergence}, a distance between those agents used for a transaction cost. Thus, space was represented discretely and abstractly, and agents could not move. The environments each used two goods and did not involve money; instead, agents exchanged goods with each other when trading would immediately improve both of their utilities. The experiments did not include agent choices about production (agents were granted a random mix of the two goods at the start of each episode), consumption (each agent $i$ optimized for a known Cobb-Douglas utility function, $U^i = g_1^i * g_2^i$, with all items consumed at the end of the episode), or the prices that they traded goods at (the price in each trade is automatically set to the midpoint of the two agents' marginal utilities for the goods). In~\cite{alvin1992decentralized}, the agents were arranged in a circle, and received a random mix of each good, with 100 items total. The agents could choose to pay a cost to advertise (within $r$ steps around the circle) their intent to trade, and two nearby agents wanting different items would reveal their marginal utilities and trade goods at the midpoint price, making both better off. Trading continued in this way until no agents chose to advertise offers. This decentralized approach involves trades happening at prices other than the equilibrium price (which would equally value both goods), and wastes some goods through advertising, but still converges to an allocation of goods across the population that is more rewarding than the initial random allocation. In~\cite{wilhite2001}, a similar arrangement of agents in a circle is used, but the emphasis is on the effort required to pick trading partners: similar to the computational cost of $O(n^2)$ for two nested for loops to consider all pairs of agents. If all agents are adjacent then the allocation of goods is efficient, but it is computationally expensive to find trading partners; when agents are arranged in local neighborhoods each exchange requires less computation to find the best partner, but more total trades are required for goods to flow through the network. By adding a small number of edges across the graph to reduce its diameter (analogous to merchants who connect otherwise distant areas), the goods are allocated quickly and efficiently. \cite{venkat2010emergence} examines a case where agents are arranged in a grid, and the random mixture of goods that each agent starts with varies between the east and west halves of the map. Agents can trade with anyone but with a linear cost determined by the distance between them. This distance cost largely determines how efficient the resulting allocation of goods is: as the cost increases local prices persist (caused by the initial skewed distribution of goods), and the collective utility of the agents drops quickly. 

\cite{manson2020methodological} presents a survey of ACE models that involve dimensions of time and space, and notes several of the challenges in using such models in practice. The authors describe different approaches for how space can be represented in the environment and observed by the agents. For example, space can be represented in an absolute way (coordinates, fixed locations for objects and entities), in a relative way (focusing on the relationships between nearby objects), or abstractly as a graph or network~\citep[Section 3]{manson2020methodological}. Specifically, the authors note possible representations of space for the agents: a tiling to discretize space and represent the entities contained in each tile, or a ``vector data model'' that represents entities as objects in the object oriented programming sense: a collection of coordinates and data elements, perhaps using lines or polygons to describe regions. This is very different from the standard approach in recent deep multi-agent reinforcement learning, where the environment's internal state representation might be entirely independent of, and not at all similar in format to the agents' observations or their internal representations. For example, a MARL environment may indeed use an object oriented approach to represent the agents and objects it contains; however, the agents are normally given a vector, image (pixel matrix), or multi-modal observation on each timestep, and the agent's internal representation of that observation is learned from experience. Any internal representation of space, therefore, is not chosen by the programmer but is instead learned and possibly different for each agent in the population.

The challenges in using spatial agent-based models decribed by~\cite{manson2020methodological} are collected from twenty years of prior work. Recurring themes include validating that models reflect the desired real world scenario, difficulty in programming the models and agents and integrating them with data sources such as GIS systems, and in communicating or sharing the models with other researchers. Similar problems are also described by Richiardi, who describes agent-based models as often lacking empirical grounding, being poorly documented and hard to replicate, and being hard to program and re-use~\cite[Page 2]{richiardi2017future}. These difficulties in programming and reuse are not surprising if the agents in agent-based models are implemented as a hand-coded collection of rules and state transitions. The use of reinforcement learning agents may help in some of these aspects and cause extra difficulty in others. For example, programming a deep reinforcement learning agent is initially a difficult programming task, but once implemented can potentially be shared with other researchers and reused in a wide range of environments with no further programming time, at the cost of requiring training time in each environment. On the other hand, such MARL agents would have no guarantees at all about producing behaviour aligned with the real-world behaviour being modelled, making validation and interpretation of results even more difficult.

The papers we have mentioned thus far largely use simple and handcrafted agents instead of such MARL agents. This is unsurprising as the above papers predate the advances in deep learning, deep reinforcement learning, and MARL throughout the 2010s. Tesfatsion does mention reinforcement learning, deep learning, and neural networks in the 2021 survey and ACE resource webpage~\citep{tesfatsion2021, tesfatsion2022webpage}. However, with a few exceptions, in our literature review we have largely not found recent application of deep learning or deep reinforcement learning techniques. One exception is a survey by~\cite{vanderhoog2017deep}, which proposes possible applications of deep learning to agent-based economic models. These proposals largely focus on training a deep learning model to emulate the policy of an agent from data, or to emulate an entire agent-based model~\citep[Page 2]{vanderhoog2017deep}. In particular, an agent-based model using millions of agents would be computationally expensive to operate, and a single deep learning model might be trained to approximate the behaviour of that multi-agent system, at a much lower computational cost. This would use deep learning to \textit{avoid} a computationally expensive multi-agent simulation, whereas in this work we use (computationally expensive) deep learning for each agent \textit{within} such a simulation. However, the survey also briefly mentions uses such as ours: creating agents with rich cognitive structure and internal models of the environment, allowing for social interaction between agents~\cite[Page 6]{vanderhoog2017deep}.

\subsection{AI Economist}
\label{sec:related:ai_economist}

Of the related work in the field of agent-based computational economics, the recent ``AI Economist'' work by~\citet{zheng2020ai} is the most closely related to this paper. Both our paper and theirs use deep reinforcement learning for agents in a 2D environment, thus emphasizing environmental richness and agent quality. From our perspective, we see our paper as ``Economics for MARL'' whereas the AI Economist paper is ``MARL for Economics''. This difference in objective leads to different choices in environments and agent training. The goal of the AI Economist work is to discover optimal tax policies to impose on a population of agents, in order to find the Pareto-optimal tradeoffs between the population metrics of equality and productivity. To do this, two types of agents are trained at once: a population of low-level agents that learn to produce, buy, sell, and consume resources, and a high-level planner agent that designs tax brackets to trade off between the population metrics. The main contribution of the work is to the field of economics, demonstrating that the high-level planner agent can choose tax brackets that better balance productivity and equality than conventional methods, while also being robust to behaviour changes by the low-level agents. Overall, the work demonstrates that deep learning agents can be a powerful component of an agent-based model for economics research.

To support that goal and contribution, the authors made several reasonable choices for the environment and agents. In the environment, the players buy and sell two types of building material (wood and stone) using coins, and coins grant reward when carried but cannot otherwise be used or consumed. The building materials are used to build houses, and the builder is given coins by the environment for doing so. Thus, the environment builds in \textit{a priori} knowledge of currency and an incentive for players to collect it. To facilitate trading between agents, the environment operates a global order book, allowing them to enter bids and asks that are fulfilled automatically, without requiring the trading agents to be nearby or to interact directly. This global order book also helps agents to explore the bid and ask actions to discover trade: the agents' joint exploration problem is easier when each agent can individually try their actions when separated in space and time, instead of having to explore when nearby and acting simultaneously. However, this also cuts off some interesting areas of investigation, such as the possible emergence of different prices for goods in different regions reflecting the local demand or abundance of resources, discovery of conventions by agents such as a ``marketplace'' area where they might meet up to trade, or the necessity of labour to transport goods from the production site to a trade partner and from there to the house building site.

On the agents side, the training procedure only trains a single set of agent parameters that is shared by all of the low-level agents. Each agent in the environment has their own private state (skills, observations, hidden state, and so on) and so still act differently from each other in order to maximize their individual reward. However, since all agents share one set of learned parameters, there is no notion of one agent learning a behaviour before, or more effectively, than their competitors. Training a single agent that interacts with many copies of itself, as they do, also simplifies the learning problem in environments that require joint exploration to discover a convention (\eg a standard price). Once two copies of the agent randomly explore their bid and ask actions and experience a trade for the first time, all of the other copies of the agent can immediately begin exploring those actions intentionally and simultaneously, making future trades much more likely to occur. If many individual agents were trained instead, like we did, then each agent would have to discover the behaviour through their own exploration, albeit with later agents being more likely to find a trade partner because early adopters would already be trading with each other. 

All of these environmental and agent choices are reasonable ways to make agent training efficient. They may also encourage the development of economically rational and interpretable behaviour in the low-level agents and the tax policy planner.

\subsection{Comparison to this work}
\label{sec:related:comparison}

Our goal in this work is to study the emergence of trading and rudimentary microeconomic behaviour with as little \textit{a priori} knowledge introduced as possible. Following the bullet points in~\ref{sec:related:ace}, we emphasize the aspects of environmental richness (spatial and temporal dimensions, and with visual observations), requiring agents to learn about production, consumption, prices, quantities, and partners, and using a population of independent state-of-the-art deep reinforcement learning agents. Unlike the related work that we have described, we do not require that our agents' behaviour after training matches any real-life or theoretically optimal behaviour. We are interested in if and how current agents can learn these skills, not whether they closely converge to an expected equilibrium behaviour.

While we do hope that our agents will learn behaviour that appears economically rational, there is no guarantee that this approach will succeed in producing a useful economic model. State-of-the-art agents could fail to learn any useful behaviour\footnote{In Section~\ref{sec:ablation:agent} we present results using an alternative agent architecture, only a few years older than the state-of-the-art agent we mainly investigate. This older agent architecture largely obtains the worst possible episodic reward in our environment. So learning to trade in Fruit Market is not at all a trivial AI problem.}, could earn reward but only as individuals that produce and then consume goods and not through interacting to trade those goods, could learn to trade but not to adapt their prices to differing environmental conditions, or could in fact learn to make decisions in accord with microeconomics of supply and demand. All of these outcomes are interesting for MARL research, but only the last could potentially make contact with rationality-oriented economics research.

Our environment differs from the related agent-based computational economics work, and in particular from the AI Economist environment, due to our goal of learning microeconomic behaviours from scratch. In our environment, agents learn to barter goods for goods, with no hard-coded notion of currency, coins, or of one good being a special \textbf{numeraire} good used as the basis for valuing other goods, unless the agents adopt such a convention on their own (something they did not yet do, but perhaps could in future work). Rewards are granted instantly for fundamental causes such as consuming tasty fruit or suffering hunger pains, instead of reward being granted at episode end through a known Cobb-Douglas function, or granted on each timestep for carrying coins which abstractly represent an agent's overall well-being (\eg, coins representing the ability to buy food and avoid hunger). In particular, we will highlight a difficulty in agents learning to barter with a consumable good: once an agent learns that they can consume a good for reward, they will begin to do so whenever possible, and then have no goods left over with which to explore trading them for something better. This difficulty cannot arise when the only use for coins is to spend them.

Our environment does introduce some domain knowledge by facilitating exchanges between agents who make offers. It is somewhat similar to the global order book used in the AI Economist work~\citep{zheng2020ai}. However, instead of one global order book, our environment considers potential offers in a small radius around each agent. This captures a more  embodied guiding intuition where trades and prices are local and entail physical interaction between nearby agents. For instance, agent i hands an apple to agent j, accepting a banana in return. This permits effects such as the emergence of different prices in different parts of the map, agents having to labour to transport goods across the map to a buyer, and agents needing to find and closely approach a trade partner in order to exchange goods with them. The agents learn this mechanism even when agents must be directly adjacent to observe each others' offers and trade resources, requiring intentional actions by agents to trade with one specific partner. Finally, unlike in AI Economist, our population of agents is independent, each agent learns their own policy through only their own stream of experience. They share no parameters and never experience episodes where they interact with copies of themselves. This permits each agent to learn a unique policy, for some agents to learn a behaviour before or more effectively than others, and for some agents to discover a niche created by other agents' behaviours. For example, in some of our experiments, a subset of the agents discover an ``arbitrage niche'' by transporting goods between parts of the map to exploit a persistent price difference. This behavior can only emerge once other agents have already learned to trade goods in a way that, as a byproduct, creates the persistent price difference which renders arbitrage rewarding.

\section{Background}
\label{sec:background}

\subsection{Markov Games}
\label{sec:background:markov}

We consider multi-agent reinforcement learning in partially observable general-sum Markov games \citep{shapley1953stochastic, littman1994markov}. In each game state, agents take actions based on a partial observation of the state space and receive an individual reward. The rules of the game are not assumed given; agents must explore to discover how the environment can be controlled and which behaviours lead to reward. Thus it is simultaneously a game of \emph{imperfect} information---each player possesses some private information not known to their adversary (as in card-games like poker)---and \emph{incomplete} information---lacking common knowledge of the rules \citep{harsanyi1967games}. Agents must learn through experience an appropriate behavior policy while interacting with one another.

We formalize this as an $N$-player partially observable Markov game $\mathcal{M}$ defined on a finite set of states $\mathcal{S}$. The observation function $\mathcal{O} : \mathcal{S} \times \{1, \dots , N\} \rightarrow \mathbb{R}^d$, specifies each player's $d$-dimensional view on the state space. In each state, each player $i$ is allowed to take an action from its own set $\mathcal{A}^i$. Following their joint action $(a^1, \dots , a^N) \in \mathcal{A}^1 \times \! \dots \! \times \mathcal{A}^N$, the state changes obeys the stochastic transition function  $\mathcal{T} : \mathcal{S} \times \mathcal{A}^1 \times \! \cdots \! \times \mathcal{A}^N \rightarrow \Delta(\mathcal{S})$, where $\Delta(\mathcal{S})$ denotes the set of discrete probability distributions over $\mathcal{S}$, and each player $i$ receives an individual reward defined as $r^i: \mathcal{S} \times \mathcal{A}^1 \times \dots \times \mathcal{A}^N \rightarrow \mathbb{R}$.

\subsection{Reinforcement learning}
\label{sec:background:rl}

Let $\mathcal{D}(\Omega)$ denote the space of distributions over the space $\Omega$. A Markov Decision Process (MDP) \citep{howard1960dynamic,sutton2018reinforcement} is a tuple $\langle \mathcal{S}, \mathcal{A}, T, r, \gamma \rangle$ where $\mathcal{S}$ is a set of states, $\mathcal{A}$ is a set of actions, $T: \mathcal{S} \times \mathcal{A} \to \mathcal{D}(\mathcal{S})$ is a transition function, $r: \mathcal{S} \times \mathcal{A} \to \mathbb{R}$ is a reward function, and $\gamma \in [0,1]$ is a discount factor. A mapping $\pi : \mathcal{S} \to \mathcal{D}(\mathcal{A})$ is called a stochastic policy. 

A partially observable Markov Decision Process (POMDP) is defined by the tuple $\langle \mathcal{O}, \mathcal{A}, T, r, \gamma \rangle$, where each element of $\mathcal{O}$ is a partial observation of a true underlying state in $\mathcal{S}$. Typically, multi-agent settings are automatically POMDPs because each agent does not have access to the observations, actions, policies or rewards of their co-players \citep{littman1994markov}.

Given a policy $\pi$ and an initial state $s_0$, we define the value function $V_\pi(s_0) = \mathbb{E} [\sum_{t=0}^\infty\gamma^t\; r(s_t, \pi(s_t))]$ where $s_t$ is a random variable defined by the recurrence relation $s_t = T(s_{t-1}, \pi(s_{t-1}))$. Reinforcement learning seeks to find an optimal policy $\pi^*$ which maximises the value function from an initial state $s_0$. We assume that the agent experiences the world in episodes of finite length $T$. During training, our RL agent receives many episodes of experience, and updates its policy to become closer to the optimal policy. We use distributed training, running several environment instances in parallel and aggregating the experience in batches for learning, resulting in a shorter wall-clock time to convergence. 

Reinforcement learning methods can be classified according to whether they represent the value function as a table of exact values (tabular) or learn it as a parametric function (function approximation). Although tabular methods have better convergence guarantees, they are impractical when the state space is large, as is the case in our 2D environment. As we will present in Table~\ref{tab:env:obs_spec}, on each timestep our agents observe a mix of visual data (a [15,15,3] matrix of pixels) and numerical data (representing inventories, offers, etc), making any tabular approach too coarse to be of use. Therefore, we employ function approximation in an actor-critic architecture, representing the policy and value function by neural networks. The reinforcement learning update rule then becomes an objective which is optimized using backpropagation. 

The standard actor-critic architecture for our setting processes visual observations of the environment with convolutional and then multi-layer perceptron (MLP) layers. Any non-visual observations are flattened, and then concatenated with the processed visual output to form a vector. This vector is fed into an LSTM layer, which allows the network to retain information through time. The output of this ``torso'' is fed into a policy head MLP and a value head MLP, which produce action probabilities and state value estimates used for training, respectively. A listing of the network architecture and agent and training hyperparameters are provided in Appendix~\ref{app:agent}.

We update our agent's policy using value-based maximum a posteriori policy optimization (\textbf{V-MPO}, \cite{song2019v}). This method is an approximate policy iteration algorithm
which uses expectation maximization (EM) under certain constraints to estimate an improved policy. Approximate policy iteration has two steps, policy evaluation and policy improvement. For policy evaluation, a value function is learned online for a policy which is fixed for a certain amount of experience $T_{\textrm{target}}$. The loss function for learning the parametric value function $V_\phi^\pi(s)$ with parameters $\phi$ is 

\begin{equation}
    \mathcal{L}_V(\phi) \propto \sum_{s_t} \left( V_\phi^\pi(s_t) - G_n^\pi (s_t) \right)^2 \, ,
\end{equation}

where the $s_t$ are drawn from a dataset of trajectories using the policy $\pi$ and $G_n^\pi(s_t)$ is the $n$-step bootstrapped return from $s_t$, which uses the trajectory rewards for the first $n$ steps, and subsequently bootstraps using the value function. From the learned value function, we define the advantage function $A^\pi(s_t, a_t) = G_n^\pi(s_t) - V^\pi(s_t)$ for each pair $(s_t, a_t)$ in the dataset of trajectories.

The policy improvement step seeks to find the maximum a posteriori estimate over policy parameters $\theta$ conditioned on the event that the new policy is an improvement. To optimize towards this objective, V-MPO takes a two step approach akin to expectation maximization under constraints. The expectation step optimizes the tightness of a lower bound on the probability that the policy is improved. The maximization step maximizes this lower bound, subject to a trust region constraint that the improved policy should not deviate from the old policy too much, as measured by the KL divergence. The policy loss turns out to be a weighted version of the familiar policy gradient loss, viz.

\begin{equation}
\mathcal{L}_\pi(\theta) = -\sum_{s,a} \psi(s,a) \log\pi_\theta(a|s) \, , \qquad \psi(s,a) = \frac{\exp \left(\eta^{-1} A^\pi(s,a) \right)}{\sum_{s,a} \exp \left(\eta^{-1} A^\pi(s,a) \right)} \, ,
\end{equation}

where $\eta$ is a hyperparameter, automatically tuned by another loss function. We refer the reader to the original paper for a full account of all loss functions. In our ablation studies, we compare V-MPO with advantage actor critic (A2C) \citep{mnih2016a3c}. Similar to V-MPO, this is an earlier distributed policy-gradient algorithm using deep neural networks for function approximation. The sequence of layers in our V-MPO and A2C networks are the same, although our V-MPO agent uses more neurons in each layer, which we found to result in higher reward with V-MPO but had no benefit for A2C.

\subsection{Multi-agent reinforcement learning}
\label{sec:background:marl}

In multi-agent reinforcement learning, a population of reinforcement learning agents learn through interactions with each other in a shared environment. In the literature, a range of options have been explored for how the agents are represented and trained. For example: is there just one shared environment, or several running in parallel; does each agent have their own set of parameters, or are some parameters shared during training; how large is the population of agents being trained, in comparison with the number of players participating in each episode of the environment.

In this paper, we follow the independent reinforcement learning approach which is standard in the recent MARL literature on sequential social dilemmas~\citep{leibo2017ssd}. We will train a population of 16 independent agents, which learn only from their own stream of experience and do not share any parameters with each other. Each agent is represented by a neural network and is trained using the V-MPO algorithm. To train the population, we use a set of 800 environment processes running asynchronously in parallel. When each environment process starts an episode, it randomly samples without replacement a set of 10 agents from the population to participate as players. The streams of experience from these many parallel episodes are sent back to the agents, who train on it to update their policy so as to maximize reward.

Each experiment that we will present in this work involved training a new population of agents from scratch, with no reuse of experience from earlier experiments. In each experiment, we ran our distributed training framework until the agents had experienced an average of \num{8e8} training steps.

To keep our terminology precise, we will use the term \textbf{agent} to refer to one instance of an algorithm and its set of learned parameters. In contrast, we will use the term \textbf{player} to refer to a position in the environment or game. For example, an agent such as AlphaZero can learn to play as the white player or the black player in chess; in any one game, AlphaZero is playing as the black player or as the white player. The rules of the game of chess describe what actions players are allowed to take, but have nothing to say about how agents should choose those actions. Thus, when describing our environment's mechanics, observations, and actions, we will refer to players; when describing learning, behaviours, and performance metrics, we will refer to agents. Since an agent may play as many different players across their episodes, we will describe an agent's performance using metrics such as ``average reward per episode''.

\subsection{Supply and Demand}
\label{sec:background:microeconomics}

In microeconomics, supply and demand curves provide a way of thinking causally about the aggregate effects of individual production, consumption, and trading decisions. Some number of agents produce a good for sale, and some others are interested in purchasing that same good; these agents need not be mutually exclusive. We refer to the interactions between these agents as a \emph{market}. For any set of environmental conditions (\eg, the abundance of goods, the reward for consuming them, and so on), we expect learning agents (both human and artificial) to converge to some equilibrium behaviour of production, consumption, and the prices that goods are exchanged at. 

\textbf{Comparative statics} is a way to study how changes to these environmental conditions affect the set of equilibria that a population of agents can reach. For example, we might run one experiment with an environment's default conditions, and measure the population's equilibrium behaviour in terms of metrics measuring production, consumption, and the price in exchanges. We might change the environmental conditions (\eg, by making apples more plentiful), train a new population of agents from scratch, and measure the new equilibrium behaviour. Comparative statics is the practice of comparing these static equilibrium points from different populations, without considering how one might transition to the other under changing conditions. 

\begin{figure}
    \centering
    \includegraphics[width=0.4\textwidth]{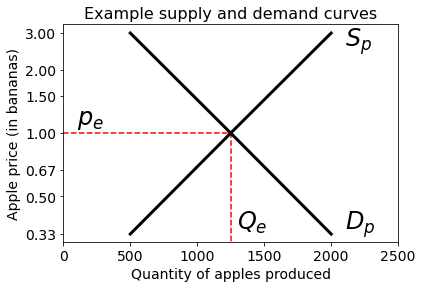}
    \caption{Theoretical supply demand curve for our setting. The x-axis measures the amount of apples produced, while the y-axis measures the price: the ratio of apples demanded per banana. While the y-axis would usually measure prices in a currency (\eg, dollars), our environment uses barter, where goods are valued in terms of each other.}
    \label{fig:basic-supply-demand}
\end{figure}

Supply and demand graphs are graphical depictions of possible equilibrium points that a population might converge to under different conditions, relating a population's willingness to produce or consume goods to the price they are be paid, or will pay, to do so. Figure~\ref{fig:basic-supply-demand} shows an example of a supply and demand graph, and Chapter 4 of~\cite{mankiw2020principles} is an excellent overview of the subject. Note that the quantity of apples produced and consumed is on the x-axis, while the price of apples is on the y-axis. The supply curve, $S_p$, demonstrates how the number of apples produced changes with price: when the price of apples is high individuals should produce more, and when the price is low individuals should produce less. If apples are expensive, an individual can get more in return for selling apples thus justifying additional labour to produce them; likewise, an individual who desires apples might choose to produce more apples themselves instead of paying the high cost. The demand curve $D_p$ relates price and consumption: when apples are expensive individuals will consume less \emph{ceteris paribus}, and if inexpensive will consume more. The intersection point of the supply and demand curves, indicated by $P_e$ and $Q_e$, is the \textbf{equilibrium point} where supply equals demand, and is the behaviour that we expect agents to converge to.

A supply and demand graph allows one to predict how a population might respond to changing environmental conditions. If apples are made more abundant, the supply curve will shift right, leading to higher production at every price. If nothing changes the desire to consume apples, the demand curve will not change, and the new intersection point between the curves will have more production and a lower price. If apples became less rewarding to individuals, we could predict that all points on the demand curve would shift to the left: lower consumption at every price, reflecting a lower willingness to pay for an apple. The equilibrium will shift to a lower price with less production. See~\cite{mankiw2020principles} for a more detailed overview.

To generate a supply and demand graph as in in Figure~\ref{fig:basic-supply-demand}, we use an empirical approach. We choose an environmental factor related to supply, such as the prevalence of apples, and run a number of experiments with differing values. For each value, we train a new population of agents and measure the resulting equilibrium behaviour. Every equilibrium point reveals an intersection of the supply and demand curves; thus, by shifting the supply curve, the intersection points reveal the shape of the \emph{demand} curve. Similarly, we vary an environmental factor related to demand to shift the demand curve, and the intersection points reveal the shape of the \emph{supply} curve.

Introducing these sources of variation is non-trivial. For one, price is not an exogenous variable that the experimenter can manipulate directly; the experimenter must manipulate other experimental measures and observe the resulting change. For another, the supply and demand curves demonstrate the relationship between equilibrium values, which require the individuals which constitute the market to have performed sufficient price discovery to determine the final price. In the real world, this is not a well-defined concept, as prices are constantly changing as the underlying conditions change. Empirical estimation of these curves is an active research topic with fractal complexity. The curves predict what happens if the price were to change \textit{ceteris paribus}: if the price were to increase, more suppliers would perform work, leading to more production, but the consumers would consume less, as each additional unit is more expensive.

When we discuss these points as the outcome at equilibrium, what we mean is that there is a process of negotiation in the marketplace which determines the market clearing price, \ie the price at which the demand for a good by consumers is equal to the number of goods that suppliers will provide for the given price. One can imagine a merchant appearing at, say, a farmer’s market to sell vegetables as an illustrative example. The merchant must guess how many vegetables will sell at a given price. At the end of each appearance, the merchant notes whether they have sold all of their vegetables or if they have remaining vegetables. If they have a surplus of vegetables, the merchant will lower their price or reduce the amount of vegetables they grow; if they had to turn customers away as they ran out of vegetables, they will either raise their price or produce more vegetables. They will continue to adjust the price they charge and the number of vegetables they supply until they either find the market clearing price, supplying the precise number of vegetables demanded by consumers, or run into some external constraint; for example, closing their shop if consumers are not willing to pay a price that is higher than the merchants cost, or being unable to sell more vegetables as they cannot increase production. 

At least, this is what the suppliers and consumers \textit{want} to do. In practice, they might not be able to. For instance, if there are no more apples to be harvested, the suppliers are not able to produce more, and price could increase without an increase in production. Similarly, if a price is already sufficiently high for individuals to spend all of their time producing, then an increase in price cannot incentivise further production. On the consumer side, as we use a barter environment, prices are calculated as the ratio of integer quantities of goods exchanged in trade. Thus, prices are discrete options and not a continuous range that can be fine-tuned. This may slow or prevent convergence. One agent might be willing to increase the price they would pay for a good, but as their only option is to offer another unit or request one less unit, the resulting price may change more than they are willing to pay. 

There is also the question as to what our reinforcement learning agents will actually do. Our agents learn from a stochastic stream of experience, must explore behaviours in order to learn their value, approximate the value of those behaviours with a neural network, and attempt to optimize their reward with respect to other members of the population, who are similarly not perfectly rational. Some joint behaviours, such as the sequence of actions by two parties required to exchange goods, might be highly mutually rewarding but still unlikely to occur through random exploration, and so agents might (suboptimally) learn to earn reward by other means without ever discovering trade. We are studying an agent-based simulation that may be \textit{prima facie} similar to a standard economics case that is analytically solvable, but in practice is a much richer interaction between complex agents.

Finally, it is important to note that supply and demand curves are a concept that may help us interpret results and predict agent behaviour after environmental changes, but are not in any way programmed into our environment or agents. Our agents learn only to maximize their individual rewards, and we will show that in many cases, their equilibrium production, consumption, and pricing behaviour does indeed move in the direction predicted by supply or demand shifts. But our goal is not for our agents to closely match an \textit{a priori} supply and demand curve or to converge to an analytically-derived optimal price. Instead, our goal is to discover what factors affect the emergence of trading behaviour in agents, and to see if their behaviour matches our rough microeconomic predictions.

\section{The Fruit Market Environment}
\label{sec:environment}

\begin{figure}
    \centering
    \begin{subfigure}{1.0\textwidth}
    \centering
    \includegraphics[height=2.9in]{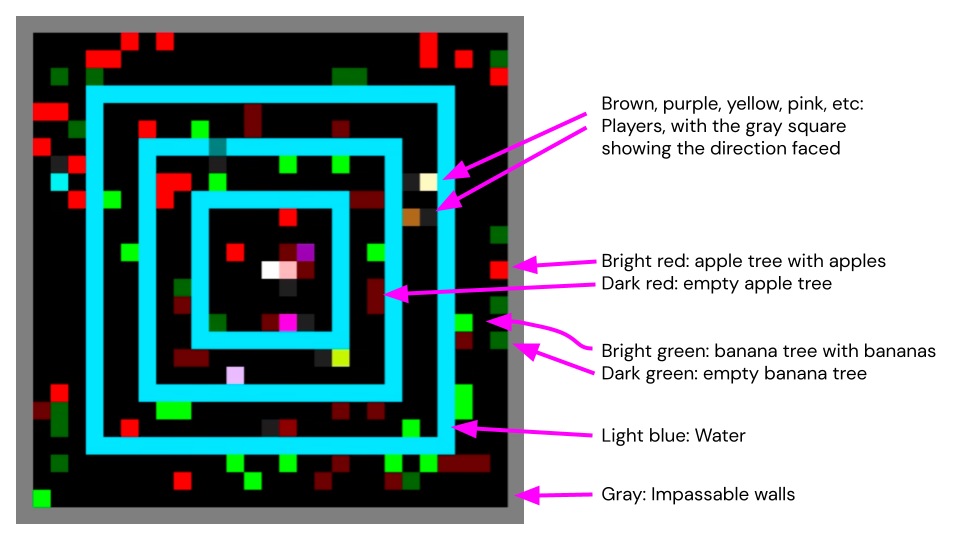}
    \caption{Version used in this paper.}
    \label{fig:map:minilab}
    \end{subfigure}
    
    \begin{subfigure}{1.0\textwidth}
    \centering
    \includegraphics[height=2.9in]{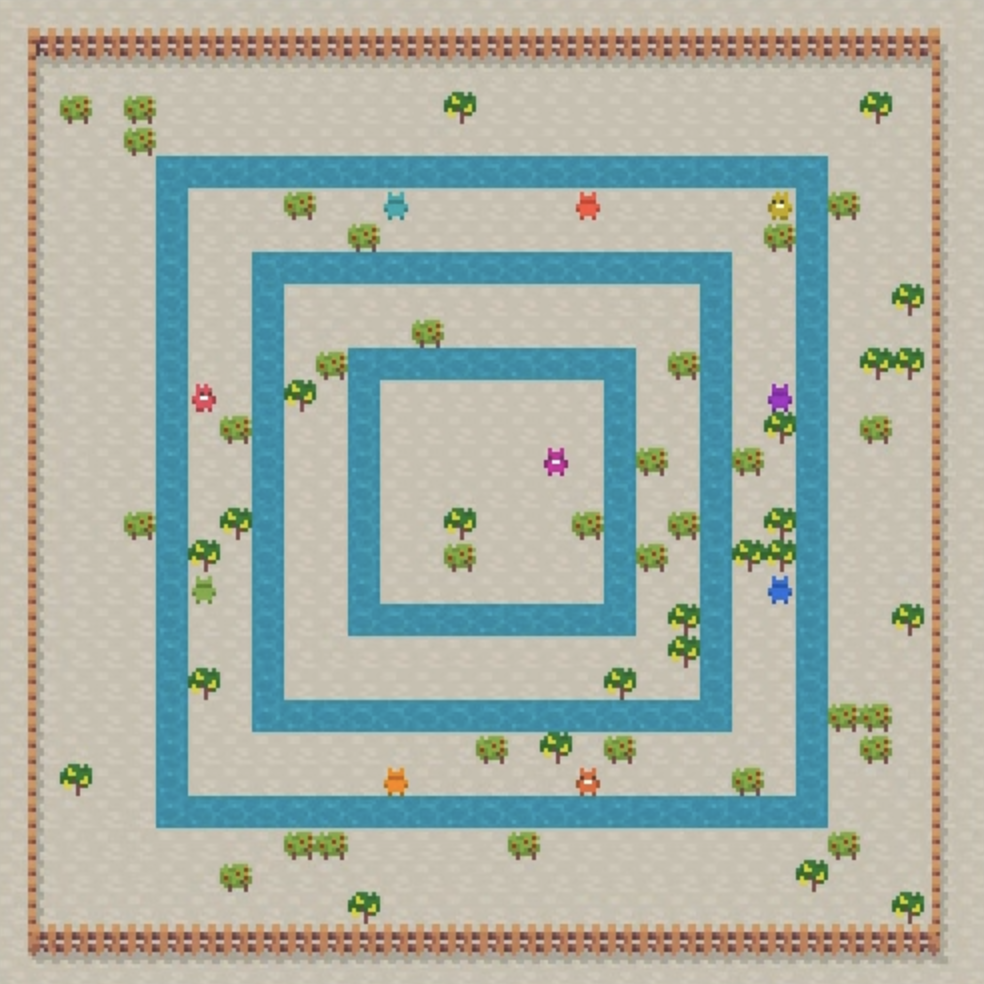}
    \caption{Upcoming Melting Pot version.}
    \label{fig:map:meltingpot}
    \end{subfigure}
    \caption{The Fruit Market environment. (a) depicts the pixel version used in the experiments in this paper. Apples and Bananas (light red and green respectively) grow on trees, and the trees turn to a darker shade after being harvested. The white square in the middle represents an optional marketplace (disabled unless otherwise mentioned) which may trade goods with the players. The pale blue lines represent water that the players can cross with a negative reward. The remaining colored squares (brown, purple, yellow, pink, \etc) are the player avatars. Agents observe a subset of this map: a $[15, 15, 3]$ patch of pixels (width, height, RGB color channels), with their avatar in the center of the bottom row, showing the pixels in the direction their avatar faces. (b) depicts a graphical version which will be released in the open source Melting Pot package. The graphical version uses sprites for each entity instead of single pixels, both for visualising the map for humans, and for the agent observations.}
    \label{fig:map}
\end{figure}

Fruit Market is an episodic multi-player game in a partially observable 2D world, where players can produce, barter, and consume fruit (apples and bananas) to earn reward. Each player's goal is to maximize their own total reward earned per episode. The environment is designed to elicit microeconomic behaviour from the players: each player must make decisions about what type of fruit it wants to produce and consume, and which offers---phrased as ``I'll give X apples to get Y bananas''---it will use to trade with nearby players. The environment is configured such that trading resources should be very mutually beneficial for both parties, \textit{if} those players, who are presumably controlled by reinforcement learning agents exploring the environment with no \textit{a priori} knowledge, can discover \textit{how} to trade.

A high level summary of the intended interaction is that half of the players (called \textbf{Apple Farmers}) are skilled at producing apples but earn more reward for consuming bananas, and the other half (\textbf{Banana Farmers}) are skilled at producing bananas but earn more reward for consuming apples. Each player takes actions to move around the map, and can produce either fruit and can consume either fruit. While Apple Farmers can slowly produce their own bananas and Banana Farmers can slowly produce their own apples---and in fact this is what happens in practice, before the agents learn to trade---it is much more rewarding for all participants if the players instead specialize in producing what they are good at, and then meet up to negotiate a trade and exchange goods. The players use actions to negotiate what quantities of apples and bananas they want to trade, allowing a player to demand a high price (if they can find a partner willing to pay it) or offer a low price to undercut their competition. This assignment of roles (\ie constant production abilities and consumption preferences) is of course not the only way to create the conditions for trade, but it is simple, has results that are easy to interpret, and is meant to replicate introductory Microeconomics textbook examples where participants have differing resources and desires\footnote{An example of an alternative role-free approach for incentivising trade would be to have homogeneous players with diminishing returns for consuming each type of fruit (or a preference for consuming bundles of an apple and a banana together) in combination with having apple trees and banana trees located in separate areas of the map. This could result in players learning to harvest apples, start to carry them to the banana area, but meet a banana-carrying player on the way and trade with them. However, we have found that in practice, reinforcement learning agents will much more easily learn to satisfy their own desires independently by producing all of their own goods, than they will jointly learn to over-produce one good and trade for what they lack. Even if the latter behaviour is more rewarding for all players, the additional difficulty of joint learning means agents are less likely to discover it. Our use of fixed roles gives additional incentive for agents to discover trading behaviour: they \textit{can} produce their desired goods independently, but very inefficiently, making trading more rewarding by comparison. See Section~\ref{sec:ablation:hunger} for further insight into environmental conditions affecting the emergence of trade.}.

Figure~\ref{fig:map} shows two top-down views of the environment. Figure~\ref{fig:map:minilab} shows the pixel version used in the experiments presented in this paper, while Figure~\ref{fig:map:meltingpot} shows a graphical version that will be available in a future release of the open-source Melting Pot package~\citep{leibo2021scalable}. The Fruit Market map is made up of discrete tiles. In Figure~\ref{fig:map:minilab}, the light and dark red tiles are apple trees, and the light and dark green tiles are banana trees. Black tiles represent empty space that players can move through easily, and the light blue rings of tiles are bodies of water that players can cross with effort. The outer gray tiles are walls that players cannot move through. The other colored tiles in Figure~\ref{fig:map:minilab}, such as yellow, purple, brown, and so on, are the ten players that participate in the game. Players cannot move onto a tile containing a wall or another player, but can move onto a tile containing a tree or water.

Each episode of Fruit Market begins by randomly spawning apple and banana trees across the empty black tiles of the map. Such procedural generation of map layouts has been shown in other settings to improve reinforcement learning agent generalization~\citep{risi2020increasing}. For our experiments we can choose the placement of each type of tree independently, both in distribution (\eg, uniform, gaussian, skewed to the left, skewed to the right, and so on) and in frequency (\eg, 15\% apples and 15\% bananas, 30\% apples and 15\% bananas, 5\% apples and 10\% bananas, \etc). For the moment, we will assume that trees are placed at uniform random with an equal probability of 15\% of each type of tree per black tile, which were the parameters used to generate Figure~\ref{fig:map:minilab}. Ten players (five Apple Farmers and five Banana Farmers) are then spawned onto the map at predetermined starting locations. On each timestep the players are shown the observations listed in Table~\ref{tab:env:obs_spec} and then submit one action from Table~\ref{tab:env:act_spec}; both of these tables will be described in detail below. The episode ends after 1000 timesteps.

\begin{table}
    \centering
    \begin{tabular}{|l|l|l|}
        \hline
        Name & Shape & Description \\
        \hline
        Vision & $[15, 15, 3]$ & \makecell[l]{A rectangular visual field with the agent centered at the bottom.\\14 tiles ahead, 7 tiles to either side, RGB color channels.} \\ 
        \hline
        Inventory & $[2]$ & Number of apples and bananas the agent is carrying. \\ 
        \hline
        Needs & $[1]$ & Agent's ``hunger satiation'' level. A scalar ranging from 0 to 30. \\ 
        \hline
        Own Offer & $[2]$ & Vector of [apples, bananas] representing the player's current offer. \\ \hline
        Offers & $[P, 2]$ & \makecell[l]{Matrix of observed offers from nearby players. \\$P$ is the number of players in the environment. \\ Each row is an offer vector of [apples, bananas]. \\ Rows are $[0, 0]$ when the other player is out of range.} \\
        \hline
        Previous Action & [1] & Index of the action chosen on the previous timestep. \\
        \hline
        Reward & [1] & Total reward earned since the previous timestep.\\
        \hline
    \end{tabular}
    \caption{Observation Specification for players in Fruit Market. Players are provided with these observations on each timestep, and the agent acting as that player may use them however they wish (\eg, flatten, concatenate, and use as input to a neural net) to choose an action.}
    \label{tab:env:obs_spec}
\end{table}

\begin{table}
    \centering
    \begin{tabular}{|l|l|l|}
        \hline
        \multicolumn{2}{|c|}{Action} & Description \\ 
        \hline
        \multirow{6}{*}{Moving (7)} & Stand Still & Do nothing.\\ 
        \cline{3-3}
        & Left & \multirow{4}{*}{Move one tile.} \\
        & Right & \\
        & Forward & \\
        & Backward & \\
        \cline{3-3}
        & Turn Left & \multirow{2}{*}{Turn 90 degrees.} \\
        & Turn Right & \\
        \hline
        \multirow{2}{*}{Consuming (2)} & Eat Apple & \multirow{2}{*}{Consume one fruit from inventory.} \\
        & Eat Banana & \\
        \hline
        \multirow{19}{*}{Offers (19)} & Cancel & Set offer vector to $[0, 0]$. \\
        & 1a:1b & Set offer vector to $[-1, 1]$. \\
        & 1a:2b & Set offer vector to $[-1, 2]$. \\
        & 2a:1b & Set offer vector to $[-2, 1]$. \\
        & 2a:2b & Set offer vector to $[-2, 2]$. \\
        & 1a:3b & Set offer vector to $[-1, 3]$. \\
        & 2a:3b & Set offer vector to $[-2, 3]$. \\        
        & 3a:1b & Set offer vector to $[-3, 1]$. \\
        & 3a:2b & Set offer vector to $[-3, 2]$. \\
        & 3a:3b & Set offer vector to $[-3, 3]$. \\
        & 1b:1a & Set offer vector to $[1, -1]$. \\
        & 1b:2a & Set offer vector to $[2, -1]$. \\
        & 2b:1a & Set offer vector to $[1, -2]$. \\
        & 2b:2a & Set offer vector to $[2, -2]$. \\
        & 1b:3a & Set offer vector to $[3, -1]$. \\
        & 2b:3a & Set offer vector to $[3, -2]$. \\
        & 3b:1a & Set offer vector to $[1, -3]$. \\
        & 3b:2a & Set offer vector to $[2, -3]$. \\
        & 3b:3a & Set offer vector to $[3, -3]$. \\
        \hline
    \end{tabular}
    \caption{Action Specification for players in Fruit Market. Players have 28 discrete actions, and choose one to take on each timestep. Fruit is produced by using the movement actions to stand on a tree tile, and not moving away until it is automatically picked up into the player's inventory. Players can still consume fruit or make offers while waiting to harvest fruit.}
    \label{tab:env:act_spec}
\end{table}

\begin{table}
    \centering
    \begin{tabular}{|c|r|r|r|r|r|r|}
        \hline
        Name & \multicolumn{2}{c|}{Production Quantity} & \multicolumn{2}{c|}{Production \%} & \multicolumn{2}{c|}{Consumption Rewards} \\
        & \multicolumn{1}{c|}{Apple} & \multicolumn{1}{c|}{Banana} & \multicolumn{1}{c|}{Apple} & \multicolumn{1}{c|}{Banana} & \multicolumn{1}{c|}{Apple} & \multicolumn{1}{c|}{Banana} \\
        \hline
        Apple Farmer (AF) & 2 & 2 & 100\% & 5\% & 1 & 8 \\
        Banana Farmer (BF) & 2 & 2 & 5\% & 100\% & 8 & 1 \\
        \hline
    \end{tabular}
    \caption{Role production and consumption constants. Players of each role produce two of each fruit when they successfully harvest a tree. Apple Farmers have a 100\% probability per timestep of harvesting ripe apple trees, but only a 5\% probability per timestep of harvesting ripe banana trees. Apple Farmers earn 1 reward for consuming an apple, and 8 for consuming a banana. Banana Farmers have these constants reversed.}
    \label{tab:env:role_constants}
\end{table}

\begin{table}
    \centering
    \begin{tabular}{|l|l|l|}
    \hline
    Constant & Value & Description \\
    \hline
    Fruit Ripening Time & 50 & Number of timesteps for fruit to regrow after harvest. \\
    \hline
    Movement Penalty & -0.25 & Reward per tile moved, representing exertion. \\
    \hline
    Water Penalty & -1.0 & Reward per timestep touching water, representing exertion. \\
    \hline
    Hunger Level & 0 -- 30 & \makecell[l]{Satiation, starting at 30, and decreasing by 1 per timestep. \\ Refilled to 30 after consuming any fruit.} \\
    \hline
    Hunger Penalty & -1 & Reward per timestep if Hunger Level is 0. \\
    \hline
    Trade Radius & 4 & \makecell[l]{Euclidean distance radius another player making a \\complementary offer must be within for an exchange to happen.} \\
    \hline
    Offer Visibility Radius & 4 & \makecell[l]{Euclidean distance radius another player must be within \\ to observe their offer.} \\
    \hline
    \end{tabular}
    \caption{Additional environmental and player constants that are not role-dependent.}
    \label{tab:env:shared_constants}
\end{table}

\subsection{Movement, Production, and Consumption}
\label{sec:env:prod_con}

Each player has seven movement actions: one to stand still, four to take a step (forwards, backwards, left, right) and two to turn (left, right). The players do not observe the complete map as shown in Figure~\ref{fig:map:minilab}, but instead observe a smaller patch of it: an egocentric $[15,15,3]$ matrix with their own avatar in the middle of the bottom row, showing what is ahead of them in the direction they are facing (7 tiles to either side, 14 tiles ahead). As they move and turn, the world appears to move around them in their visual observation. Since they can only see a small portion of the map, the players need to move around to discover where they can find apples and bananas, and where other players are, so that they can move away to avoid competing for the same resources, or approach them to trade goods.

Each player, regardless of their role, can produce and consume both apples and bananas. Each role's constants for production and consumption are listed in Table~\ref{tab:env:role_constants} and other related constants in the environment are listed in Table~\ref{tab:env:shared_constants}, all of which we will now describe in detail. Production occurs when a player moves on top of a tree bearing ripe fruit. With a fixed probability per timestep, the player harvests two fruit from the tree and places them into their inventory. This inventory mechanic is not found in most reinforcement learning environments, where an agent would typically immediately consume a rewarding object on touch. Apple Farmers have a 100\% probability per timestep of harvesting apples and a 5\% probability per timestep of harvesting bananas, while Banana Farmers have these constants reversed. After fruit is harvested from a tree, the tree is empty and requires 50 timesteps to grow new fruit. The difference between a tree with ripe fruit and an empty tree is displayed visually, by switching from a light shade of green or red to a dark shade, and then back to a light shade when the tree is ready to be harvested again.

Consumption occurs when a player has apples or bananas in their inventory and uses the ``Consume Apple'' or ``Consume Banana'' action, which consumes one fruit. An Apple Farmer earns 1 reward for consuming an apple and 8 reward for consuming a banana, while Banana Farmers have these values reversed. Our environment uses constant rewards for consumption, and the difference between the fruit is a simple model of an Apple Farmer's likely satiation of their desire for apples\footnote{A more realistic satiation model would involve diminishing reward for repeated consumption of each good, and then gradual recovery over time. This would lead to a ``natural'' lower reward for apples in Apple Farmers, without requiring role-specific rewards. It would also give Apple Farmers a reason to produce some of each fruit for their own consumption even before they discover trading, instead of specializing in producing whichever type of fruit had the higher expected value per timestep (harvesting probability times reward, for example). We intend to explore such nonlinear reward models in future work.}. In addition to earning reward, consuming fruit also addresses a player's \textbf{hunger} needs. Each player has a \textbf{hunger level} that they observe, which starts at 30 (full) and decreases at 1 per timestep towards 0 (starving). Eating one fruit of any type resets their hunger level to 30. If it reaches 0, then the player suffers a reward of -1 per timestep until they consume any fruit. In the experiments that we will present, trained agents learn to produce and consume fruit frequently enough to almost entirely avoid the hunger penalty. However, it plays a critical role in the emergence of trading behaviour, which we will explore later in Section~\ref{sec:ablation:hunger}.

\subsection{Offers and Exchanges}
\label{sec:env:trading}

Finally, we can describe the actions and mechanics that allow players to trade resources with each other. First, we will note that there is a broad range of mechanics we could use for this, trading off realism, precise control over price, quantities, and trading partners, and the difficulty for agents to learn the mechanics. For example, we could implement simple and fundamental actions such as picking up and dropping fruit, or giving one fruit to an adjacent player. Players could then drop an apple next to a trade partner, wait for them to drop a banana, and then each pick up the other's fruit. Alternatively, we could use an abstract mechanism where agents choose to enter a bartering interface, similar to computer role-playing games. We could choose a mechanism that allows players to explore any behaviour, including ``bad'' behaviour such as giving items away for nothing or running away with the partner's offered goods while giving nothing in return, or we could constrain the players' behaviour to allow only reasonable offers and disallow theft. Finally, we could choose a trade mechanism that uses domain knowledge such as currency (\eg selling apples for dollars and buying bananas with dollars) which is a convention that humans have already discovered, or have the environment facilitate exchanges by pairing up players who want to buy and sell the same good at mutually acceptable prices and automatically exchanging their goods.

The trade mechanism we will present strikes a balance, with our goal being to create a system that is simple, gives agents control over their trade quantities, prices, and partners, and includes minimal domain knowledge, while still being learnable by our current agents such that trading behaviour emerges. When we have compromised in these aspects, it is with the intent that future research can use this work as a benchmark where agents consistently do learn to trade, and then pursue the emergence of trade with fewer compromises. At the end of the paper, in Section~\ref{sec:ablation:trade}, we will re-examine our decisions by presenting a series of simpler and more expressive trade mechanisms, and show that our current agents do not yet demonstrate rational trading behaviour in those settings.

Our trade mechanism involves a set of actions that allow players to make \textbf{offers} to trade quantities of apples and bananas. An offer is represented by a vector of $[\textrm{apples}, \textrm{bananas}]$, such as $[-1, 1]$, which means ``I will give 1 apple to get 1 banana''. The two elements of the vector represent the player's desired change in inventory of apples and bananas. Put another way, negative numbers are what a player is willing to give, and positive numbers are what a player wants in return. The offer $[1, -1]$ means ``I will give 1 banana to get 1 apple'', and is the inverse of the previous offer: two players making these offers should be happy to trade, as they are each willing to give exactly what the other wants. The offer $[-2, 1]$, which means ``I will give 2 apples to get 1 banana'' thus gives more, and $[-1, 2]$ or ``I will give 1 apple to get 2 bananas'' demands more. Throughout the text, we will describe offers like $[-1, 1]$ with a phrase such as ``Give 1 apple for 1 banana'' or the short name ``1a:1b'', where what is being given is on the left of the colon, and what is demanded is on the right.

Each player has a persistent \textbf{offer vector} which is ``advertised'' to other nearby players within an \textbf{offer visibility radius} of 4 tiles. Each player observes their own offer vector and other nearby offer vectors on each timestep. The default value of an offer vector is $[0, 0]$, the null offer, which means that the player does not want to trade right now. Players can change the value of their offer vector by taking one of nineteen \textbf{offer actions}, listed in Table~\ref{tab:env:act_spec}. Each offer action sets the player's offer vector to a specific value, and eighteen of the offer actions enumerate all possible exchanges of apples and bananas up to a maximum quantity of 3 items of each type. The final ``cancel offer'' action resets a player's vector back to $[0, 0]$. Once set to a value, the player's offer vector persists through time and is advertised to other players as they move around the world to produce and consume items. It only changes when the player uses a different offer action to set it to a new value, cancels their offer to set it to $[0, 0]$, or fulfills their offer by trading with another player, after which it resets to $[0, 0]$. A player can only set their offer vector to an offer that they can fulfill: they cannot offer to give away three apples if they only have one. Attempting to do so instead resets their offer vector back to $[0, 0]$. Similarly, if a player is already making an offer and then consumes an item such that they can no longer fulfill it, their offer vector is reset to $[0, 0]$.

We use the term \textbf{compatible} to describe a pair of offers that satisfy each other. Specifically, two offers are compatible if each gives \textit{at least} as many items of each type as the other requests. When two offers exactly satisfy each other, such as $[-1, 1]$ and $[1, -1]$, we call them \textbf{inverse} offers, which are a subset of compatible offers. But what if the first player is unnecessarily generous by offering $[-2, 1]$, thus willing to give two apples while the second player's offer of $[1, -1]$ only requested one apple? These offers are not inverse, but are still compatible, because both players would be happy to trade using either offer's quantities. With either offer, one player would simply get a better deal than expected: either getting an extra apple, or only having to give away one apple, respectively. Thus, the inverse offers of $[-1, 1]$ and $[1, -1]$ are compatible, as are $[-2, 1]$ and $[1, -1]$, or even $[-2, 1]$ and $[1, -2]$ where both players offer more than the other requests. The offers $[-1, 2]$ and $[1, -1]$ are not compatible, since the first player wants two bananas, and the second player is only willing to give one.

An \textbf{exchange} of goods occurs when two players are simultaneously making compatible offers and are within a \textbf{trade radius} of 4 tiles\footnote{Note that the \textbf{offer visibility radius} for observing another player's offer, and the \textbf{trade radius} for exchanging goods with another player, are both set to 4 by default. There is no need for these constants to be equal: two people might speak across a room to negotiate a deal, but have to move close together in order to exchange goods. Later, in Section~\ref{sec:experiments:trade_radius}, we will investigate the impact of shrinking or growing both of these constants on the agents' ability to learn to trade.}. This exchange of goods is automatic (the players do not take any additional ``exchange'' action) and atomic (the items are swapped between their inventories in one step). This is an abstraction of the physical task of exchanging items: the players state their willingness to trade through their offers, and then the environment intercedes by exchanging their goods when it detects two nearby players have compatible offers. An alternative mechanism could require agents to take a sequence of actions to exchange goods, but might be too difficult for our current reinforcement learning agents to learn through joint exploration.

The precise exchange process is as follows. At the end of each timestep, the environment loops over all players making offers in a random order, so as to not give any advantage to lower-indexed players (\eg, player 1 versus player 10). Next, there is a two step task: selecting which partner (if any) that player will trade with, and then determining the quantities of goods exchanged. To select the partner, the environment makes a list of all potential partners in their trade radius that are making compatible offers. Any of these potential partners whose offers are \textbf{dominated} by other offers in the list are removed. For example, if player A offers $[-1, 1]$, potential partner B offers $[1, -1]$ and potential partner C offers $[1, -2]$, then C's offer dominates B's by offering an extra banana, and so B is removed from A's list of potential trade partners. The resulting list contains the most generous partners' offers from the player's perspective. For each partner in the list, the environment generates a similar list of compatible undominated offers from the partner's perspective; if the player is not also among the \textit{partner's} most generous offers, then the partner is removed from the player's list. If multiple partners remain in the list, the environment breaks ties by distance between the players, and then randomly selects one.

The second step is to determine the quantity of goods exchanged. Because we selected a pair of players whose offers were compatible, and not strictly inverse, there could be a range of quantities agreeable to both players. In our implementation, we exchange the minimum quantity of goods that satisfies both parties, and this might be different from either or both players' offers. For example, if player A offers $[-2, 1]$ and players B offers $[1, -1]$, then we exchange $[-1, 1]$ from player A's perspective, which happens to be the inverse of player B's offer. If player A offers $[-2, 1]$ and player B offers $[1, -2]$, then both players are willing to give more than the other requests, and we exchange $[-1, 1]$ from player A's perspective. 

Having selected both a partner and a quantity, the environment then atomically exchanges the items between the player's and partner's inventories, and resets their offers back to $[0, 0]$. The environment then continues its randomized loop over the players to check for other exchanges to perform.

The exchange mechanism described above is similar to a spatially local order book, where nearby bids and asks are paired together. Players have fine control over how they barter in public, can observe nearby offers to accept or compete with before making their own, and can increase what they will give or decrease what they demand in order to make a deal. This system also has useful properties for reinforcement learning agents, who have no \textit{a priori} knowledge about how their offer actions work and only discover their use through trial and error. First, performing exchanges between the broad range of compatible offers instead of only the one inverse offer makes it more likely that agents will experience actually exchanging goods as they jointly explore their actions. Second, the partner selection mechanism's preference for generous offers (by ignoring dominated offers) encourages agents to explore in the space of offers to either give more or request less than their neighbors are offering. That is, an agent can explore making a more generous offer, and then experience trading more often but at a worse ratio. Unfortunately, this mechanism also encodes some domain knowledge (\eg, that a generous offer is preferable to a lower one, and that stating an offer is a sufficient representation for the finer steps of actually handing goods back and forth without theft). Later, in Section~\ref{sec:ablation:trade}, we will consider alternative mechanisms that remove the environment's involvement in partner and quantity selection.

Before continuing we have two final comments on our trade mechanism, both relating to the concept of money. Throughout this work, offers will only involve apples and bananas: the two goods that agents can produce through labor. Our trading method is thus a barter system, where agents will be able to negotiate deals by offering to give more items away, or request fewer items in return. We have intentionally not introduced a third resource to represent money, and this is for two reasons. First, currency is itself domain knowledge that we do not want to encode. In one view, it is a solution to the problem of resources being discrete quantities and the difficulty of finding a trade partner whose desires simultaneously coincide with one's own assets. Second, in future work involving environments with more than two resources, we want to leave open the possibility of agents jointly learning a convention to use one resource as a numeraire good in which other goods are valued and exchanged (\eg, sell apples for chocolate, and then buy bananas with chocolate). We feel that the emergence of such a convention would be a significant milestone for reinforcement learning agents, and also an opportunity to study how resource properties influence its selection or not as the numeraire, potentially yielding a useful testing ground for classical theories of the emergence of money (e.g.~the theories reviewed in~\cite{smit2011money}).

However, as people who are used to thinking in terms of money, words such as \textbf{price} are a convenient shorthand for describing the value of things. For example, a player's offer of $[-1, 3]$ or ``Give 1 apple for 3 bananas'' describes that player's valuation of an apple as being worth 3 bananas. It is convenient for us, as observers of the environment, to describe this as a \textit{high price} for apples: it costs 3.0 bananas to buy one apple. Similarly, an offer of $[-3, 1]$ or ``Give 3 apples for 1 banana'' is a \textit{low price} for apples: it costs 0.33 bananas to buy an apple. This is a casual use of the term  ``price'' as no currency (\eg dollars) is involved. Further, neither ``apples'' or ``bananas'' are a numeraire good to the agents, who only set and observe offer vectors where these are simply goods A and B. In our experiments, when we describe results such as the population's average price for apples, we mean only the average ratio of bananas per apple in the population's exchanges.

\subsection{Opportunity Costs}
\label{sec:environment:opportunity_costs}

Players in our environment encounter two final sources of reward: a \textbf{movement penalty} of -0.25 reward per tile moved, and a \textbf{water penalty} of -1 per timestep spent on a water tile. Both of these penalties are included to provide an opportunity cost to players: a value for laziness, by making labour costly. These penalties are important in order for changes in the population's offer behaviour to influence the population's production behaviour. For example, without these penalties, Apple Farmer agents might learn to maximize reward by producing as many apples as possible, trading them for bananas using the $[-1, 1]$ offer, and then eating the bananas. However, if Banana Farmers started offering $[1, -2]$, thus giving more bananas per apple, this could not result in any increase in apple production: Apple Farmers were already producing as many apples as possible. Thus the population could still learn to trade, but their behaviour would not be economically rational.

With the movement and water penalties, the agent should learn to balance the (now non-zero) marginal cost of harvesting the next most convenient apple, against the future reward it could obtain after trading them for bananas, with the alternative to harvesting being to stand still. A higher price for apples could then incentivise more production, and a lower price could incentivise more idleness and fewer apples produced. Of course, the emergence of such behaviour is not guaranteed, as it depends on the agents accurately learning to evaluate labour and laziness from their own experience.

\subsection{Distributed Training}
\label{sec:environment:distributed}

As we described in Section~\ref{sec:background:rl}, our experiments will use the distributed training paradigm to train our reinforcement learning agents in Fruit Market. Each experiment will train a population of 16 agents, each permanently assigned a role of either an Apple Farmer or a Banana Farmer. Each episode of Fruit Market will be played by 10 of these agents by sampling 5 Apple Farmer agents and 5 Banana Farmer agents at uniform random without replacement. Our distributed training framework runs 800 independent episodes of Fruit Market in parallel, each sampling different agents, and sending the resulting stream of experience back to the agents for training.

In our empirical results we will use ``Average Agent Steps'' on the x-axis of graphs as a measurement of training time. This metric measures the average number of timesteps experienced by the population of agents at one moment; because of the random sampling of agents in each episode, there will be some minor variation in the amount of training each agent has had. Unless otherwise noted, our experiments run until agents have experienced an average of $8e8$ (or 800,000,000) training steps. Each episode of Fruit Market lasts 1000 timesteps, and so this is equivalent to each agent training on an average of $8e5$ (or 800,000) episodes. Overall, approximately 1.3 million episodes were used for training, since only 10 out of 16 agents participate in each episode.

\subsection{Overview}
\label{sec:environment:overview}

We will conclude by describing the high-level behaviour that we hope our agents will learn. Recall that each agent is assigned a permanent role as either an Apple Farmer or a Banana Farmer. If we changed the environment to disable trading, or if the agents did not learn how to use their offer actions to trade, then we would expect each agent to learn how to produce fruit and then consume it for reward. There are several reasonable strategies: Apple Farmers could learn to quickly produce apples for a small consumption reward each, or could slowly produce bananas for a large consumption reward each, or (most likely) could harvest both, depending on what fruit was nearby and how imminent their hunger penalty was.

If the agents do learn how to trade, however, then we would hope to see a different joint behaviour: Apple Farmers should efficiently produce apples, Banana Farmers should efficiently produce bananas, all agents should make offers to trade, and all agents should consume their preferred fruit. By trading, all agents should earn much more reward than they otherwise could in the previous scenario without trade. However, there is still room for a wide range of behaviours. With the range of offer actions we have provided, the population could learn to trade at any ratio from ``1 apple for 3 bananas'' to ``3 apples for 1 banana''. And to an Apple Farmer who earns 1 reward per apple and 8 reward per banana, \textit{any} price in that range is preferable to just eating apples. There is no \textit{a priori} correct price that one agent should converge to; each agent's optimal behaviour depends on the current behaviour of the rest of the population. Of course, over time as all agents adapt their prices to compete with each other, we would hope to see convergence to prices that reflect the abundance of each type of fruit.

In the rest of this paper, we will investigate what our reinforcement learning agents are able to learn about barter and trade in the Fruit Market environment. Does trading behaviour indeed emerge between our agents? If so, how quickly does it emerge? Do the agents converge towards consistent and ``fair'' offers that equally value apples and bananas when those goods are equally plentiful? What kinds of offers do the agents make over time as they explore trading? If we vary the environmental conditions to change the supply or demand of one type of fruit, does the population's production, consumption, and offer behaviour change in the directions predicted by supply and demand shifts?

In the next section, we will investigate these questions, and show that the agents' learned behaviour indeed largely aligns with microeconomic predictions. We will then explore more spatially interesting maps for Fruit Market, where apples and bananas are abundant in different regions, and will show that different local prices emerge, that the agents can also learn arbitrage behaviour by transporting goods between regions, and that this specialization in arbitrage and transportation by some agents earns more reward than producing and selling goods.

\section{Experiments}
\label{sec:experiments}

We will begin our empirical analysis with an extended example of a simple map, to build the reader's understanding of the model, its dynamics, and what kinds of behaviour we can measure. We will also demonstrate that agents quickly and consistently learn to trade as their means of maximizing reward. Next, we will adopt a microeconomic view and explore how production, consumption, and pricing behaviour changes as we exogenously perturb the environment, similar to supply and demand shifts. We also study the extent to which the population's behaviour matches the microeconomic predictions from supply and demand shifts. Finally, we will consider environments with varied geographic features in order to explore how the emergent prices of goods vary through space.

\subsection{Production, Consumption, and Trade}
\label{sec:experiments:baseline}

\begin{figure}
    \centering
    \begin{subfigure}{0.39\textwidth}
        \centering
        \includegraphics[height=2in]{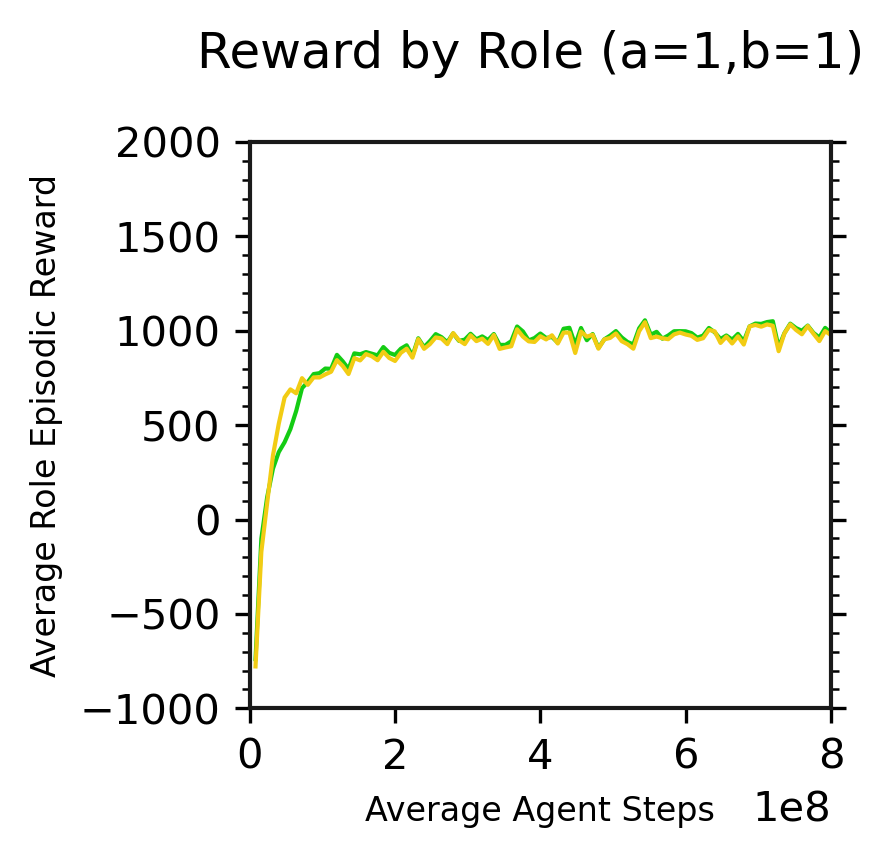}
        \caption{}
        \label{fig:baseline_reward:a1_b1_role_returns}
    \end{subfigure}%
    ~
    \begin{subfigure}{0.59\textwidth}
        \centering
        \includegraphics[height=2in]{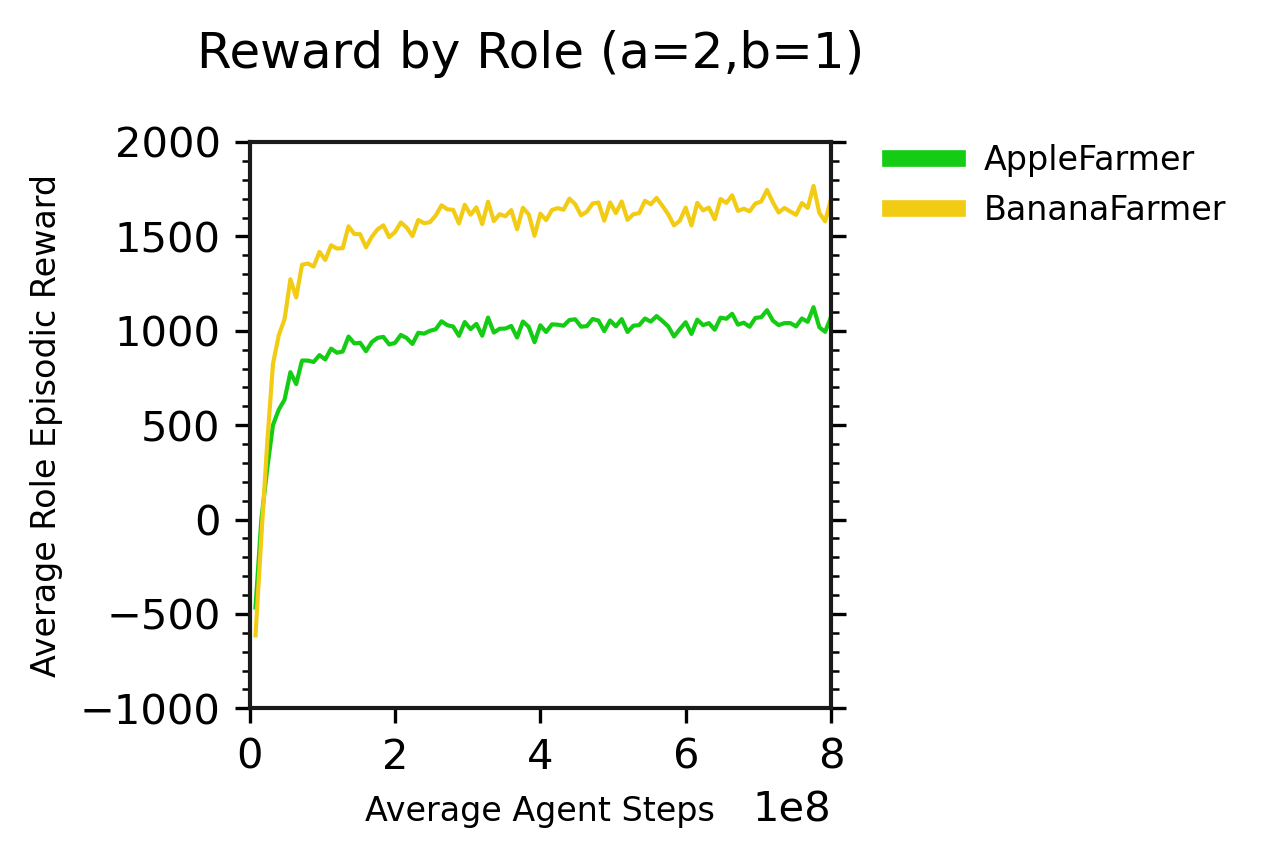}
        \caption{}
        \label{fig:baseline_reward:a2_b1_role_returns}
    \end{subfigure}
    
    \begin{subfigure}{0.39\textwidth}
        \centering
        \includegraphics[height=2in]{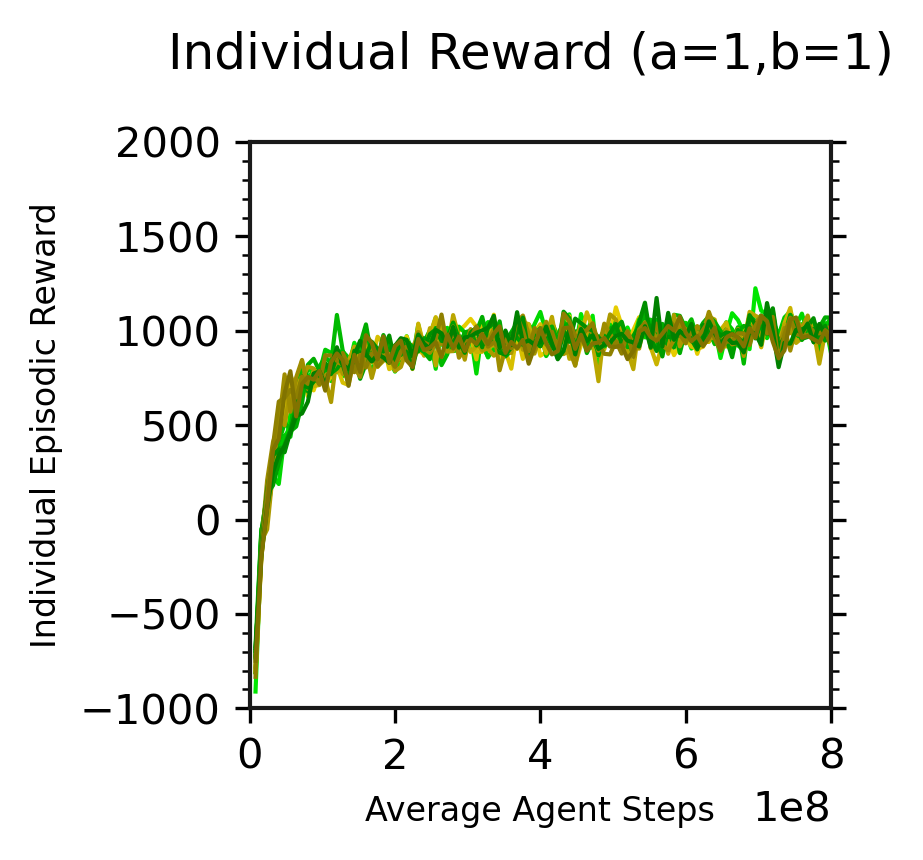}
        \caption{}
        \label{fig:baseline_reward:a1_b1_individual_returns}
    \end{subfigure}%
    ~
    \begin{subfigure}{0.59\textwidth}
        \centering
        \includegraphics[height=2in]{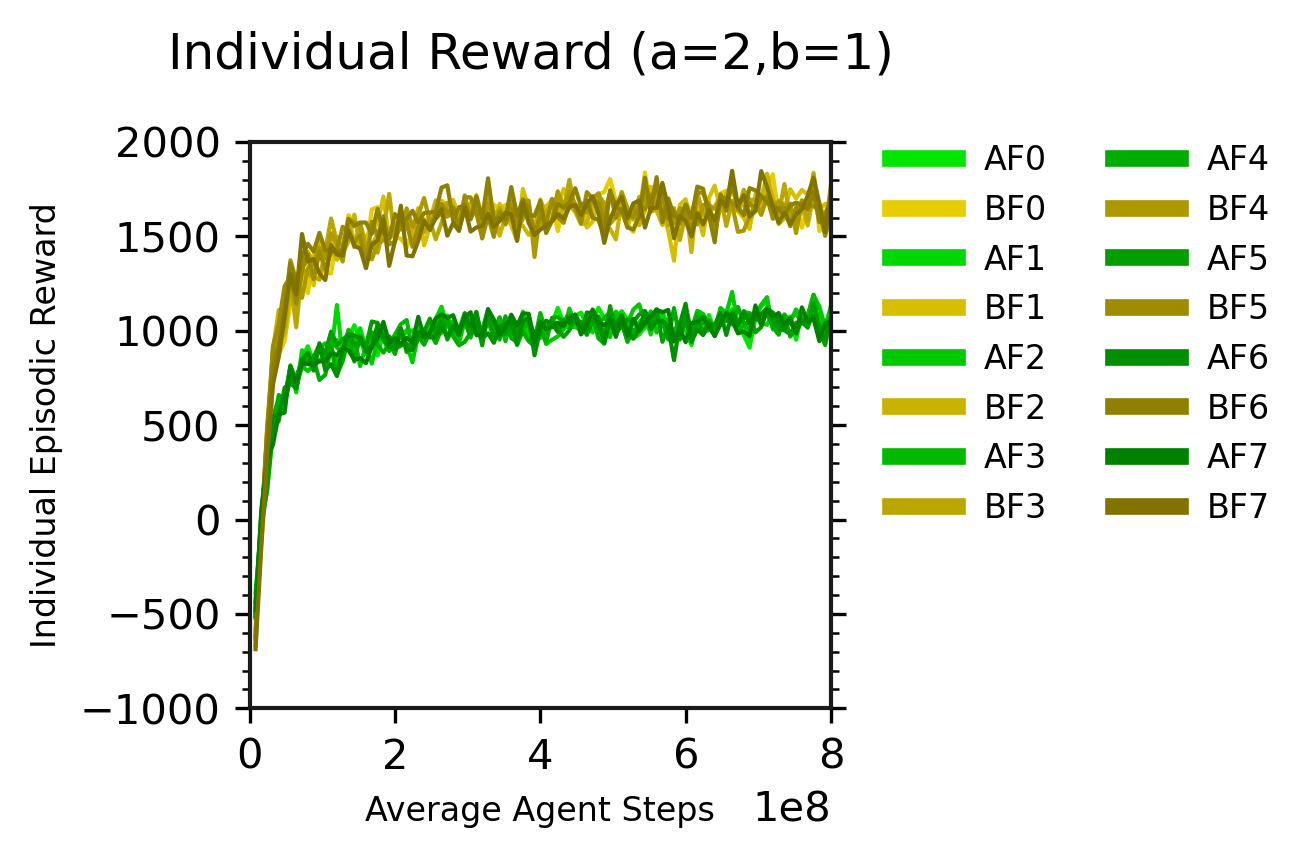}
        \caption{}
        \label{fig:baseline_reward:a2_b1_individual_returns}
    \end{subfigure}
    
    \caption{Average episodic return for roles and agents in two experiments. (a) and (c) show the default setting, where the base probability for each map location of spawning apple trees and banana trees, 15\%, is multiplied by 1 (\ie, unmodified). (b) and (d) show an alternate setting where the probability of apple trees is multiplied by 2, thus 30\% for apple trees and 15\% for banana trees. (a) and (b) show the average episodic return for all agents of one role, while (c) and (d) show each agent's performance individually, color coded by their role, with 'AF0' indicating the first Apple Farmer agent.}
    \label{fig:baseline_reward}
\end{figure}

We will begin by demonstrating what our agents can learn in a simple environment. These experiments use the map presented in Figure~\ref{fig:map:minilab}, where each type of tree appears with uniform density across the map, and the placement of trees is chosen randomly at the start of each episode. We will compare two experiments throughout this section: one where apple and banana trees are equally common, and one where apple trees are more common than banana trees, so that we can observe how the agents' behaviour changes.

In each episode, each empty map location has a base probability of spawning an apple tree of 15\%, and another 15\% probability for spawning a banana tree. We vary the environment by multiplying this base probability by a multiplier $a$ or $b$, giving a bonus or penalty to apple or banana trees. In the $(a=1,b=1)$ setting, apple and banana trees are thus equally common, and appear at the base rate of 15\% each. In the $(a=2,b=1)$ setting, apple trees spawn twice as often (\ie, with 30\% probability), with banana trees unmodified at 15\%. We trained an independent population of agents for each of these settings.

We begin exploring these two populations' behaviour with the foundational measurement of agent behaviour in reinforcement learning: reward over time. Figure~\ref{fig:baseline_reward} presents the average episodic reward over training in each setting. Figures~\ref{fig:baseline_reward:a1_b1_role_returns} and \ref{fig:baseline_reward:a2_b1_role_returns} summarize the agents' reward by averaging across each role, while \ref{fig:baseline_reward:a1_b1_individual_returns} and \ref{fig:baseline_reward:a2_b1_individual_returns} show each agent's individual return as a separate line. In both cases, the episode statistics are binned into 100 equal width bins on the x-axis and averaged, to more easily show the trend from the approximately 1.3 million episodes being summarized. From these graphs, we see that the agents consistently and quickly learn and then plateau. Figures~\ref{fig:baseline_reward:a1_b1_individual_returns} and \ref{fig:baseline_reward:a2_b1_individual_returns} demonstrate that all agents perform very similarly as the eight lines for each role largely overlap, and so for this section we will focus on the average behaviour across agents of each role. Finally, by comparing the $(a=1,b=1)$ and $(a=2,b=1)$ settings, we can see that increasing the abundance of apples is a much larger benefit to Banana Farmers (who prefer to consume apples) than to Apple Farmers (who are better at producing apples, but prefer bananas).

\begin{figure}
    \centering
    \begin{subfigure}{0.46\textwidth}
        \centering
        \includegraphics[height=3.8in]{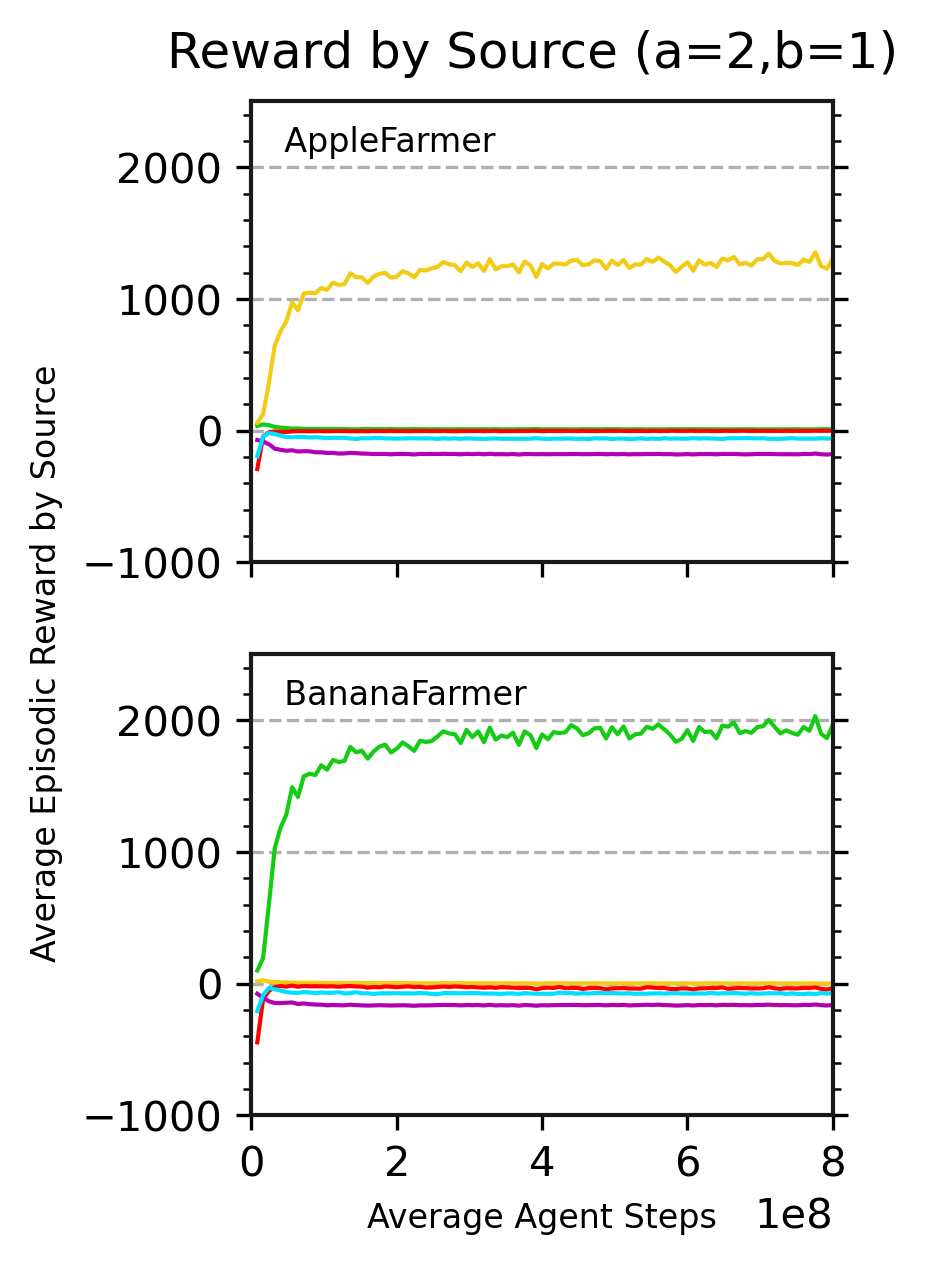}
        \caption{}
        \label{fig:baseline_reward_source:all}
    \end{subfigure}%
    ~
    \begin{subfigure}{0.52\textwidth}
        \centering
        \includegraphics[height=3.8in]{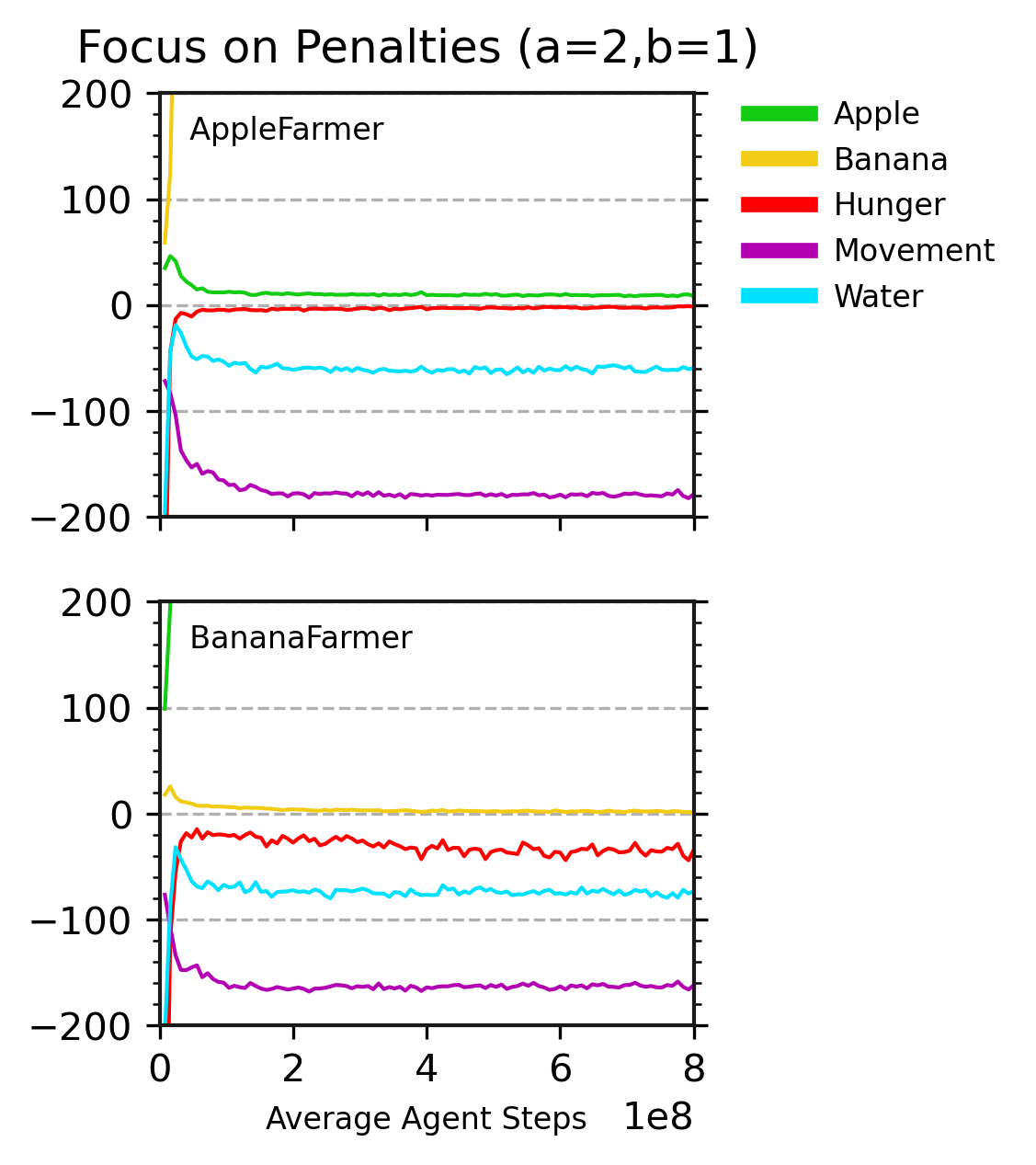}
        \caption{}
        \label{fig:baseline_reward_source:penalties}
    \end{subfigure}

    \caption{Sources of reward in the $(a=2,b=1)$ setting, averaged over the agents of each role. (a) shows the entire range of rewards, while (b) focuses on the range (-200, 200) to highlight the hunger, movement, and water penalties.}
    \label{fig:baseline_reward_source}
\end{figure}

We can dig into why Banana Farmers earn more reward than Apple Farmers when we make apples more plentiful. Using the $(a=2,b=1)$ setting, Figure~\ref{fig:baseline_reward_source:all} breaks down the episodic reward for each role by its source in the environment: how much is gained from eating apples and bananas, lost to labour from movement and crossing water, and lost to hunger pains if agents do not eat frequently enough. We see that Apple Farmers earn most of their reward from consuming bananas, and Banana Farmers earn most of their reward from consuming apples. Figure~\ref{fig:baseline_reward_source:penalties} zooms in on the less rewarding item (apples for Apple Farmers, bananas for Banana Farmers) and the penalties (hunger, movement, and water), all of which are of a small magnitude compared to the large reward gained from the more rewarding item. The $(a=2,b=1)$ setting has more apples to be consumed, and from this plot we see that it is the apple consumers -- that is, Banana Farmers -- who benefit, as they earn just under 2000 reward per episode for doing so. By comparison, Apple Farmers do eat a few apples for a small reward, but gain by far the majority of their reward from consuming just over 1000 reward worth of bananas per episode. While this graph suggests trade is occurring, it does not yet confirm if this is the case, or if each role is inefficiently producing the items they prefer.

\begin{figure}
    \centering
    \begin{subfigure}{0.44\textwidth}
        \centering
        \includegraphics[height=2.5in]{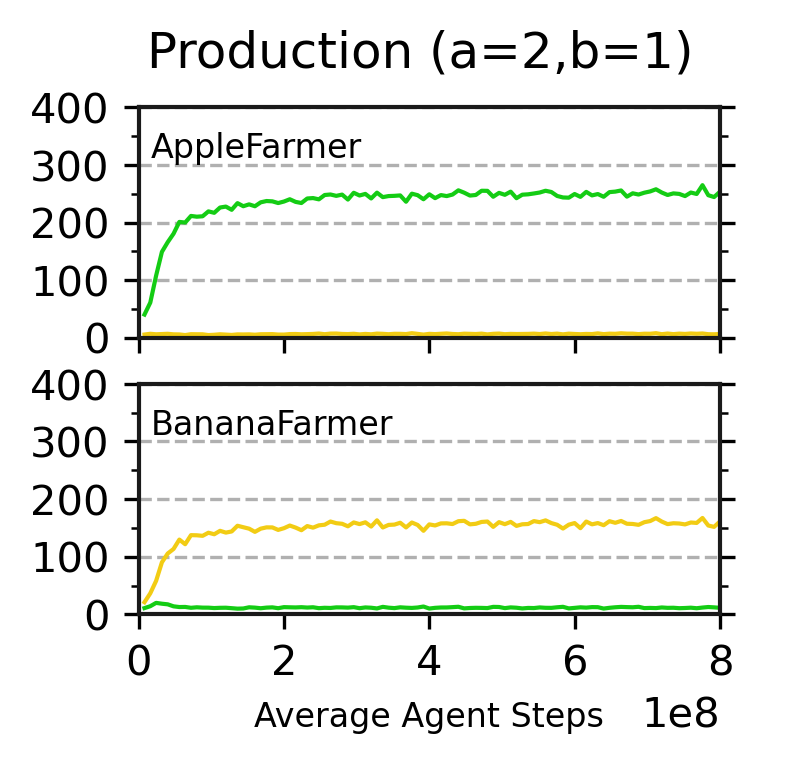}
        \caption{}
        \label{fig:baseline_prod_con:production}
    \end{subfigure}%
    ~
    \begin{subfigure}{0.54\textwidth}
        \centering
        \includegraphics[height=2.5in]{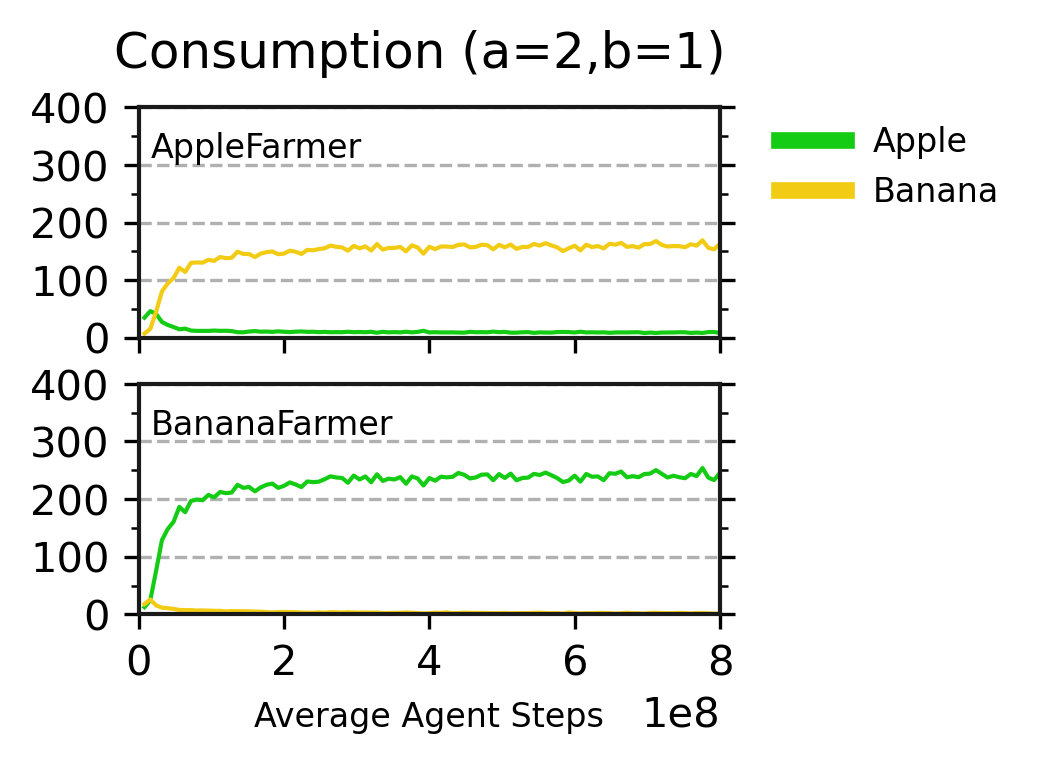}
        \caption{}
        \label{fig:baseline_prod_con:consumption}
    \end{subfigure}
    \caption{Average episodic production and consumption by role. These plots measure the quantity of fruit produced or consumed, unlike Figure~\ref{fig:baseline_reward_source} which measured the reward for consuming fruit.}
    \label{fig:baseline_prod_con}
\end{figure}

Figure~\ref{fig:baseline_prod_con} gives our first confirmation that the agents are trading goods by plotting how many items each role produces and consumes. Figure~\ref{fig:baseline_prod_con:production}, plots production by role, and shows that Apple Farmers produce mostly apples, and Banana Farmers produce mostly bananas. Figure~\ref{fig:baseline_prod_con:consumption} shows consumption by role in terms of the \textit{quantity} of items consumed, instead of our previous figure which presented the \textit{reward} for consuming each fruit. We see that while agents of each role initially consume some of the item they are specialized to produce, they quickly shift their consumption almost exclusively to the item they prefer. Since each item consumed must have been produced by someone, and Apple Farmers produce apples but consume bananas, trade must be occurring.

\begin{figure}
    \centering
    \begin{subfigure}{0.44\textwidth}
    \includegraphics[height=2.5in]{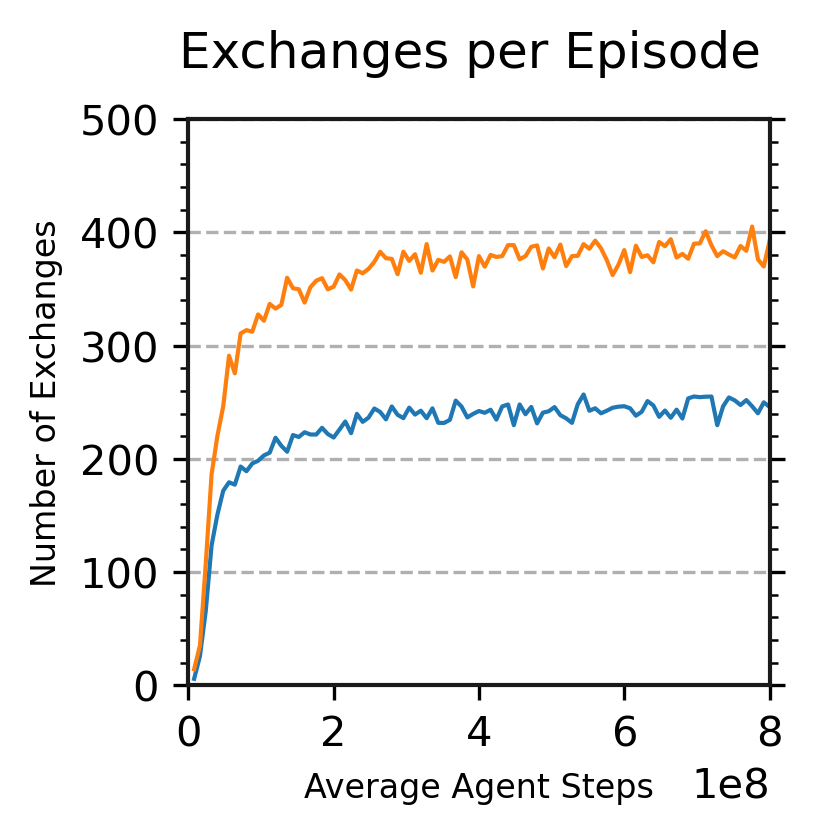}
    \caption{}
    \label{fig:baseline_exchanges_price:exchanges}
    \end{subfigure}%
    ~
    \begin{subfigure}{0.54\textwidth}
    \includegraphics[height=2.5in]{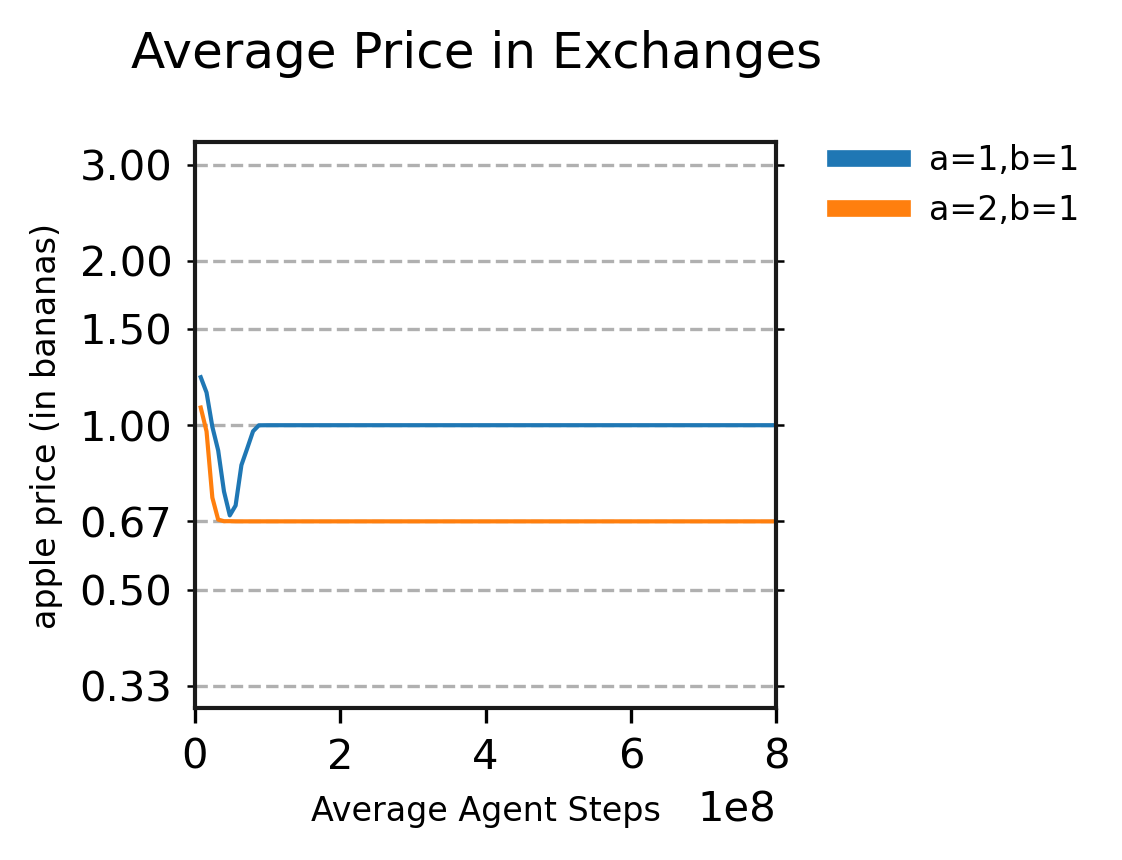}
    \caption{}
    \label{fig:baseline_exchanges_price:prices}
    \end{subfigure}
    
    \caption{Quantity and price of exchanges, for the (a=1,b=1) and (a=2,b=1) settings.}
    \label{fig:baseline_exchanges_price}
\end{figure}

Next, we can investigate this trading behaviour directly: what offers are the players making, what exchanges occur, and how does this change between the $(a=1,b=1)$ and $(a=2,b=1)$ settings. Figure~\ref{fig:baseline_exchanges_price} presents a high-level summary of the number of exchanges that occur per episode (between any partners and of any quantity of items), and the average ratio of bananas per apple in those exchanges. Figure~\ref{fig:baseline_exchanges_price:exchanges} shows that in both settings, the agents quickly discover resource trading as a collective behaviour. Figure~\ref{fig:baseline_exchanges_price:prices} shows that in both settings, the average price shifts initially while the number of exchanges is ramping up, before settling on a stable price, with apples being less valued in exchanges when apple trees are more common.

Note that these prices are not mandated by the environment: the role rewards of 1 for the resource an agent can produce efficiently or 8 for the resource they prefer are constructed such that trading with any available offer from 1a:3b to 3a:1b should be preferable to both agents over just consuming the fruit an agent can efficiently produce. If the agents in the $(a=2,b=1)$ setting had also arrived at an average ratio of 1 apple for 1 banana, or even at a ratio of 1 apple for 2 bananas where apples are expensive, those could possibly still be local optima for a population; it is interesting that they have instead arrived at a price that matches our intuitions, where a more common good trades at a lower price. As we will see throughout the rest of the paper, this behaviour is not rare or selectively chosen from our results, but is instead the usual outcome.

\begin{figure}
    \centering
    \begin{subfigure}{0.34\textwidth}
        \includegraphics[height=2in]{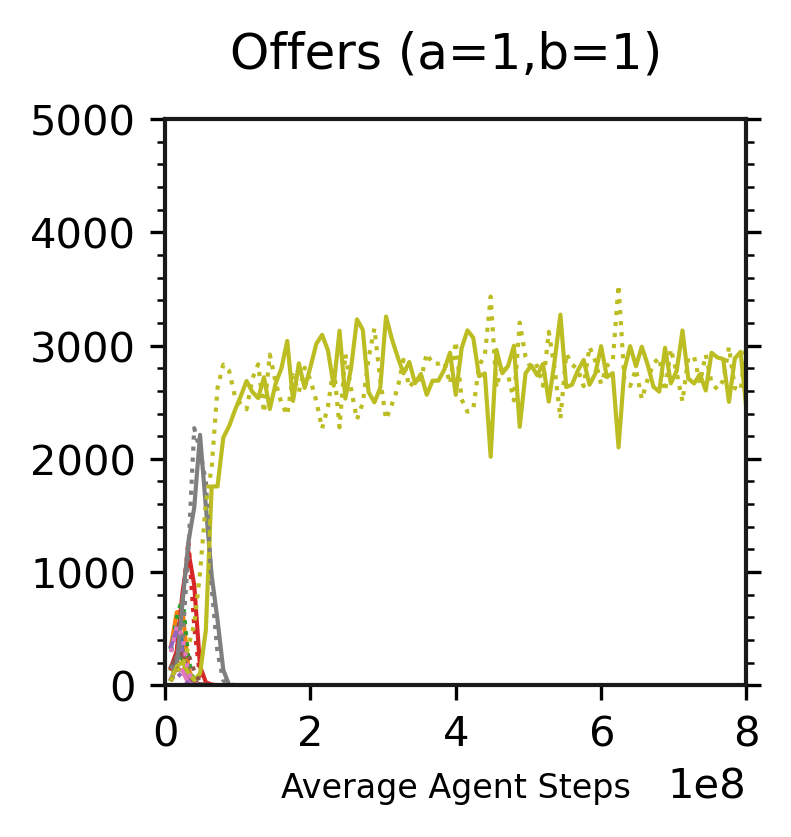}
        \caption{}
        \label{fig:baseline_offers_a1:offers}
    \end{subfigure}%
    ~
    \begin{subfigure}{0.64\textwidth}
        \includegraphics[height=2in]{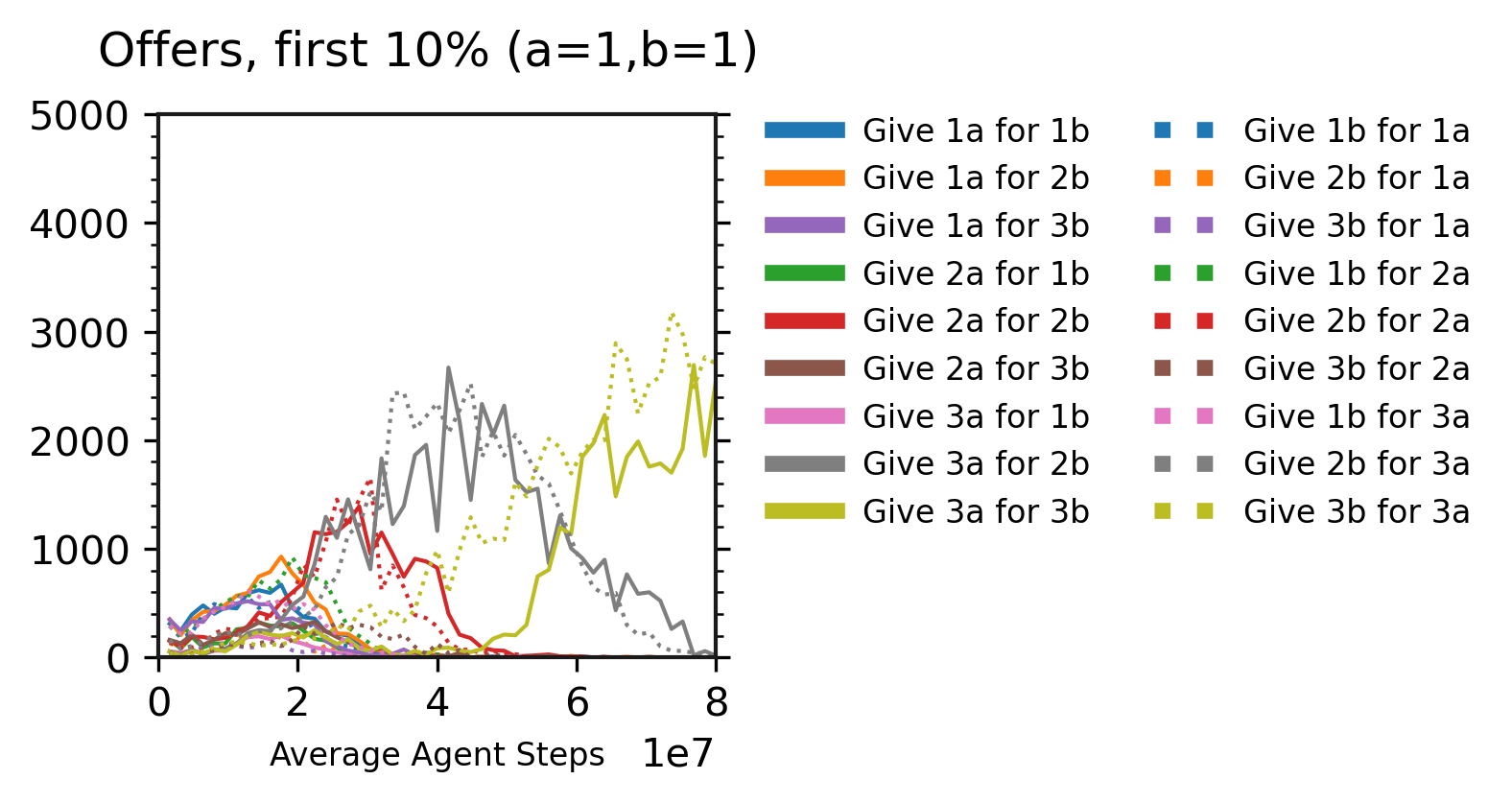}
        \caption{}
        \label{fig:baseline_offers_a1:offers_early}
    \end{subfigure}
    
    \begin{subfigure}{0.34\textwidth}
        \includegraphics[height=2in]{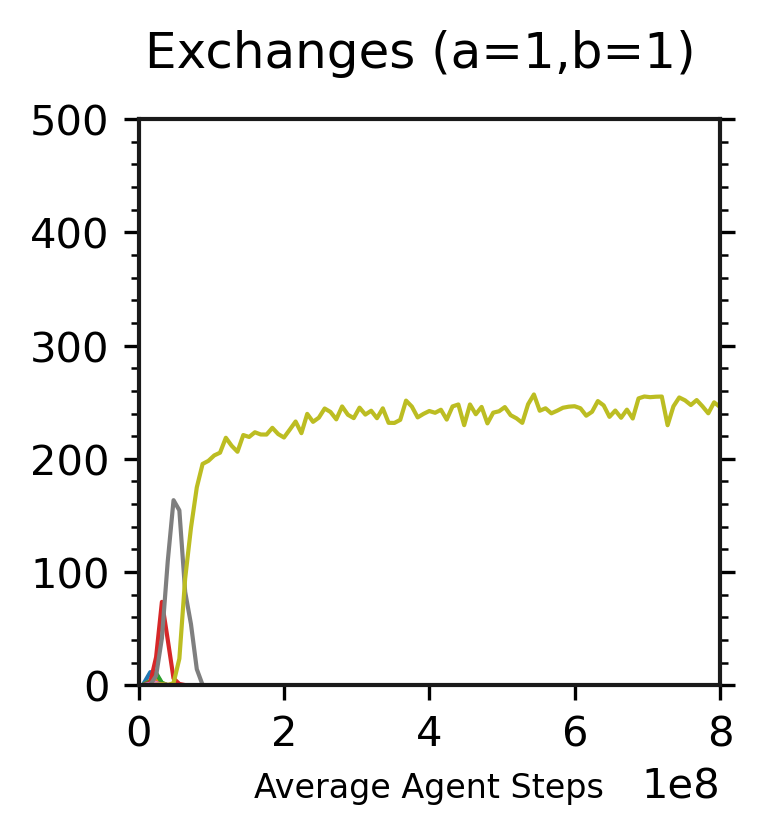}
        \caption{}
        \label{fig:baseline_offers_a1:exchanges}
    \end{subfigure}%
    ~
    \begin{subfigure}{0.64\textwidth}
        \includegraphics[height=2in]{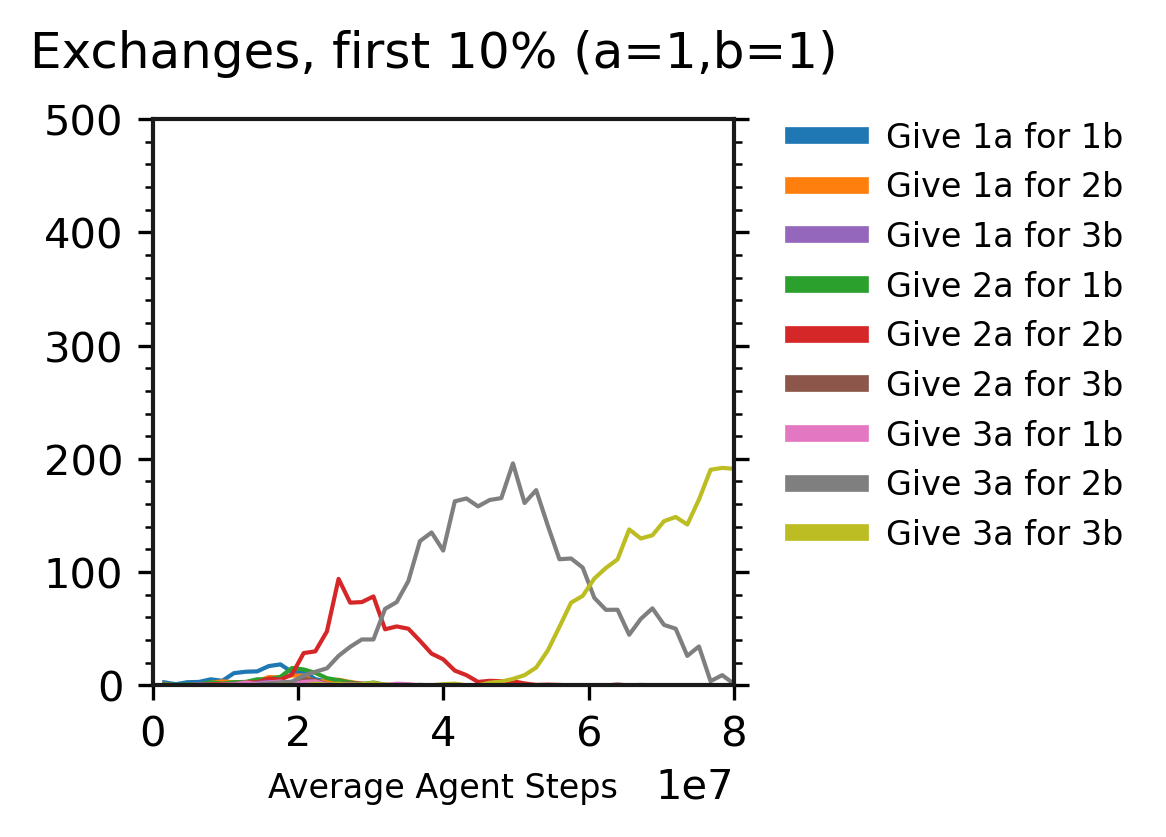}
        \caption{}
        \label{fig:baseline_offers_a1:exchanges_early}
    \end{subfigure}

    \caption{Average quantity of each offer and exchange in the $(a=1,b=1)$ setting. (b) and (d) zoom in on the first 10\% of the experiment.}
    \label{fig:baseline_offers_a1}
\end{figure}

\begin{figure}
    \centering
    \begin{subfigure}{0.34\textwidth}
        \includegraphics[height=2in]{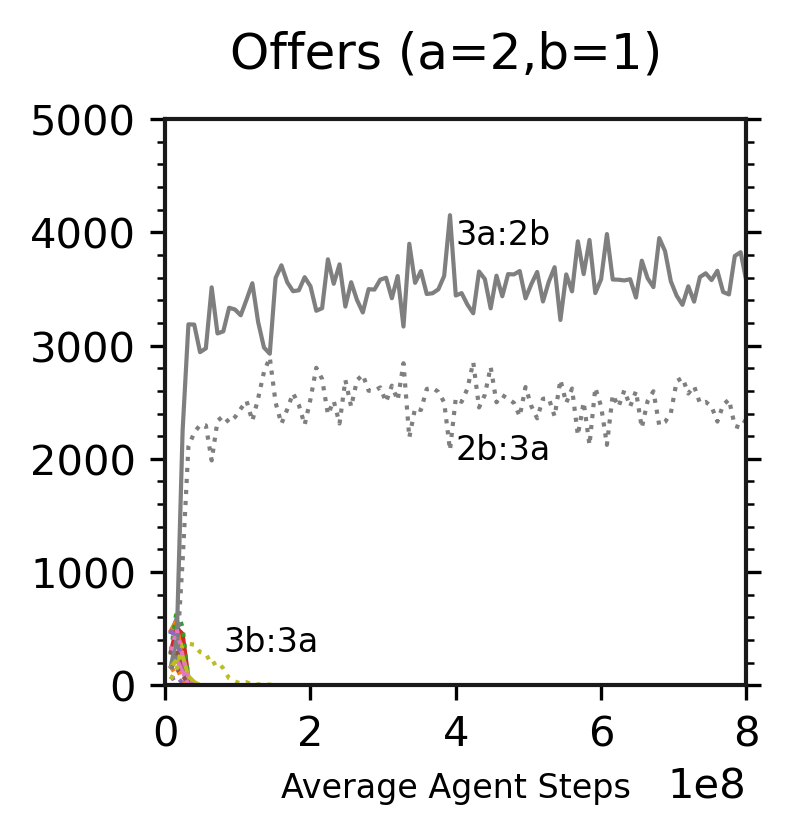}
        \caption{}
        \label{fig:baseline_offers_a2:offers}
    \end{subfigure}%
    ~
    \begin{subfigure}{0.64\textwidth}
        \includegraphics[height=2in]{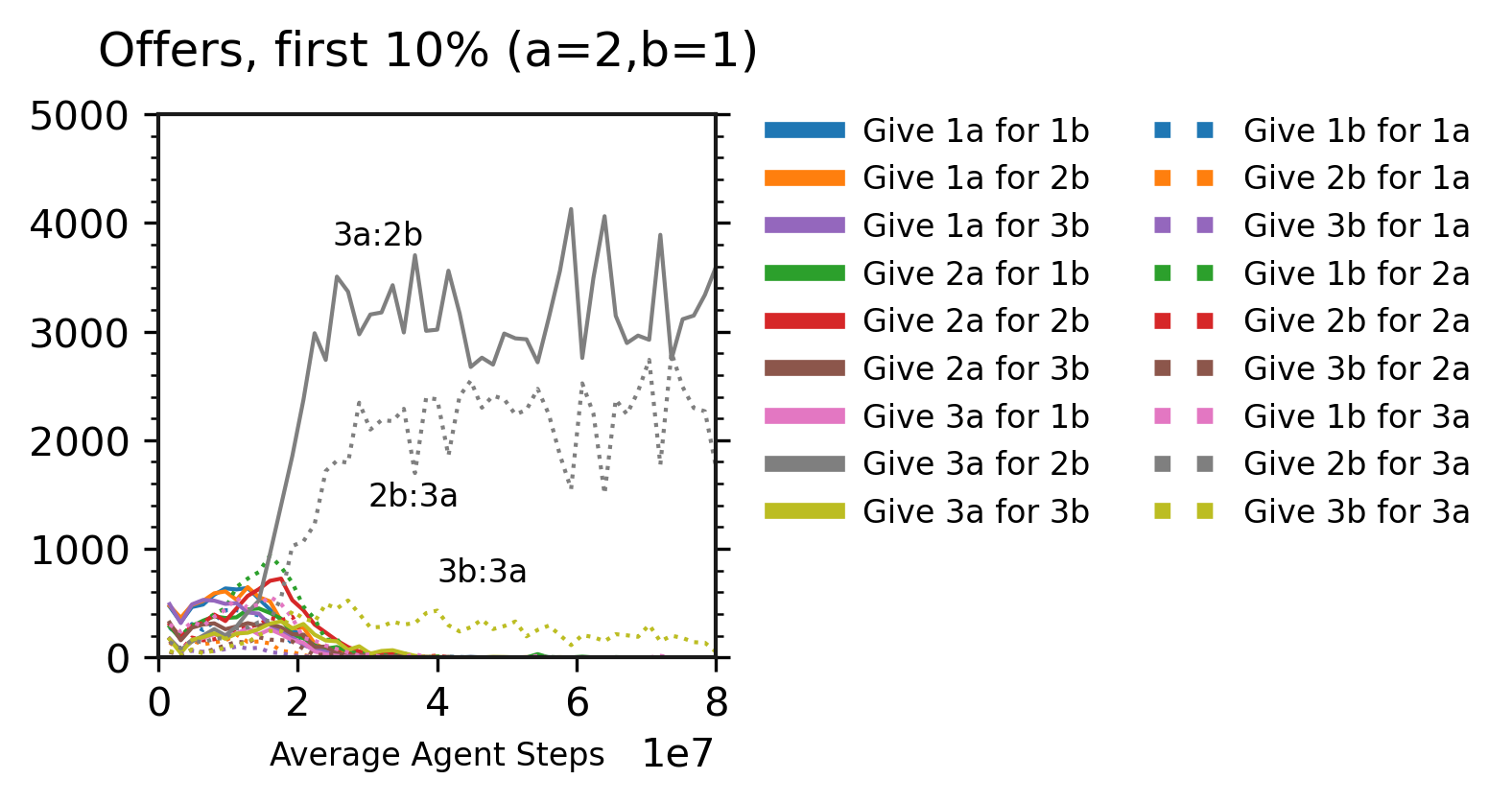}
        \caption{}
        \label{fig:baseline_offers_a2:offers_early}
    \end{subfigure}
    
    \begin{subfigure}{0.34\textwidth}
        \includegraphics[height=2in]{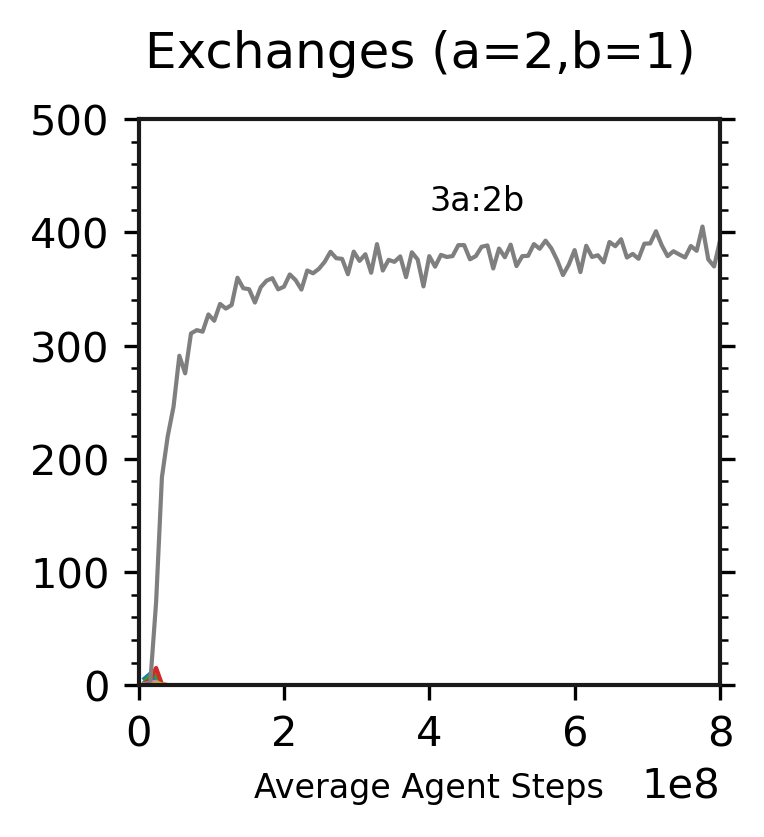}
        \caption{}
        \label{fig:baseline_offers_a2:exchaanges}
    \end{subfigure}%
    ~
    \begin{subfigure}{0.64\textwidth}
        \includegraphics[height=2in]{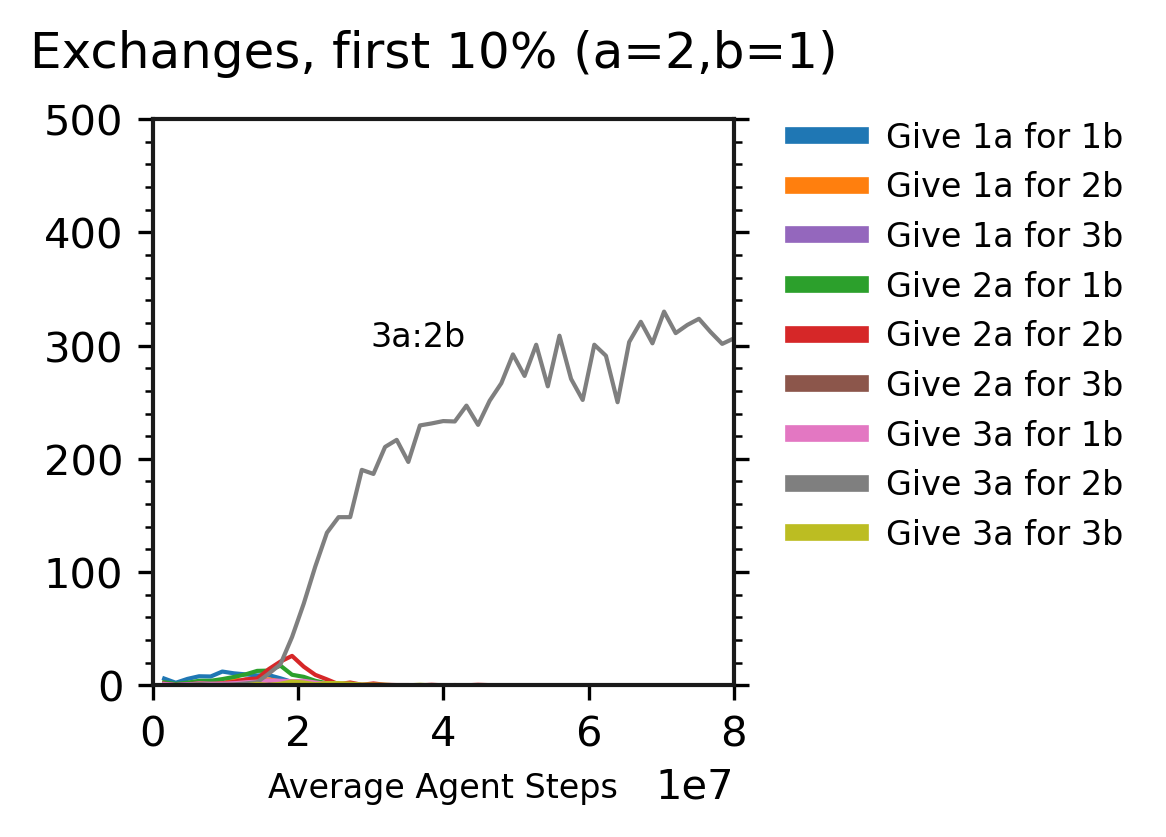}
        \caption{}
        \label{fig:baseline_offers_a2:exchanges_early}
    \end{subfigure}

    \caption{Average quantity of each offer and exchange in the $(a=2,b=1)$ setting. (b) and (d) zoom in on the first 10\% of the experiment.}
    \label{fig:baseline_offers_a2}
\end{figure}

We can now look deeper at the low-level offers that agents choose to make. Recall from Section~\ref{sec:environment} that agents use a set of 18 actions to propose exchanges, such as ``Give 3 apples for 1 banana'' (or 3a:1b), ``Give 2 bananas for 3 apples'' (or 2b:3a), and so on. In these next results, we will plot the total usage of each offer per episode, and the total number of exchanges using each offer per episode. This is a more detailed view into the trade behaviour than the average price of exchange results that we saw previously in Figure~\ref{fig:baseline_exchanges_price:prices}.

Figures~\ref{fig:baseline_offers_a1} and \ref{fig:baseline_offers_a2} investigate this for the $(a=1,b=1)$ and $(a=2,b=1)$ settings. Subfigures (a) and (b) plot the average quantity of offers made of each type per episode, summed across all players and timesteps in the episode, with (b) focusing on the first 10\% of training. To make the presentation of 18 offers as lines on one graph clearer, inverse offers such as ``Give 1 apple for 1 banana'' and ``Give 1 banana for 1 apple'' are assigned the same colour, with apple sales presented as a solid line and apple purchases as a dashed line. (c) and (d) present the quantity of exchanges of each type that occur, with (d) similarly showing the first 10\% of training. Since the number of ``Give 1 apple for 1 banana'' exchanges is exactly equal to the number of ``Give 1 banana for 1 apple'' exchanges, we only present the former.

Figure~\ref{fig:baseline_offers_a1:exchanges_early}, showing the $(a=1,b=1)$ exchanges, is particularly interesting. The population moves through four different offers with increasing frequency: 1a:1b, then 2a:2b, then 3a:2b, and then finally 3a:3b. The brief dominance of the ``Give 3a for 2b'' exchange explains the oscillation in price that we saw earlier in Figure~\ref{fig:baseline_exchanges_price:prices}. 

The population's movement between prices before converging to one is a promising sign for future experiments, as it suggests that agents are exploring and negotiating, instead of settling for the first offer that results in exchanges. We believe this is aided because the environment's mechanism for pairing offers into an exchange uses compatible (instead of inverse) offers and a preference for the most generous offers. If the population's exchanges are occurring at the 2a:2b ratio (the red line in Figure~\ref{fig:baseline_offers_a1:exchanges_early}), an agent selling apples can increase their offer to 3a:2b (the grey line) and still trade with the population -- in fact, their offers get accepted before other nearby 2a:2b offers. If the environment used only inverse offers, an agent switching to 3a:2b would have to wait for 2b:2a agents to notice and change their own offers to match, which could take a long time and thus disincentivize exploration in offers. If the environment used compatible offers but didn't prioritize higher offers by eliminating lower dominated offers, then the exploring agent could still trade with the rest of the population would see no advantage for offering 3a:2b. The ``compatible and non-dominated'' mechanism used here helps agents explore different offers, at the cost of injecting some domain knowledge; we will revisit this choice in Section~\ref{sec:ablation:trade:inverse}.

Figure~\ref{fig:baseline_offers_a2} shows the $(a=2,b=1)$ setting where apples are easier to acquire, and we see the population quickly adopt the 3a:2b and 2b:3a offers. Figure~\ref{fig:baseline_offers_a2:offers_early} shows some brief exploration of other offers before 3a:2b and 2b:3a emerge, and only 3b:3a continues being used before also dropping off. Here, the 3b:3a offer is more generous than the more frequently used 2b:3a offer, and is thus prioritized by the environment while still resulting in a 2b:3a exchange, since compatible offers exchange with the lowest quantities that satisfy each party. However, this drops off over the first 10\% of training, with all agents adopting the 3a:2b and 2b:3a offers.

\begin{figure}
    \centering
    \includegraphics[width=0.8\hsize]{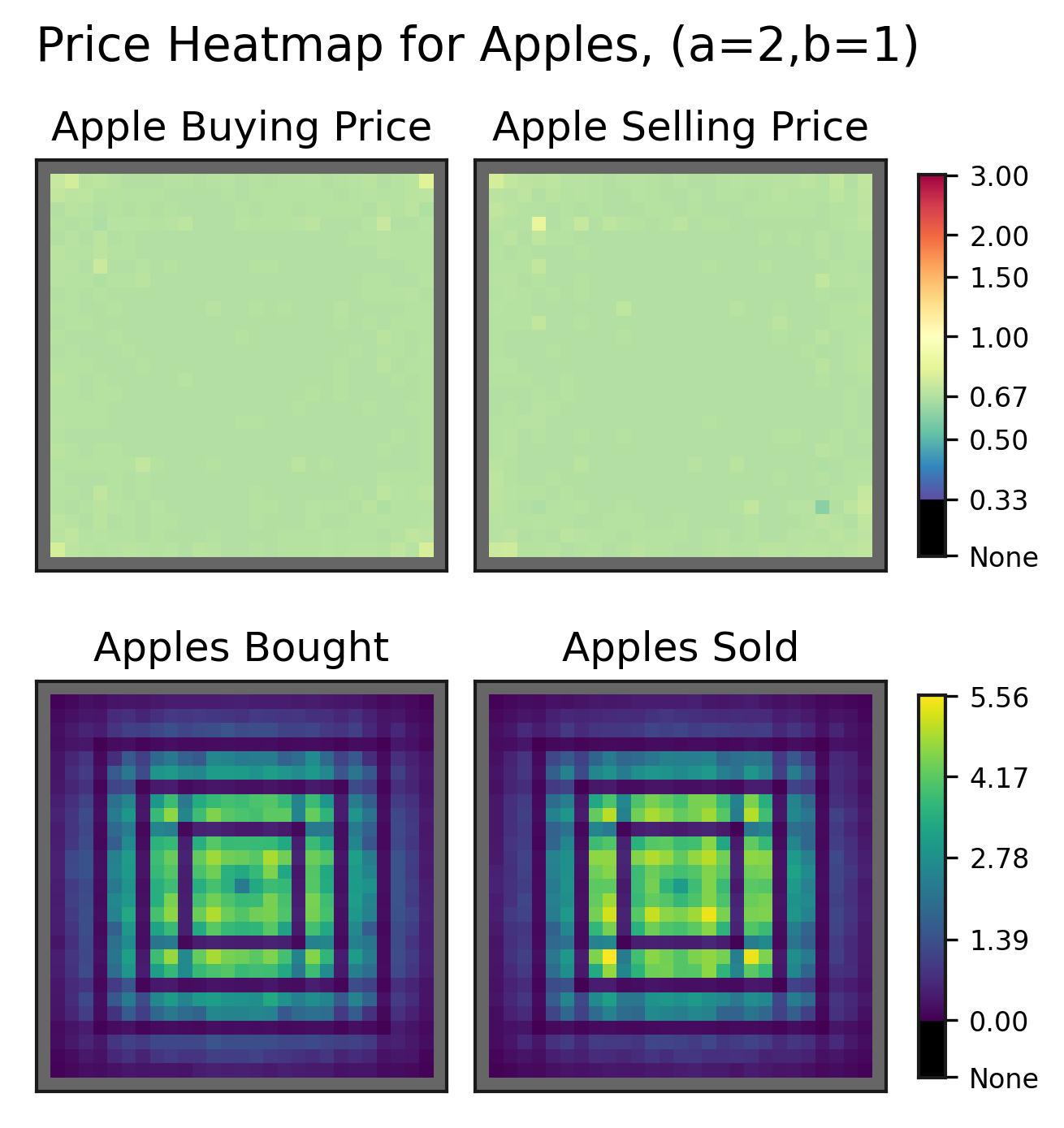}
    \caption{Price and frequency of exchanges over space, in the $(a=2,b=1)$ setting. The ``Apple Buying Price'' and ``Apple Selling Price'' plots show the average price (ratio of bananas to apples) in all exchanges over all episodes, when the player buying or selling apples was in each map location. The ``Apples Bought'' and ``Apples Sold'' heatmaps show the average quantity of apples bought or sold per episode from each map location. The 3a:2b exchange (with a price of 2/3 or 0.66) was approximately uniform across the map, and apples are more frequently bought and sold near the center.}
    \label{fig:baseline_pricemap_a2b1}
\end{figure}

We can also examine the population's trading behaviour spatially. Figure~\ref{fig:baseline_pricemap_a2b1} shows an example in the $(a=2,b=1)$ setting. The first two plots, ``Apple Buying Price'' and ``Apple Selling Price'', show the ratio of bananas per apple averaged over all exchanges that occur in each map location. Here, we see that the 3a:2b exchange (with a ratio of 2 bananas per 3 apples, or 0.66 bananas per apple) is used fairly uniformly across the map. The bottom two plots, ``Apples Bought'' and ``Apples Sold'', show the average quantity of apples bought and sold per episode from each map location. Agents buy and sell more apples from the center of the map than the edges, and the rings of water on the map show up as locations where agents are unlikely to trade goods from. We will return to this style of plot in Section~\ref{sec:experiments:regions} where we examine maps where apple and banana trees grow in different locations, and the agents converge to different prices in different parts of the map.

This concludes our initial set of experiments demonstrating how agents produce, consume, and trade resources, and how we can visualize that behaviour. In our baseline environment, the population quickly moves to an equilibrium where agents produce the goods their role is specialized in, and trade for the goods they gain more reward for consuming. By comparing the $(a=1,b=1)$ and $(a=2,b=1)$ settings, we have seen early evidence that the offers that agents converge to is affected by the relative abundance of goods in the environment. Next, we will present several targetted experiments, starting with supply and demand shifts.

\subsection{Supply and Demand Shifts}
\label{sec:experiments:sd}

\begin{figure}
    \centering
    \begin{subfigure}{\textwidth}
        \centering
        \includegraphics{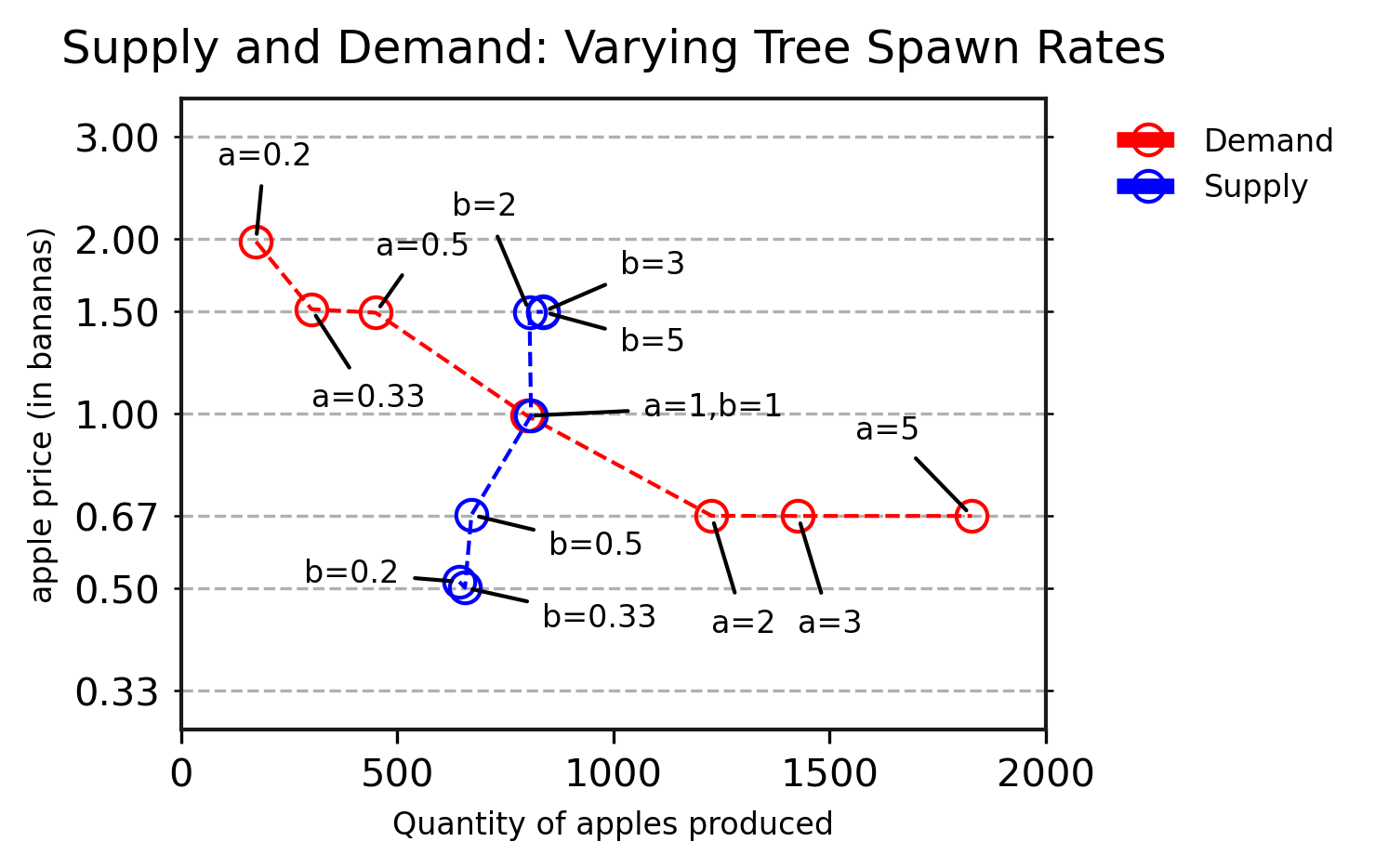}
        \caption{Supply and Demand, with apples produced on the x-axis.}
        \label{fig:sd-spawn-spawn:produced}
    \end{subfigure}
    
    \begin{subfigure}{\textwidth}
        \centering
        \includegraphics{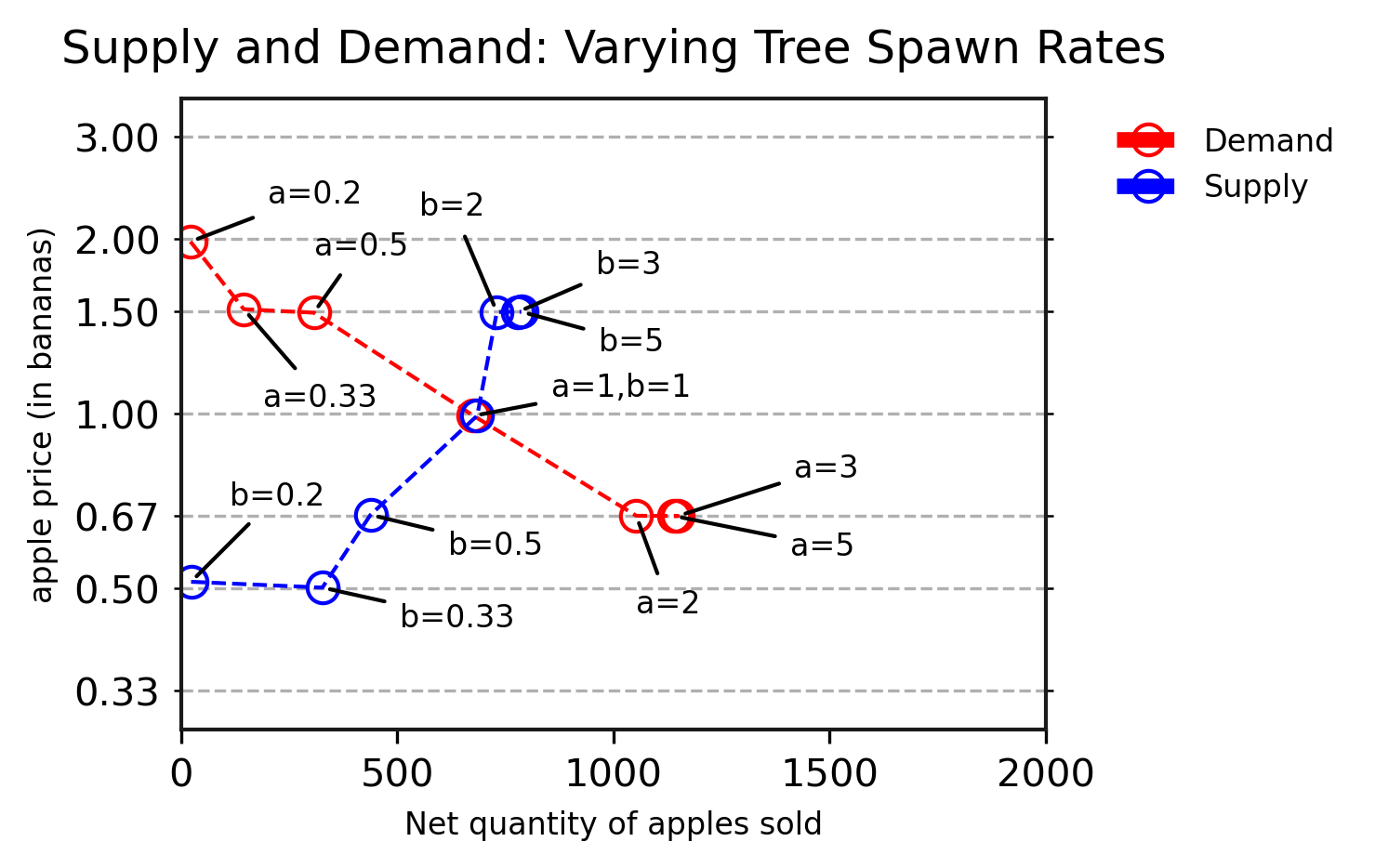}
        \caption{Supply and Demand, with net apples traded on the x-axis.}
        \label{fig:sd-spawn-spawn:traded}
    \end{subfigure}
    
    \caption{Supply and Demand curves for Apples as the spawn rate of apple trees and banana trees is varied. Each datapoint represents an experiment with an independent population of agents. The label indicates the modifier to the apple tree or banana tree spawn rate; a=0.5 indicates apple trees appearing with half of their default probability of 15\% per map tile. The y-axis measures the average
    ratio of bananas per apple, averaged over all exchanges in all episodes. In (a), the x-axis measures the average quantity of apples produced per episode. (b) shows the same experiments, but with the x-axis measuring net apples produced and then traded. Varying the spawn rate of apple trees shifts the supply curve, and reveals the shape of the demand curve. Varying the spawn rate of banana trees affects the ability of agents to buy apples and indirectly the equilibrium price of apples, shifting the demand curve and revealing the shape of the supply curve.}
    \label{fig:sd-spawn-spawn}
\end{figure}

As we described in Section~\ref{sec:background:microeconomics}, supply and demand curves are a useful model for predicting how environmental changes should affect an economy's equilibrium production, consumption, and prices. An equilibrium point in some experiment is the intersection of the supply and demand curves. By adjusting the environment to shift the supply curve, we find new equilibrium points that reveal the shape of the demand curve. Similarly, shifting the demand curve reveals the shape of the supply curve.

In this section, we will present a set of experiments to investigate whether our agents' learned behaviour can be well described by supply and demand curves. Specifically, we can see if the standard microeconomic predictions hold: if the supply of apples increases, does the equilibrium price for apples go down? Does a higher price incentivize more production and less consumption?

Recall from our earlier discussion that our experiments use comparative statics: an analysis of equilibrium behaviour from different populations trained under different conditions, without modelling how a population might move from one equilibrium to another. In each experiment, we will perform two parameter sweeps: one to affect the supply of goods, and the other to affect the demand for goods. By plotting each intersection point, the supply sweep will reveal the demand curve, and the demand sweep will reveal the supply curve. Finally, as we noted in Section~\ref{sec:background:microeconomics}, note that supply and demand graphs are is an abstraction used to understand and predict behaviour: there is no such thing as a ``supply curve'' or ``demand curve'' in the environment or the agents. We are not attempting to discover \textit{the} supply curve in the environment; instead, our goal is to examine our agents behaviour, to see if it contains a pattern usefully described as supply or demand curves.

Figure~\ref{fig:sd-spawn-spawn} presents our first supply and demand graph, produced by sweeping the probability of each type of tree appearing at the start of each episode, similar to our earlier $(a=2,b=1)$ results. The y-axis in each graph measures the average ratio of bananas to apples in exchanges across all episodes. The x-axis in Figure~\ref{fig:sd-spawn-spawn:produced} measures the quantity of apples produced per episode, averaged across all episodes. Figure~\ref{fig:sd-spawn-spawn:traded} presents the same set of experiments, but with the x-axis instead showing the quantity of apples produced \textit{and then traded}. This is an alternative view that excludes apples that an agent produces for their own consumption, which do not affect the price on the y-axis\footnote{Note that the opposite is not true: if apples are expensive to buy, agents might respond by producing more apples for their own consumption.}. 

\begin{figure}
    \centering
    \begin{subfigure}{0.3\textwidth}
    \centering
    \includegraphics[height=1.5in]{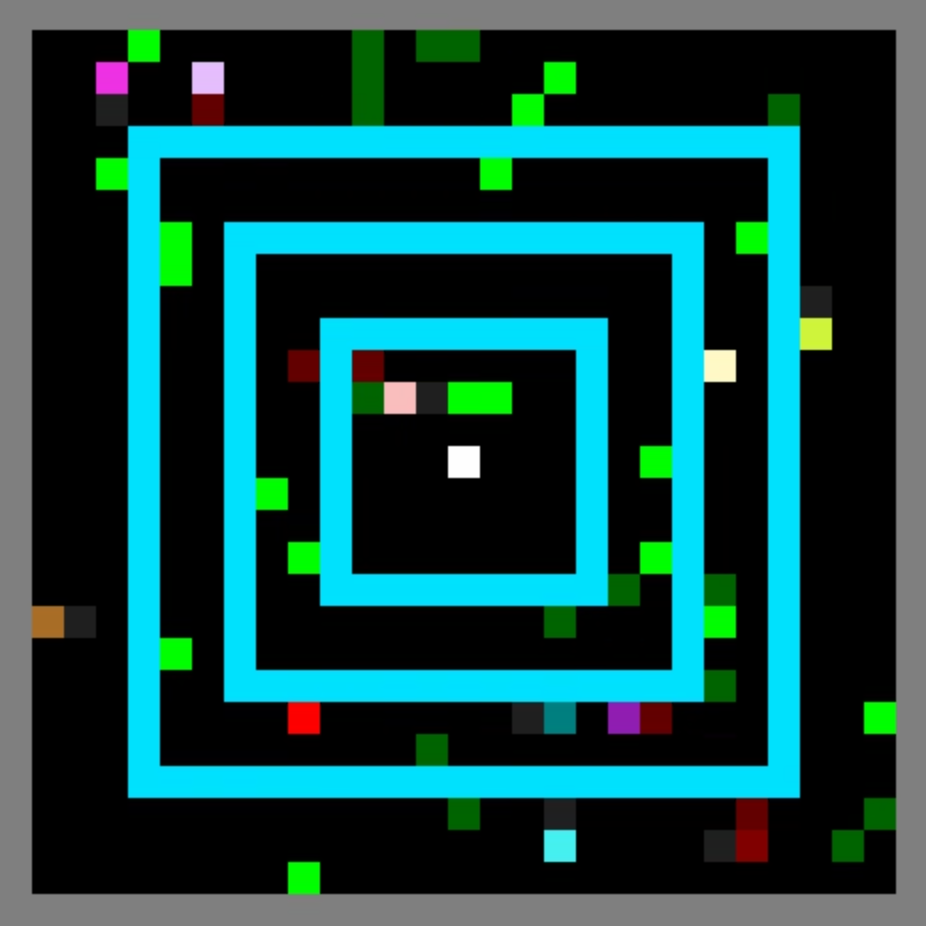}
    \caption{a=0.2}
    \label{fig:sd-spawn-spawn-map:a02}
    \end{subfigure}%
    ~
    \begin{subfigure}{0.3\textwidth}
    \centering
    \includegraphics[height=1.5in]{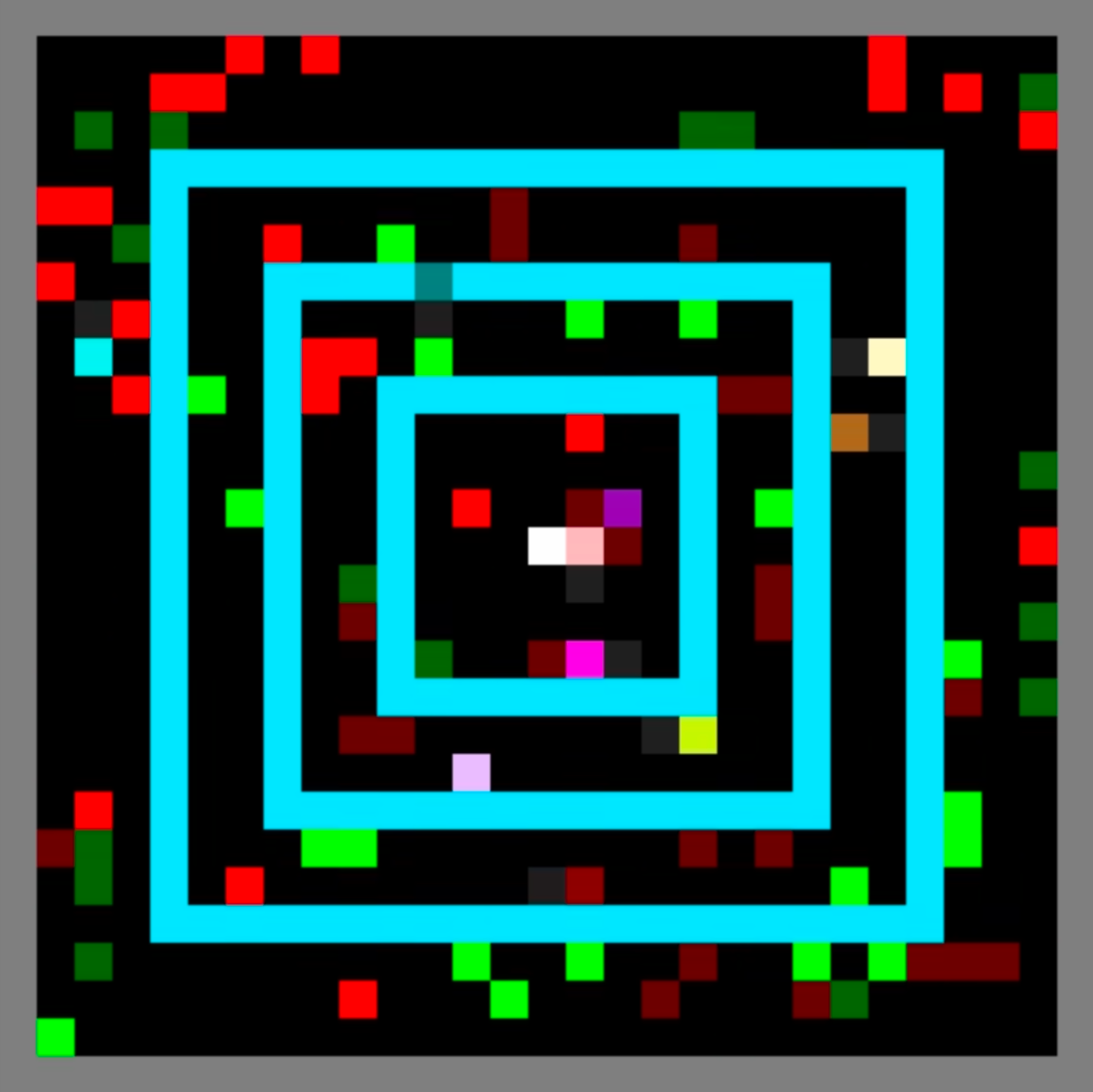}
    \caption{a=1}
    \label{fig:sd-spawn-spawn-map:a1}
    \end{subfigure}%
    ~
    \begin{subfigure}{0.3\textwidth}
    \centering
    \includegraphics[height=1.5in]{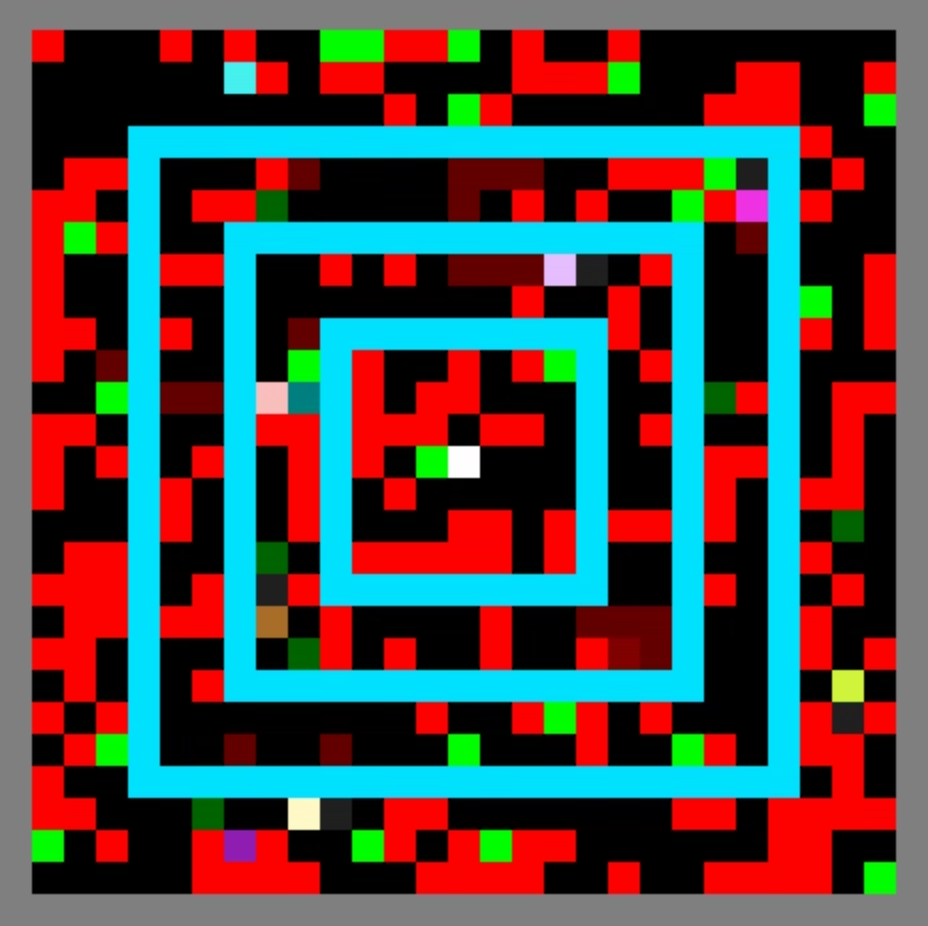}
    \caption{a=5}
    \label{fig:sd-spawn-spawn-map:a5}
    \end{subfigure}
    
    \caption{Example maps sampled from the $a=0.2$, $a=1$, and $a=5$ settings. The 0.2 and 5 settings should be viewed as extreme cases of sparsity and abundance.}
    \label{fig:sd-spawn-spawn-map}
\end{figure}

The $a=x$ datapoints are a supply sweep, which multiplies the default apple tree spawn rate of 15\% by a modifier from 0.2 to 5.0. This sweep directly affects the ability of agents to produce apples. The $b=x$ datapoints are a demand sweep, by similarly sweeping the spawn rate of banana trees in the same range. This does not directly affect how many apples can potentially be produced, but it does affect the ability of agents to produce bananas to pay for apples, which shifts the price for apples and may thus influence the production of apples, if agents learn to respond in that way. Note that the extreme multipliers of 0.2 and 5 in each sweep are drastic. Figure~\ref{fig:sd-spawn-spawn-map} presents example maps sampled from the $a=0.2$, $a=1$, and $a=5$ settings, demonstrating the difference between only a few apple trees existing on the map, versus apple trees filling most of the available space on the map.

We will now explore Figure~\ref{fig:sd-spawn-spawn:produced} in more detail. First, the supply and demand curves slope in the directions predicted by elementary microeconomics. The supply curve slopes (steeply) upwards, and the demand curve slopes gradually downwards. When prices are high, production is high and consumption is low; when prices are low, production is low and consumption is high. Note that while we show only apple production on the x-axis in these graphs, it is a strict upper bound on apple consumption (since apples must be produced in order to be consumed), and in these experiments the consumption quantities are only slightly lower than the displayed production quantities (and not exactly equal because at episode end some fruit have not been consumed).

The supply intervention (which reveals the demand curve) has a direct effect on the quantity of apples produced: when there are more apple trees to harvest, it is not at all surprising that agents produce more apples. Even if we removed the Offer actions to make trading impossible, we would still expect to see more apples produced when apple trees are more plentiful. What is not as obvious, however, is the offers and prices that the agents converge to. When apple trees spawn less often than banana trees, the agents learn to make offers that value apples at a higher price; likewise, when apple trees are plentiful, the average price of an apple drops below 1. This is the population-level result of the learned behaviours of agents attempting to maximize their long-term individual rewards, and is not at all a required outcome of the experiment. The agents could potentially have not learned to trade at all, or could have converged to the 1.0 price in all datapoints, changing only the quantity of apples produced. In fact, the environmental constants we use---8 reward for the preferred fruit and 1 reward for the efficiently produced fruit, and a range of exchanges from 1:3 to 3:1---are such that producing and then trading at \textit{any} available price is more rewarding for an agent than producing and eating their own fruit. Thus, the intuitive correlation in these results between scarcity and price is an emergent effect produced by the agents' learning dynamics, and not an outcome forced by the environment or the experiment.

The demand intervention (revealing the supply curve) has a less direct effect on apple production and prices, as the number of apple trees and the difficulty to produce apples is unchanged. Varying the availability of bananas affects the agents' ability to trade for apples; the agents respond by adjusting their prices, and higher or lower apple prices then incentivize the production of more or fewer apples. When bananas are rare ($b=0.2$, $b=0.33$, $b=0.5$), the agents converge to a low price for apples, and fewer apples are produced. When bananas are plentiful ($b=2$, $b=3$, $b=5$), the agents converge to a higher price for apples, and slightly more apples are produced in the $b=3$ and $b=5$ cases. The slope of the supply curve is quite steep; as we noted in Section~\ref{sec:environment:opportunity_costs}, there are only minor incentives (the movement and water penalties) for agents to \textit{not} produce as many apples as possible, so a high price can only have a limited effect.

By comparison, Figure~\ref{fig:sd-spawn-spawn:traded} also helps us understand the very steep supply curve slope in Figure~\ref{fig:sd-spawn-spawn:produced}, by changing the x-axis to only measure apples produced and then traded, ignoring apples produced and then eaten by the same agent\footnote{Specifically, the x-axis measures $\sum_p (s_p - b_p)^+$: the sum over all players $p$ of the difference between their apples sold $s_p$ and bought $b_p$, floored at zero. Each apple an agent sells must have either been produced or bought by that agent, and so their net sales ignores apples that they consumed or did not use. Summing the positive net sales across agents thus counts each produced apple as being sold at most once, even if it is bought and sold several times by different agents between production and consumption.}. At extreme conditions where apples are cheap, such as $b=0.2$ or $a=5$, agents may learn to produce some or all of their apples for their own consumption, instead of for sale. Comparing the two plots, note the $b=0.2$ datapoint shifting far left to indicate almost zero apples are sold, and the a=5 datapoint shifting left to indicate that about one third of apples are consumed instead of sold. These apples that are consumed by the producing agent cannot affect the price of apples, and the resulting supply curve has a much more gradual slope, suggesting that higher prices can incentivize greater production of apples \textit{for sale} instead of for consumption.

\begin{figure}
    \centering
    \begin{subfigure}{\textwidth}
        \centering
        \includegraphics{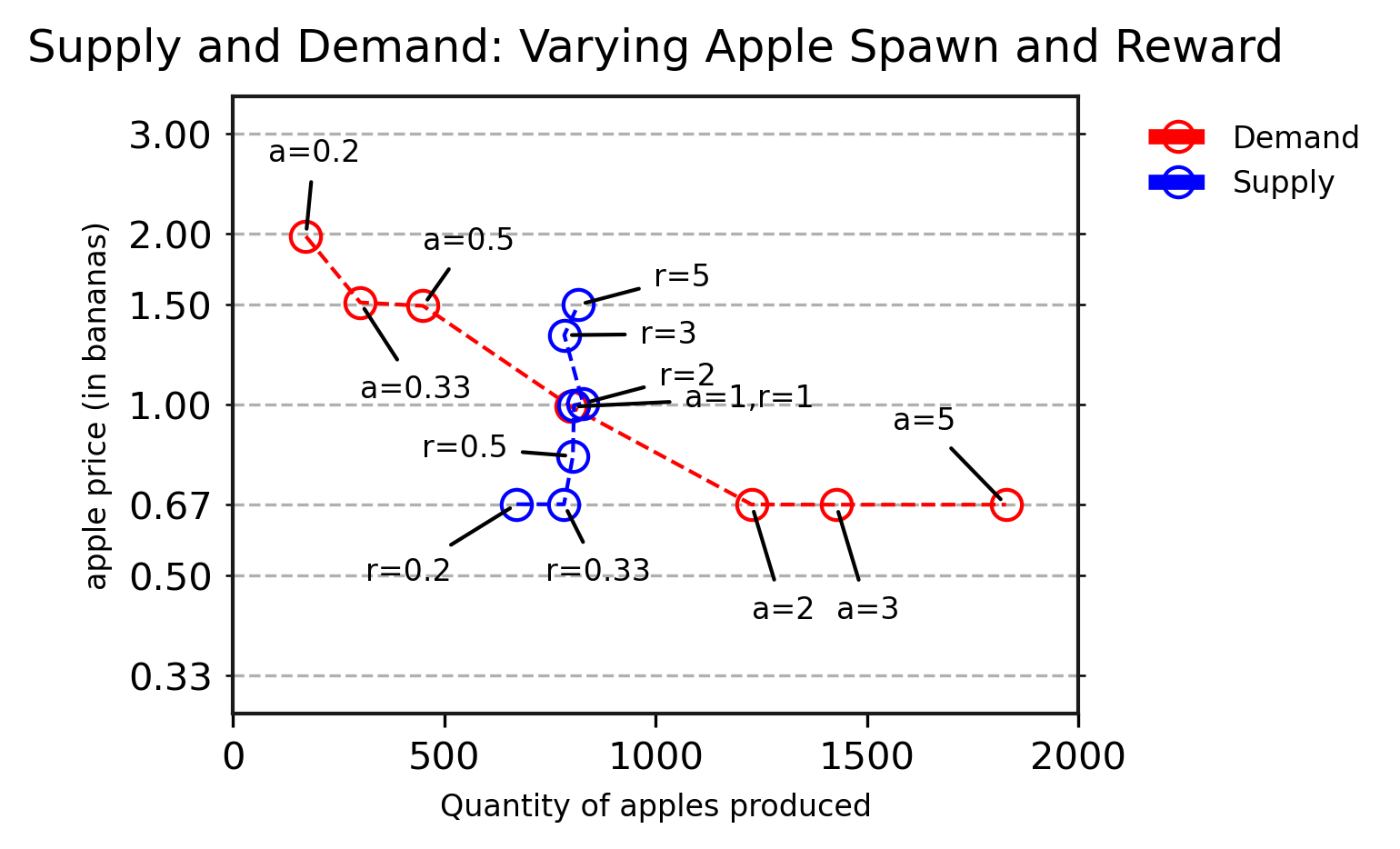}
        \caption{}
        \label{fig:sd-spawn-reward:all:produced}
    \end{subfigure}
    
    \begin{subfigure}{\textwidth}
        \centering
        \includegraphics{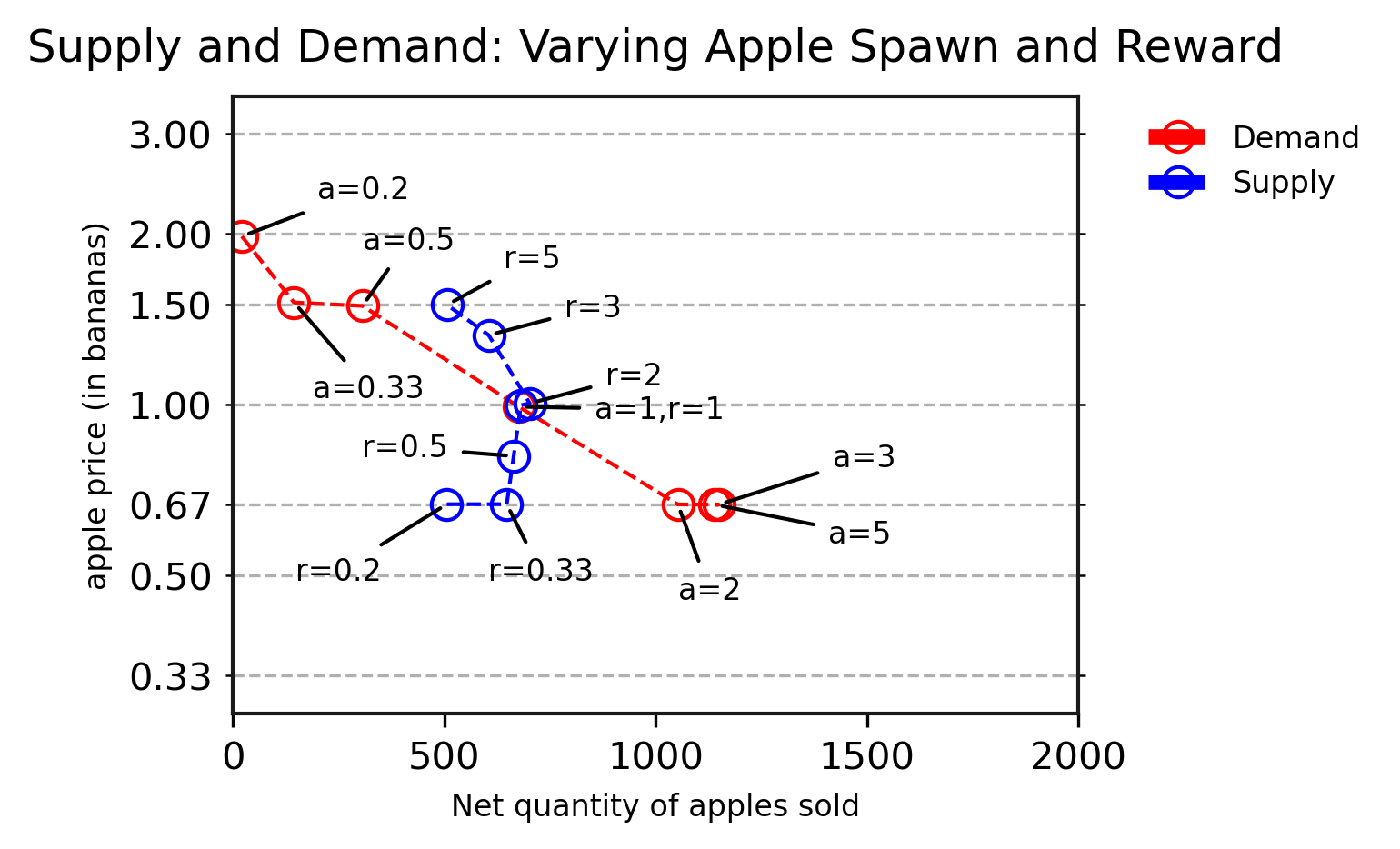}
        \caption{}
        \label{fig:sd-spawn-reward:all:traded}
    \end{subfigure}
    \caption{Supply and Demand curves for Apples, as the spawn rate of apple trees and the reward granted by apples to all players is varied. The datapoint labels `r=x' indicate a multiplier to the normal reward of given for apple consumption. In `r=5', Apple Farmers would gain 5 reward instead of 1 for eating an apple, while Banana Farmers would gain 40 reward instead of 8. The reward for bananas is left at the default values for all agents. (a) shows production on the x-axis, while (b) shows net apples traded.}
    \label{fig:sd-spawn-reward:all}
\end{figure}

\begin{figure}
    \centering
    \begin{subfigure}{\textwidth}
        \centering
        \includegraphics{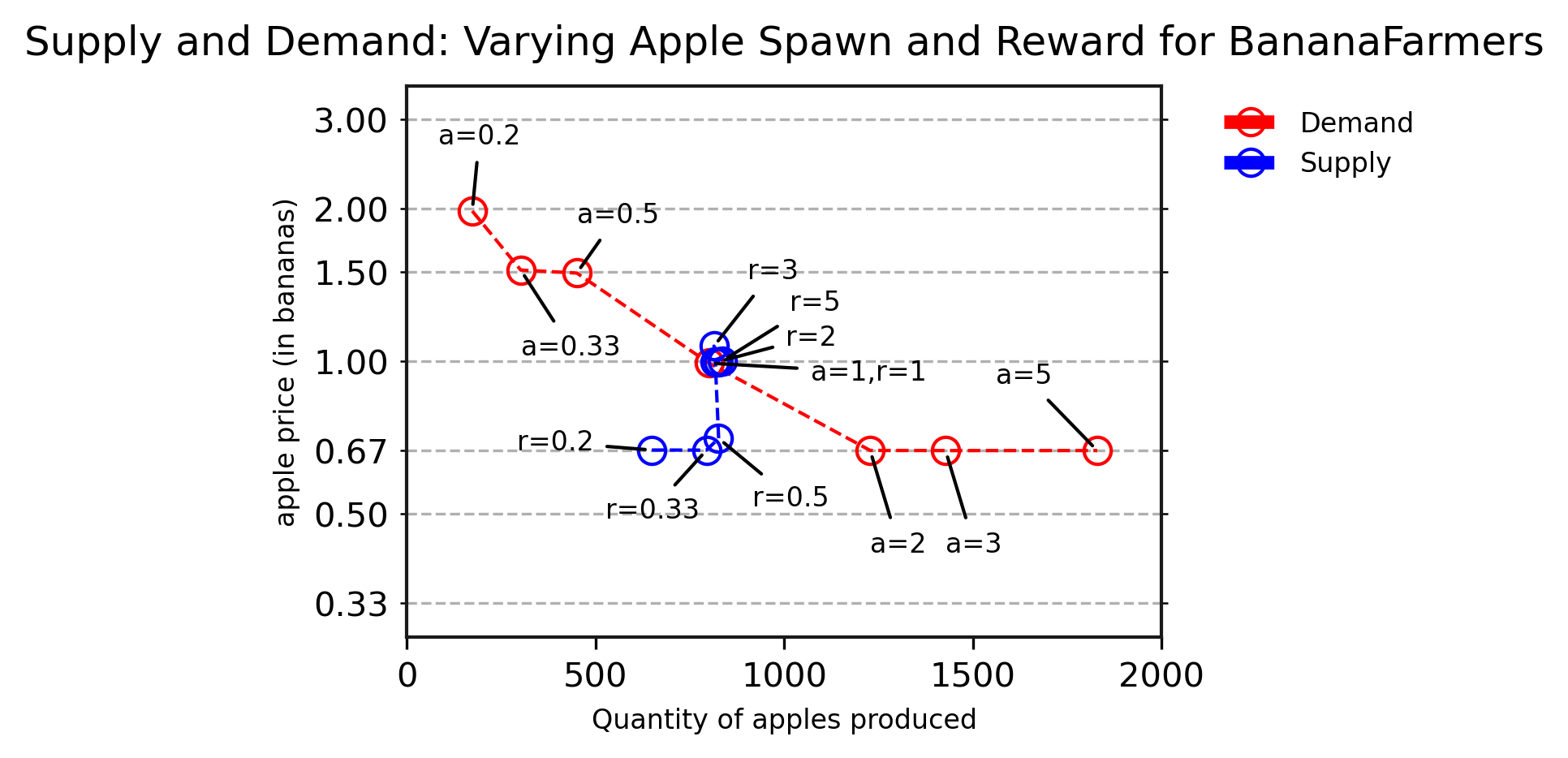}
        \caption{}
        \label{fig:sd-spawn-reward:bfs:produced}
    \end{subfigure}
    
    \begin{subfigure}{\textwidth}
        \centering
        \includegraphics{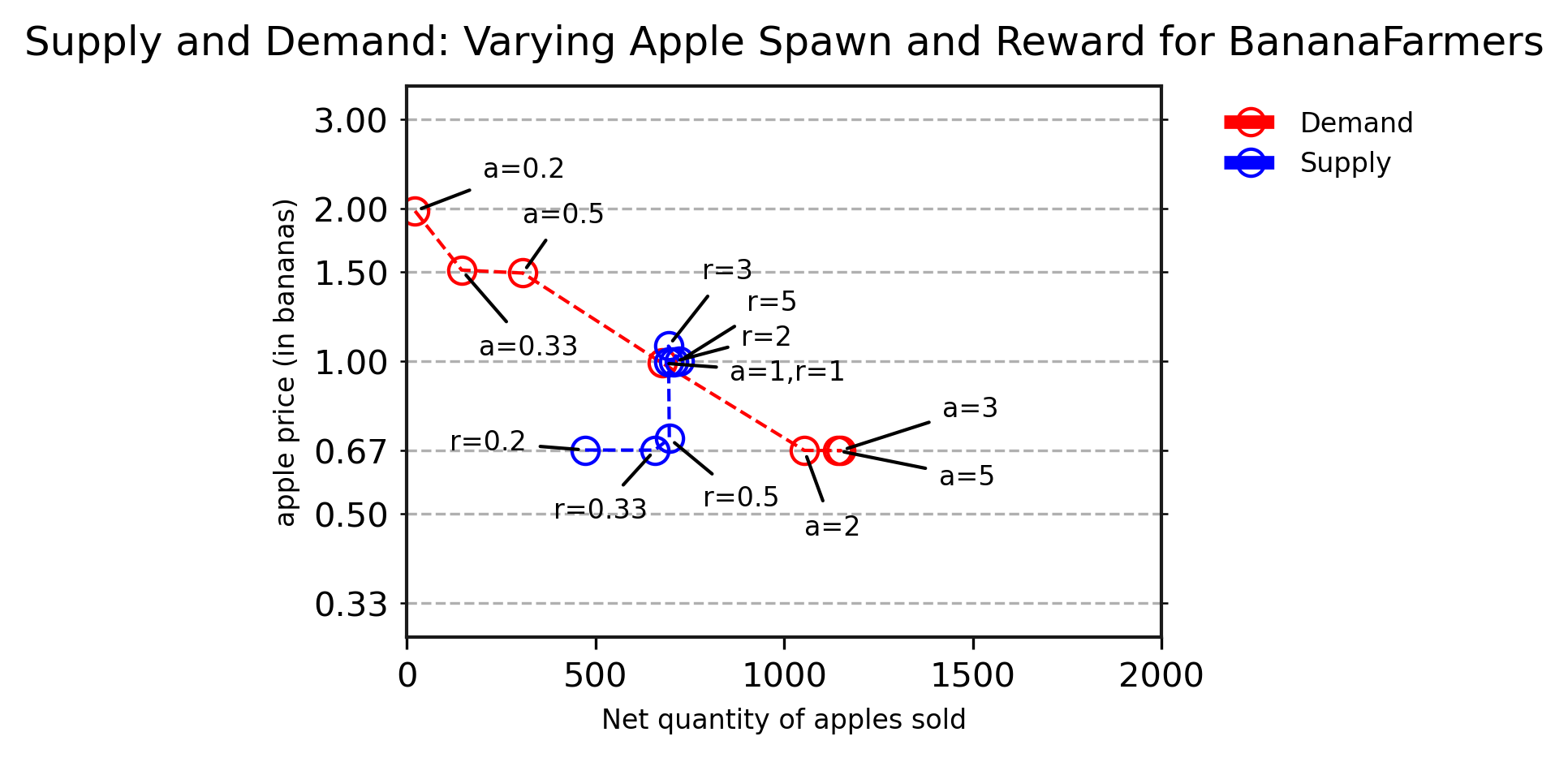}
        \caption{}
        \label{fig:sd-spawn-reward:bfs:traded}
    \end{subfigure}
    \caption{Supply and Demand curves for Apples, as the spawn rate of apple trees and the reward granted by apples to Banana Farmers is varied. The datapoint labels `r=x' indicate a multiplier to the normal reward given for apple consumption. In `r=5', Apple Farmers would still gain 1 reward for eating an apple, while Banana Farmers would gain 40 instead of 8. The reward for bananas is left at the default values for all agents. (a) shows production on the x-axis, while (b) shows net apples traded.}
    \label{fig:sd-spawn-reward:bfs}
\end{figure}

Varying the spawn rates of trees is just one option for affecting supply and demand. In Figures~\ref{fig:sd-spawn-reward:all} and~\ref{fig:sd-spawn-reward:bfs}, we change the demand sweep (to reveal a supply curve) by instead sweeping a direct notion of demand: the reward for consuming apples. Figure~\ref{fig:sd-spawn-reward:all} changes the apple consumption reward for all agents (both Apple Farmers and Banana Farmers), while Figure~\ref{fig:sd-spawn-reward:bfs} only modifies the reward for Banana Farmers, who already prefer to consume apples. In both cases, the datapoint label $r=x$ is a multiplier to the apple consumption reward in the range $[0.2, 0.33, 0.5, 1.0, 2.0, 3.0, 5.0]$. In both graphs, the supply sweep (revealing the demand curve) is identical to the previous experiment, and the same datapoints are shown. The reward for consuming bananas is unchanged in these experiments.

These experiments are an example of an environmental change that does not match our microeconomic intuitions about how agents should respond. In Figure~\ref{fig:sd-spawn-reward:all}, we see that modifying the apple reward for all agents does have a consistent effect on the apple price, but virtually no effect on production: high apple prices and high apple rewards at $r=5$ do not result in significantly more apple production than in $r=0.33$. This is somewhat surprising: at $r=5$ an Apple Farmer would obtain 5 reward for consuming an apple or 8 reward for consuming a banana, making an even 1a:1b exchange profitable, aside from the overhead of trading. If the agents had adopted the 1a:3b price, Apple Farmers could trade to earn 24 reward of bananas instead of 5 of apples, and Banana Farmers could earn 40 reward of apples instead of 3 of bananas: clearly favourable to both sides. However, Figure~\ref{fig:sd-spawn-reward:all:traded} shows that at $r=5$, many of the apples produced are consumed by the same agent instead of sold, producing a bent supply curve.

To eliminate the temptation for Apple Farmers to eat their own apples, Figure~\ref{fig:sd-spawn-reward:bfs} explores only affecting the Banana Farmers' reward for apple consumption. The results here are equally surprising in the agents' lack of adaptation. High apple rewards do not affect apple price or production, and low apple rewards of $r=0.5$ or $r=0.33$ reduce the price but do not affect production. We had predicted that Banana Farmers would compete with each other: offering 2b:1a instead of 1b:1a would value apples more highly, but also be prioritized before any nearby 1b:1a offers. However, this does not appear to have happened.

Overall, the agent behaviour in all of three supply and demand graphs is a partial success. Varying the spawn rate of apple and banana trees produced agent behaviour that was interpretable as supply and demand curves, albeit with a quite steep supply curve. Sweeping demand by affecting the agent behaviour did not produce behaviour suggestive of a supply curve. 

\subsection{Marketplaces}
\label{sec:experiments:marketplace}

In our Supply and Demand graphs, our goal was to discover the relationship between production, consumption, and price. One interpretation is as a counterfactual: \textit{if} the price of apples was 0.5, how many apples would the population choose to produce (on the supply curve) and consume (on the demand curve)? In the previous section's experiments, we used only indirect ways to create this scenario: we could vary the environmental conditions through tree spawn rates and fruit rewards, and then see what combination of price and production we arrived at.

Our simulated environment gives us other options for investigating this counterfactual. In this section we will explore influencing (but not \textit{setting}) the population's price through an environmental feature we call a \textbf{marketplace}. A marketplace is an entity in the environment, placed at a specific map location, which agents observe as a white square. Marketplaces make offers, and agents can trade with them using exactly the same actions and mechanics that they trade with each other. The environment uses the same mechanism for resolving compatible offers into an exchange, regardless of if one participant is a marketplace or an agent; both parties have to be within the usual trade radius of each other, and high offers are given priority over low offers.

Each marketplace constantly makes one or more fixed offers that we can configure for each experiment. For example, we could place a marketplace in the middle of the map, and set it to constantly make the offer ``Give 1 banana for 3 apples''. This marketplace could then be a potentially infinite source and infinite sink of fruit\footnote{Note that we could easily implement other kinds of marketplaces that are not infinite sources and sinks. For example, a marketplace could start each episode with a small inventory, and make offers inverse to its current contents. However, the marketplaces we use here are useful for strongly influencing the population's price.}, depending on how agents chose to trade with it; at this low valuation of apples, our agents would likely prefer to trade with each other instead, using the 1a:1b and 1b:1a offers. However, if we instead configured the marketplace to offer both the ``Give 1 banana for 3 apples'' and ``Give 3 apples for 1 banana'' offers, Banana Farmers would likely strongly prefer the marketplace, as it would offer the highest possible price, always be in the same location, and always have an offer available to trade with. Apple Farmers would then have to adopt the ``Give 3 apples for 1 banana'' offer in order to make any trades with the marketplace or with other agents.

Marketplaces thus give us a powerful way to influence the price that the population will converge to, without restricting the offer actions they can use or overriding their decisions. Agents can still choose to trade with each other, and likely will do so when the marketplace is out of their trade radius, but the availability of the marketplace's price \textit{somewhere} on the map might still affect their pricing behaviour everywhere.

\begin{figure}
    \centering
    \begin{subfigure}{\textwidth}
        \centering
        \includegraphics{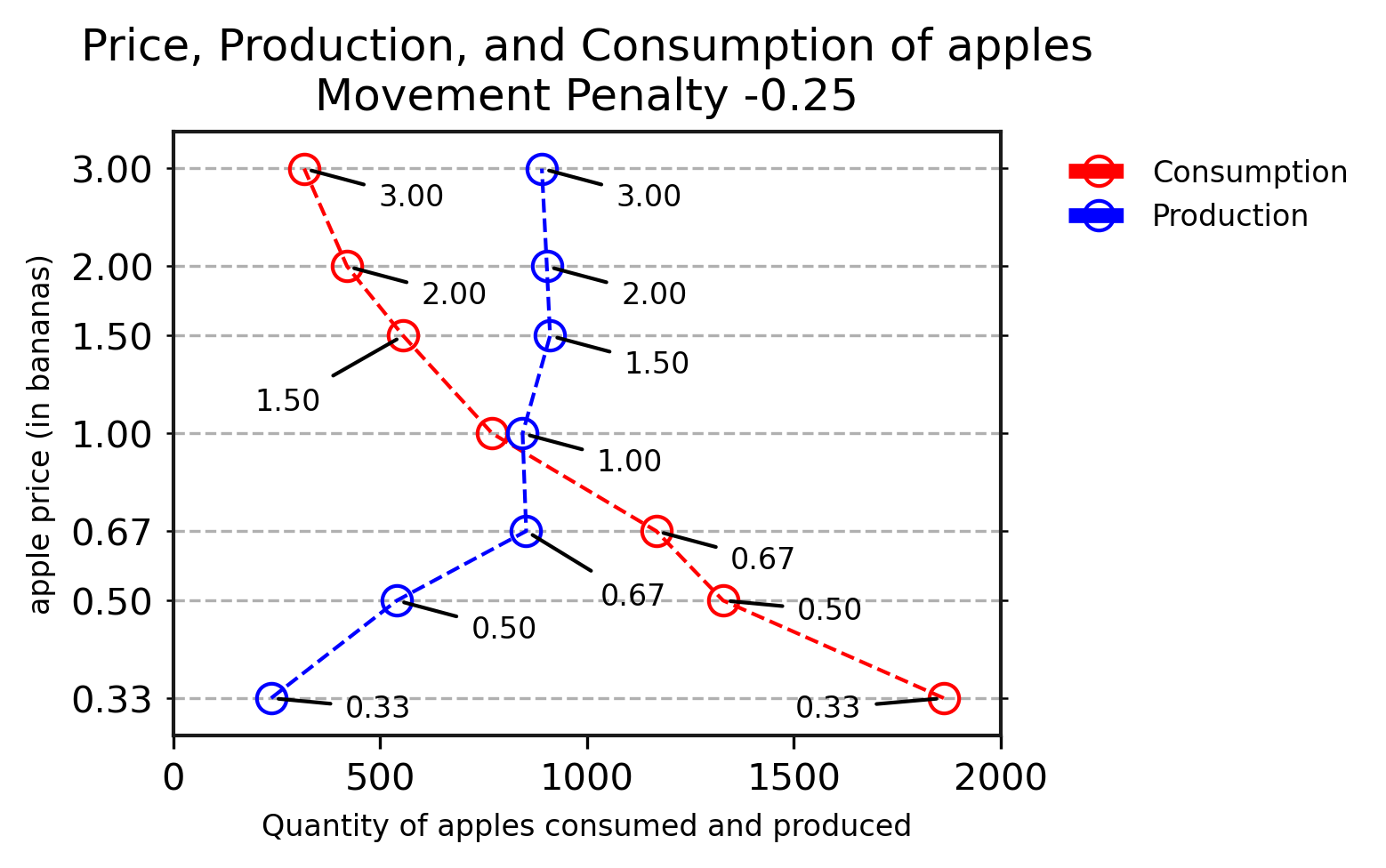}
        \caption{}
        \label{fig:marketplace:m025}
    \end{subfigure}
    
    \begin{subfigure}{\textwidth}
        \centering
        \includegraphics{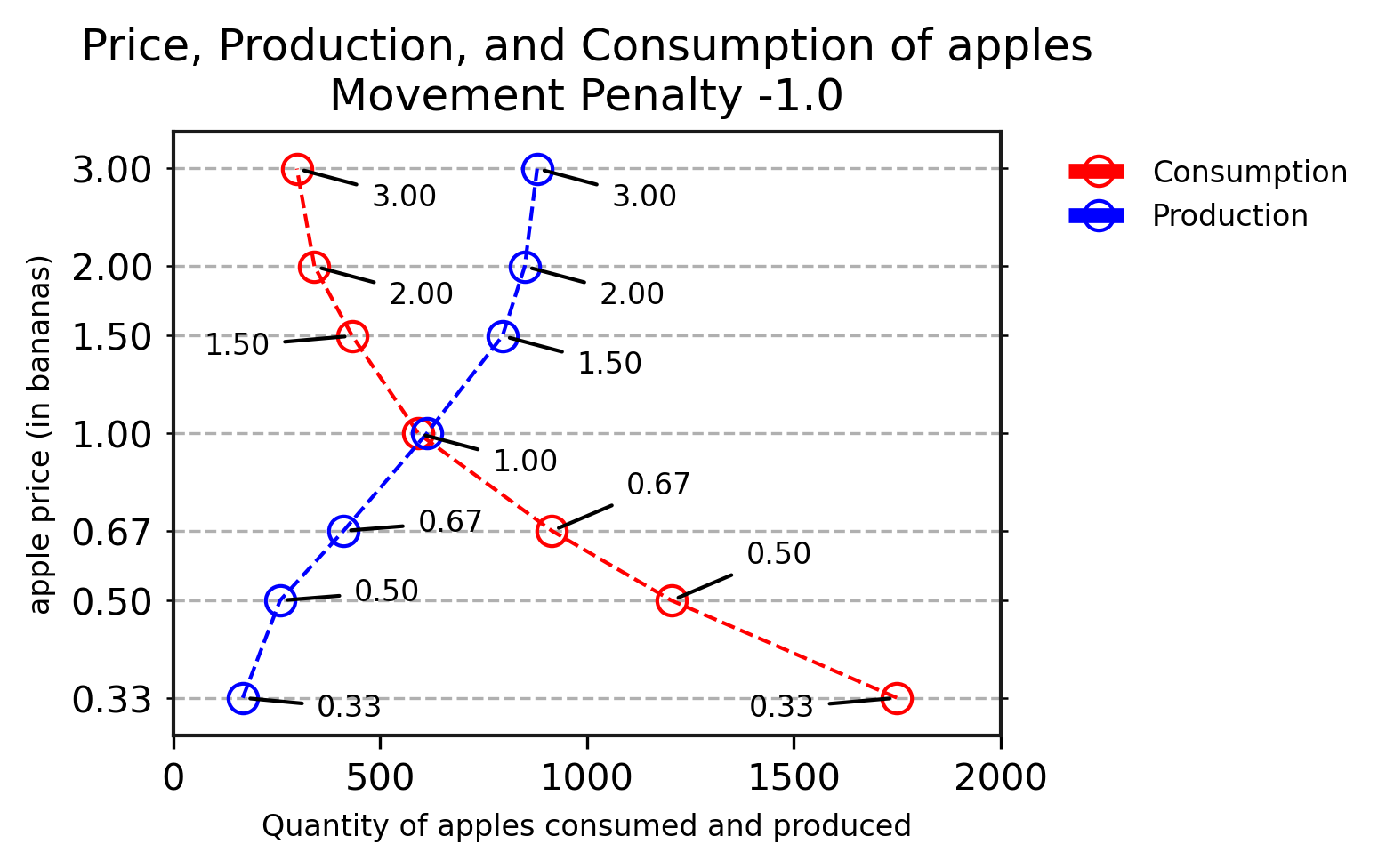}
        \caption{}
        \label{fig:marketplace:m1}
    \end{subfigure}
    \caption{Price, production, and consumption of apples with a marketplace. The marketplace's price is indicated by the datapoint labels, and each pair of datapoints that share a label are results from the same experiment. (a) shows the agents' behaviour when the environment's movement penalty is $-0.25$ per tile, which is our default throughout all other results in this paper. (b) changes the movement penalty to $-1.0$ per tile which makes producing goods more costly, and ``laziness'' a better alternative when prices are low.}
    \label{fig:marketplace}
\end{figure}

In this section, we will present a similar experiment to our earlier Supply and Demand experiments, by using a single marketplace to influence the price and then measure the populations' production and consumption behaviour. This experiment uses the same map as in our earlier results, except with a single marketplace at the center. The marketplace constantly makes two offers on each side of one price, such as simultaneously making both the ``Give 3 apples for 1 banana'' and ``Give 1 banana for 3 apples'' offers. We can then perform a set of independent experiments where we sweep the marketplace price over seven values, from 3a:1b (cheap apples) to 1a:3b (expensive apples), and measure how many apples agents choose to produce and consume in response.

Figure~\ref{fig:marketplace} presents two such sets of experiments. Figure~\ref{fig:marketplace:m025} presents a sweep where the agents' movement penalty is at its default value of -0.25, the same value used in our earlier Supply and Demand experiments. Each of the seven experiments in this sweep set the marketplace price to one ratio, and we extracted both a ``Consumption'' and ``Production'' datapoint. Each datapoint is labelled with the marketplace's price for that experiment, which may appear redundant in this first figure, as each datapoint sits exactly on that price on the y-axis. This illustrates the influence that the marketplace has over the population's price: an extreme marketplace price of 3-to-1 is good for one of the roles, and the other role has to follow in order to trade.

The Production and Consumption datapoints only show the apples produced or consumed \textit{by agents}: any apples bought or sold by the marketplace do not count towards these figures. The resulting curves are similar to Supply and Demand curves, showing how many apples would be produced or consumed if the price was at a certain value. There is a caveat: more apples could be consumed than the environment could normally produce, if most of the apples come from the marketplace. Similar to our initial Supply and Demand graph in Figure~\ref{fig:sd-spawn-spawn}, the curves slope in the direction predicted by microeconomics: high prices cause more production and less consumption, and low prices are the opposite. The Production curve is also steep at marketplace prices from $0.67$ and above, but shows a more gradual slope at $0.33$ and $0.5$. Unlike our earlier Supply and Demand graphs, the Consumption and Production curves are smoother, and reach the full range of available prices from $0.33$ to $3.0$.

Figure~\ref{fig:marketplace:m1} explores a possible reason why the Production and Supply curves in our earlier results were steep or vertical. Here, the agents' movement penalty is changed from its default value of $-0.25$ per timestep to $-1.0$ per timestep, making movement -- and thus production -- very costly. This gives a more meaningful opportunity cost to labour. Apples near the marketplace are worth harvesting even at low prices, because Apple Farmers still need to eat for reward and to avoid the hunger penalty. However, once the convenient apples are exhausted, the marginal costs of production increase, as agents need to move farther and cross more water, and only a high price for apples makes that labour worthwhile. 

With the higher movement penalty, we see smooth Consumption and Production curves, with Production's slope not nearly as steep as in our earlier results. This demonstrates that each price increase can incentivize more production. However, this high movement penalty does have a disadvantage, which is why we did not use it in our earlier results. With the marketplace, agents have the benefit of a guaranteed trading partner. Moving is expensive, but also not risky, and agents quickly learn the trading behaviour shown in Figure~\ref{fig:marketplace:m1}. When we train a population with the movement penalty of $-1.0$ but \textit{without} a marketplace, very little trading occurs and the price of those few exchanges is near 1.0 regardless of the spawn rate of apples and bananas. Thus, our agents struggle to learn how trade at all with that movement penalty, unless the marketplace is present to help it develop. We will revisit the effect of the movement penalty in Section~\ref{sec:ablation:movement}.

Overall, using the marketplace to strongly influence the price gives us a convenient tool for measuring how our agents' production and consumption behaviour varies with price. As compared to the earlier interventions we took for Supply and Demand graphs, such as sweeping tree spawn rates and fruit rewards, influencing the price with a marketplace feels less drastic and gives more interpretable results. However, as the movement penalty $-1.0$ results show, it also serves as a curriculum\footnote{Early in our research when we used the A2C agent architecture instead of our current V-MPO architecture, we considered introducing a marketplace that only made pairs of poor offers, such as ``Give 1 apple for 3 bananas'' and ``Give 1 banana for 3 apples'', as a ``market of last resort''. Agents would then be able to learn to trade with the marketplace as a curriculum for learning how to make offers, and then discover that trading with each other was more profitable for both parties, and thus stop using the marketplace. Fortunately, the V-MPO agents consistently discover trade on their own without needing this curriculum.} to help agents learn how to trade with the marketplace and with each other.

\subsubsection{Moving the Marketplace}
\label{sec:experiments:marketplace:moving}

Adding a marketplace to the map gives us a powerful way to influence the prices that the agents converge to. For example, in Figure~\ref{fig:marketplace}, every experiment resulted in the agents converging to exactly the same price that the marketplace provided. This is not surprising, since the marketplace was positioned in the center of the map and always had resources available to trade. But now it is interesting to ask: just how much power \textit{does} the marketplace have over the agents' price? How often do agents trade with the marketplace instead of with each other, and how inconvenient would the marketplace's location have to be for its influence to wane?

\begin{figure}
    \centering
    \includegraphics[width=2in]{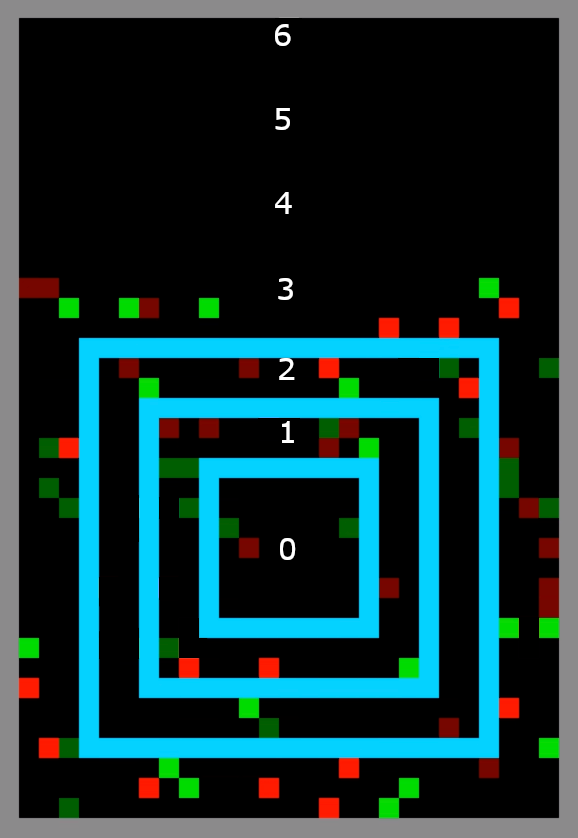}
    \caption{Seven increasingly inconvenient locations where the marketplace can be located.}
    \label{fig:moving_market_locations}
\end{figure}

Figure~\ref{fig:moving_market_locations} presents an example map, and seven increasingly inconvenient locations for the marketplace. Location 0 is its initial position at the center of the map. Locations 1 through 3 move it across the rings of water, but still in the area where trees grow. Locations 4 to 6 move it off of the normal map, into a barren area where no trees grow. At each step, agents have to incur more movement and water costs to reach the marketplace, and also have a larger opportunity cost, as the travel time spent reaching the market is time not spent producing, trading, or consuming.

\begin{figure}
    \centering
    \begin{subfigure}{0.4\textwidth}
    \centering
    \includegraphics[height=1.9in]{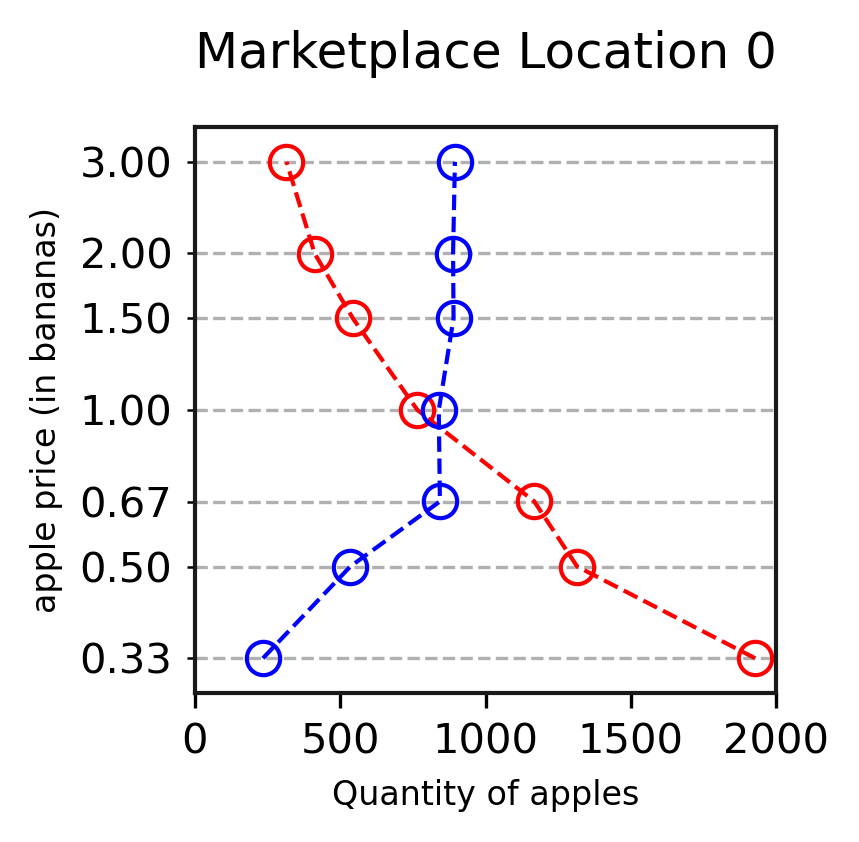}
    \caption{}
    \label{fig:moving_market:loc0}
    \end{subfigure}%
    ~
    \begin{subfigure}{0.59\textwidth}
    \centering
    \includegraphics[height=1.9in]{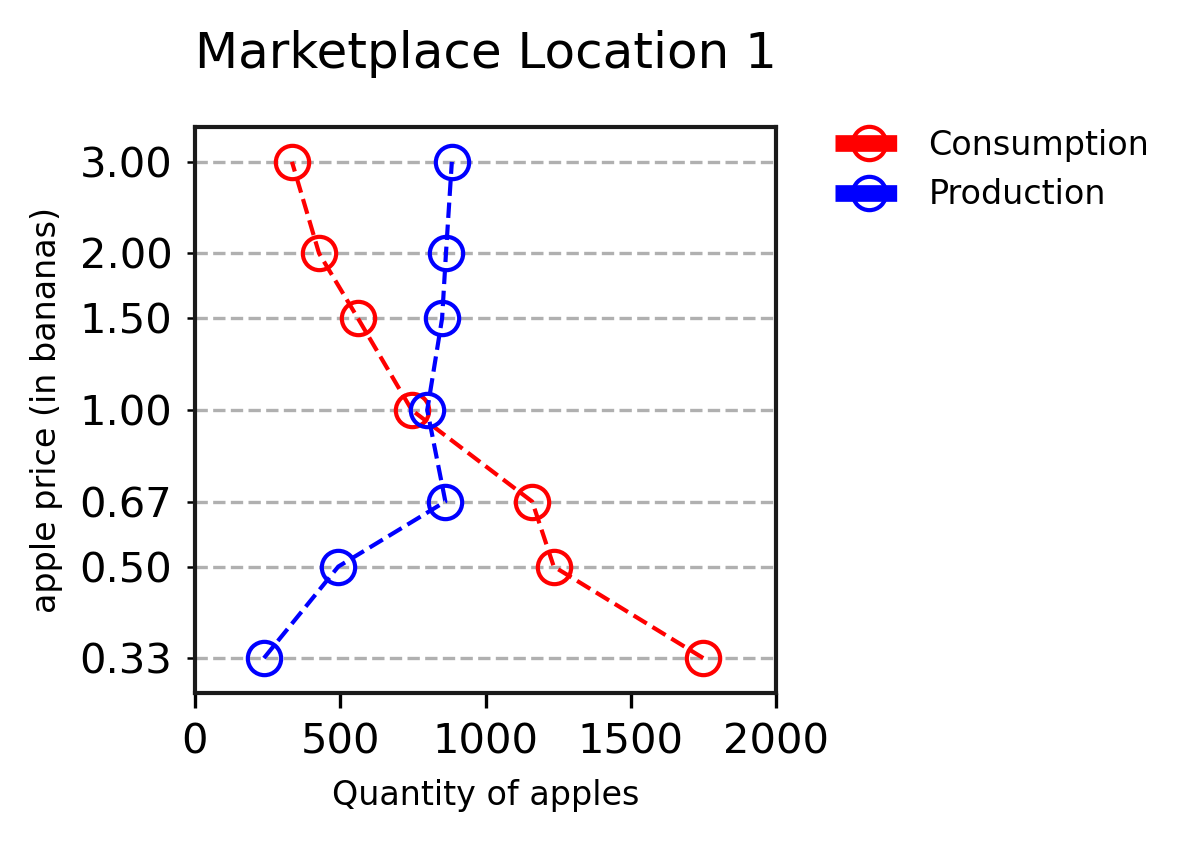}
    \caption{}
    \label{fig:moving_market:loc1}
    \end{subfigure}
    
    \begin{subfigure}{0.4\textwidth}
    \centering
    \includegraphics[height=1.9in]{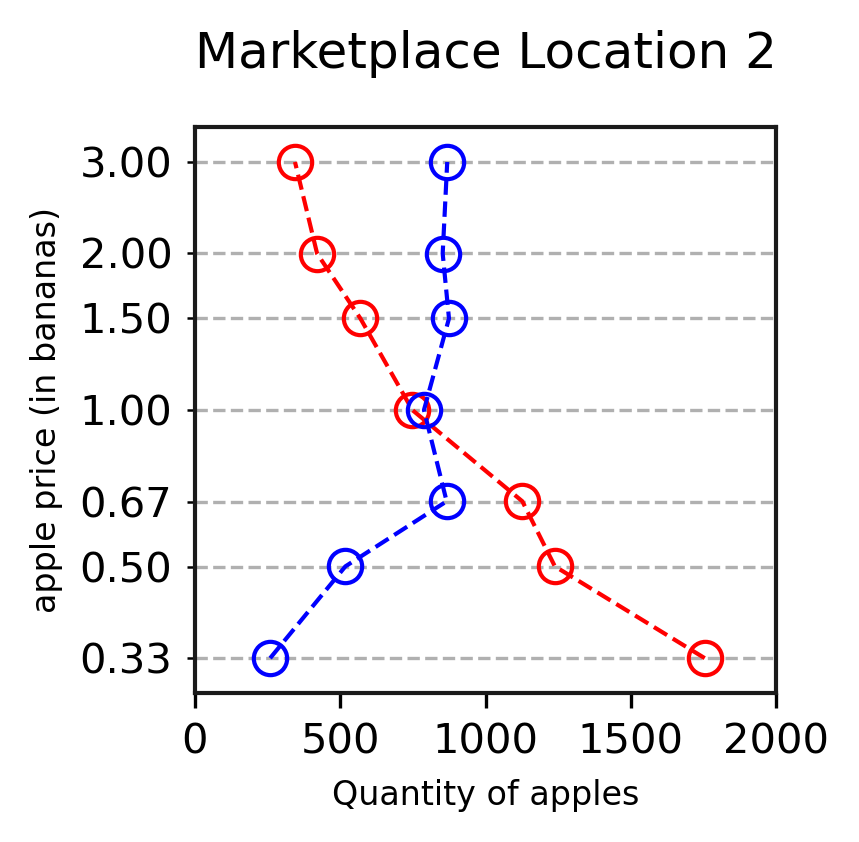}
    \caption{}
    \label{fig:moving_market:loc2}
    \end{subfigure}%
    ~
    \begin{subfigure}{0.59\textwidth}
    \centering
    \includegraphics[height=1.9in]{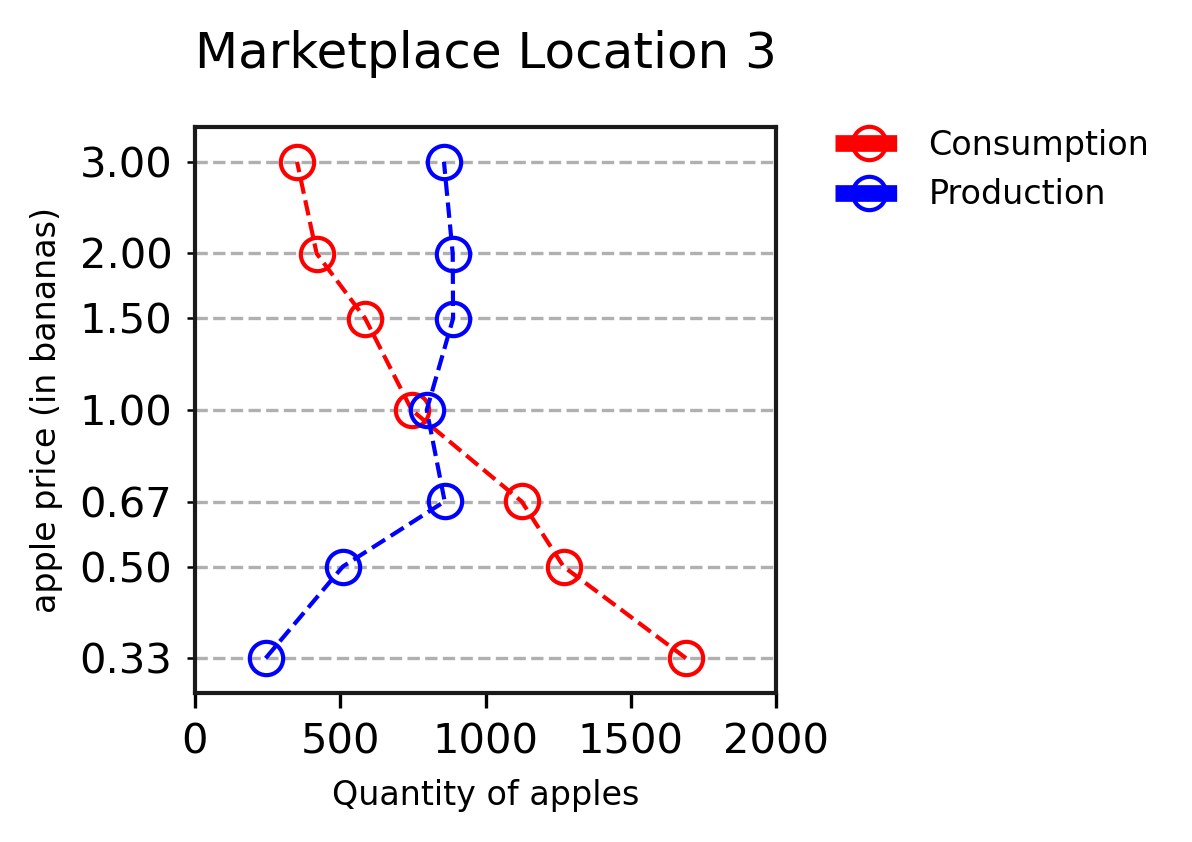}
    \caption{}
    \label{fig:moving_market:loc3}
    \end{subfigure}
    
    \begin{subfigure}{0.4\textwidth}
    \centering
    \includegraphics[height=1.9in]{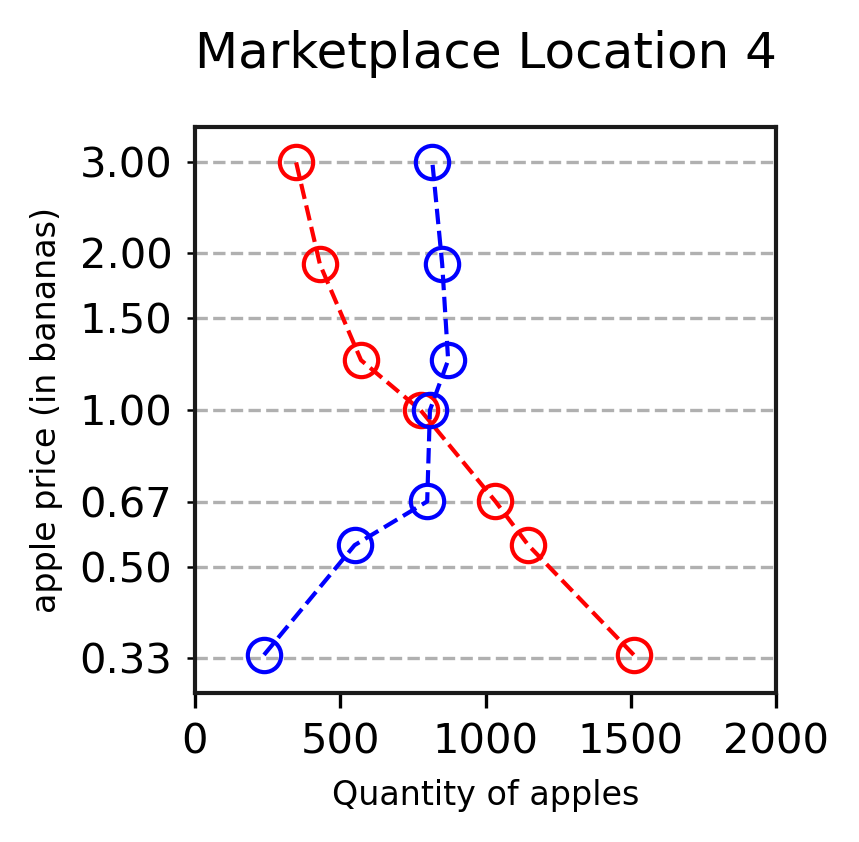}
    \caption{}
    \label{fig:moving_market:loc4}
    \end{subfigure}%
    ~
    \begin{subfigure}{0.59\textwidth}
    \centering
    \includegraphics[height=1.9in]{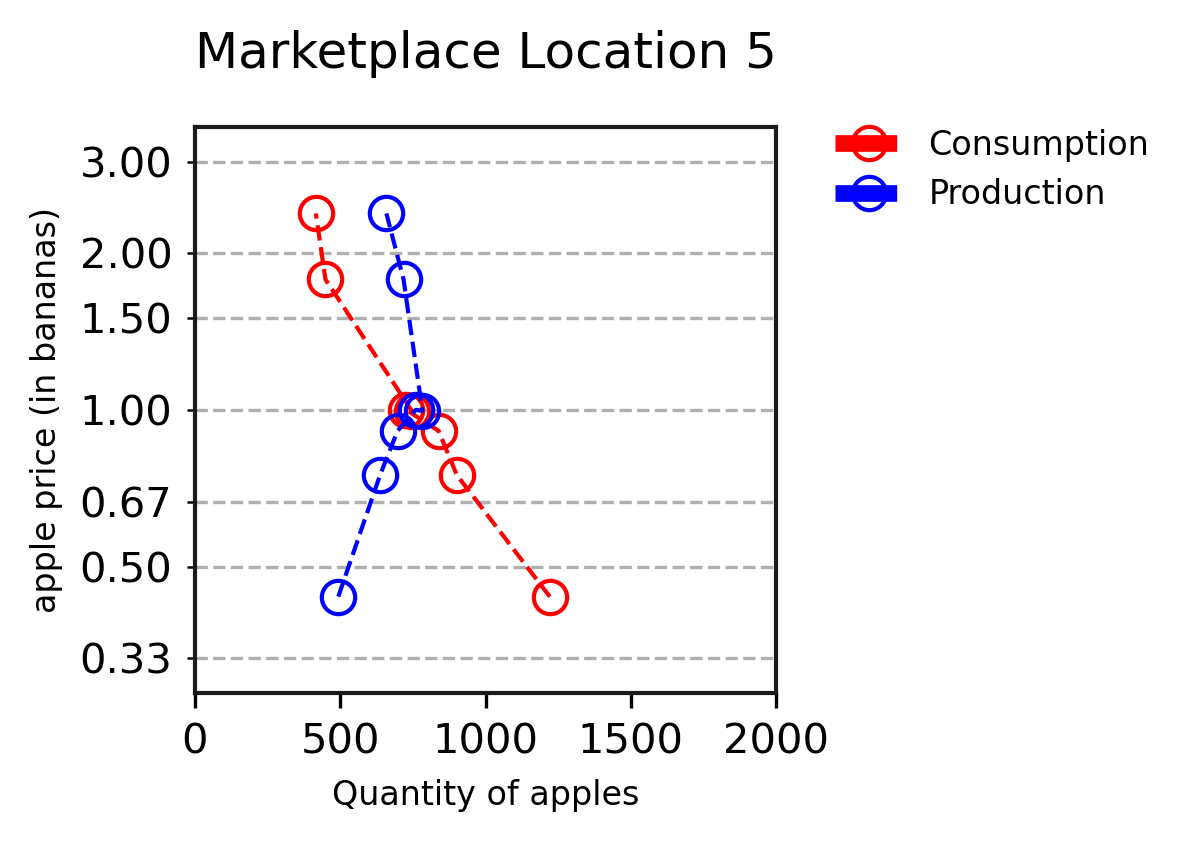}
    \caption{}
    \label{fig:moving_market:loc5}
    \end{subfigure}
    
    \begin{subfigure}{0.5\textwidth}
    \centering
    \includegraphics[height=1.9in]{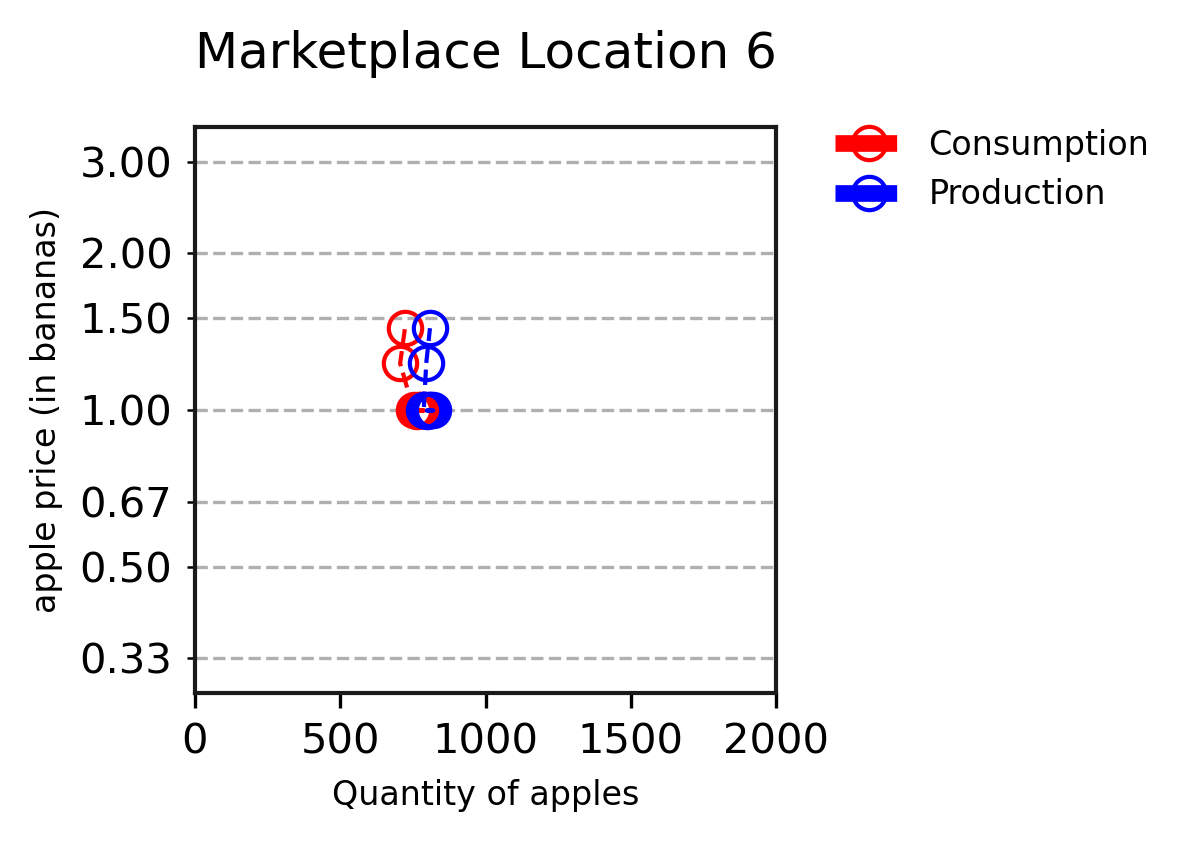}
    \caption{}
    \label{fig:moving_market:loc6}
    \end{subfigure}

    \caption{Production, Consumption, and Price at seven different marketplace prices per location.}
    \label{fig:moving_market}
\end{figure}

Figure~\ref{fig:moving_market} presents the equilibrium production, consumption, and price behaviour of the agents, similar to our previous marketplace experiment in Figure~\ref{fig:marketplace}. For each of the seven marketplace locations, we performed seven independent experiments where the marketplace offered both sides of one price. In all of these experiments, apple and banana trees spawn with the same probability. 

Figure~\ref{fig:moving_market:loc0}, with the marketplace in the center of the map, presents the same datapoints as in Figure~\ref{fig:marketplace}. As the marketplace moves towards the edge of the map in Figures~\ref{fig:moving_market:loc1} through~\ref{fig:moving_market:loc3}, we see minor changes in the horizontal positions of the datapoints (\eg, slight changes in production and consumption), but their vertical positions lie exactly on the y-axis rules matching the marketplace's price, showing that all agents' exchanges occur at the marketplace's price.

As the marketplace moves off the edge of the map in Figures~\ref{fig:moving_market:loc4} through~\ref{fig:moving_market:loc6}, this behaviour begins to change quickly. When the marketplace is just off the map in Figure~\ref{fig:moving_market:loc4}, all datapoints aside from the extreme 3.0 and 0.33 marketplace prices move slightly away from the y-axis rules towards the 1.0 price. Thus, while the marketplace still has a strong influence over the price, the agents are now also trading amongst themselves using prices closer to 1.0.

This continues in Figure~\ref{fig:moving_market:loc5}, when all datapoints shift towards the 1.0 price. Finally, in Figure~\ref{fig:moving_market:loc6} when the marketplace is truly inconvenient, almost all of the datapoints have shifted to the 1.0 price. The only outliers, when the marketplace price was 3.0 and 2.0, have similar production and consumption and only slightly higher average prices. Overall, these results demonstrate that the marketplace's influence over pricing behaviour is nonlinear: it remains strong while the marketplace is not too inconvenient to reach, and falls off quickly thereafter.

\begin{figure}
    \centering
    \begin{subfigure}{0.5\textwidth}
    \centering
    \includegraphics[height=2in]{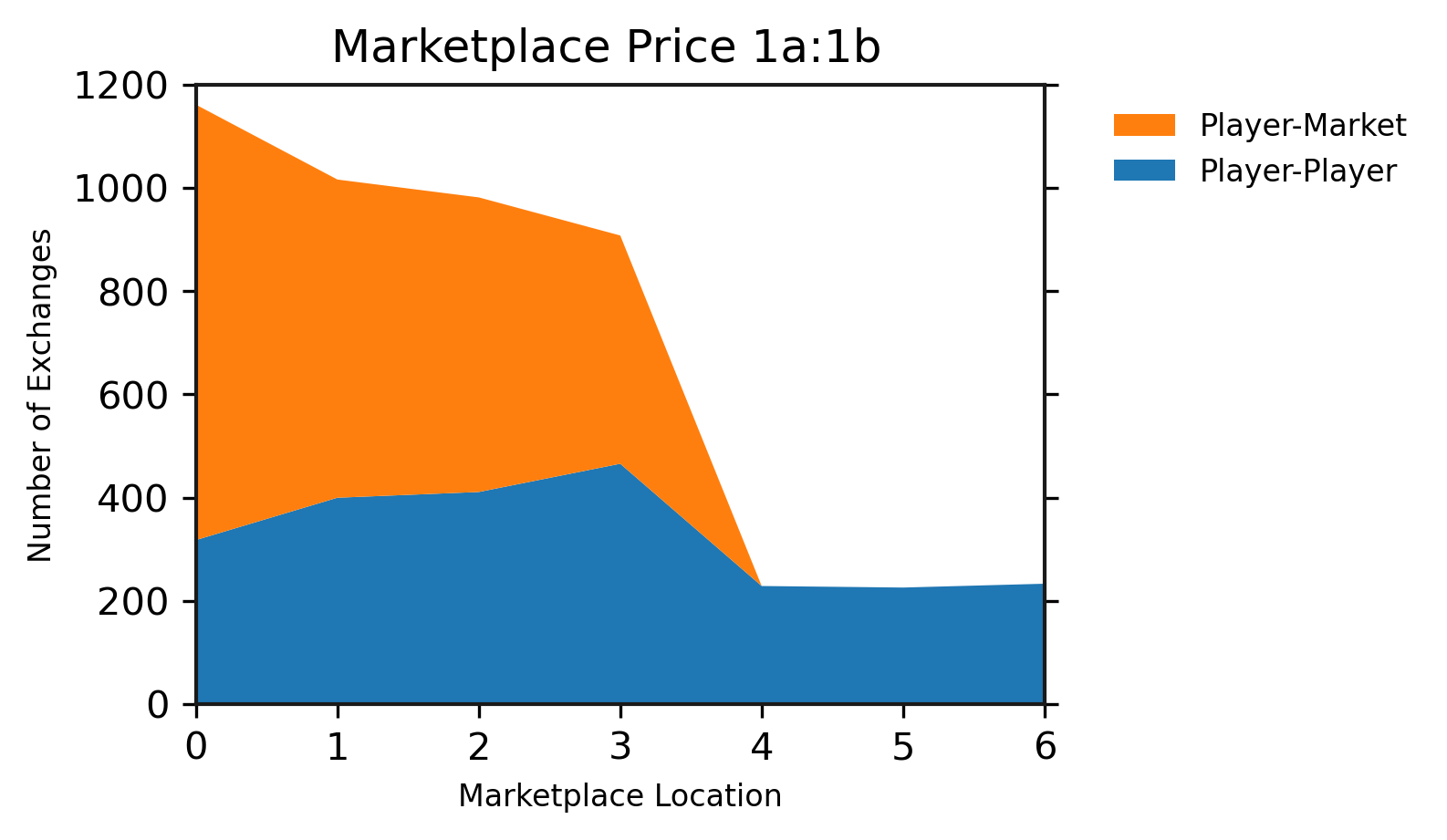}
    \caption{}
    \label{fig:moving_market_partner_price:partner_1a_1b}
    \end{subfigure}%
    ~
    \begin{subfigure}{0.5\textwidth}
    \centering
    \includegraphics[height=2in]{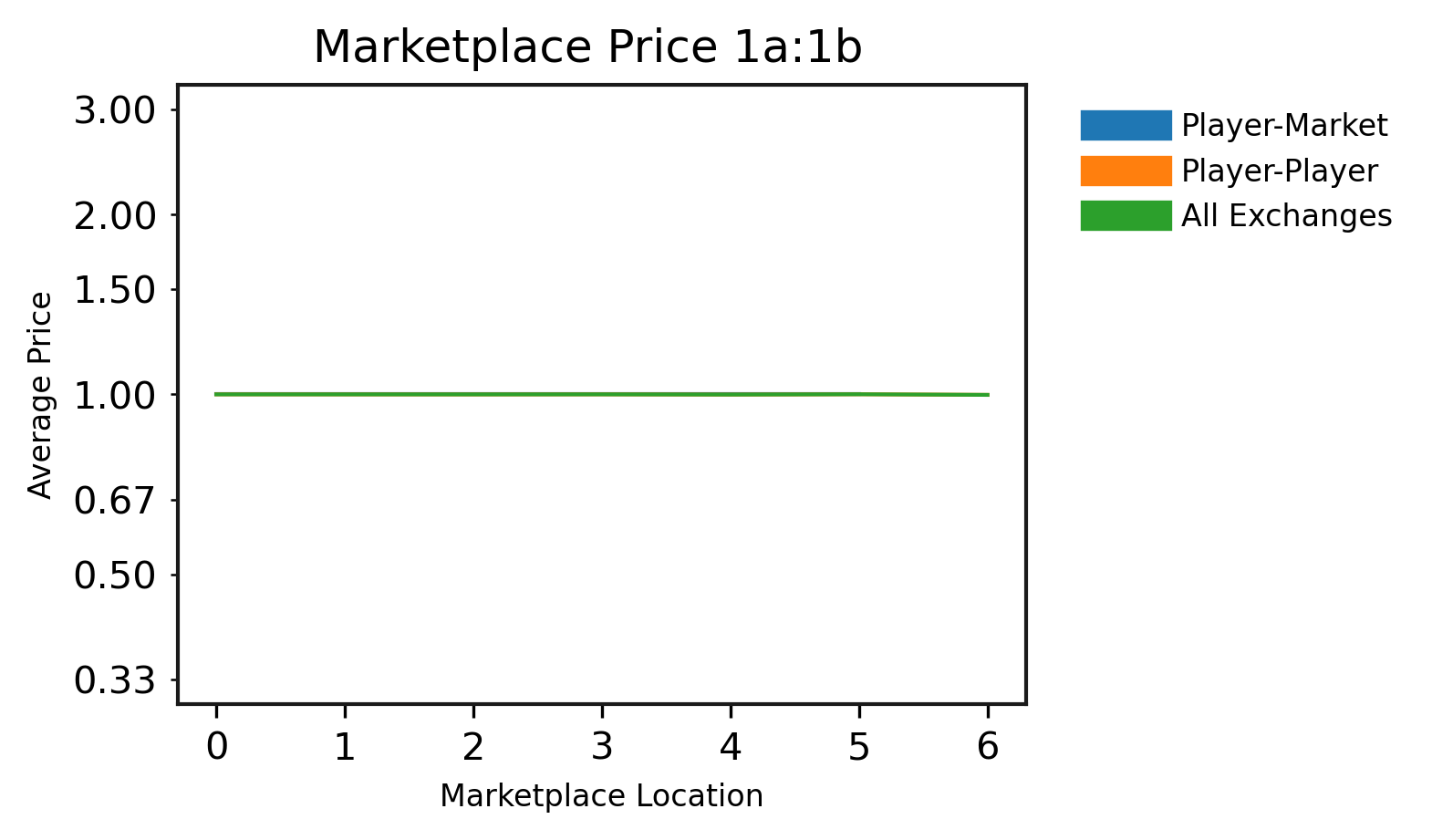}
    \caption{}
    \label{fig:moving_market_partner_price:price_1a_1b}
    \end{subfigure}
    
    \begin{subfigure}{0.5\textwidth}
    \centering
    \includegraphics[height=2in]{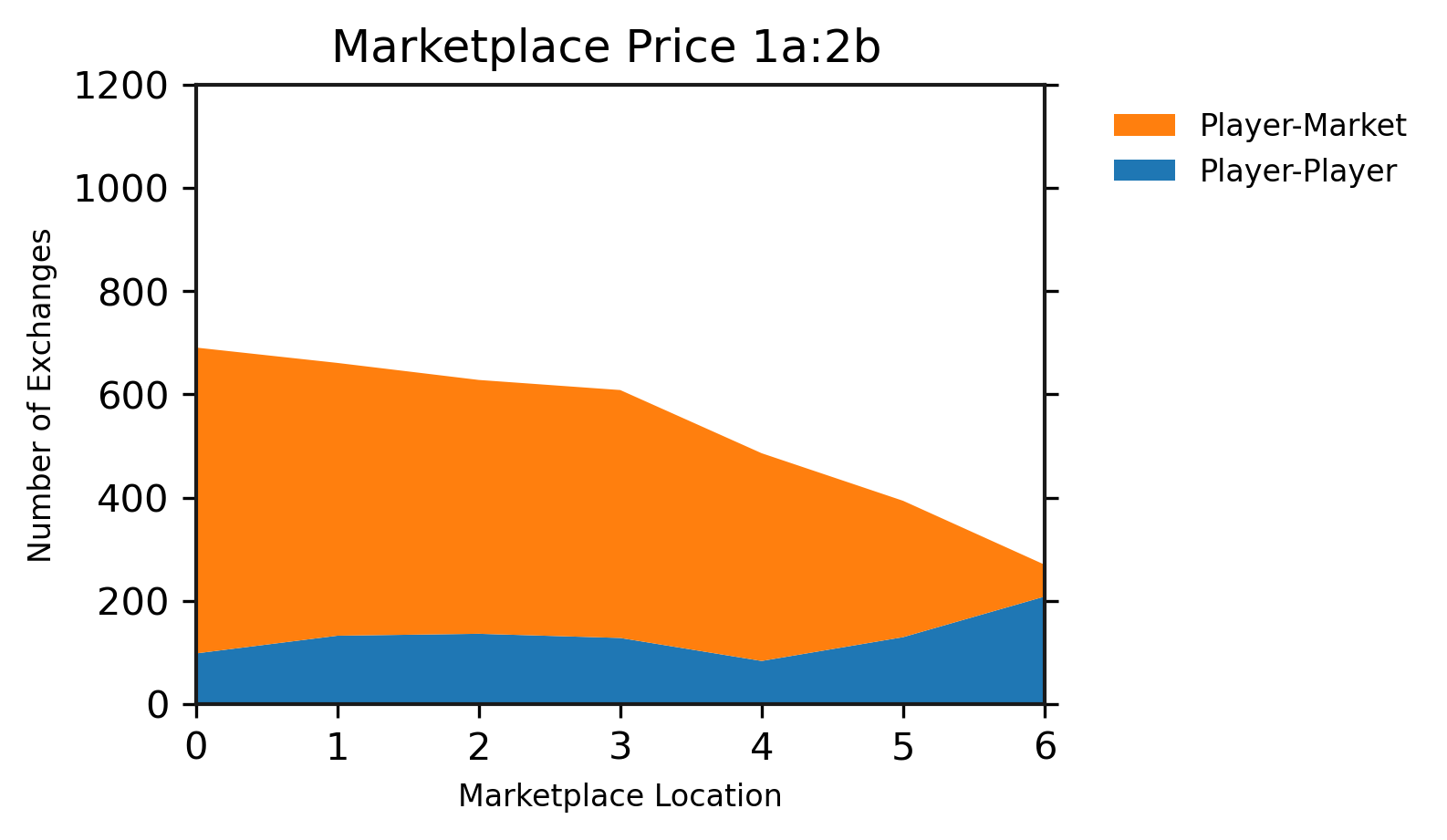}
    \caption{}
    \label{fig:moving_market_partner_price:partner_1a_2b}
    \end{subfigure}%
    ~
    \begin{subfigure}{0.5\textwidth}
    \centering
    \includegraphics[height=2in]{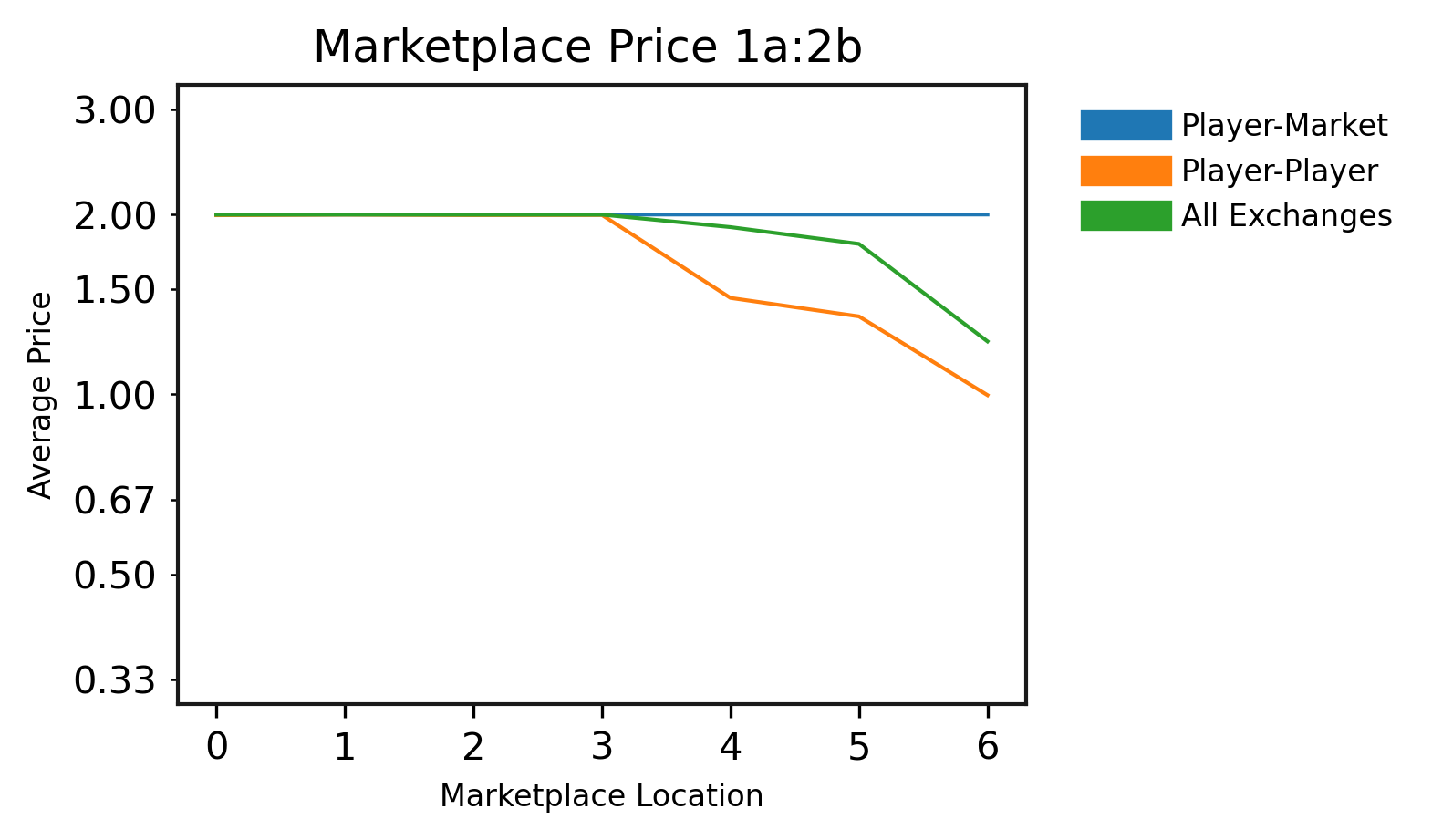}
    \caption{}
    \label{fig:moving_market_partner_price:price_1a_2b}
    \end{subfigure}
    
    \begin{subfigure}{0.5\textwidth}
    \centering
    \includegraphics[height=2in]{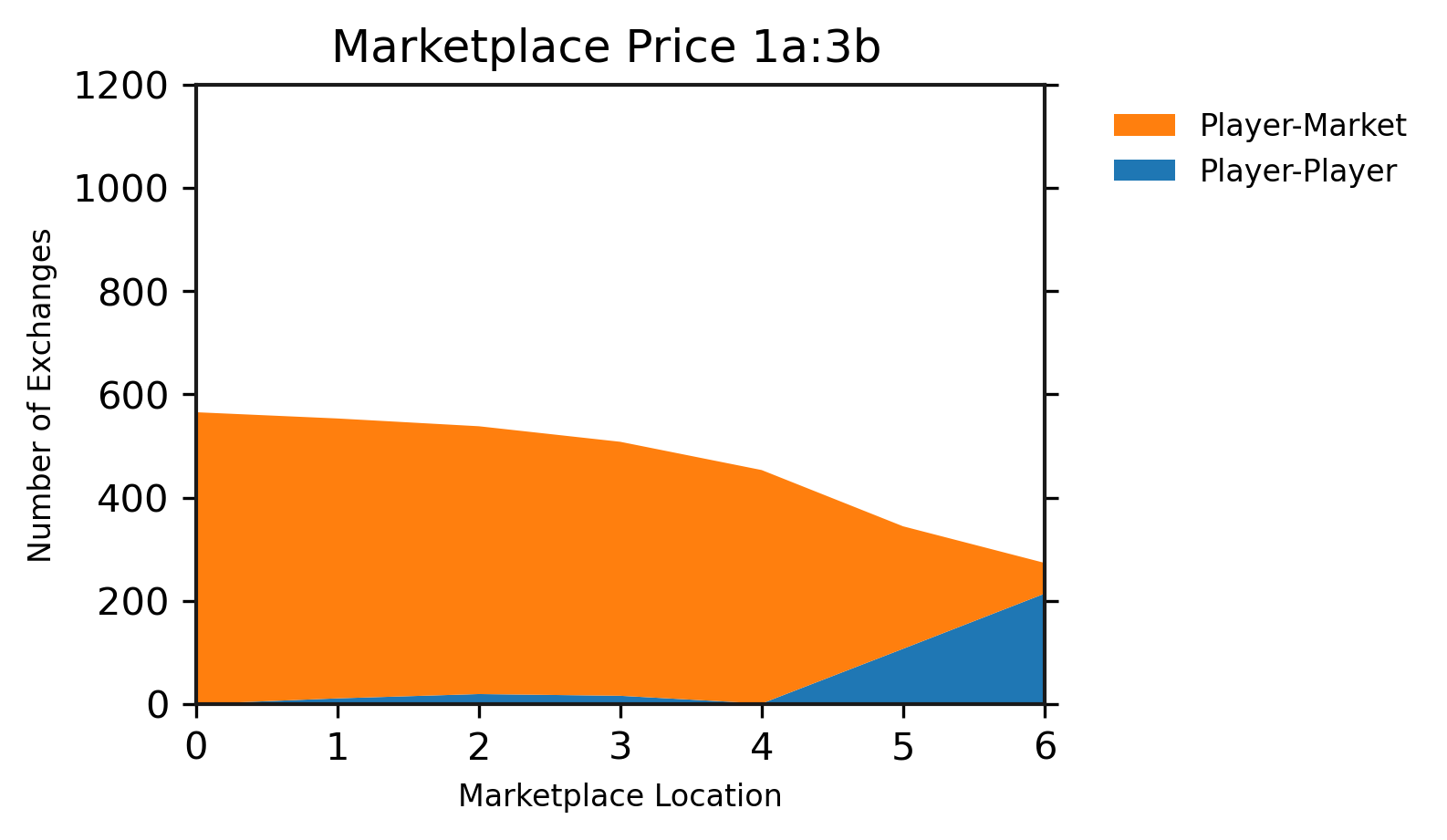}
    \caption{}
    \label{fig:moving_market_partner_price:partner_1a_3b}
    \end{subfigure}%
    ~
    \begin{subfigure}{0.5\textwidth}
    \centering
    \includegraphics[height=2in]{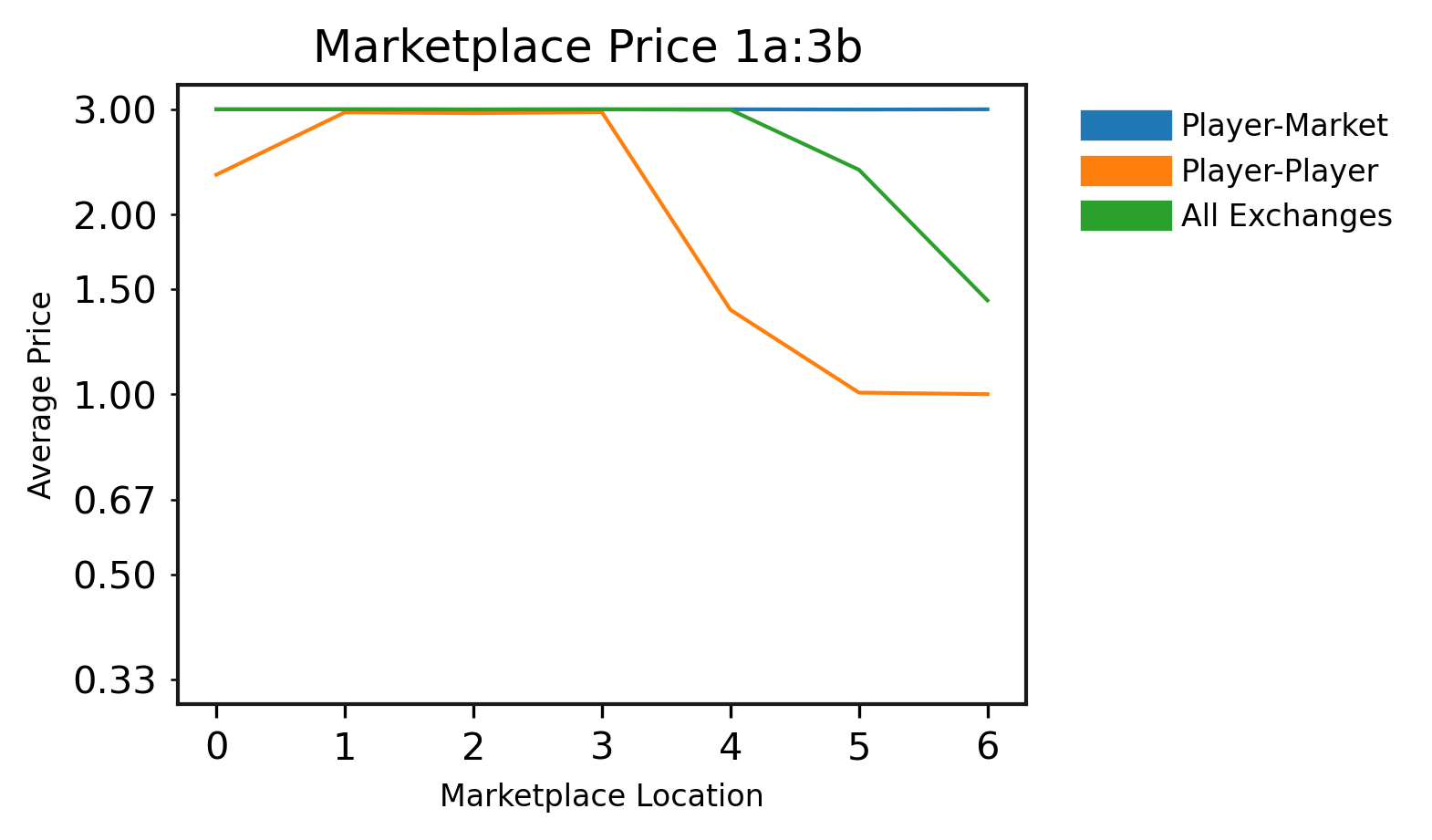}
    \caption{}
    \label{fig:moving_market_partner_price:price_1a_3b}
    \end{subfigure}
    
    \caption{Trading partners and prices, when the marketplace offers 1a:1b, 1a:2b, and 1a:3b prices. (a), (c), and (e) show stackplots of the quantity of player-marketplace and player-player exchanges, showing how exchanges shift from mostly player-market to mostly player-player as the marketplace becomes inconvenient to reach. (b), (d), and (f) show the average price used in player-market, player-player, and all exchanges. As the marketplace becomes inconvenient to reach, the player-player price reverts towards 1.0.}
\label{fig:moving_market_partner_price}
\end{figure}

Figure~\ref{fig:moving_market_partner_price} explores these results further, by plotting how many exchanges happen between pairs of players as opposed to players and the marketplace, and the average price used in those exchanges. Each subfigure shows the marketplace location on the x-axis, and we highlight the marketplace prices of 1a:1b, 1a:2b, and 1a:3b. Subfigures~\ref{fig:moving_market_partner_price:partner_1a_1b}, \ref{fig:moving_market_partner_price:partner_1a_2b}, and \ref{fig:moving_market_partner_price:partner_1a_3b} are a stackplot, showing the player-player quantity of trades on top of the player-market trades; Figure~\ref{fig:moving_market_partner_price:partner_1a_1b} should thus be read at marketplace 0 as showing approximately 1200 total exchanges, divided into 300 exchanges between players, and 900 between players and the marketplace.

These plots highlight how the marketplace's influence over the population's price falls off. In Figure~\ref{fig:moving_market_partner_price:partner_1a_1b}, for example, we see that players mostly trade with the market when it is near trees, but exchanges with the market fall off to zero as soon as it is off of the map\footnote{Interestingly, the number of exchanges between players \textit{also} drops sharply from about 400 per episode at location 3 to around 250 per episode at locations 4 and beyond. In earlier results without a marketplace such as Figure~\ref{fig:baseline_exchanges_price:exchanges}, we also saw convergence to 250 exchanges per episode. While it is possible that the presence of the marketplace helps the agents learn how to trade more even with each other (\eg, learning is easier when the marketplace is always in the same location, always making a consistent offer, always has goods to trade, etc), we believe that the explanation is more mundane. Note that the price in Figure~\ref{fig:moving_market_partner_price:price_1a_1b} remains at 1.0 at locations 4-6 when all trades are between players, and that the production and consumption quantities at the 1.0 price are roughly the same in Figures~\ref{fig:moving_market:loc0} and~\ref{fig:moving_market:loc6} at around 800 per episode. So, price, production, and consumption are about the same between locations 0 and 6, although the quantity of trades is higher when the marketplace is convenient. The solution is likely that the marketplace offers ``Give 1 apple for 1 banana'' and its reverse, while the players switch to the more efficient ``Give 3 apples for 3 bananas'' offer, resulting in more items transferred per exchange at the same price.}.

\begin{figure}
    \centering
    \includegraphics{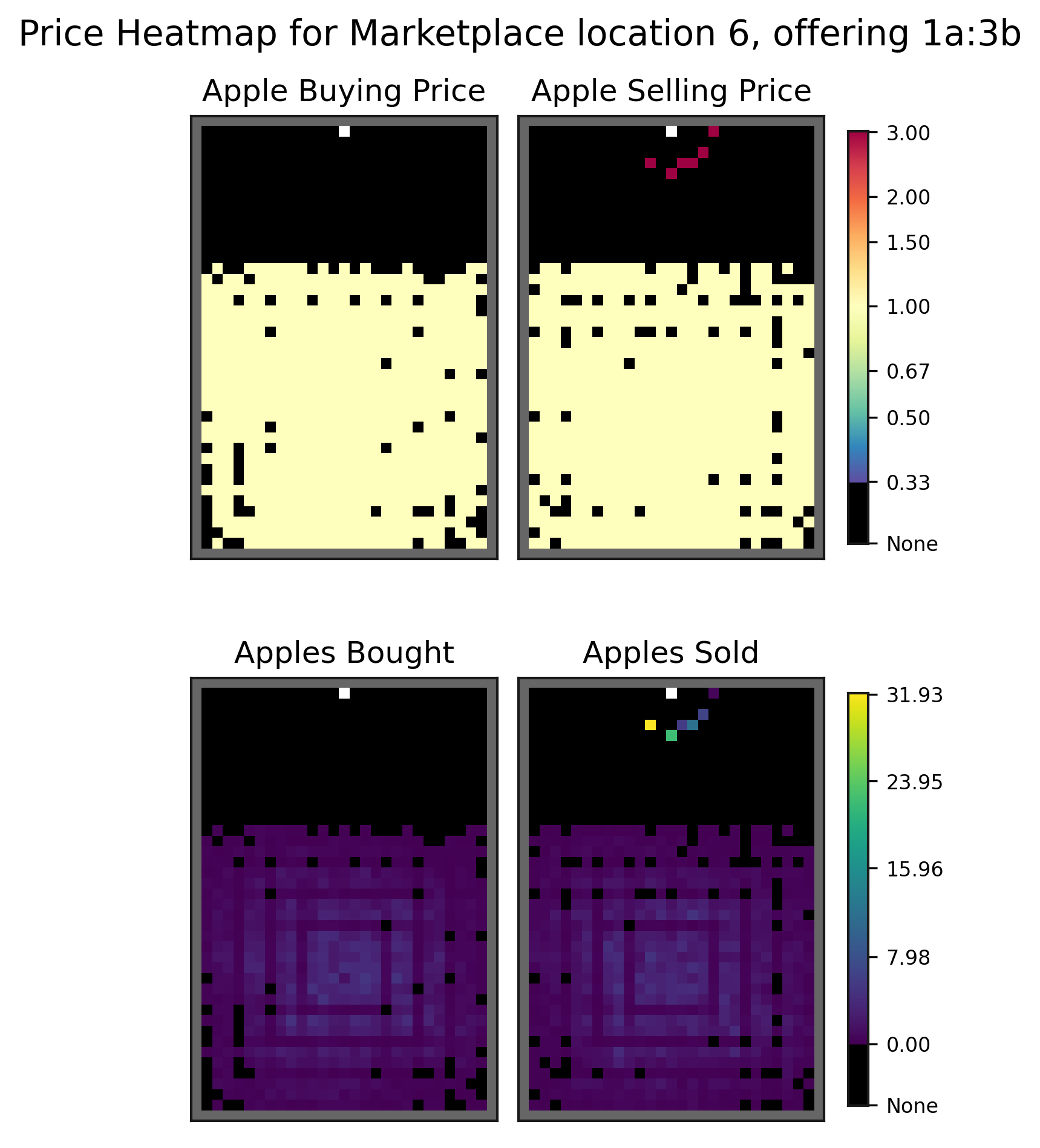}
    \caption{Heatmap of apple prices and quantities, when a Marketplace is at location 6 and offering to buy and sell apples at 1 apple for 3 bananas. Results are averaged over the final 25\% of training. Note that two price regions emerge: apples are sold at a high price near the marketplace but are never bought there, and are bought and sold at the 1.0 price across the region where trees grow.}
    \label{fig:moving_market_loc6_1a3b_pricemap}
\end{figure}

When the marketplace offers higher prices such as ``1 apple for 2 bananas'' and ``1 apple for 3 bananas'', player-marketplace exchanges dominate player-player exchanges until the marketplace reaches its most inconvenient location. At locations 0 to 3, when the marketplace is on the edge of the trees, the player-player exchanges happen at the same price that the marketplace offers. But once the marketplace moves out of the trees and off of the normal map, we see both an increase in player-player exchanges, and their exchanges deviate from the marketplace price towards the 1.0 price. Thus, some agents continue making the trip to the marketplace to extract its high price, but the rest quickly revert to an equal price. Figure~\ref{fig:moving_market_loc6_1a3b_pricemap} expands on this by presenting heatmaps of apple price and quantity when the marketplace is both inconvenient and making the 1a:3b and 3b:1a offers. Here, we see just two prices emerge: exchanges at the 3.0 price a few steps away from the marketplace, and exchanges at the 1.0 price everywhere else. 

These are our first results demonstrating that different prices can emerge in different parts of the map, and that agents can learn to adjust their pricing behaviour in response to their local conditions and the costs required to travel. We will revisit this idea more organically, without marketplaces, in Section~\ref{sec:experiments:regions}, where we will demonstrate the emergence of multiple prices and arbitrage behaviour.

\subsection{Trade Radius}
\label{sec:experiments:trade_radius}

Each player has a \textbf{trade radius} and a \textbf{offer visibility radius}, both set to 4 tiles by default, that specifies how close players must be to exchange goods and observe each others' offers. This value of 4 tiles trades off between making the mechanic easier for agents to learn (at high values) and more emphasis on inter-player interaction and the emergence of local prices (at low values). In this section, we will explore varying both of these radii together, to see how this choice impacts the emergence of trade. Can our agents still learn to trade with a radius of 1, when trade partners must be adjacent?

\begin{figure}
    \centering
    \begin{subfigure}{0.4\textwidth}
        \centering
        \includegraphics[height=2.5in]{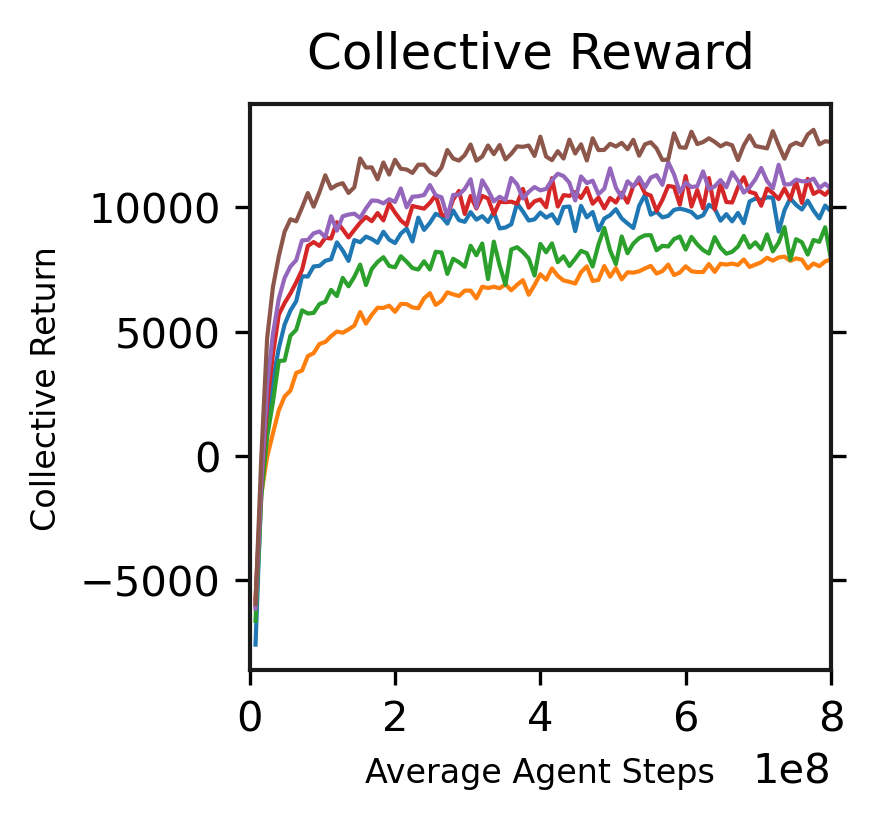}
        \caption{}
        \label{fig:trade-radius:reward}
    \end{subfigure}%
    ~
    \begin{subfigure}{0.6\textwidth}
        \centering
        \includegraphics[height=2.5in]{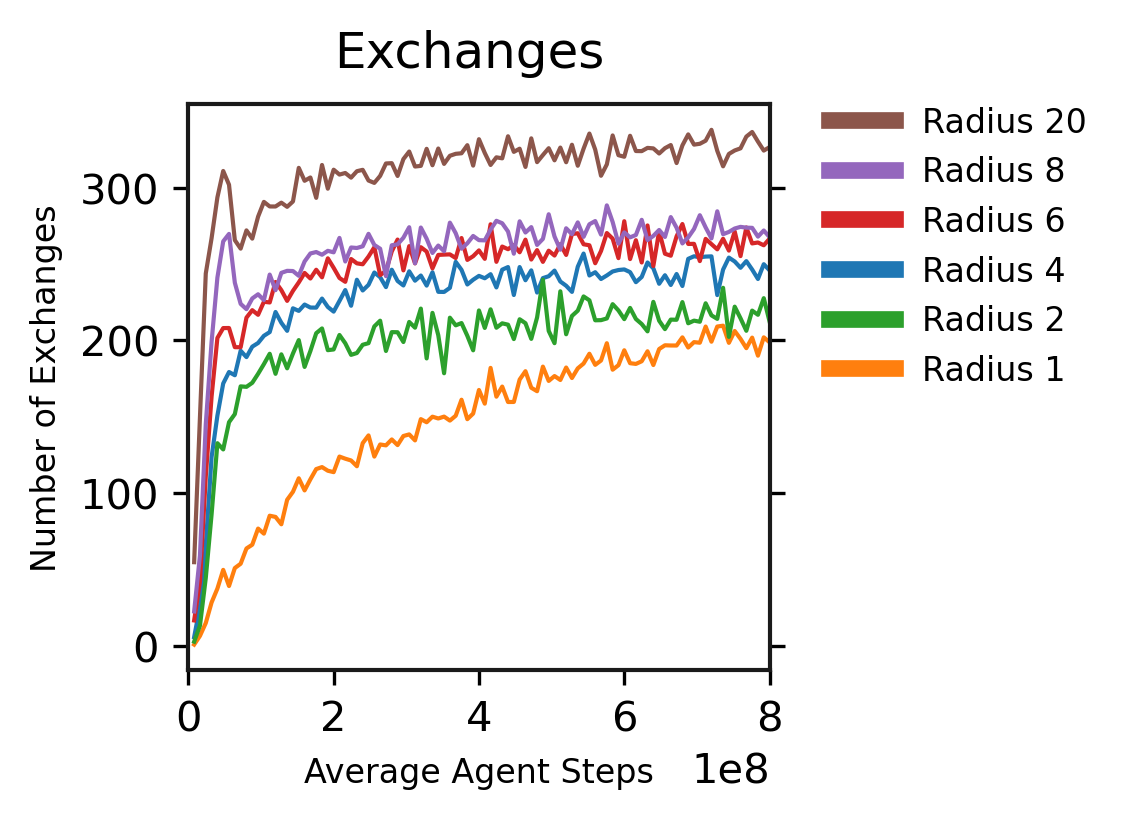}
        \caption{}
        \label{fig:trade-radius:exchanges}
    \end{subfigure}
    
    \caption{Varying the trade radius. As the trade radius and offer visibility radius are varied from their default value of 4, we see that trading behavior still emerges reliably.}
    \label{fig:trade-radius}
\end{figure}

Figure~\ref{fig:trade-radius} shows that the answer is yes. In the $(a=1,b=1)$ setting, our agents learn to trade with every radius from 1 (orthogonally adjacent) to 20 (nearly spanning the map). 
In terms of reward, in Figure~\ref{fig:trade-radius:reward}, each increase in radius enables higher collective reward. This is unsurprising, as a larger radius means less time spent transporting goods, and more time producing and consuming them. In terms of exchanges, in Figure~\ref{fig:trade-radius:exchanges}, we see that the agents learn to trade very quickly in all cases aside from the radius of 1, although the total number of exchanges tops out at different values. Even though the Radius 1 population learns to trade more slowly, by the end of training they are trading nearly as often as with a larger radius. 

This ability to learn with a small radius, even the extreme case where players must be adjacent, is a promising result for future experiments where we want to detect local variations in price across the map. These local prices might not emerge if a large trade radius was required for our agents to learn how to trade; fortunately, this is not necessary.

\subsection{Regions, Borders, and Merchant Behavior (Arbitrage)}
\label{sec:experiments:regions}

Our experiments thus far have examined one map in which each type of tree spawns with uniform density across space. We have shown that agents converge to different pricing, production, and consumption behaviour as we adjust the abundance of each type of tree, but this uniform density limits the scope of behaviours that our agents can learn. For example, it is not possible to learn to ``buy low and sell high'' when the population converges to one price across space and time.

In this section, we will explore alternative maps that have apple-rich and banana-rich regions. Our goal is to investigate how this affects the agents' learned behaviour: which regions will agents of each role visit, where will exchanges take place, and will one price become dominant across the map, or will the price vary across the map reflecting the local abundance of goods?

We will demonstrate some settings in which local prices do emerge, and further, that some agents take advantage of this price difference by buying and selling both apples and bananas, and transporting the goods between the regions. The agents discover new niches over time that depend on the rest of the population's behaviour: first producing items to consume, then producing items to sell, and then buying items to sell. This learned behaviour is related to arbitrage in that it involves making transactions across multiple markets at different prices to extract a profit. Note however, the term arbitrage often has the connotation of an instantaneous and risk-free set of transactions, whereas in our setting the transactions are separated by both time (required to transport goods between regions) and space (the distance between regions), and involve additional risks (uncertainty about the ability to find a trade partner in the next region) and costs (the movement penalty for carrying goods). Thus, we will call it \textbf{merchant behaviour}, as it involves the agents learning to buy, sell, and transport goods, instead of only producing fruit for sale as the title ``farmer'' has suggested until now.

\begin{figure}
    \centering
    \includegraphics[width=\hsize]{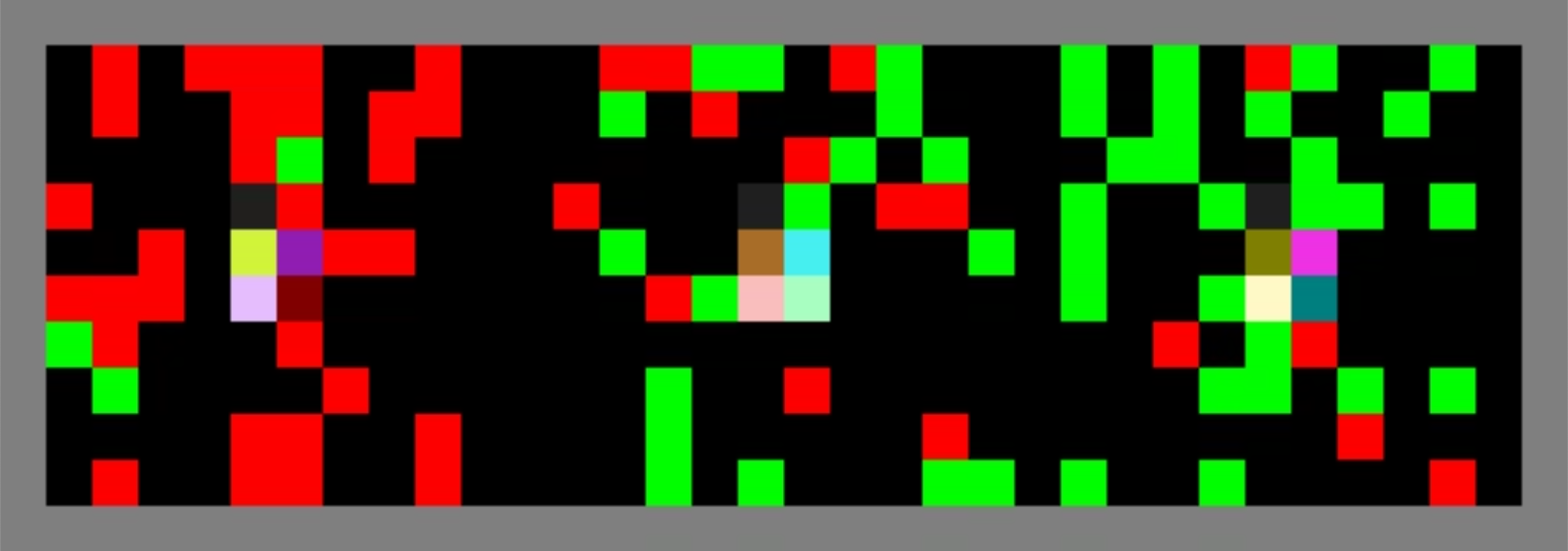}
    \caption{An example of the ``No Walls'' map, in which players can freely move through three regions with varying density of trees: an apple-rich region on the left, a banana-rich region on the right, and a neutral region in the middle. The trees are placed at uniform random within each region at the start of each episode, and map shown here is one example from that distribution. The four colored squares in the center of each region are the players, who start each episode in these locations.}
    \label{fig:regions:map_regions}
\end{figure}

Figure~\ref{fig:regions:map_regions} presents a map called ``No Walls''. It has an apple-rich region on the left, a banana-rich region on the right, and a neutral region in the middle. As with our previous results, the trees are placed at uniformly random locations within each region at the start of each episode. In the left and right regions, the base probability of a tree appearing in each map location is 30\%, with 90\% of those being for the plentiful fruit and 10\% being for the rare fruit. In the middle region, the base probability of trees is also 30\%, with 50\% being apple trees and 50\% being banana trees.

Each episode is played by twelve players, with two Apple Farmers and two Banana Farmers starting in the center of each region. Similar to our earlier ``Apple Farmer'' and ``Banana Farmer'' roles for agents, in this map each agent is permanently assigned both a role and starting region. Thus, each agent is an ``Apple Region - Apple Farmer'', ``Apple Region - Banana Farmer'', ``Neutral Region - Apple Farmer'', and so on, for the six combinations of two roles and three starting regions. We trained a population of 24 agents (four of each role-region pair), and sampled twelve without replacement (two of each region and role combination) for each episode.

In the ``No Walls'' map, each player starts off in their designated region, but then can move wherever they choose during the episode. We will also explore two more maps, called ``Walls'' and ``Thick Walls''. The ``Walls'' map adds walls one tile wide to separate each region, which represent semi-permeable borders: players cannot move through them, but can trade across them so long as another player is within the trade radius of four spaces. The walls also do not block the player's top-down view of the map in front of them, so a player can see if a potential trading partner is across the wall, and can see the trees available in that region. The ``Thick Walls'' map thickens these walls to eight spaces, thus preventing both movement and trade. Since each agent is assigned a permanent starting region, the presence of the walls will lock an agent into that region for their entire experience. In all maps, each region has 96 tiles where trees can grow (a $10\times10$ square region, with four tiles reserved for the agent starting locations in the middle), and thus the presence or thickness of walls does not affect the number of fruit trees available in each region.

By comparing the agents' learned behaviour in these three maps, we can explore when and how different trading behaviour and prices might emerge over the map. For example, do agents learn to trade goods close to where they are produced, or do they transport them a common location such as the center of the map, or does trade occur everywhere? Do the agents converge to one global price across the map, or does the price depend on the local abundance of fruit?  Do different prices emerge when agents can move everywhere, or only when thin walls create distinct regions, or only when thick walls separate the agents into independent economies?

\begin{figure}
    \centering
    \begin{subfigure}{0.42\textwidth}
        \centering
        \includegraphics[height=2.5in]{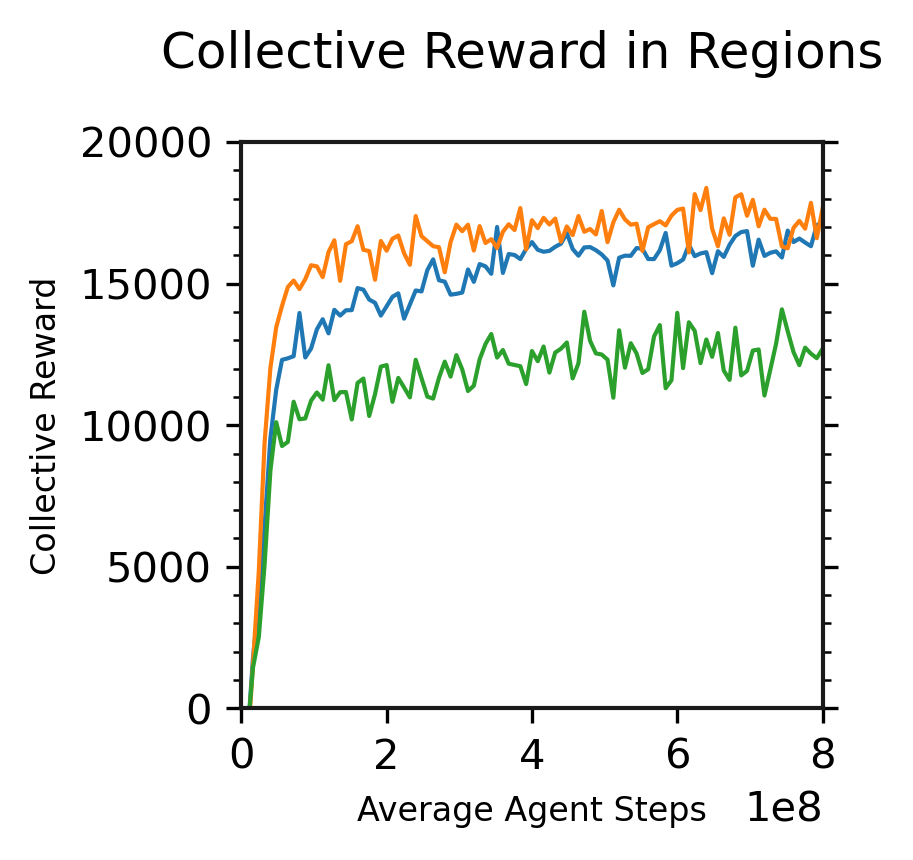}
        \caption{}
        \label{fig:regions:return_exchanges:return}
    \end{subfigure}%
    ~
    \begin{subfigure}{0.58\textwidth}
        \centering
        \includegraphics[height=2.5in]{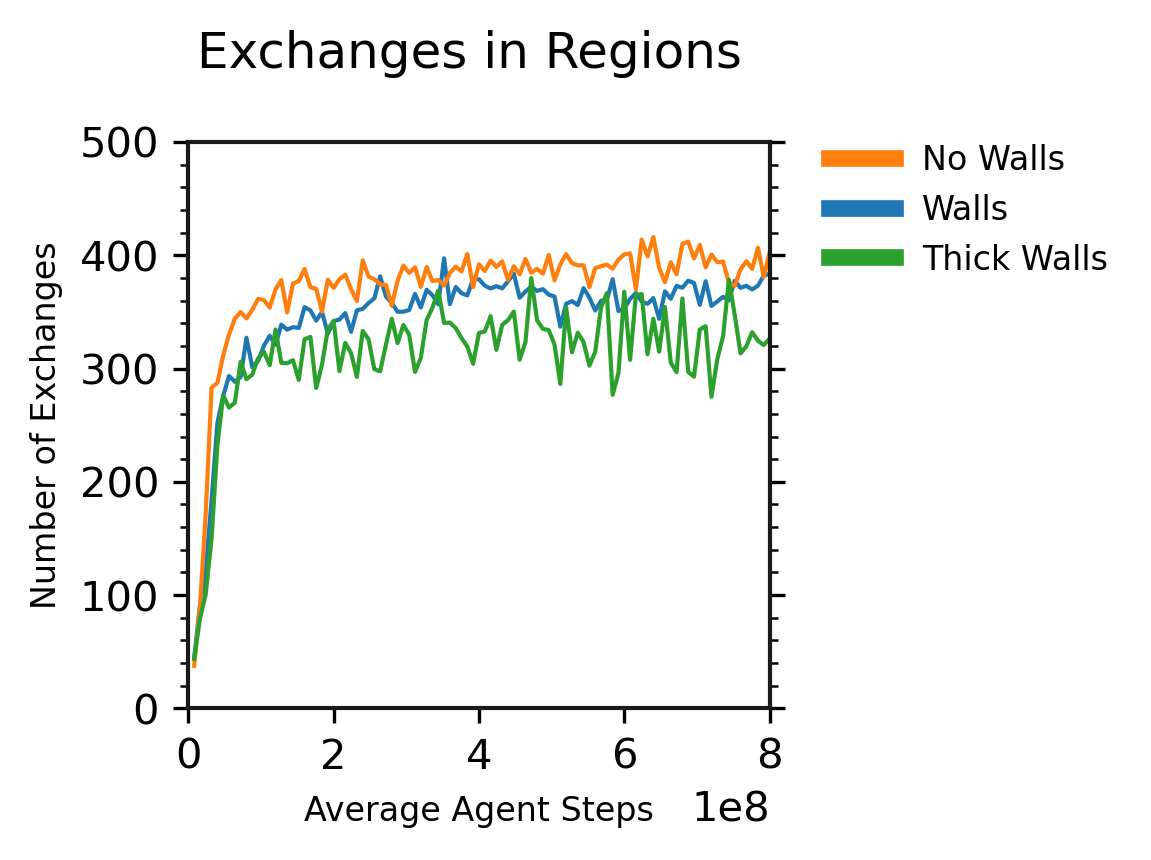}
        \caption{}
        \label{fig:regions:return_exchanges:exchanges}
    \end{subfigure}

    \caption{Collective reward and exchanges per episode in the three Regions maps.}
    \label{fig:regions:return_exchanges}
\end{figure}

Figure~\ref{fig:regions:return_exchanges} begins our analysis by comparing the collective reward and quantity of exchanges in populations trained in each map. Perhaps unsurprisingly, both metrics are highest in the ``No Walls'' map, where agents are free to produce and trade wherever they choose. The ``Walls'' agents are close behind in both metrics. In ``Thick Walls'', trade still emerges between the two Apple Farmers and two Banana Farmers sharing each region, although with fewer exchanges and a lower collective reward. Comparing ``Walls'' and ``Thick Walls'', we conclude that the agents would collectively earn more reward, despite being trapped in their regions, if they were able to trade goods across the borders.

\begin{figure}
    \centering
    \begin{subfigure}{\textwidth}
        \centering
        \includegraphics[height=2.5in]{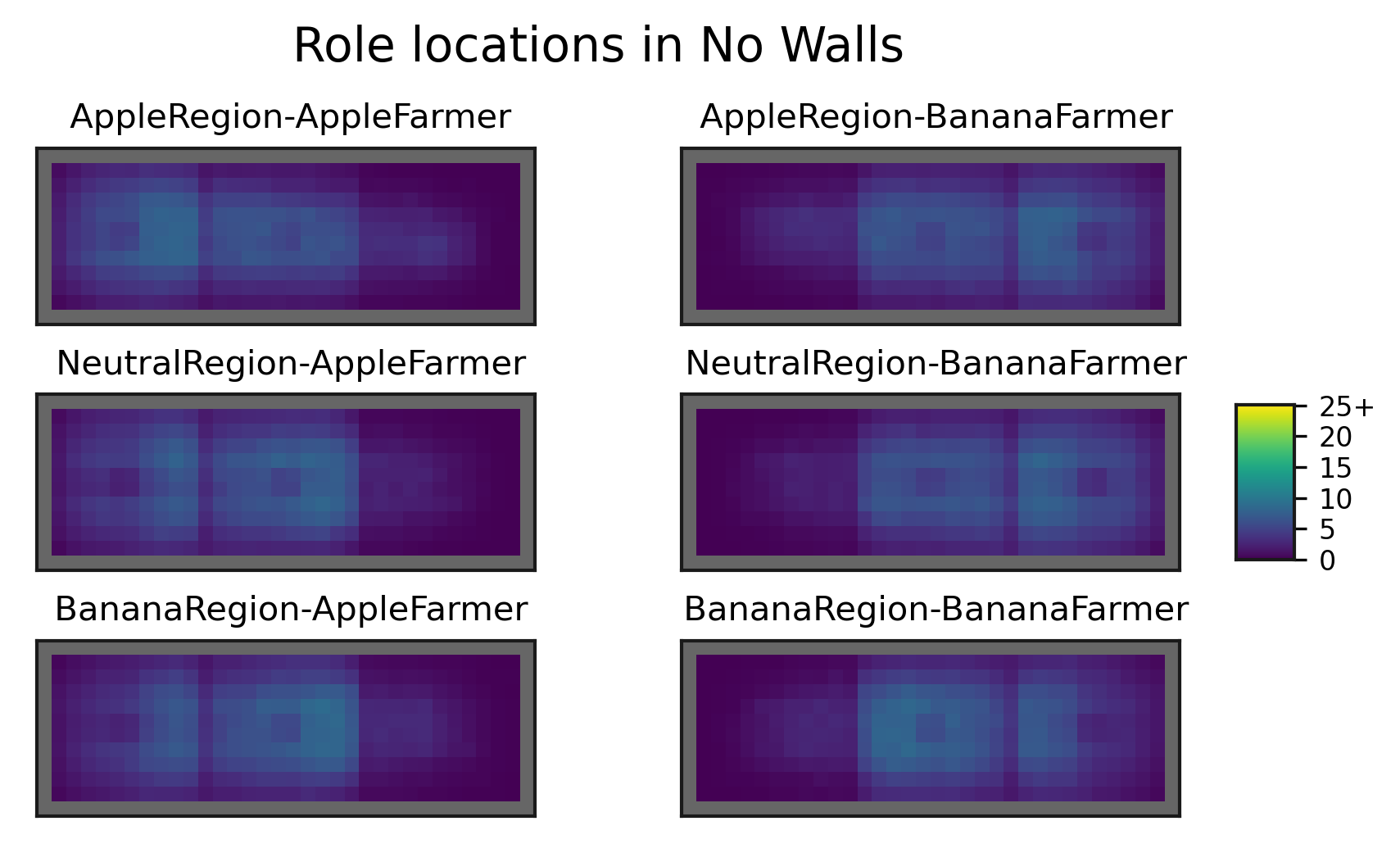}
        \caption{}
        \label{fig:regions:location_heatmaps:no_walls}
    \end{subfigure}
    
    \begin{subfigure}{\textwidth}
        \centering
        \includegraphics[height=2.5in]{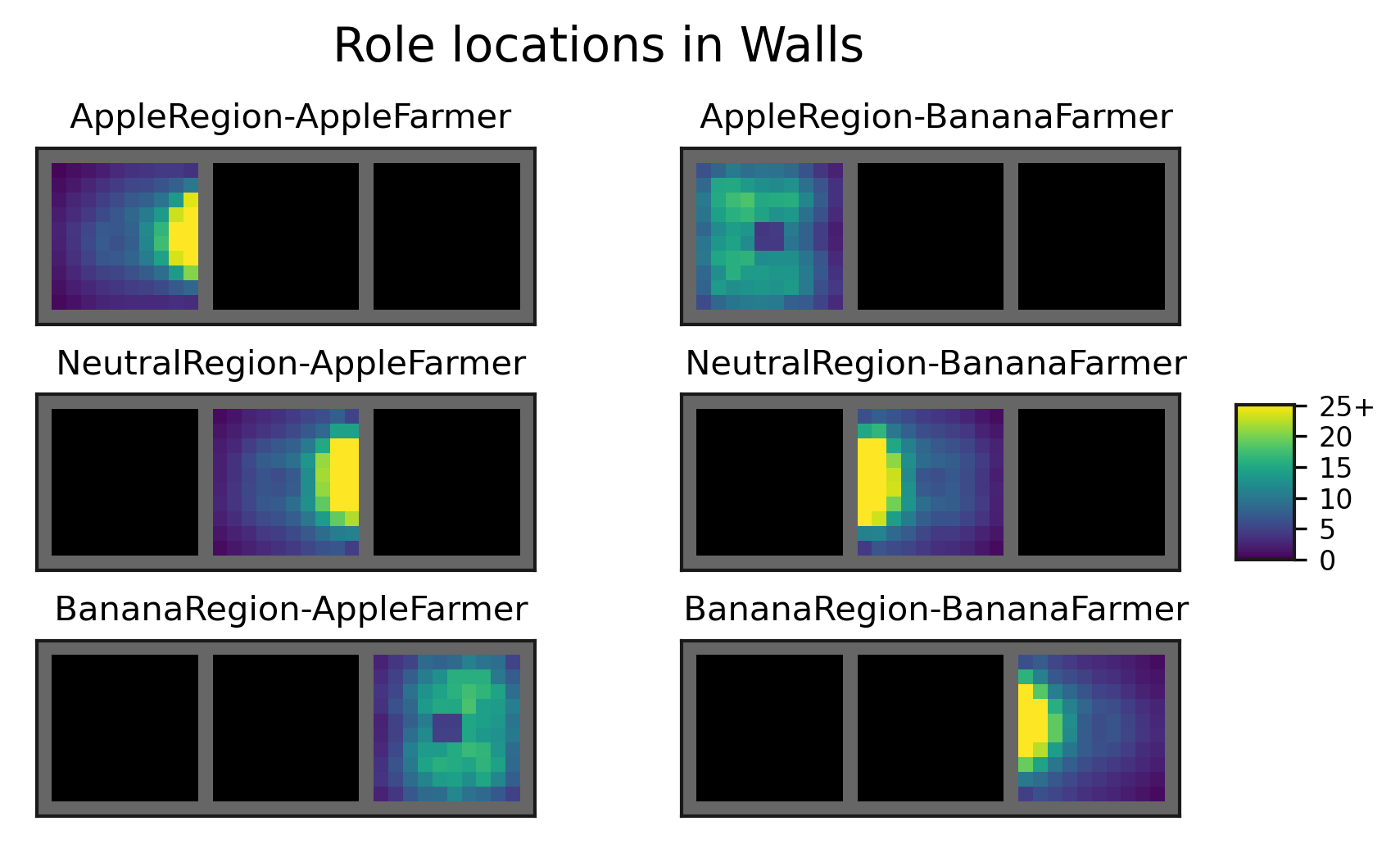}
        \caption{}
        \label{fig:regions:location_heatmaps:walls}
    \end{subfigure}
    
    \begin{subfigure}{\textwidth}
        \centering
        \includegraphics[height=2.5in]{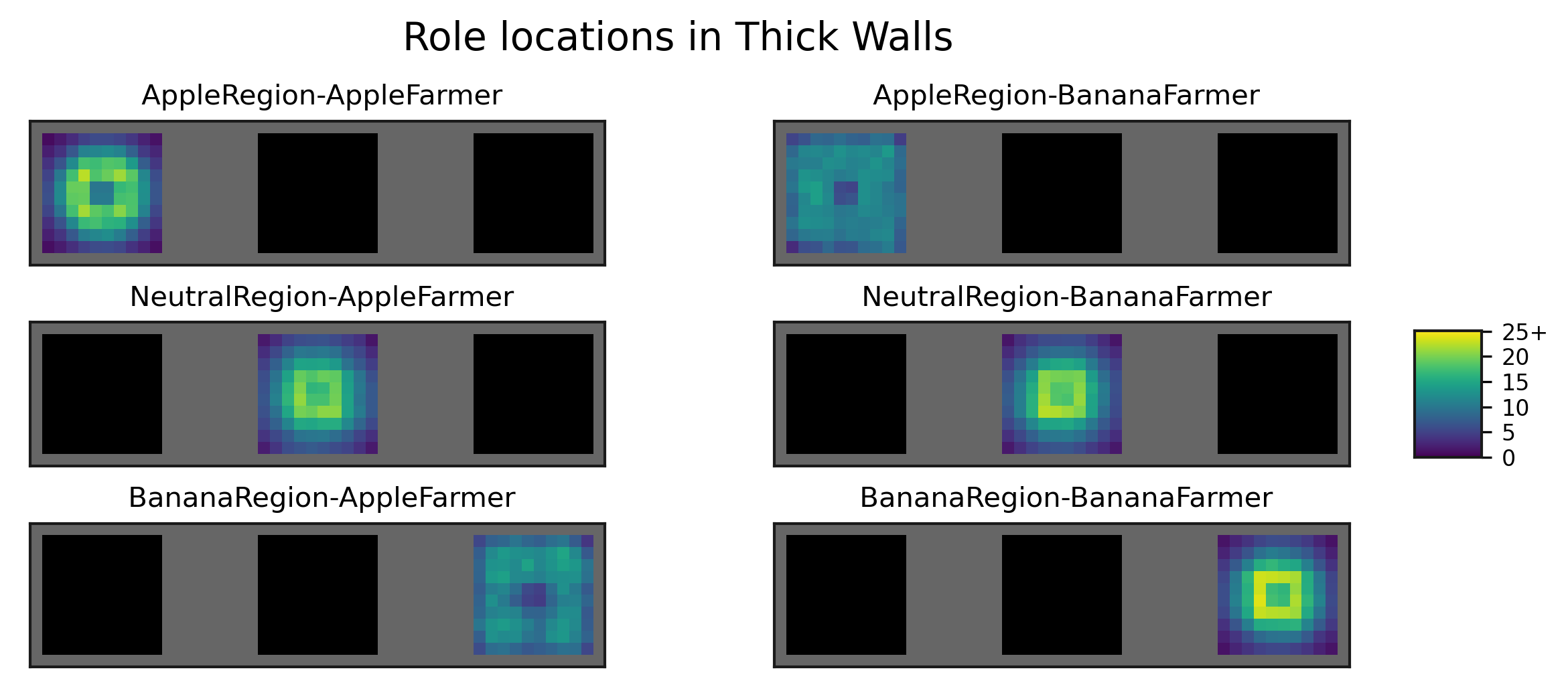}
        \caption{}
        \label{fig:regions:location_heatmaps:thick_walls}
    \end{subfigure}
    
    \caption{Location visitation heatmaps for each role in the ``No Walls'', ``Walls'', and ``Thick Walls'' maps. The color of each map location indicates the average number of timesteps per agent and episode spent in that location. The average is calculated over the final 25\% of training.}
    \label{fig:regions:location_heatmaps}
\end{figure}

Figure~\ref{fig:regions:location_heatmaps} presents the frequency with which agents of each role visit each map location, and each role's behaviour in each map is entirely distinct. In the ``No Walls'' map, we see that all roles visit the entire map, but Apple Farmers (regardless of starting region) primarily visit the apple-rich and neutral regions, and Banana Farmers primarily visit the banana-rich and neutral regions. In the ``Walls'' map, we see that four of the roles -- those specialized to produce a good that is rarer in an adjacent region -- spend most of their time near the border to that region.\footnote{We have set the color map range from 0 to 25+ to more clearly visualize all of the roles' visitation patterns. In the ``Walls'' map, the AppleRegion-AppleFarmer and BananaRegion-BananaFarmer agents spend much of their time -- approximately 100 timesteps per episode -- in just two map locations: the two center rows, in the column adjacent to the wall with the neutral region. Their behaviour is to quickly run through the region to collect many of the fruit they are specialized to produce, and then return to one of those two map locations for many timesteps to sell them. Presenting this heatmap with a linear scale from (0, 100) would obscure the visitation pattern for all of the other roles.} Finally, in the ``Thick Walls'' map, we see visitation that looks roughly gaussian (aside from the center tiles), except for producers of rare goods (AppleRegion-BananaFarmers and BananaRegion-AppleFarmers) whose visitation is more uniform. The center 2-by-2 tiles of each region are reserved for agent spawn points; trees never appear there, and thus players visit these locations less often.

\begin{figure}
    \centering
    \begin{subfigure}{\textwidth}
        \centering
        \includegraphics[height=2in]{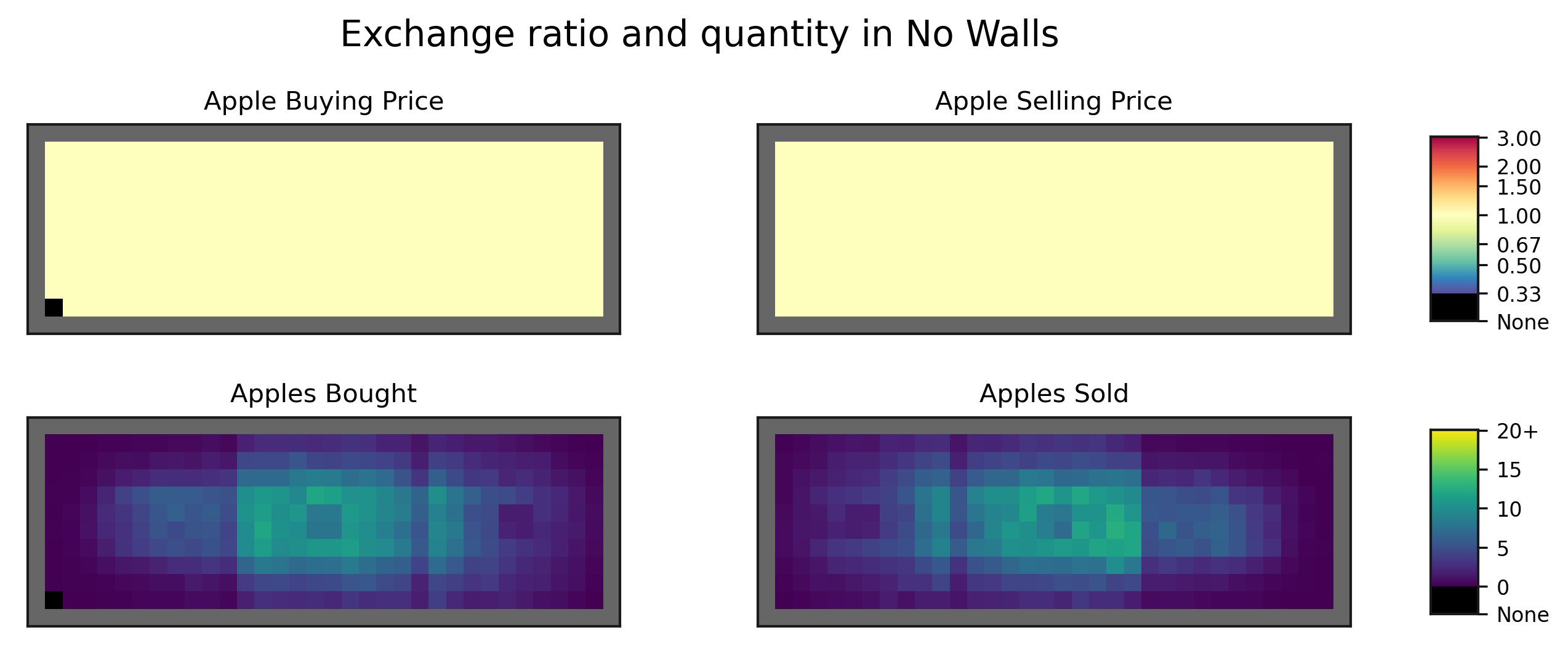}
        \caption{}
        \label{fig:regions:pricemaps:no_walls}
    \end{subfigure}
    
    \begin{subfigure}{\textwidth}
        \centering
        \includegraphics[height=2in]{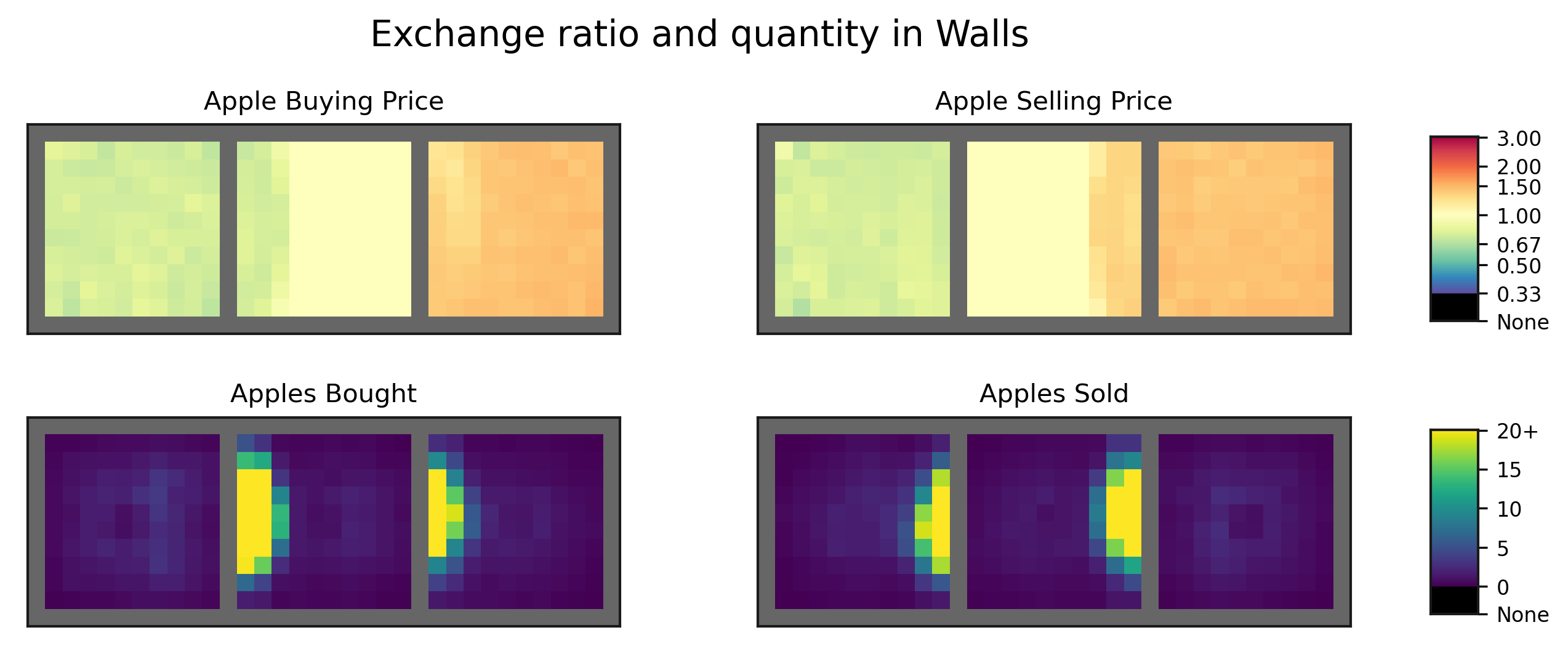}
        \caption{}
        \label{fig:regions:pricemaps:walls}
    \end{subfigure}
    
    \begin{subfigure}{\textwidth}
        \centering
        \includegraphics[height=2in]{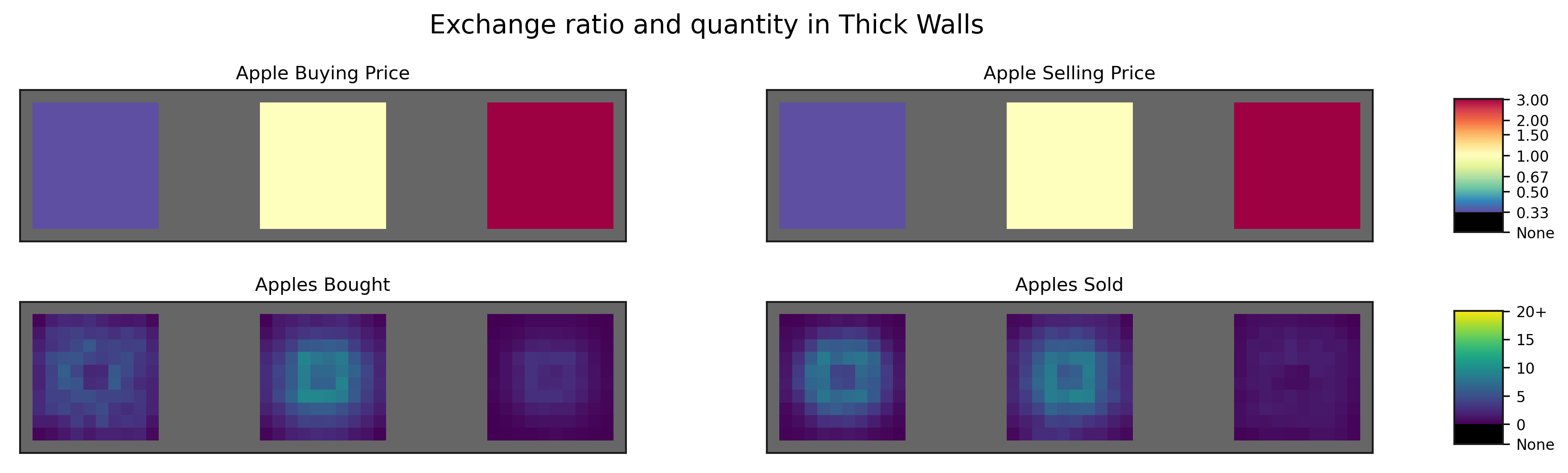}
        \caption{}
        \label{fig:regions:pricemaps:walls_separated}
    \end{subfigure}
    
    \caption{Price and quantity of apples exchanged over space in the ``No Walls'', ``Walls'', and ``Thick Walls'' maps. The price heatmaps indicate the average price (\ie, ratio of bananas per apple) across all exchanges where apples are bought and sold from each map location. The quantity heatmaps indicate the average number of apples bought and sold from each map location, by all players, per episode. Only the final 25\% of training is used for these results, to show the equilibrium behaviour.}
\label{fig:regions:pricemaps}
\end{figure}

The varying visitation patterns in these maps is explained by where trading partners and favourable prices can be found. Figure~\ref{fig:regions:pricemaps} shows heatmaps of the price and quantity of apples exchanged from each map location. In the price heatmaps, the color map uses yellow to indicate prices near 1 where apples and bananas are equally valued, warm colors (orange and then red) to indicate high prices for apples, and cool colors (green and then blue) to indicate low prices for apples.

In the ``No Walls'' price heatmaps, we see that one price emerges across the entire map: one apple for one banana. In the quantity heatmaps, we see that apples are bought and sold across the entire map, but with highest volume in the neutral region, even though apples are much more plentiful in the adjacent apple-rich region. Thus, it appears that the agents learn to produce goods where they can, but then transport them to the neutral region where they are more likely to cross paths with a buyer.

In the ``Walls'' price heatmaps, we see that three price areas emerge: apples are cheap in the apple-rich region and the closest columns of the neutral region, expensive in the banana-rich region and the closest columns of the neutral region, and of equal price in the middle of the neutral region. However, the ``Walls'' quantity heatmaps show that \textit{by far} most of these exchanges occur across the borders of each region, as apples are sold from the apple-rich region to buyers in the neutral region, and from the neutral region to buyers in the banana-rich region. Extremely few exchanges happen in the middle of the neutral region, and so the exchanges are best described as happening at two prices: close to 3 apples for 2 bananas on the left, and 2 apples for 3 bananas on the right.

Finally, the ``Thick Walls'' heatmaps show us that these three regions converge to very different prices if cross-region trading is not possible: apples are traded at the lowest available price of 3 apples for 1 banana in the apple-rich region, the highest available price of 1 apple for 3 bananas in the banana-rich region, and an equal price in the neutral region. Half of the players in each region would thus get a more favourable price if they could trade with players in another region.

\begin{figure}
    \centering
    \begin{subfigure}{0.3\textwidth}
        \begin{flushright}
            \includegraphics[height=2in]{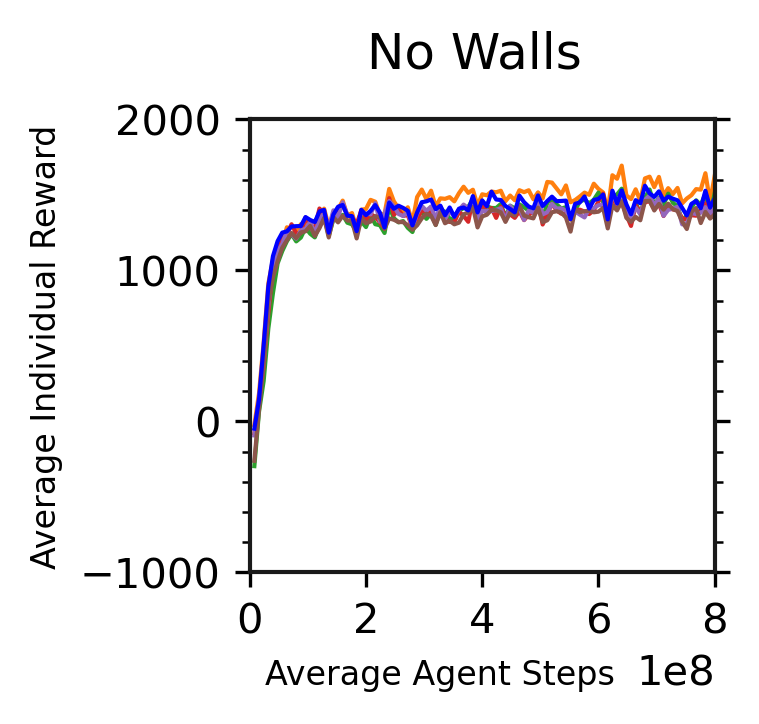}
            \caption{}
            \label{fig:regions:role_returns:no_walls}
        \end{flushright}
    \end{subfigure}%
    ~
    \begin{subfigure}{0.3\textwidth}
        \centering
        \includegraphics[height=2in]{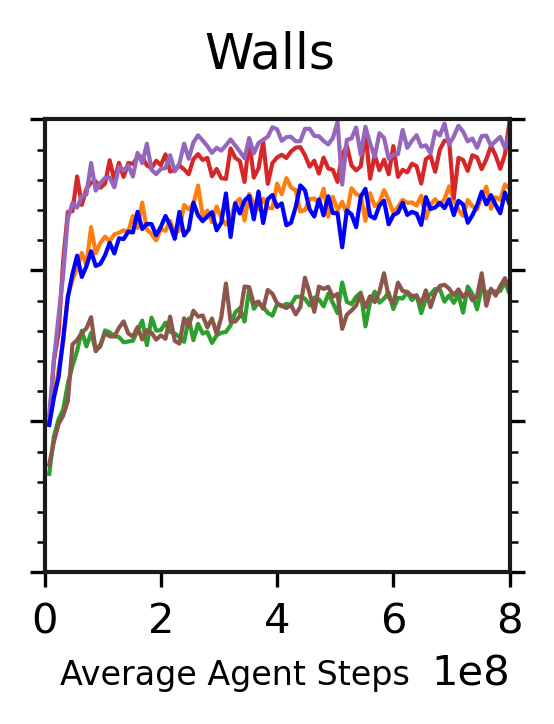}
        \caption{}
        \label{fig:regions:role_returns:walls}
    \end{subfigure}%
    ~
    \begin{subfigure}{0.4\textwidth}
        \begin{flushleft}
            \includegraphics[height=2in]{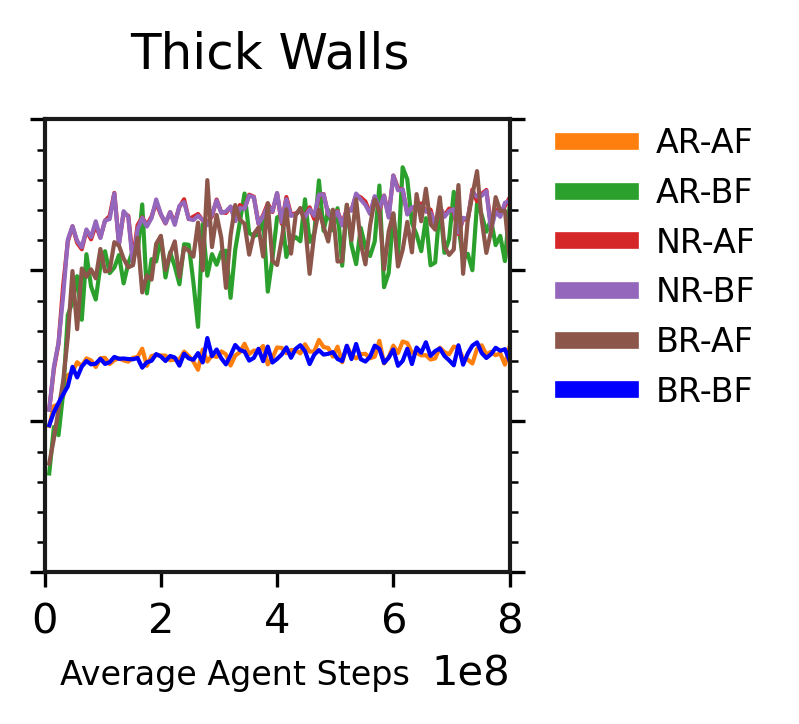}
            \caption{}
            \label{fig:regions:role_returns:thick_walls}
        \end{flushleft}
    \end{subfigure}
    
    \caption{Average individual return by role in the ``No Walls'', ``Walls'', and ``Thick Walls'' maps. Note that the thickness of the walls inverts the ranking of producers of common items (AR-AF, BR-BF) and rare items (AR-BF, BR-AF).}
    \label{fig:regions:role_returns}
\end{figure}

However, while reducing or removing the walls would increase \textit{collective} reward, it would not increase \textit{everyone's} reward. Figure~\ref{fig:regions:role_returns} presents the average individual reward for each role in each map. Note that the producers of rare goods, Apple Region Banana Farmers (AR-BF) and Banana Region Apple Farmers (BR-AF), earn much more reward when cross-region trading is impossible in ``Thick Walls'' than they do in ``Walls''. Similarly, producers of common goods, Apple Region Apple Farmers (AR-AF) and Banana Region Banana Farmers (BR-BF), earn more reward when cross-region trading is possible in either ``Walls'' or ``No Walls'' than in ``Thick Walls''. Neutral Region agents earn the most reward (or tie) in all three maps, but perform best in ``Walls'', when only they have the positional advantage of having the ability to trade with all of the other agents.

\begin{figure}
    \centering
    \begin{subfigure}{\textwidth}
        \centering
        \includegraphics[width=\hsize]{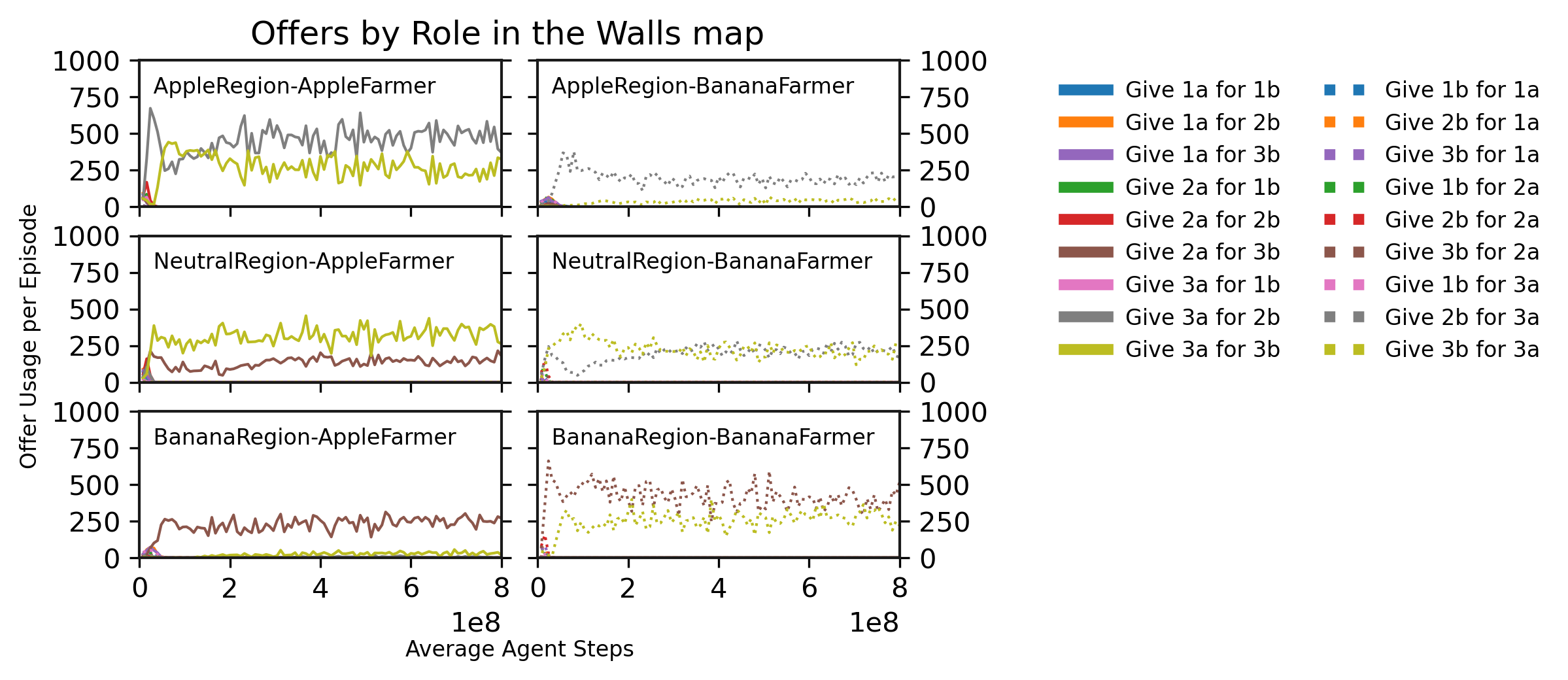}
        \caption{}
        \label{fig:regions:walls_trading:offers}
    \end{subfigure}
    
    \begin{subfigure}{\textwidth}
        \centering
        \includegraphics[width=\hsize]{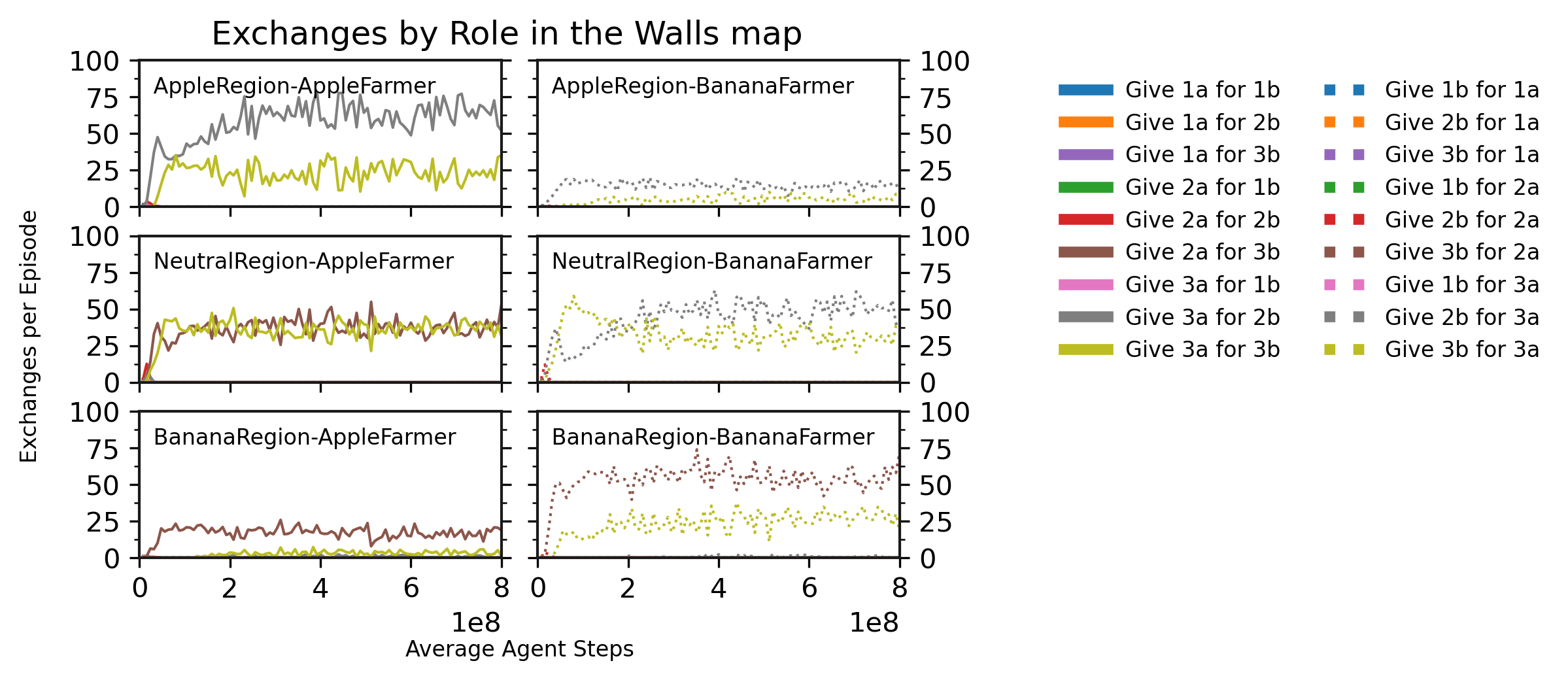}
        \caption{}
        \label{fig:regions:walls_trading:exchanges}
    \end{subfigure}
    
    \caption{Average usage of offers and quantity of exchanges of each type, by role, in the ``Walls'' map.}
    \label{fig:regions:walls_trading}
\end{figure}

We will focus on the ``Walls'' agents, since their trading behaviour is more sophisticated than in the other maps. Figure~\ref{fig:regions:walls_trading} presents the usage of each offer and exchange per episode in ``Walls'', separated out by role. There are several interesting results here. First, focusing on the exchange results in Figure~\ref{fig:regions:walls_trading:exchanges}, we can now see the underlying exchanges that created the price heatmaps presented earlier in Figure~\ref{fig:regions:pricemaps:walls}. Apples are traded with a mixture of offers in each region, resulting in an average price between ``Give 3a for 2b'' (or 0.66) and ``Give 3a for 3b'' (or 1.0) in the Apple Region, and between ``Give 2a for 3b'' (or 1.5) and ``Give 3a for 3b'' (or 1.0) in the Banana Region. Second, we can now see why the Neutral Region roles earn the most reward of all roles. A Neutral Region Apple Farmer performs about three times more exchanges per episode than a Banana Region Apple Farmer, albeit at a lower average price by mixing between exchanges at 3a:3b and 2a:3b instead of only exchanging at 2a:3b. They perform fewer exchanges than an Apple Region Apple Farmer, but those exchanges occur at a better price, since Apple Region Apple Farmers mix between the ``Give 3a for 2b'' (0.66) and ``Give 3a for 3b'' (1.0) offers.

Thus, the Neutral agents have a positional advantage that only they can exploit. However, note that these Neutral agents are still acting as farmers by producing fruit in the neutral region and selling it to a neighboring region. Another strategy, not discovered by these agents, would be to act as merchants by selling apples to the Banana Region Banana Farmers at the ``Give 2a for 3b'' price, and then saving some of those bananas to sell to the Apple Region Apple Farmers at the ``Give 2b for 3a'' price, resulting in a net gain. The agent could then consume some of the excess apples or bananas (according to their role's preference) in payment for their transportation labour, or save them in order to trade in increasing volume in each loop between the regions, thus amortizing their movement costs across more exchanges. 

We know that the Neutral agents have not discovered this behaviour, because it would appear in Figure~\ref{fig:regions:walls_trading:exchanges} as one role's plot containing both solid and dotted lines: buying and selling both goods. Instead, all six of the roles either only buy apples or only sell apples (\ie, only have solid lines or dotted lines), but not both. There are several reasons why the Neutral agents behaviour might still be reasonable, however. For example, consider a Neutral Region Apple Farmer: if enough apples are easily produced in the Neutral region to satisfy demand in the Banana Region, then there would be no benefit in selling some precious bananas to the Apple Region in order to get even more apples. Further, buying and selling apples between the regions might only be profitable enough to cover transportation costs if the difference in prices is wide, and these results show that the prices in each region only slightly differ from 1.0. 

\subsubsection{Quantifying the Neutral Region Advantage}
\label{sec:experiments:regions:neutral}

In this section we will quantify the Neutral agents' positional advantage by making their region less abundant in resources. Specifically, we will \textbf{penalize} the probability of trees appearing in the neutral region, by multiplying apple and banana tree spawning probabilities by a constant such as $\times1.0$ (no penalty), or $\times0.5$ (half as many trees as normal). This lets us state the positional advantage in terms of the underlying productivity of the region: for example, the positional advantage is about equal to only having one quarter the number of trees to harvest. This experiment will also reveal cases where our agents do learn merchant behaviour: when agents cannot easily produce their own goods to sell, and fewer exchanges lead to a wider price difference between the markets, the neutral agents begin acting as merchants.

\begin{figure}
    \centering
    \includegraphics[height=2.5in]{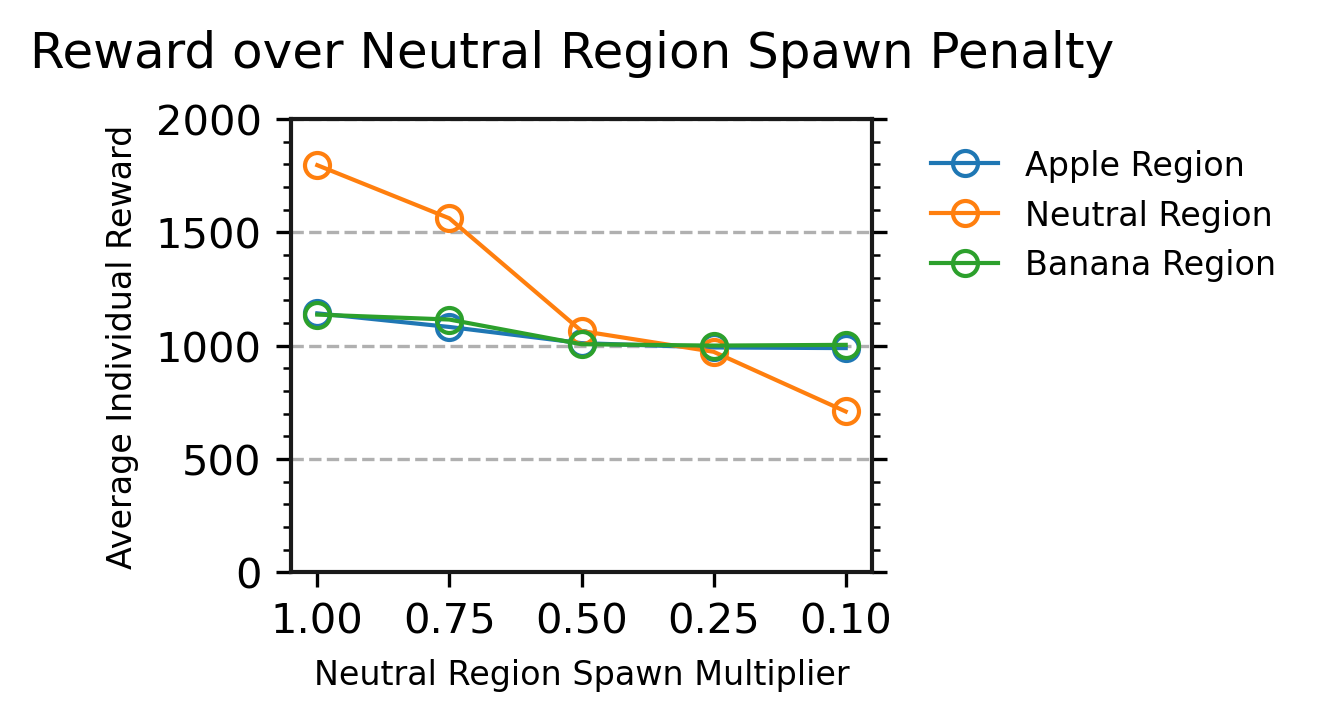}
    \caption{Average reward of agents in each region, as the Neutral region's tree spawn rate is penalized. Average individual reward is averaged over the final 25\% of training, when agents have experienced between $6e8$ and $8e8$ timesteps, to highlight their equilibrium behaviour.}
    \label{fig:regions:neutral_range}
\end{figure}

Figure~\ref{fig:regions:neutral_range} presents the average episodic reward of agents in each region, as we vary the neutral region tree penalty term. The reward values are computed over the final 25\% of training (\ie when agents have experienced $6e8$ to $8e8$ timesteps). We found this graph surprising, as the neutral agents' positional advantage was stronger than we had expected. 
Even with just one quarter the trees, the Neutral agents earn just slightly less reward than their neighbours. In exact numbers, the Neutral agents drop from 1797 reward per episode at $\times1.0$ to 973 reward per episode at $\times0.25$: more than half the reward, with one quarter the trees. Even reducing the trees by one quarter at $\times0.75$ causes a noticeable decrease in the Neutral agents' reward, so it is not true that the $\times1.0$ case provided far more trees than the agents could practically harvest. 

Even with the extreme $\times0.1$ penalty, the Neutral agents still earn 709 per episode: 39\% of what they earn at $\times1.0$. Note that the $\times0.1$ multiplier is quite harsh. Each region has 96 tiles where trees can spawn; at a base spawn rate for any type of tree of $\times0.3$ and a penalty of $\times0.1$, only an average of 2.88 trees (half apple, half banana) will appear in the entire Neutral region, to support four players. The production of those trees is not even enough to stave off the hunger penalty that the four players will face.\footnote{From the environment constants described in Section~\ref{sec:environment}, each tree produces at most 2 fruit every 50 timesteps. Thus, 2.88 trees will produce 0.12 fruit per timestep, if harvested instantly. Four players require 1 fruit each every 30 timesteps to avoid the hunger penalty, requiring 0.13 fruit per timestep.} 

The step from $\times0.5$ to $\times0.25$ is also interesting: the Neutral agents have half as many trees to draw from, yet their reward only decreases from just above 1000 to just below 1000 per episode. As we will see in our next results, this is due to a phase shift in the neutral agents' behaviour from primarily farming goods for sale at $\times0.5$ and above, to primarily buying, selling, and transporting goods at $\times0.25$ and below.

\subsubsection{Emergence of Merchant Behaviour}
\label{sec:experiments:regions:merchant}

\begin{figure}
    \centering
    \includegraphics[width=\textwidth]{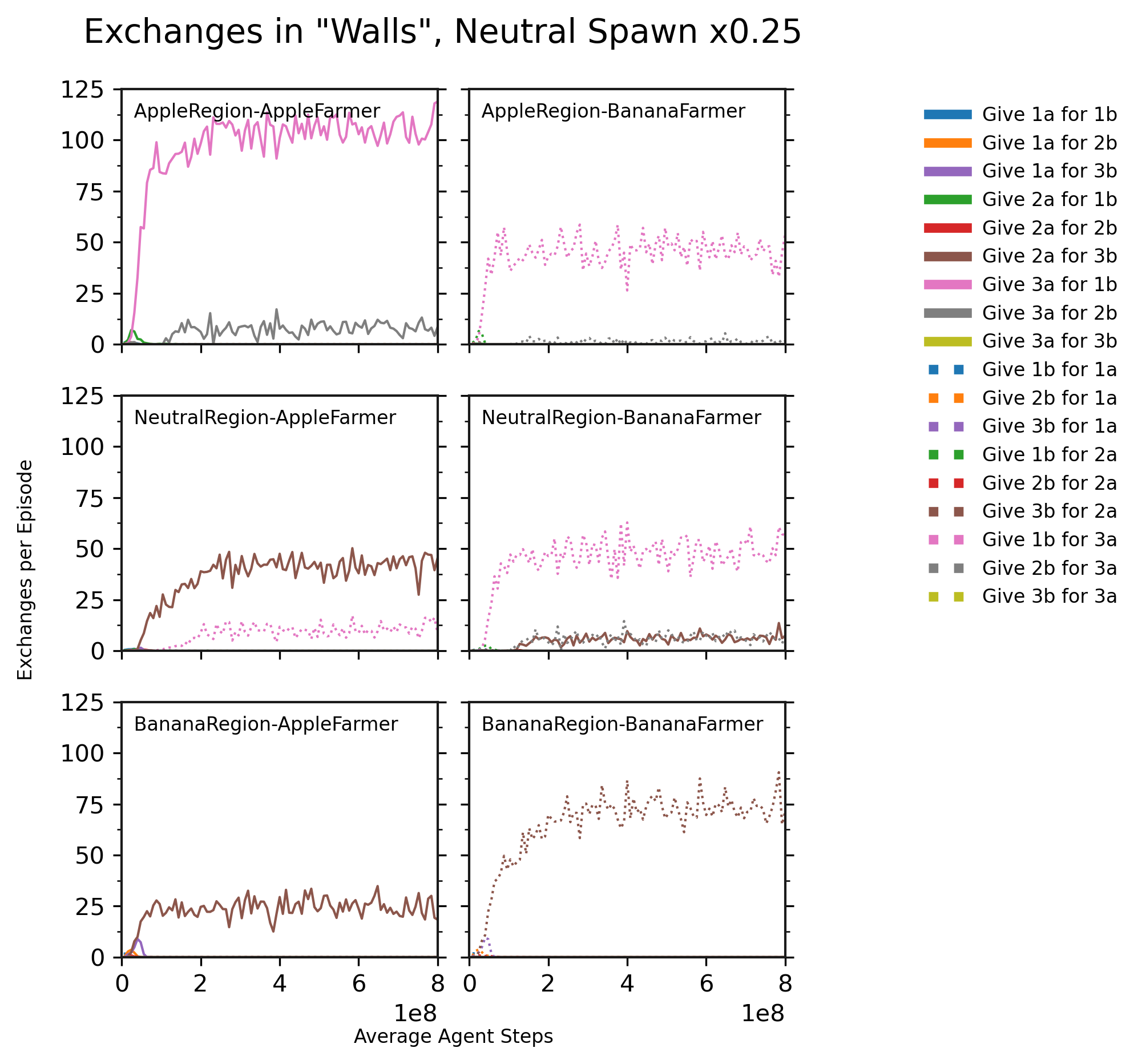}
    \caption{Exchanges using each offer per episode, by role, when the neutral penalty is $\times0.25$.}
    \label{fig:regions:neutral_25_role_exchanges}
\end{figure}

Our next results will focus on the $\times0.25$ setting where this phase shift occurs. Figure~\ref{fig:regions:neutral_25_role_exchanges} presents the exchanges made by agents of each role under this neutral penalty. Recall that in our earlier ``Walls'' $\times1.0$ results in  Figure~\ref{fig:regions:walls_trading:exchanges}, where each role either bought or sold each good, this was visualized by each plot containing only solid or dotted lines, but not both. However, with the $\times0.25$ penalty, we now observe different behaviour: the Neutral Apple Farmers sell apples using the ``Give 2 apples for 3 bananas'' offer, but also buy some apples using the ``Give 1 banana for 3 apples'' offer. Neutral Region Banana Farmers sell bananas mostly with the ``Give 1 banana for 3 apples'' offer and sometimes with the ``Give 2 bananas for 3 apples'' offer, but also buy some bananas using ``Give 2 apples for 3 bananas''.

Note that the difference in height between lines in the plots may understate the importance of agents buying the goods their role can produce, because of the quantities involved in each exchange. Consider the Neutral Region Apple Farmers. Over the final 25\% of training, these agents on average perform 42.9 exchanges per episode with the ``Give 2 apples for 3 bananas'' offer (thus selling 85.8 apples to buy 128.7 bananas), and 11.0 exchanges with the ``Give 1 banana for 3 apples'' offer (thus selling 11.0 bananas to buy 33.1 apples). These agents produce 56.7 apples per episode, and so 33.1 out of 89.8, or 37\%, of the apples obtained by Neutral Apple Farmers are \textit{bought} instead of produced. A complete table of these numbers is presented in Table~\ref{tab:regions:wall_penalty_25}.

The Neutral Banana Farmers' behaviour is similar. On average, they perform 48.7 exchanges of ``Give 1 banana for 3 apples'', 6.7 exchanges of ``Give 2 bananas for 3 apples'', and 6.4 exchanges of ``Give 2 apples for 3 bananas'', resulting in 166.4 apples bought, 12.7 apples sold, 19.1 bananas bought, and 62.2 bananas sold. They produce 49.1 bananas per episode, and thus 28\% of bananas sold were obtained through trade instead of production.\footnote{Note that for Apple Farmers, apples produced plus apples bought is greater than apples sold, and the same is true for Banana Farmers and bananas. The two quantities are not equal because Apple Farmers still consume some apples (\eg, to ward off hunger, or to gain the small reward for apple consumption), and the episode might end with some fruit left in the agent's inventory.}

\begin{figure}
    \centering
    \includegraphics[width=\textwidth]{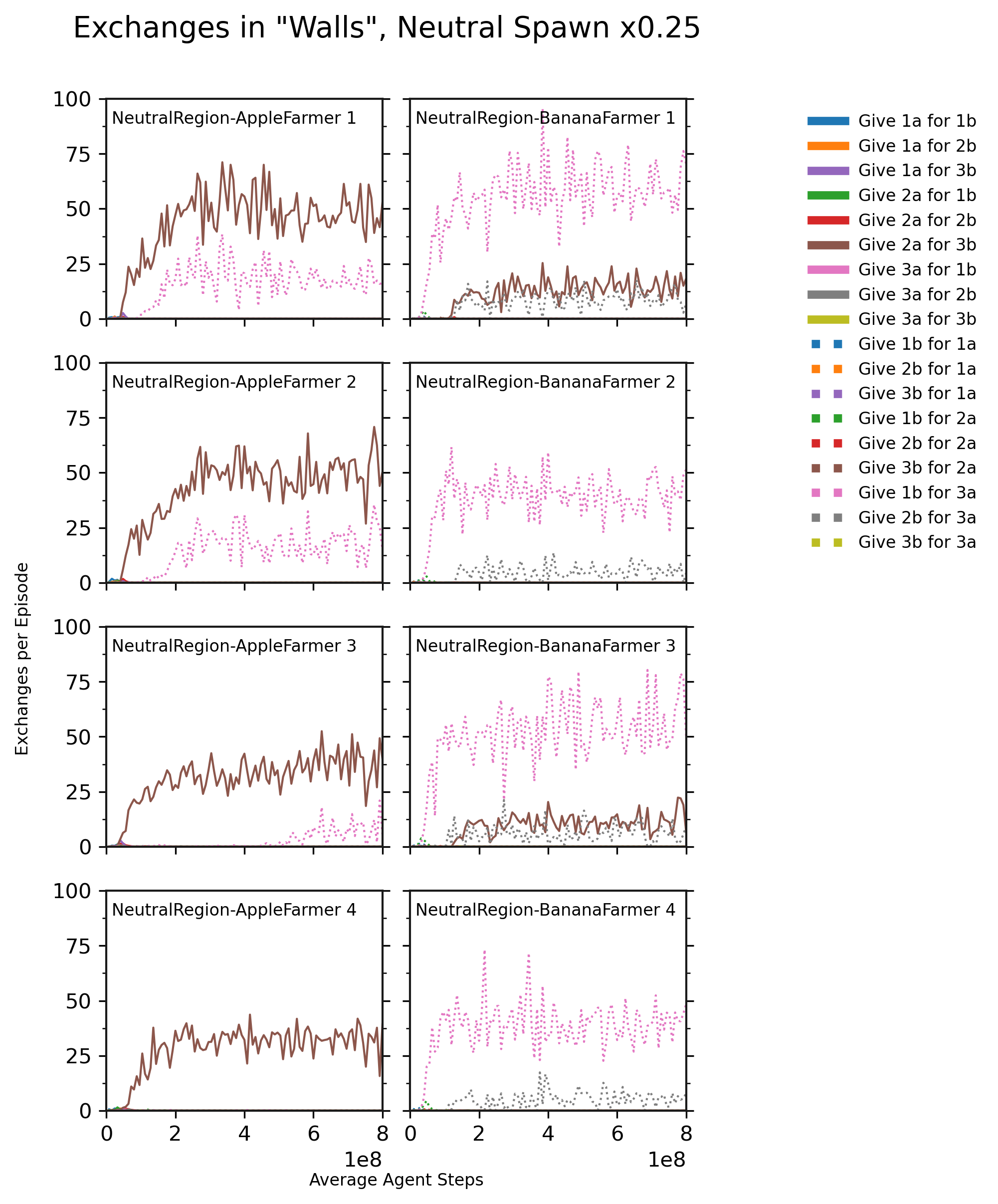}
    \caption{Exchanges using each offer per episode, per Neutral agent, when the neutral penalty is $\times0.25$.}
    \label{fig:regions:neutral_25_agent_exchanges}
\end{figure}

However, these results are more interesting when viewed for each individual agent, instead of averaging together those agents into roles. Figure~\ref{fig:regions:neutral_25_agent_exchanges}, shows all four neutral Apple Farmers in the left column, and all four neutral Banana Farmers in the right column. Note that three out of four Apple Farmers learn to buy and sell apples (NR-AF 1 and NR-AF 2, with NR-AF 3 starting late in training), and one only sells apples (NR-AF 4). Similarly, two of the Banana Farmers learn to buy and sell bananas (NR-BF 1 and 3), while the other two only sell bananas (NR-BF 2 and 4). Each of the agents that learns to buy goods to supplement their production is also able to sell much more than those who only produce goods. In the final 25\% of training, the Neutral Apple Farmers sell 94.5, 103.1, 78.5, and 66.4 apples respectively, with the first two agents having learned this behavior early, and the third learning it late. The Neutral Banana Farmers sell 78.5, 51.0, 69.4, and 50.6 bananas respectively, with the first and third being the agents that buy bananas. The statistics describing this flow of goods is  provided below in Table~\ref{tab:regions:wall_penalty_25}.

\begin{figure}
    \centering
    \begin{subfigure}{\textwidth}
        \centering
        \includegraphics[height=4in]{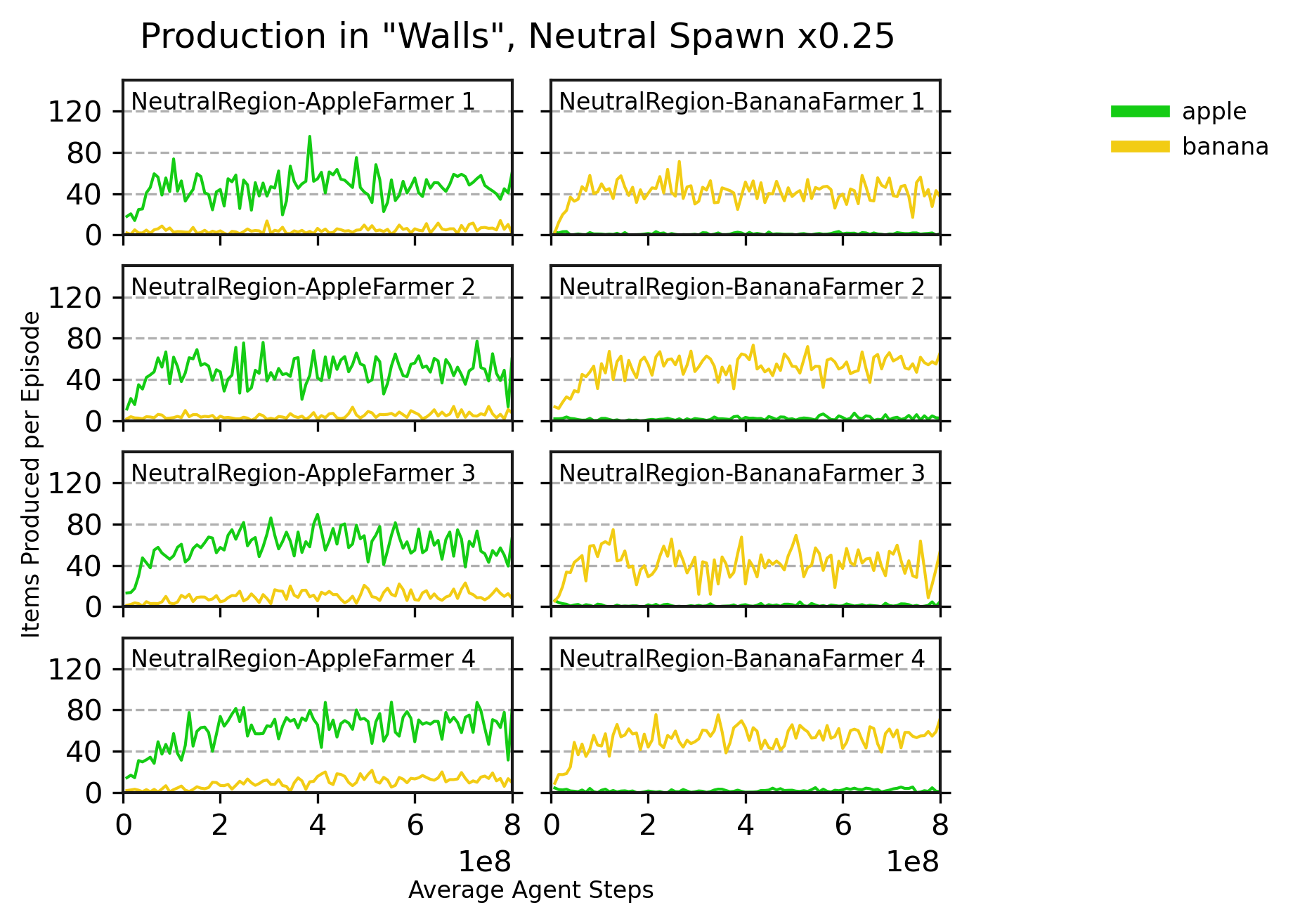}    
        \caption{}
        \label{fig:regions:neutral_25_prod_con:production}
    \end{subfigure}
    
    \begin{subfigure}{\textwidth}
        \centering
        \includegraphics[height=4in]{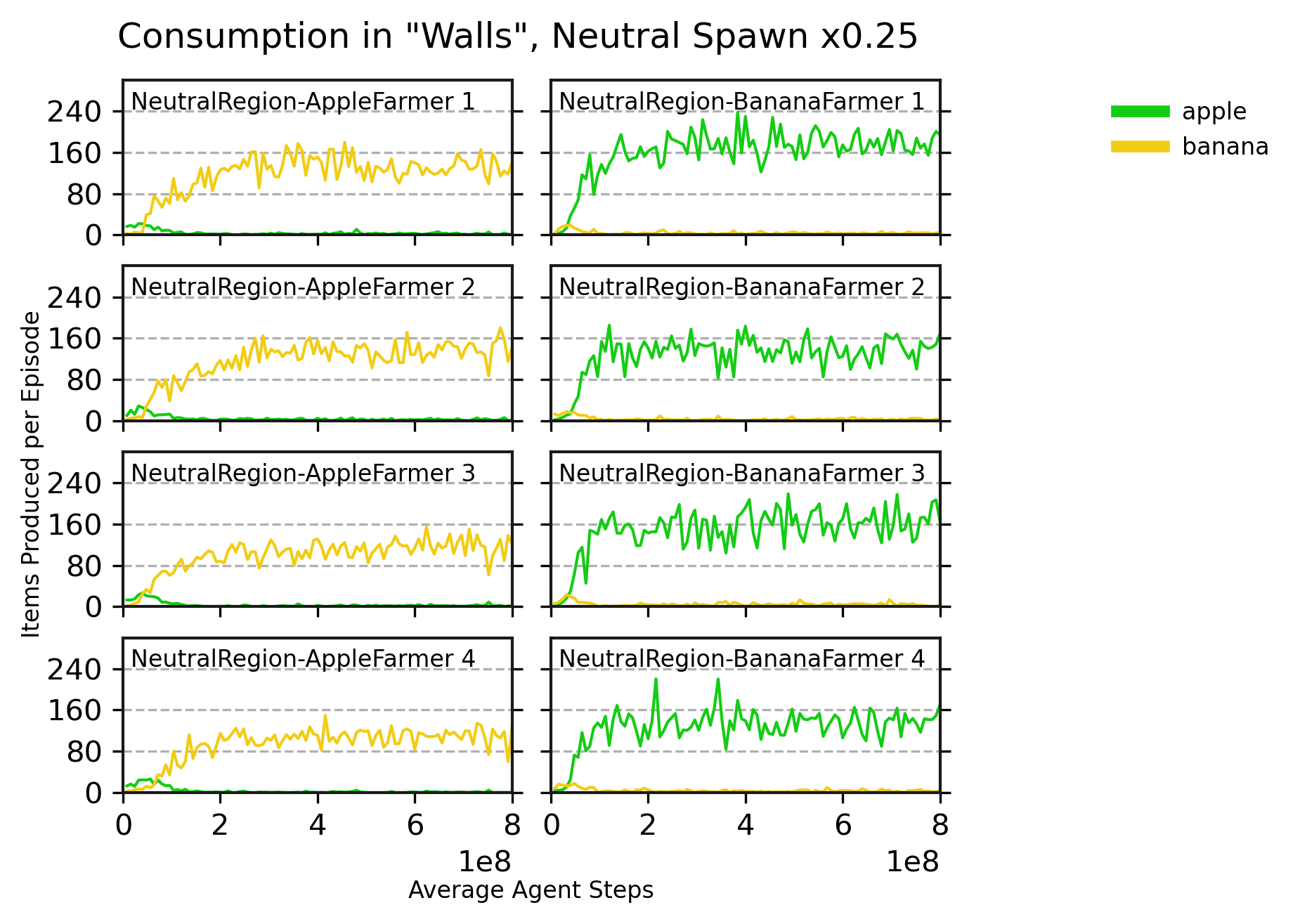}    
        \caption{}
        \label{fig:regions:neutral_25_prod_con:consumption}
    \end{subfigure}
    \caption{Production and Consumption of Neutral agents when the Neutral penalty is $\times0.25$.}
    \label{fig:regions:neutral_25_prod_con}
\end{figure}

Is this purchasing behaviour a significant part of the agents' strategy? Yes. Figure~\ref{fig:regions:neutral_25_prod_con} shows the number of apples and bananas produced and consumed by each neutral agent. In the final 25\% of training the Apple Farmers who buy apples, NR-AF 1 and 2, produce 47.7 and 51.3 apples and buy 51.8 and 56.8, respectively. The Apple Farmers who do not (or only begin late in training), NR-AF 3 and 4, produce 58.7 and 68.6 apples, and buy 23.7 and 0.0 apples, respectively. Apple Farmers 1 and 2 consume 129.1 and 140.3 bananas, while 3 and 4 consume 120.0 and 112.0 bananas. Thus, the agents that learn to buy apples also produce fewer apples; overall, Apple Farmers 1 and 2 gain 52\% and 53\% of their apples through purchase instead of production. This suggests that the apple buying is not merely incidental, a behaviour used in addition to production, but is instead a reallocation of how they spend their time. The apple-buying agents also consume more bananas than their peers.

The trend for Banana Farmers is the same: agents 1 and 3 produce 40.9 and 43.7 bananas, buy 44.6 and 33.2 bananas, and consume 177.4 and 165.5 apples; agents 2 and 4 produce 55.6 and 55.5 bananas, buy 0.0 and 0.0 bananas, and consume 137.7 and 137.2 apples. Banana Farmers 2 and 4 gain 52\% and 43\% of their total bananas, respectively, through purchase instead of production. The Banana Farmers that learn to buy bananas also produce fewer bananas and consume more apples than their peers.

\begin{figure}
    \centering
    \begin{subfigure}{\textwidth}
        \centering
        \includegraphics{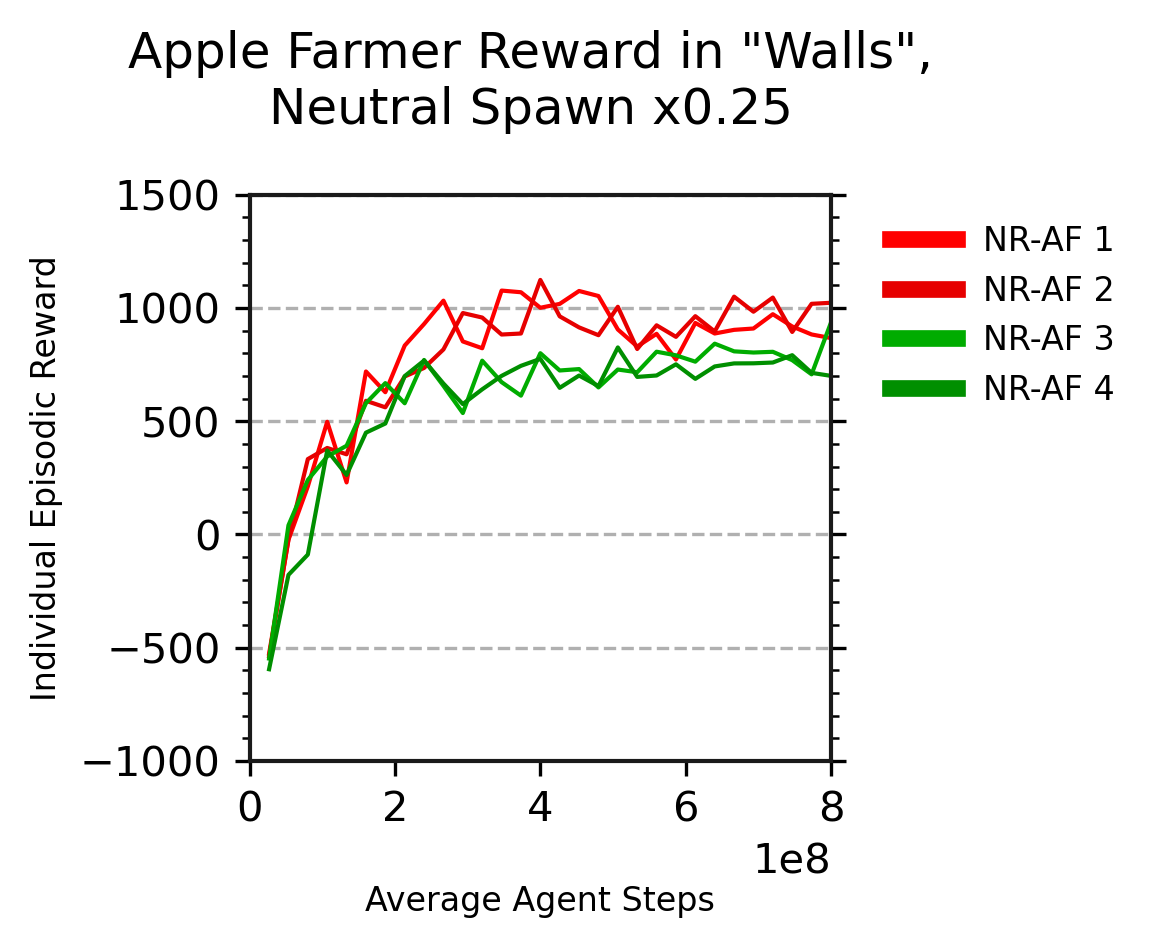}
        \caption{}
        \label{fig:regions_neutral_25_agent_reward:af}
    \end{subfigure}
    
    \begin{subfigure}{\textwidth}
        \centering
        \includegraphics{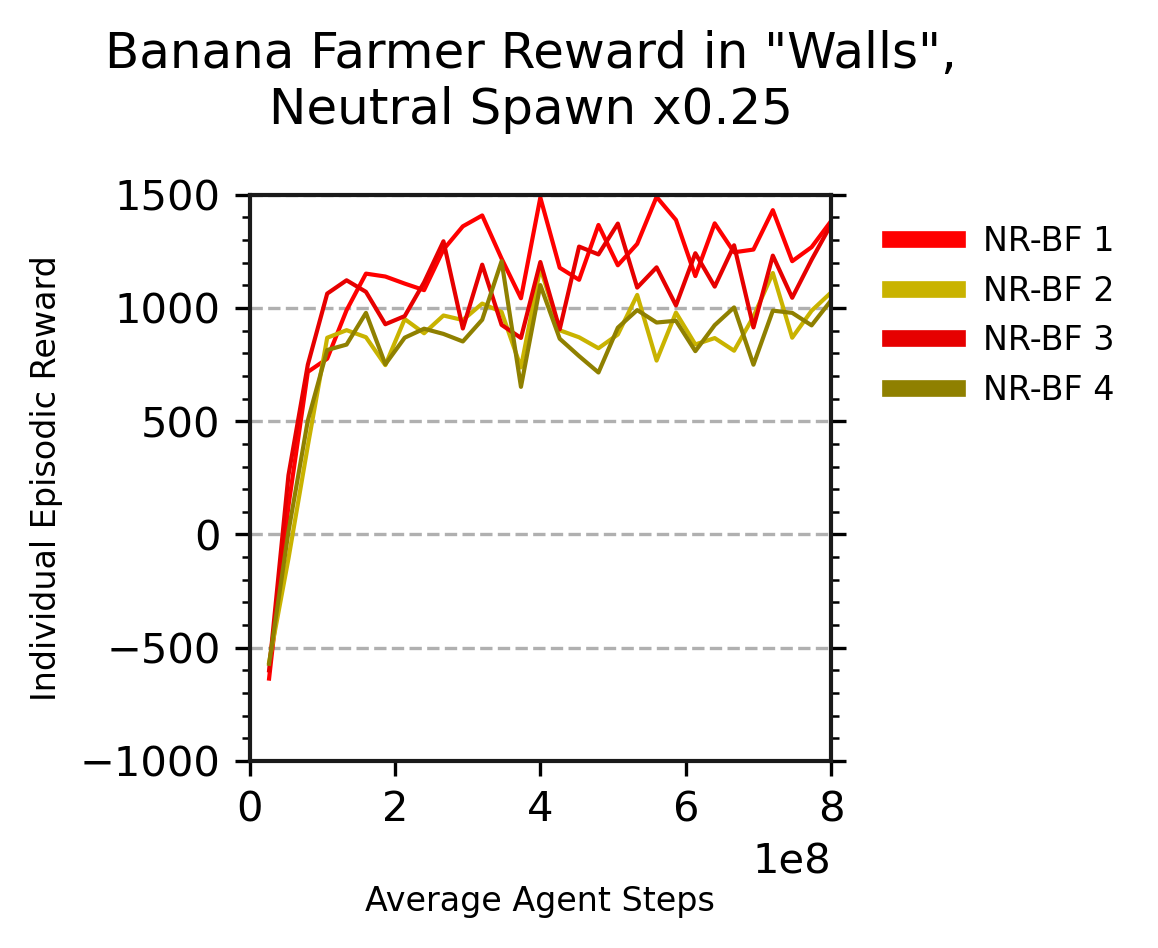}
        \caption{}
        \label{fig:regions_neutral_25_agent_reward:bf}
    \end{subfigure}
    \caption{Episodic reward of Neutral Apple Farmers (a) and Banana Farmers (b) when the Neutral penalty is $\times0.25$. The agents indicated by red lines are those whose behaviour includes substantially buying the good that their role can produce efficiently. Agents indicated by green lines and yellow lines are those that only sell the apples or bananas, respectively, that they produce themselves.}
    \label{fig:regions_neutral_25_agent_reward}
\end{figure}

One last question remains: is this apple-buying and banana-buying beneficial? From our calculations above, the agents that buy and produce goods also consume more of the goods that they prefer, which suggests that they will earn more reward. However, buying and selling could also have higher associated costs: for example, from the movement penalty for repeatedly crossing the neutral region, or possibly from suffering the hunger penalty since they must save some fruit for sale instead of consumption. Figure~\ref{fig:regions_neutral_25_agent_reward} answers this question by presenting the average episodic reward of each Neutral Apple Farmer and Banana Farmer. The agents that learned to buy items (NR-AF 1 and 2, NR-BF 1 and 3) are shown with red lines, while the other agents are shown by green and yellow lines depending on their role. 

All of the agents who learned to both buy and produce their goods earned substantially more reward. Apple Farmers 1 and 2 earned 900.2 and 990.1 reward per episode (average 945.2) compared to 3 and 4 who earned 804.2 and 747.9 (average 776.1); Banana Farmers 1 and 3 earned 1289.0 and 1185.1 (average 1237.1), while 2 and 4 earned 940.8 and 927.4 (average 934.1)\footnote{The asymmetry in these figures, where the neutral Banana Farmers as a group earned more than the neutral Apple Farmers, is explained by the difference in prices that the other regions has converged to, which we presented in Figure~\ref{fig:regions:neutral_25_role_exchanges}. In the Apple Region, the exchanges are mixed between predominantly ``3 apples for 1 banana'' with some ``3 apples for 2 bananas''. In the Banana Region, the exchanges instead exclusively occur at the ``3 bananas for 2 apples'' price. There is no inherent reason why both sides must or should converge to the same price: they are distinct populations who learn through a stochastic process. These particular prices happen to be more beneficial for the neutral Banana Farmers, who consume apples bought cheaply from the Apple Region, than for the neutral Apple Farmers, who consume bananas bought at twice the price from the Banana Region.}.

In summary, we have now shown that with a neutral penalty of $\times0.25$, this ``merchant behaviour'', or (using the term informally) arbitrage behaviour, emerges among half of the Neutral agents of each role. The agents who develop it obtain about half of their items through purchase instead of production, and they obtain substantially more reward as a result: 22\% more for Apple Farmers, and 32\% more for Banana Farmers. Even though the environment assigned these agents the role of Apple Farmers and Banana Farmers, from their learned behaviour, they would be better described as Apple Merchants and Banana Merchants. We consider the emergence of this behaviour to be a key result in this work. 

In our earlier results in this paper, we demonstrated that agents could learn to produce and consume goods. After the population had learned this, the agents could learn the further behaviour of producing their role's specialized goods efficiently, in order to trade them for goods that were more rewarding. And after trading behaviour emerged in the population, agents could go farther by changing their offers to demand better prices for rare goods, or offer cheaper prices for plentiful goods, to earn even more reward. The arbitrage results in this section demonstrate another behaviour: once the population has negotiated regional prices reflecting local abundance, some agents can specialize in transporting goods to exploit that price difference. This requires an agent to learn to give up the most rewarding item they could possibly consume, a short term loss, in order to obtain a higher long-term reward. While this is exactly the trade-off that reinforcement learning agents are designed to make, by learning to approximate the long-term discounted reward of different behaviours, it can be quite difficult to elicit this behaviour in practice. But here, our agents learn all of these layers on layers of microeconomic behaviour, only from their stream of observations, actions, and rewards, with very little domain knowledge injected into the environment (including the actions and observations) and none into the agents themselves.

\subsubsection{Merchant Behaviour in Other Settings}
\label{sec:experiments:regions:merchant_other}

We have shown that only one global price emerges in ``No Walls'', local prices do emerge in ``Walls'' with a $\times1.0$ neutral penalty but merchant behaviour does not emerge, and local prices and merchant behaviour does emerge in ``Walls'' with a $\times0.25$ neutral penalty. Following the same analysis described above for the $\times0.25$ penalty, we have not found any emergence of merchant behaviour in ``Walls'' with penalties of $\times0.75$ or $\times0.5$. Recall from Figure~\ref{fig:regions:neutral_range} that the Neutral agents' reward only decreased slightly when the abundance of trees was cut in half in moving from $\times0.5$ to $\times0.25$. The emergence of merchant behaviour in half of the agents is the reason that the difference in reward is so small, making up for the loss of half of the reward that we might otherwise expect.

\begin{figure}
    \centering
    \begin{subfigure}{\textwidth}
        \centering
        \includegraphics{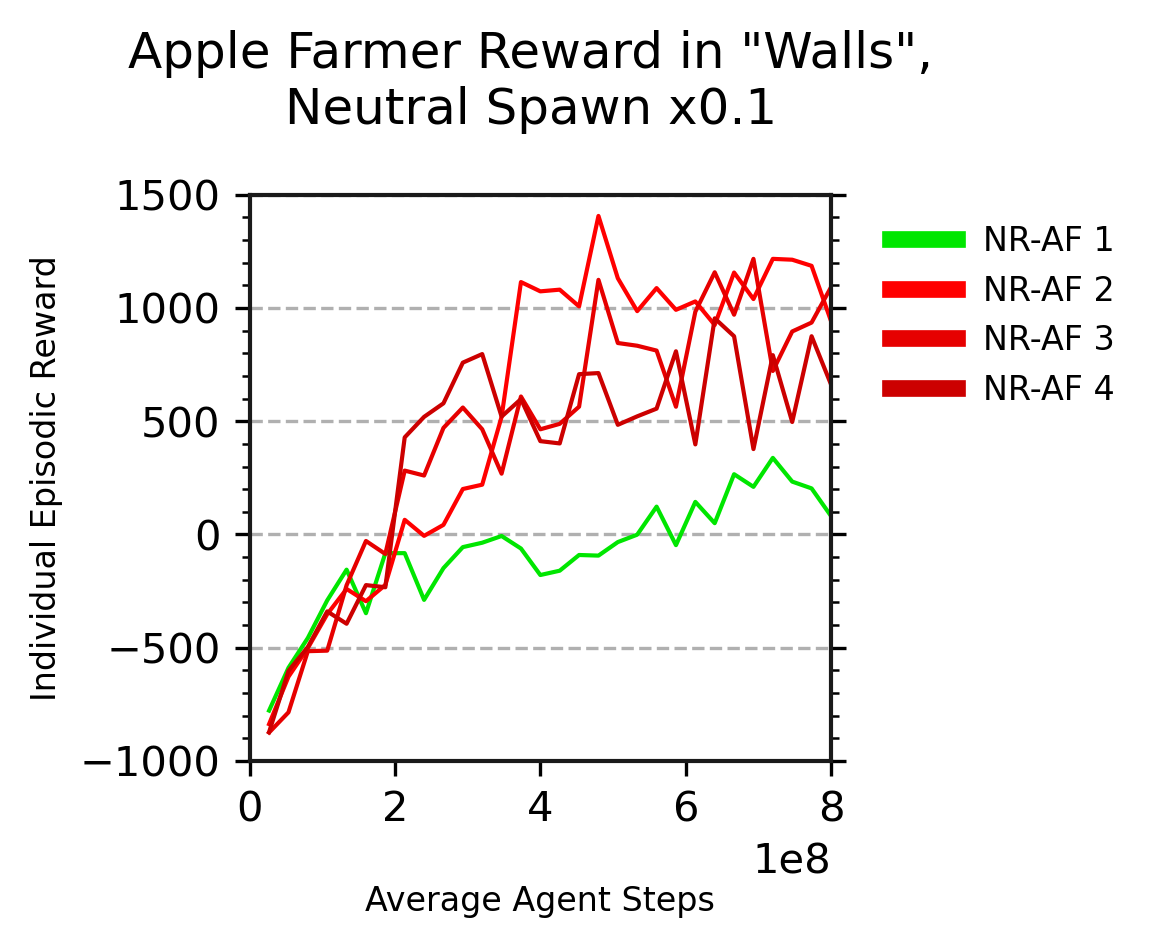}
        \caption{}
        \label{fig:regions_neutral_10_agent_reward:af}
    \end{subfigure}
    
    \begin{subfigure}{\textwidth}
        \centering
        \includegraphics{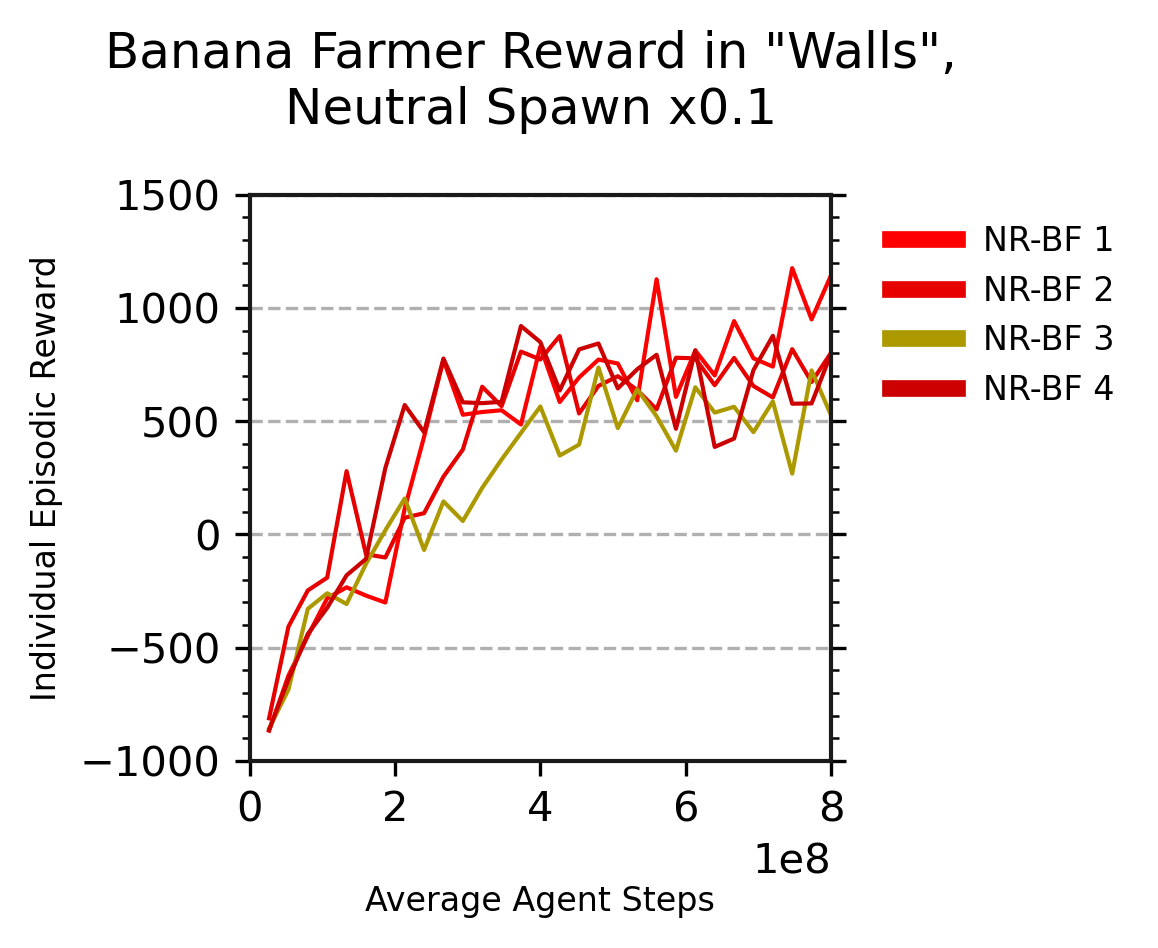}
        \caption{}
        \label{fig:regions_neutral_10_agent_reward:bf}
    \end{subfigure}
    \caption{Episodic reward of Neutral Apple Farmers (a) and Banana Farmers (b) when the Neutral penalty is $\times0.1$. The agents indicated by red lines are those whose behaviour includes buying the good that their role produces. Agents indicated by green lines and yellow lines are those that only sell the apples or bananas, respectively, that they produce themselves.}
    \label{fig:regions_neutral_10_agent_reward}
\end{figure}

Does merchant behaviour emerge in other settings? We can confirm that it also emerged in the ``Walls'' $\times0.1$ penalty experiment: three out of four neutral agents of each role developed it. Figure~\ref{fig:regions_neutral_10_agent_reward} shows that the three merchant agents of each role (represented by red lines) outperform the one agent of each role that does not (represented by green and yellow lines). Merchant behaviour is particularly important with the $\times0.1$ penalty, since only 2.88 trees of either type will appear on average in the neutral region, to be fought over by four agents. Further, since the placement and number of trees is stochastic, there will often be episodes with only two, one, or occasionally even zero trees spawned in the neutral region, leading to catastrophic hunger penalties by all Neutral agents. From the statistics presented in Table~\ref{tab:regions:wall_penalty_10}, we see that the Apple Farmers who bought apples gained 80\% of their apples through trade and 20\% through production, and Banana Farmers who bought bananas gained 69\% of their bananas through trade and 31\% through production, compared to the 53\% and 48\% from trade that we saw in the $\times0.25$ setting. This is why we focused our initial analysis on the $\times0.25$ setting, where producing goods to sell remains a viable alternative, and some agents can discover becoming merchants instead of being forced into it.

\begin{figure}
    \centering
    \begin{subfigure}{\textwidth}
        \centering
        \includegraphics[height=2in]{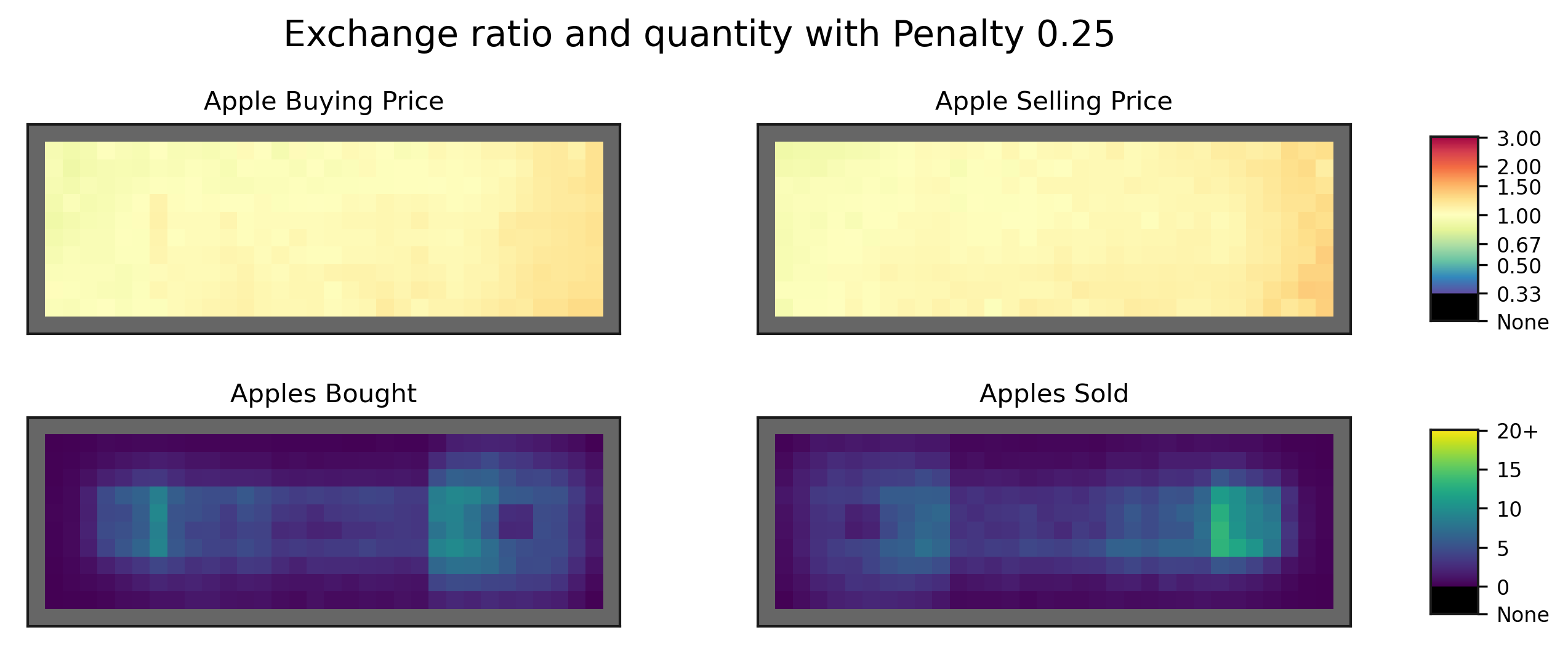}
        \caption{}
        \label{fig:regions:no_walls_penalty_pricemaps:25}
    \end{subfigure}
    
    \begin{subfigure}{\textwidth}
        \centering
        \includegraphics[height=2in]{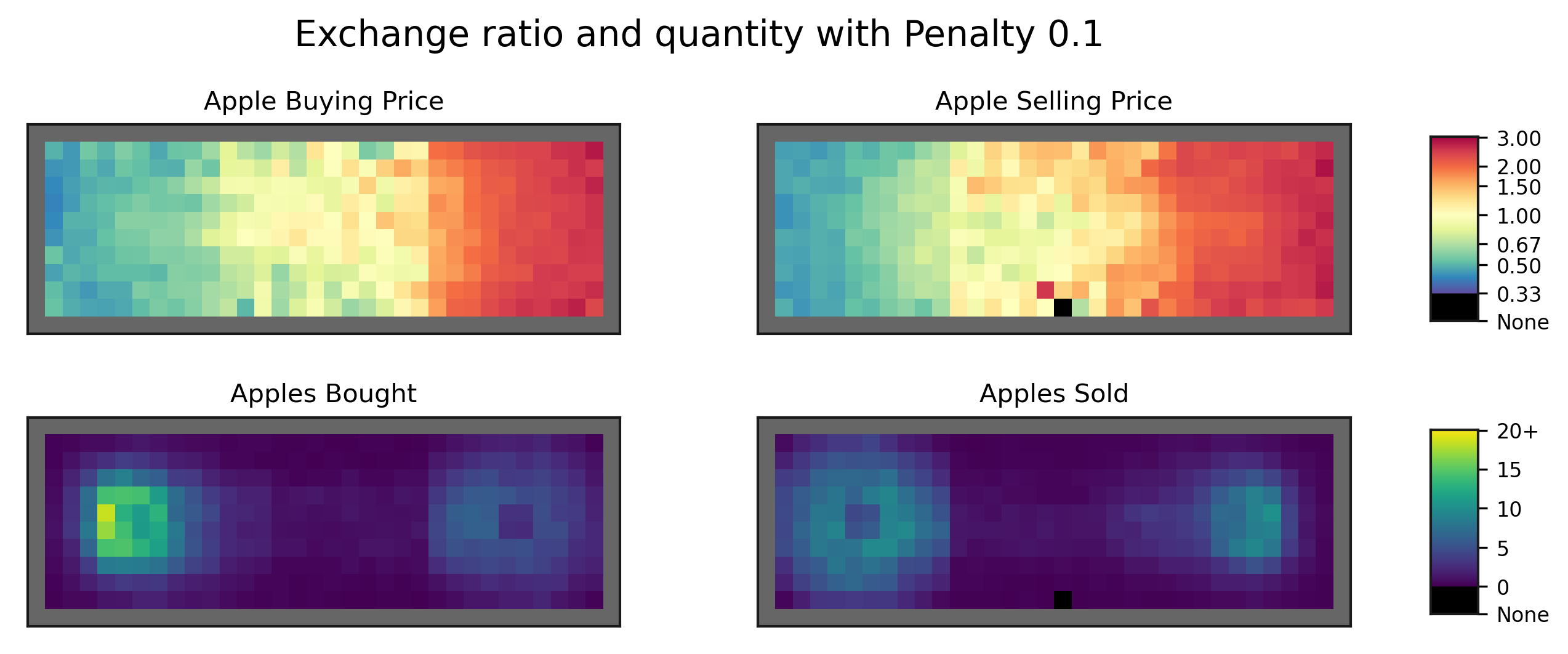}
        \caption{}
        \label{fig:regions:no_walls_penalty_pricemaps:10}
    \end{subfigure}
    \caption{Price and quantity heatmaps in the ``No Walls'' map, with Neutral region penalties of $\times0.25$ and $\times0.1$.}
    \label{fig:regions:no_walls_penalty_pricemaps}
\end{figure}

We might conclude from these results that the emergence of merchant behaviour requires both access to regions with different prices, and a relative scarcity of resources so that agents cannot simply produce all of the goods they would want to sell. If the neutral penalty was successful in creating those conditions and eliciting merchant behaviour in ``Walls'', then would a similar approach work in ``No Walls'', where all agents can traverse the entire map?  Figure~\ref{fig:regions:no_walls_penalty_pricemaps} presents price and quantity heatmaps of that experiment, with neutral penalties of $\times0.25$ and $\times0.1$. With a penalty of $\times0.25$, only a slight gradient in price emerges across the map. With a penalty of $\times0.1$ two very different prices do emerge, and exchanges almost exclusively happen in the apple-rich and banana-rich regions. Thus, the conditions are such that merchant behaviour could emerge.

\begin{figure}
    \centering
    \includegraphics{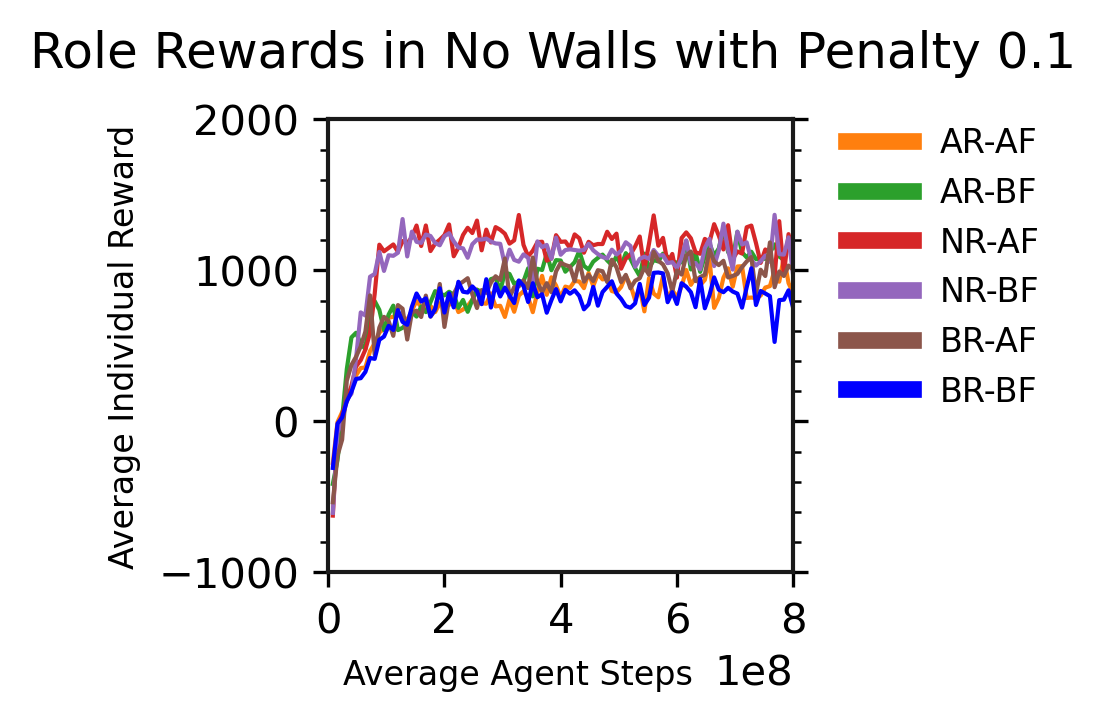}
    \caption{Average individual reward by role in the ``No Walls'' map with a neutral region penalty of $\times0.1$.}
    \label{fig:regions:no_walls_penalty_rewards}
\end{figure}

Although any agent of any role or starting region could potentially learn merchant behaviour, none of them did: all Apple Farmers only sold apples, and all Banana Farmers only sold bananas. Further, we cannot conclude that becoming merchants would actually be more rewarding than this behaviour. In Figure~\ref{fig:regions:no_walls_penalty_rewards} we present the average episodic reward by role in ``No Walls'' with a neutral penalty of $\times0.1$. The Neutral agents earn the highest reward in the final 25\% of training: 1174.2 for Apple Farmers and 1122.5 for Banana Farmers. They are followed by the rare item producing AR-BF and BR-AF roles at 1115.8 and 1012.56, and in last the common good producing AR-AF and BR-BF roles at 918.8 and 843.3. Although all agents can traverse the map, and do so in practice to some extent, we found that all agents starting in the Apple Region spend most of their time there, all agents starting in the Banana Region spend most of their time there, and all Neutral Region agents (both Apple Farmers and Banana Farmers) spend about half of their time in each of the Apple and Banana Regions. 

The higher reward for Neutral Region agents suggests that this behaviour of spending time in both regions, perhaps by choosing just one at the start of each episode, still gives them a positional advantage. However, when we compare against our earlier ``Walls'' $\times0.1$ results where merchant behaviour emerged, we found that the average reward of the ``merchant agents'' was 830.5: less than even the lowest-scoring role in the ``No Walls'' $\times0.1$ shown in Figure~\ref{fig:regions:no_walls_penalty_rewards}. Thus, while it is possible that merchant behaviour could be advantageous if learned, the agents in ``Walls'' are already earning more reward than our most similar merchants, and so we cannot conclude that their behaviour is suboptimal.

\begin{table} 
\centering 
\resizebox{\textwidth}{!}{ 
\begin{tabular}{|l|r||r|r|r|r|r|r||r|r|r|r|r|r|} \hline 
 & Episodic & \multicolumn{6}{c||}{Apples} & \multicolumn{6}{c|}{Bananas} \\ 
 & Reward & Prod & Bought & Con & Sold & Total In & Total Out & Prod & Bought & Con & Sold & Total In & Total Out \\ \hline 
\textrm{Roles} & & & & & & & & & & & & & \\ \hline 
\textrm{ \qquad NR-AF} & 1723.4 & 194.6 & 0.0 & 3.1 & 187.3 & 194.6 & 190.4 & 11.6 & 224.2 & 234.8 & 0.0 & 235.8 & 234.8 \\ \hline 
\textrm{ \qquad NR-BF} & 1870.3 & 11.2 & 243.5 & 253.6 & 0.0 & 254.7 & 253.6 & 197.1 & 0.0 & 1.6 & 192.0 & 197.1 & 193.6 \\ \hline 
\textrm{NR-AF Agents} & & & & & & & & & & & & & \\ \hline 
\textrm{ \qquad NR-AF 1} & 1747.0 & 193.7 & 0.0 & 2.8 & 187.1 & 193.7 & 189.9 & 11.4 & 227.1 & 237.5 & 0.0 & 238.5 & 237.5 \\ \hline 
\textrm{ \qquad NR-AF 2} & 1697.6 & 186.1 & 0.0 & 3.2 & 178.8 & 186.1 & 181.9 & 12.1 & 220.6 & 231.6 & 0.0 & 232.7 & 231.6 \\ \hline 
\textrm{ \qquad NR-AF 3} & 1754.2 & 192.1 & 0.0 & 2.8 & 185.1 & 192.1 & 187.9 & 11.7 & 227.5 & 238.1 & 0.0 & 239.2 & 238.1 \\ \hline 
\textrm{ \qquad NR-AF 4} & 1698.4 & 204.7 & 0.0 & 3.3 & 196.6 & 204.7 & 200.0 & 11.2 & 221.9 & 232.1 & 0.0 & 233.1 & 232.1 \\ \hline 
\textrm{NR-BF Agents} & & & & & & & & & & & & & \\ \hline 
\textrm{ \qquad NR-BF 1} & 1811.6 & 11.6 & 235.9 & 246.6 & 0.0 & 247.5 & 246.6 & 193.9 & 0.0 & 1.9 & 188.7 & 193.9 & 190.6 \\ \hline 
\textrm{ \qquad NR-BF 2} & 1894.4 & 10.9 & 246.8 & 256.4 & 0.0 & 257.6 & 256.4 & 195.8 & 0.0 & 1.6 & 190.5 & 195.8 & 192.1 \\ \hline 
\textrm{ \qquad NR-BF 3} & 1931.1 & 11.9 & 250.1 & 260.8 & 0.0 & 262.0 & 260.8 & 201.5 & 0.0 & 1.5 & 196.4 & 201.5 & 197.9 \\ \hline 
\textrm{ \qquad NR-BF 4} & 1846.2 & 10.5 & 241.5 & 251.0 & 0.0 & 252.0 & 251.0 & 197.2 & 0.0 & 1.3 & 192.6 & 197.2 & 193.9 \\ \hline 
\end{tabular} 
} 
\caption{Table of Neutral role and agent production, consumption, and exchange statistics for the ``Walls'' map with a neutral tree spawn penalty of $\times1.0$.} 
\label{tab:regions:wall_penalty_100} 
\end{table}

\begin{table} 
\centering 
\resizebox{\textwidth}{!}{ 
\begin{tabular}{|l|r||r|r|r|r|r|r||r|r|r|r|r|r|} \hline 
 & Episodic & \multicolumn{6}{c||}{Apples} & \multicolumn{6}{c|}{Bananas} \\ 
 & Reward & Prod & Bought & Con & Sold & Total In & Total Out & Prod & Bought & Con & Sold & Total In & Total Out \\ \hline 
\textrm{Roles} & & & & & & & & & & & & & \\ \hline 
\textrm{ \qquad NR-AF} & 1534.6 & 147.6 & 0.0 & 3.4 & 140.8 & 147.6 & 144.2 & 8.5 & 205.9 & 213.0 & 0.0 & 214.4 & 213.0 \\ \hline 
\textrm{ \qquad NR-BF} & 1589.6 & 8.7 & 212.5 & 219.9 & 0.0 & 221.2 & 219.9 & 151.5 & 0.0 & 2.8 & 145.2 & 151.5 & 148.1 \\ \hline 
\textrm{NR-AF Agents} & & & & & & & & & & & & & \\ \hline 
\textrm{ \qquad NR-AF 1} & 1574.7 & 150.8 & 0.0 & 3.6 & 143.9 & 150.8 & 147.5 & 8.5 & 210.0 & 217.1 & 0.0 & 218.5 & 217.1 \\ \hline 
\textrm{ \qquad NR-AF 2} & 1508.4 & 146.5 & 0.0 & 3.1 & 139.8 & 146.5 & 143.0 & 7.6 & 203.9 & 210.0 & 0.0 & 211.5 & 210.0 \\ \hline 
\textrm{ \qquad NR-AF 3} & 1537.7 & 146.2 & 0.0 & 3.2 & 139.8 & 146.2 & 143.0 & 9.4 & 205.5 & 213.4 & 0.0 & 214.9 & 213.4 \\ \hline 
\textrm{ \qquad NR-AF 4} & 1519.5 & 146.9 & 0.0 & 3.4 & 139.9 & 146.9 & 143.3 & 8.6 & 204.4 & 211.7 & 0.0 & 212.9 & 211.7 \\ \hline 
\textrm{NR-BF Agents} & & & & & & & & & & & & & \\ \hline 
\textrm{ \qquad NR-BF 1} & 1603.0 & 9.1 & 214.0 & 221.5 & 0.0 & 223.0 & 221.5 & 152.5 & 0.0 & 2.6 & 146.3 & 152.5 & 148.9 \\ \hline 
\textrm{ \qquad NR-BF 2} & 1546.3 & 8.7 & 207.2 & 214.7 & 0.0 & 216.0 & 214.7 & 148.3 & 0.0 & 3.1 & 141.7 & 148.3 & 144.7 \\ \hline 
\textrm{ \qquad NR-BF 3} & 1593.1 & 8.4 & 213.0 & 220.2 & 0.0 & 221.5 & 220.2 & 151.8 & 0.0 & 2.8 & 145.5 & 151.8 & 148.2 \\ \hline 
\textrm{ \qquad NR-BF 4} & 1613.1 & 8.5 & 215.5 & 222.9 & 0.0 & 223.9 & 222.9 & 153.3 & 0.0 & 2.8 & 147.2 & 153.3 & 150.1 \\ \hline 
\end{tabular} 
} 
\caption{Table of Neutral role and agent production, consumption, and exchange statistics for the ``Walls'' map with a neutral tree spawn penalty of $\times0.75$.}
\label{tab:regions:wall_penalty_75}
\end{table}

\begin{table} 
\centering 
\resizebox{\textwidth}{!}{ 
\begin{tabular}{|l|r||r|r|r|r|r|r||r|r|r|r|r|r|} \hline 
 & Episodic & \multicolumn{6}{c||}{Apples} & \multicolumn{6}{c|}{Bananas} \\ 
 & Reward & Prod & Bought & Con & Sold & Total In & Total Out & Prod & Bought & Con & Sold & Total In & Total Out \\ \hline 
\textrm{Roles} & & & & & & & & & & & & & \\ \hline 
\textrm{ \qquad NR-AF} & 1059.5 & 105.3 & 0.0 & 1.6 & 101.7 & 105.3 & 103.3 & 5.6 & 152.4 & 157.0 & 0.0 & 158.0 & 157.0 \\ \hline 
\textrm{ \qquad NR-BF} & 1070.1 & 5.6 & 154.5 & 159.1 & 0.0 & 160.1 & 159.1 & 106.5 & 0.0 & 1.6 & 103.1 & 106.5 & 104.7 \\ \hline 
\textrm{NR-AF Agents} & & & & & & & & & & & & & \\ \hline 
\textrm{ \qquad NR-AF 1} & 1075.1 & 104.6 & 0.0 & 1.3 & 101.5 & 104.6 & 102.8 & 5.5 & 152.3 & 156.7 & 0.0 & 157.7 & 156.7 \\ \hline 
\textrm{ \qquad NR-AF 2} & 1095.0 & 110.2 & 0.0 & 1.8 & 106.2 & 110.2 & 108.0 & 5.0 & 159.1 & 163.4 & 0.0 & 164.1 & 163.4 \\ \hline 
\textrm{ \qquad NR-AF 3} & 1033.8 & 102.0 & 0.0 & 1.2 & 98.7 & 102.0 & 99.9 & 6.6 & 147.9 & 153.5 & 0.0 & 154.5 & 153.5 \\ \hline 
\textrm{ \qquad NR-AF 4} & 1031.7 & 104.3 & 0.0 & 2.1 & 100.1 & 104.3 & 102.2 & 5.4 & 150.0 & 154.3 & 0.0 & 155.4 & 154.3 \\ \hline 
\textrm{NR-BF Agents} & & & & & & & & & & & & & \\ \hline 
\textrm{ \qquad NR-BF 1} & 1106.4 & 5.9 & 155.4 & 160.1 & 0.0 & 161.3 & 160.1 & 107.3 & 0.0 & 1.7 & 103.7 & 107.3 & 105.4 \\ \hline 
\textrm{ \qquad NR-BF 2} & 1035.3 & 5.9 & 151.1 & 156.1 & 0.0 & 157.0 & 156.1 & 104.9 & 0.0 & 1.9 & 100.8 & 104.9 & 102.7 \\ \hline 
\textrm{ \qquad NR-BF 3} & 1070.8 & 5.0 & 155.6 & 159.5 & 0.0 & 160.6 & 159.5 & 106.8 & 0.0 & 1.3 & 103.8 & 106.8 & 105.1 \\ \hline 
\textrm{ \qquad NR-BF 4} & 1068.8 & 5.5 & 156.0 & 160.4 & 0.0 & 161.5 & 160.4 & 107.2 & 0.0 & 1.4 & 104.1 & 107.2 & 105.4 \\ \hline 
\end{tabular} 
} 
\caption{Table of Neutral role and agent production, consumption, and exchange statistics for the ``Walls'' map with a neutral tree spawn penalty of $\times0.5$.} 
\label{tab:regions:wall_penalty_50} 
\end{table}

\begin{table} 
\centering 
\resizebox{\textwidth}{!}{ 
\begin{tabular}{|l|r||r|r|r|r|r|r||r|r|r|r|r|r|} \hline 
 & Episodic & \multicolumn{6}{c||}{Apples} & \multicolumn{6}{c|}{Bananas} \\ 
 & Reward & Prod & Bought & Con & Sold & Total In & Total Out & Prod & Bought & Con & Sold & Total In & Total Out \\ \hline 
\textrm{Roles} & & & & & & & & & & & & & \\ \hline 
\textrm{ \qquad NR-AF}   & 862.8  & 56.7 & 33.1  & 1.9   & 85.8  & 89.8  & 87.7  & 9.4  & 128.7 & 125.6 & 11.0 & 138.0 & 136.6 \\ \hline 
\textrm{ \qquad NR-BF}   & 1083.2 & 2.2  & 166.4 & 154.2 & 12.7  & 168.6 & 166.9 & 49.1 & 19.1  & 3.4   & 62.2 & 68.2  & 65.6 \\ \hline 
\textrm{NR-AF Agents} & & & & & & & & & & & & & \\ \hline 
\textrm{ \qquad NR-AF 1} & 900.2  & 47.7 & 51.8  & 2.3   & 94.5  & 99.6  & 96.8  & 6.5  & 141.8 & 129.1 & 17.3 & 148.2 & 146.4 \\ \hline 
\textrm{ \qquad NR-AF 2} & 990.1  & 51.3 & 56.8  & 2.5   & 103.1 & 108.0 & 105.6 & 6.5  & 154.6 & 140.3 & 18.9 & 161.1 & 159.2 \\ \hline 
\textrm{ \qquad NR-AF 3} & 804.2  & 58.7 & 23.7  & 2.1   & 78.5  & 82.4  & 80.6  & 11.2 & 117.7 & 120.0 & 7.9  & 128.9 & 127.9 \\ \hline 
\textrm{ \qquad NR-AF 4} & 747.9  & 68.6 & 0.0   & 0.9   & 66.4  & 68.6  & 67.2  & 13.3 & 99.6  & 112.0 & 0.0  & 112.9 & 112.0 \\ \hline 
\textrm{NR-BF Agents} & & & & & & & & & & & & & \\ \hline 
\textrm{ \qquad NR-BF 1} & 1289.0 & 1.4  & 208.2 & 177.4 & 29.7  & 209.6 & 207.1 & 40.9 & 44.6  & 3.5   & 78.5 & 85.4  & 81.9 \\ \hline 
\textrm{ \qquad NR-BF 2} & 940.8  & 2.8  & 135.9 & 137.7 & 0.0   & 138.8 & 137.7 & 55.6 & 0.0   & 2.9   & 51.0 & 55.6  & 53.9 \\ \hline 
\textrm{ \qquad NR-BF 3} & 1185.1 & 1.6  & 187.9 & 165.5 & 22.2  & 189.5 & 187.7 & 43.7 & 33.2  & 4.3   & 69.4 & 76.9  & 73.7 \\ \hline 
\textrm{ \qquad NR-BF 4} & 927.4  & 3.0  & 135.5 & 137.1 & 0.0   & 138.5 & 137.1 & 55.5 & 0.0   & 3.1   & 50.6 & 55.5  & 53.7 \\ \hline 
\end{tabular} 
} 
\caption{Table of Neutral role and agent production, consumption, and exchange statistics for the ``Walls'' map with a neutral tree spawn penalty of $\times0.25$.}
\label{tab:regions:wall_penalty_25} 
\end{table}

\begin{table} 
\centering 
\resizebox{\textwidth}{!}{ 
\begin{tabular}{|l|r||r|r|r|r|r|r||r|r|r|r|r|r|} \hline 
 & Episodic & \multicolumn{6}{c||}{Apples} & \multicolumn{6}{c|}{Bananas} \\ 
 & Reward & Prod & Bought & Con & Sold & Total In & Total Out & Prod & Bought & Con & Sold & Total In & Total Out \\ \hline 
\textrm{Roles} & & & & & & & & & & & & & \\ \hline 
\textrm{ \qquad NR-AF} & 749.5    & 17.6 & 43.0  & 3.0   & 54.5 & 60.6  & 57.5  & 1.3  & 142.5 & 125.9 & 15.6 & 143.8 & 141.6 \\ \hline 
\textrm{ \qquad NR-BF} & 669.2    & 2.6  & 125.1 & 114.8 & 11.2 & 127.7 & 126.0 & 20.8 & 27.7  & 3.8   & 42.2 & 48.5  & 46.0 \\ \hline 
\textrm{NR-AF Agents} & & & & & & & & & & & & & \\ \hline 
\textrm{ \qquad NR-AF 1} & 182.3  & 28.1 & 1.6   & 3.5   & 25.1 & 29.7  & 28.6  & 0.4  & 71.9  & 71.0  & 0.5  & 72.3  & 71.6 \\ \hline 
\textrm{ \qquad NR-AF 2} & 1089.7 & 12.6 & 63.3  & 2.9   & 68.9 & 76.0  & 71.7  & 2.5  & 177.6 & 156.1 & 21.1 & 180.1 & 177.2 \\ \hline 
\textrm{ \qquad NR-AF 3} & 1013.2 & 12.9 & 58.7  & 3.6   & 64.6 & 71.7  & 68.3  & 1.6  & 172.4 & 146.4 & 24.6 & 174.0 & 171.0 \\ \hline 
\textrm{ \qquad NR-AF 4} & 692.4  & 17.1 & 46.3  & 2.2   & 57.7 & 63.4  & 60.0  & 0.6  & 144.5 & 127.7 & 15.4 & 145.2 & 143.1 \\ \hline 
\textrm{NR-BF Agents} & & & & & & & & & & & & & \\ \hline 
\textrm{ \qquad NR-BF 1} & 868.4  & 0.4  & 172.0 & 146.6 & 23.1 & 172.4 & 169.7 & 14.2 & 50.1  & 2.2   & 58.1 & 64.3 & 60.4 \\ \hline 
\textrm{ \qquad NR-BF 2} & 710.8  & 1.8  & 120.0 & 111.2 & 9.0  & 121.8 & 120.2 & 21.3 & 27.1  & 6.2   & 40.0 & 48.4 & 46.2 \\ \hline 
\textrm{ \qquad NR-BF 3} & 524.8  & 7.5  & 81.7  & 88.3  & 0.0  & 89.2  & 88.3  & 31.4 & 0.0   & 3.1   & 27.3 & 31.4 & 30.4 \\ \hline 
\textrm{ \qquad NR-BF 4} & 608.6  & 0.3  & 134.1 & 118.0 & 14.3 & 134.3 & 132.4 & 15.2 & 37.3  & 3.7   & 45.8 & 52.6 & 49.5 \\ \hline 
\end{tabular} 
} 
\caption{Table of Neutral role and agent production, consumption, and exchange statistics for the ``Walls'' map with a neutral tree spawn penalty of $\times0.1$.} 
\label{tab:regions:wall_penalty_10} 
\end{table}

\section{Ablations and Tuning}
\label{sec:ablation}

Thus far in the paper, we have described the details of a reinforcement learning environment where agents could potentially learn microeconomic behaviour, and then demonstrated in detail how our agents do in fact learn it in practice. As we varied the environment in ways familiar to an undergraduate Microeconomics student, by influencing supply and demand, the agents responded with changes to the offers they made and the quantities of goods they produced and consumed. When we created maps designed to elicit the emergence of multiple prices, we found that our agents could also learn further concepts such as arbitrage.

In this section, we will reconsider many of the environmental design decisions that allowed us to reach this point. As we discussed in Section~\ref{sec:environment}, our main objective in this work has been to design a microeconomics themed environment for reinforcement learning research, while injecting as little domain knowledge as possible. We have compromised on this objective slightly in order to make the learning problem easier for our agents, and ideally future work will remove these compromises so that stronger agents can truly learn to trade from scratch. Our aim in this section is to highlight this domain knowledge and the subtle choices that have a dramatic impact on whether the agents learn to trade or not, note how the choice of agent architecture is critical, and explore several alternative trade actions and mechanisms that would have advantages over the mechanism we have presented.

\subsection{Hunger penalty}
\label{sec:ablation:hunger}

We begin with the environment's hunger penalty: a seemingly minor mechanic that could be removed but, in practice, has a dramatic effect on the emergence of trade. Recall from Section~\ref{sec:environment} that each agent has a ``satiation level'' that they observe, starting at 30 and reset to 30 whenever they eat any fruit, and otherwise decreasing by 1 per timestep. When the agent's satiation is 0, they suffer -1 reward per timestep as a ``hunger penalty''. While this notion of ``hunger'' has a natural interpretation in the real world, it also has a practical significance here as reward shaping. It incentivises agents to explore harvesting and consuming, but also to carry surplus fruit around in case they get hungry in the future. Ideally our agents would learn to explore all aspects of this environment without this reward shaping (\eg, by setting the hunger penalty to 0 reward), but as we will see, trading behaviour does not emerge \textit{at all} without it.

We will compare three environmental settings to better understand this effect. The ``Hunger'' setting uses the default hunger penalty of -1 as described above. ``No Hunger'' changes the hunger penalty to 0, thus eliminating it. Agents still observe their satiation level and it decreases over time, but there is no penalty for it reaching zero. Finally, ``Restricted No Hunger'' makes two changes: the hunger penalty is set to 0 as in ``No Hunger'', and the probability of producing an item that the agent's role is not skilled at producing is changed from 5\% to 0\%. That is, an Apple Farmer is \textit{restricted} to only harvest apples and cannot harvest bananas, and can only obtain a banana through trading with a Banana Farmer, who is similarly restricted to only harvest bananas. The reason for including this third setting will become apparent as we compare the differing behaviour of the ``Hunger'' and ``No Hunger'' populations.

\begin{figure}
    \centering
    \begin{subfigure}{0.5\textwidth}
        \centering
        \includegraphics[height=2in]{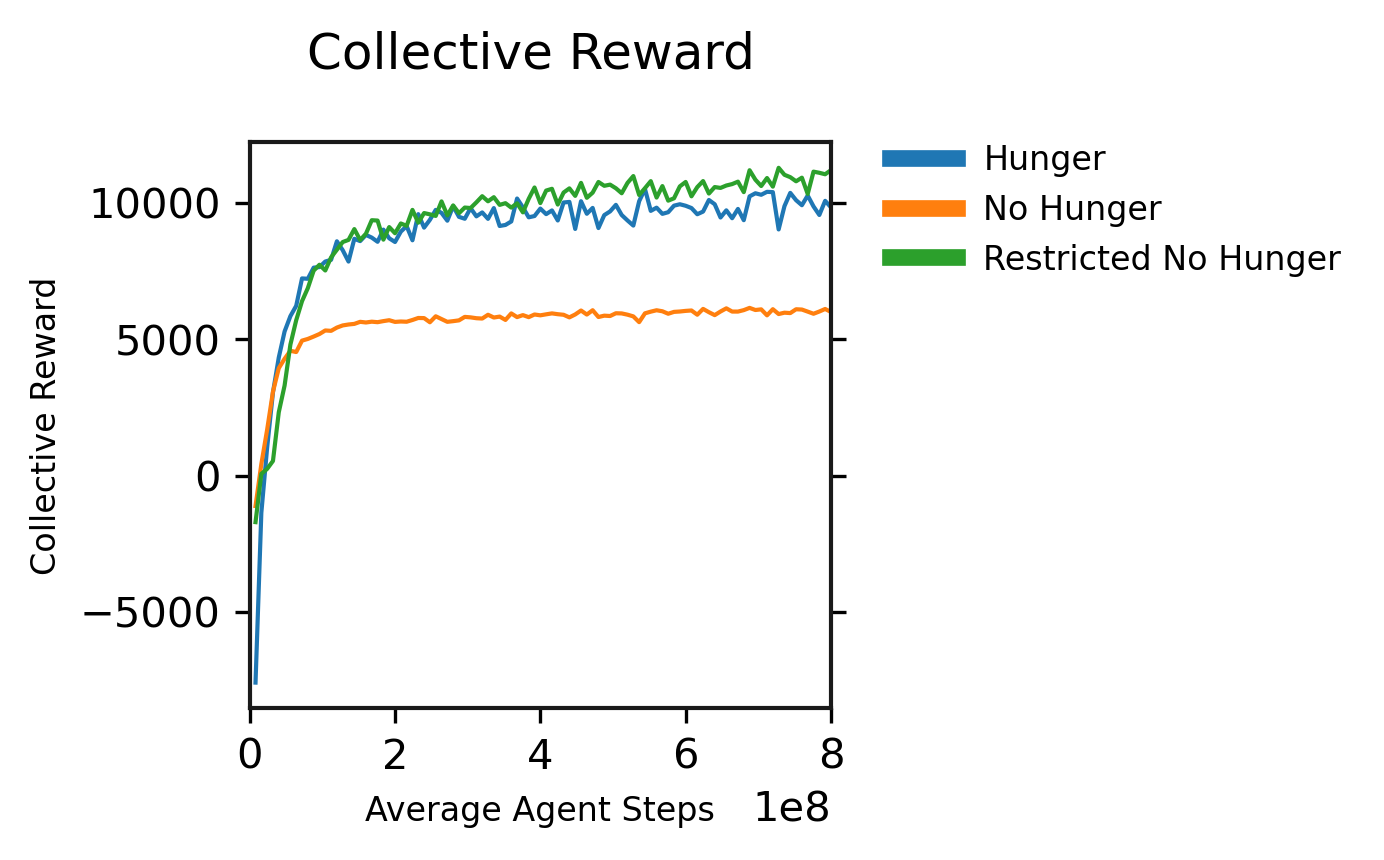}
        \caption{}
        \label{fig:ablation_hunger:summary:reward}
    \end{subfigure}
    
    \begin{subfigure}{0.5\textwidth}
        \centering
        \includegraphics[height=2in]{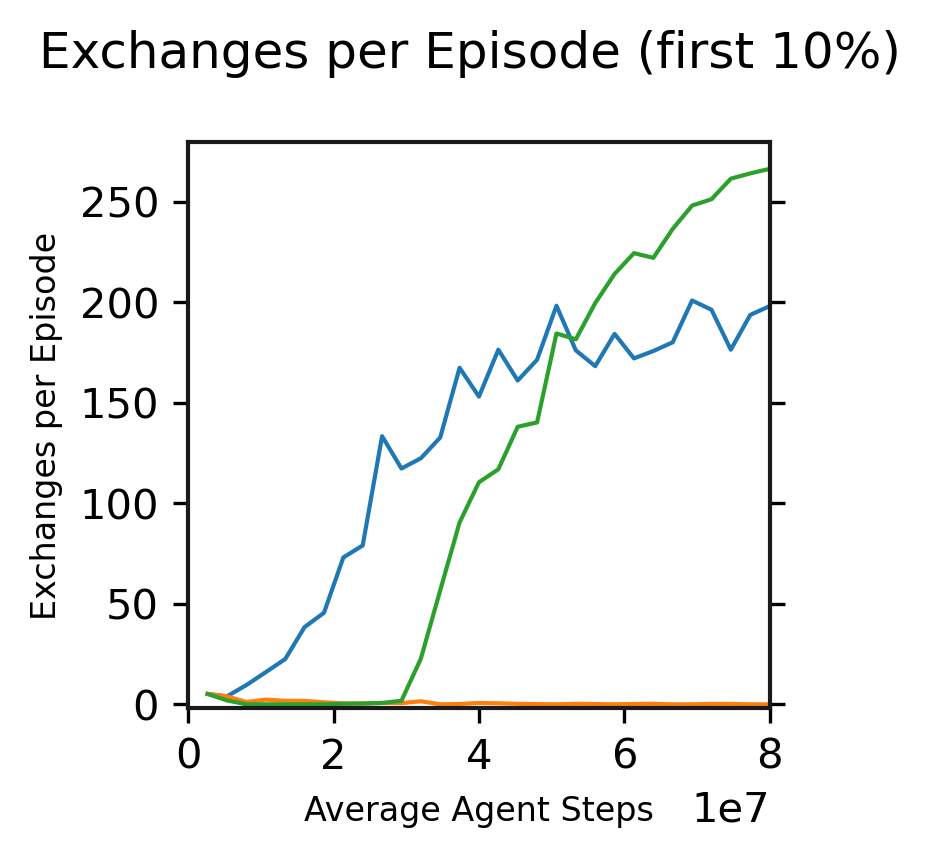}
        \caption{}
        \label{fig:ablation_hunger:summary:exchanges_early}
    \end{subfigure}%
    ~
    \begin{subfigure}{0.5\textwidth}
        \centering
        \includegraphics[height=2in]{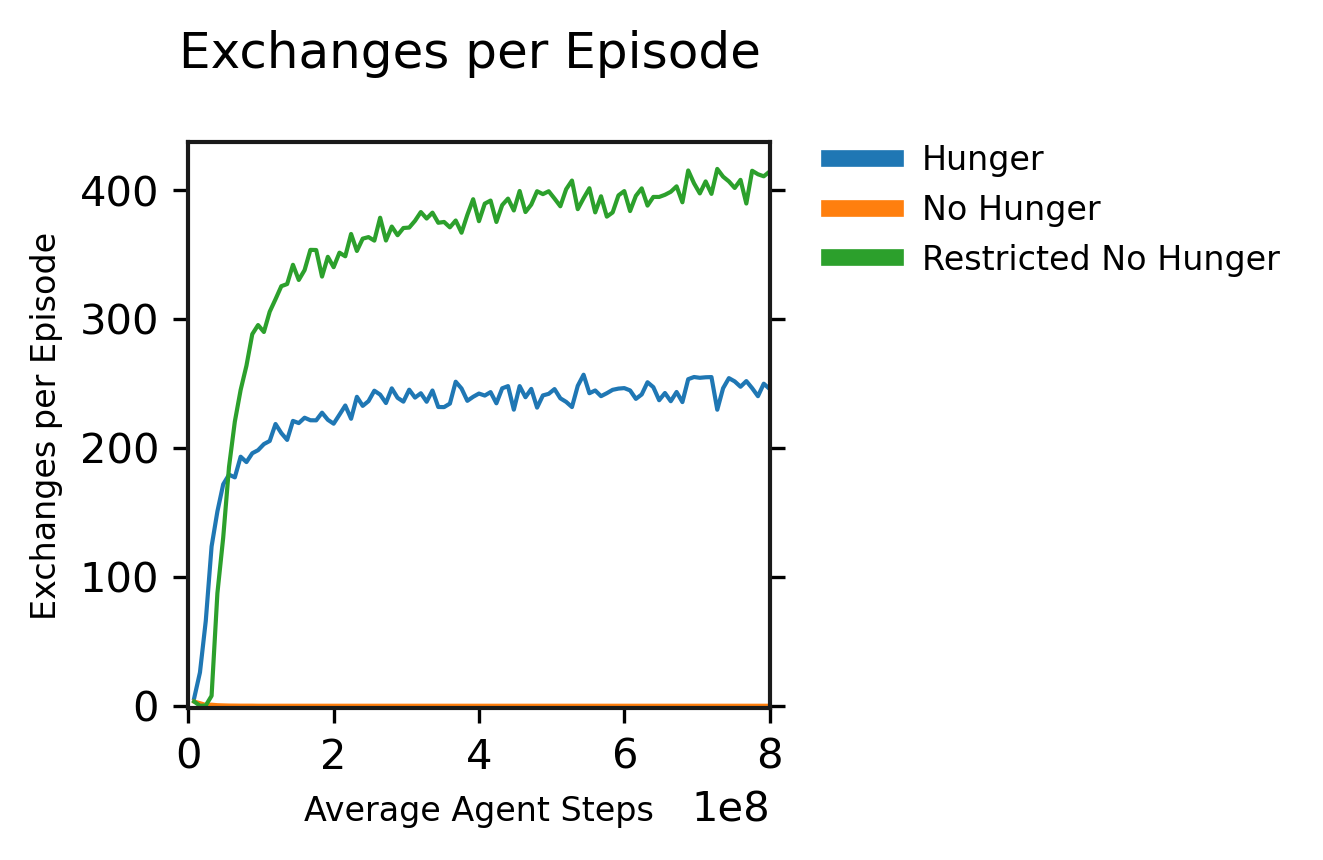}
        \caption{}
        \label{fig:ablation_hunger:summary:exchanges}
    \end{subfigure}

    \caption{Effects of the hunger penalty on agent performance and behaviour. (c) illustrates that without the hunger penalty, agents do not develop trading behaviour.}
    \label{fig:ablation_hunger:summary}
\end{figure}

Figure~\ref{fig:ablation_hunger:summary} begins our comparison. In Figure~\ref{fig:ablation_hunger:summary:reward}, we see that the absence of the hunger penalty results in a significant decrease in collective reward. This is already surprising: the hunger penalty only \textit{hurts} agents, and in Figure~\ref{fig:baseline_reward_source} we observed that the agents quickly learn to drive the hunger penalty close to zero. So how could the absence of an easily avoidable penalty lead to such a difference? Figure~\ref{fig:ablation_hunger:summary:exchanges} provides our first hint: in the ``No Hunger'' case, virtually no exchanges occur. By comparison, trade emerges very quickly in ``Hunger'', and only slightly slower in ``Restricted No Hunger''; this is clearer in Figure~\ref{fig:ablation_hunger:summary:exchanges_early}, which focuses on the first 10\% of training.

This failure to learn trading behaviour was also present in our earliest implementation of Fruit Market in 2018, which did not include a hunger penalty. Our hypothesis at that time focused on the difficulty of exploring the Offer actions. Specifically, once a reinforcement learning agent has experienced consuming fruit to gain reward, their policy will be updated towards consuming fruit as often and as quickly as possible. To have fruit and \textit{not} consume it would initially be an exploration step, as the agent would not yet have experienced any longer-horizon and more rewarding use for fruit. And that means exploring alternative uses for fruit, such as trading it away for something even more rewarding, will be difficult for the agents to explore: if the agent has already consumed their fruit, then they have no fruit to explore with. Taking an Offer action while not holding any fruit has no effect: the agent simply stands still on that timestep. This problem is even worse when exploring an alternative use for fruit requires avoiding temptation for many timesteps to explore a sequence of actions: harvesting a fruit, then carrying it to a partner of the other role, then using offer actions to choose compatible offers, and so on. This is also a joint exploration task, requiring two agents to be similarly avoiding temptation by exploring, simultaneously and nearby. Finally, exploring an Offer action while holding fruit may only have a noticeable effect several timesteps in the future when an exchange happens, and the agent must learn to assign credit to the Offer action and not any of the intermediate steps.

In summary, the hypothesis is: if agents learn to consume all of their fruit for reward, then they cannot explore making offers, and so trading behaviour is less likely to emerge. The hunger mechanic addresses this by giving agents a reason to carry some excess fruit around instead of immediately consuming it. Once the agent starts carrying even one or two extra fruit to stave off future hunger, they are able to randomly try exploring the Offer actions, and then discover the benefits of trade.

\begin{figure}
    \centering
    \begin{subfigure}{0.32\textwidth}
        \flushright
        \includegraphics[height=3.8in]{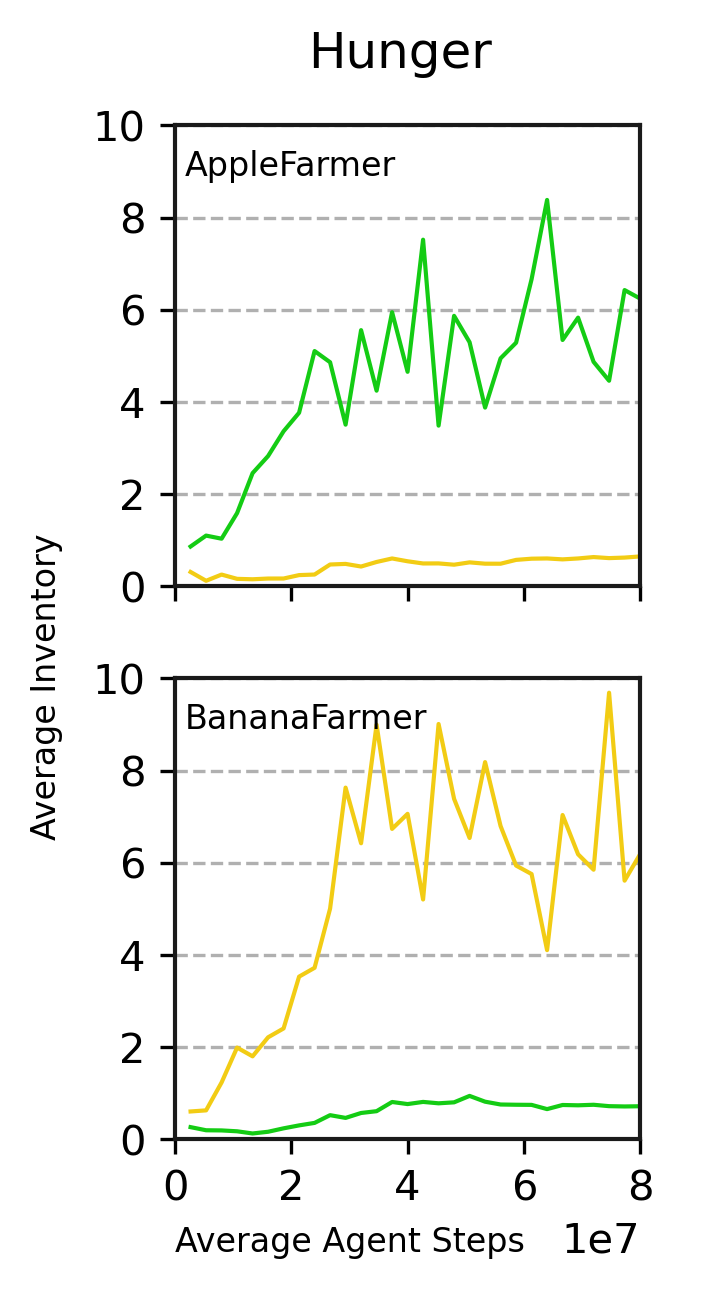}
        \caption{}
        \label{fig:ablation_hunger:inventory:hunger}
    \end{subfigure}%
    ~
    \begin{subfigure}{0.3\textwidth}
        \centering
        \includegraphics[height=3.8in]{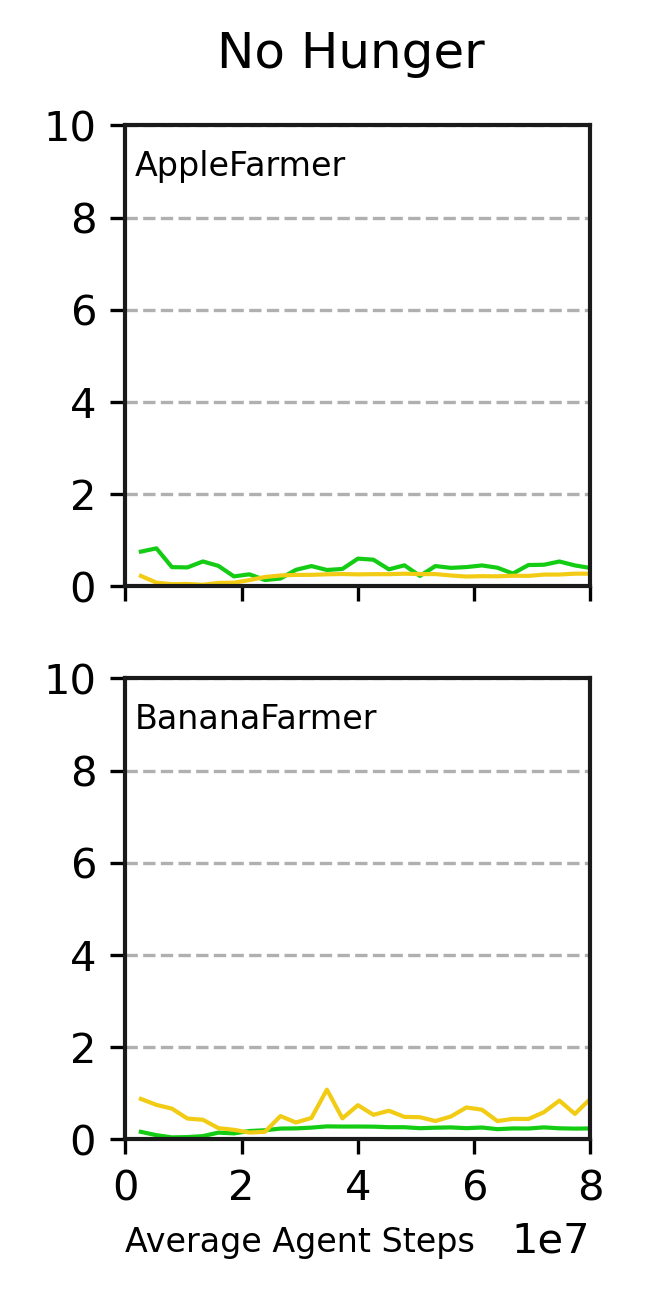}
        \caption{}
        \label{fig:ablation_hunger:inventory:no_hunger}
    \end{subfigure}%
    ~
    \begin{subfigure}{0.38\textwidth}
        \flushleft
        \includegraphics[height=3.8in]{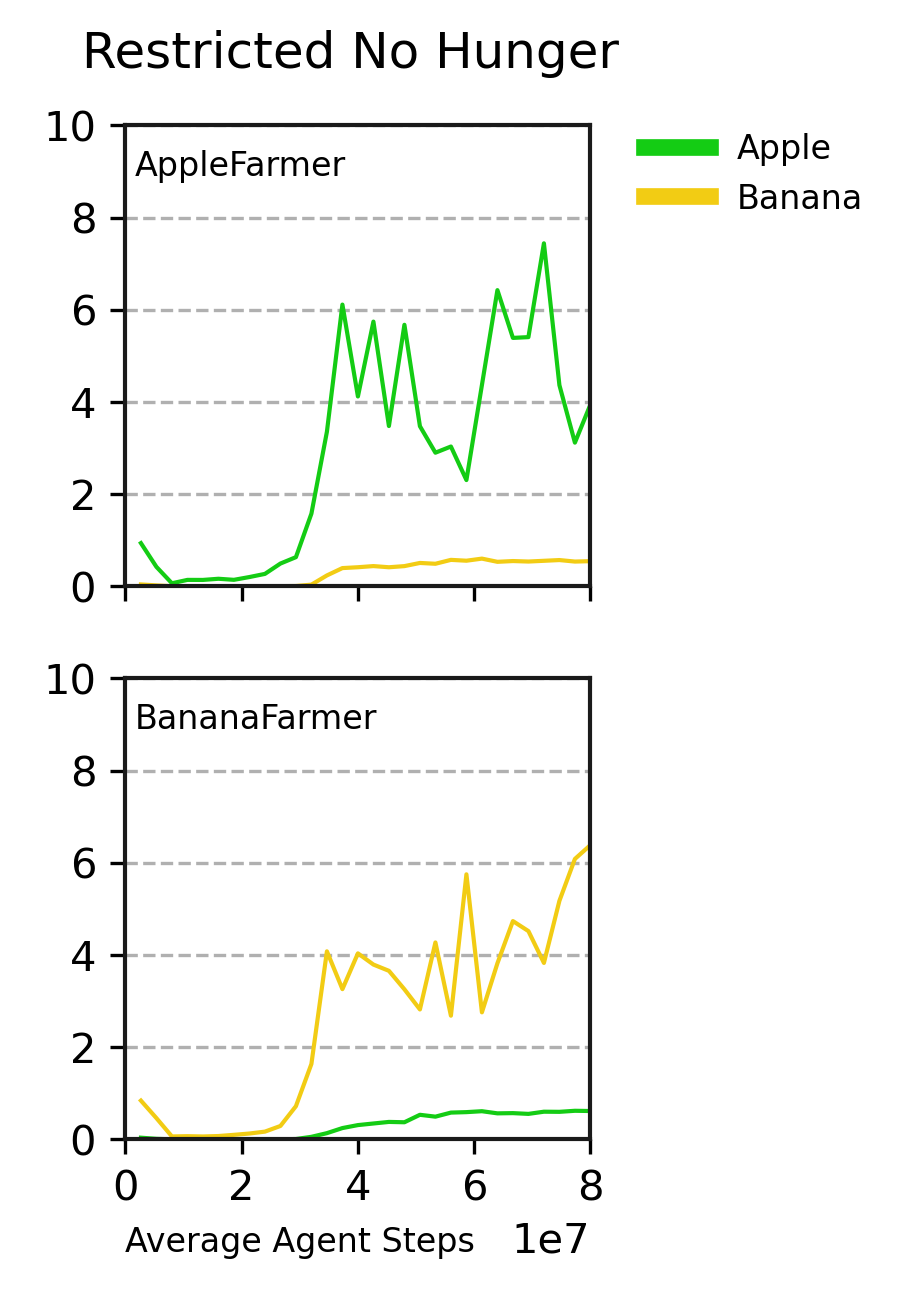}
        \caption{}
        \label{fig:ablation_hunger:inventory:restricted_no_hunger}
    \end{subfigure}

    \caption{Average inventory contents for each role, in the ``Hunger'', ``No Hunger'', and ``Restricted No Hunger'' cases. In the ``Hunger'' case, agents carry an excess of the goods their role can efficiently produce. In the ``No Hunger'' case, agents carry less than one item of each type at any time, preventing exploration of the Offer actions. In the ``Restricted No Hunger'' case, agents initially do not carry excess items, but around timestep $3e7$ begin to do so.}
    \label{fig:ablation_hunger:inventory}
\end{figure}

If the hypothesis is true, then we would expect to see agents carrying very few items in their inventory in the ``No Hunger'' case, as they would eat fruit quickly after producing it. Figure~\ref{fig:ablation_hunger:inventory} shows the average inventory contents for each role over the first 10\% of training ($8e7$ timesteps). This result supports the hypothesis: in the ``No Hunger'' setting, agents carry less than one item of each type on average during an episode. In the ``Hunger'' setting, we see that Apple Farmers learn to carry extra apples with them, and Banana Farmers do likewise with bananas. In the ``Restricted No Hunger'' case, agents initially do not carry extra items, but then begin to do so around timestep $3e7$, which is the same time that Figure~\ref{fig:ablation_hunger:summary:exchanges_early} showed trading behaviour starting to emerge.

\begin{figure}
    \centering
    \begin{subfigure}{0.3\textwidth}
        \centering
        \includegraphics[height=2in]{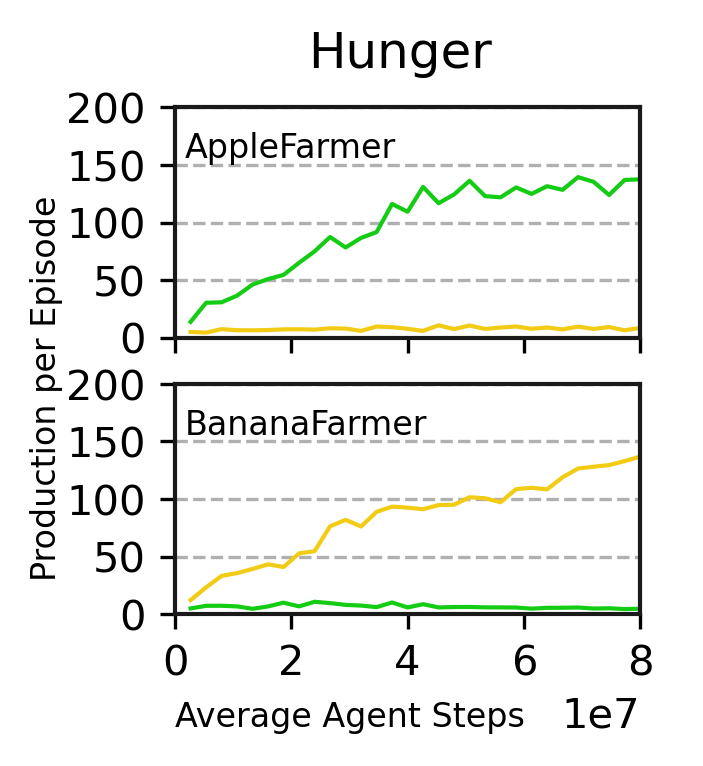}
        \caption{}
        \label{fig:ablation_hunger_prod_con:prod_hunger}
    \end{subfigure}%
    ~
    \begin{subfigure}{0.3\textwidth}
        \centering
        \includegraphics[height=2in]{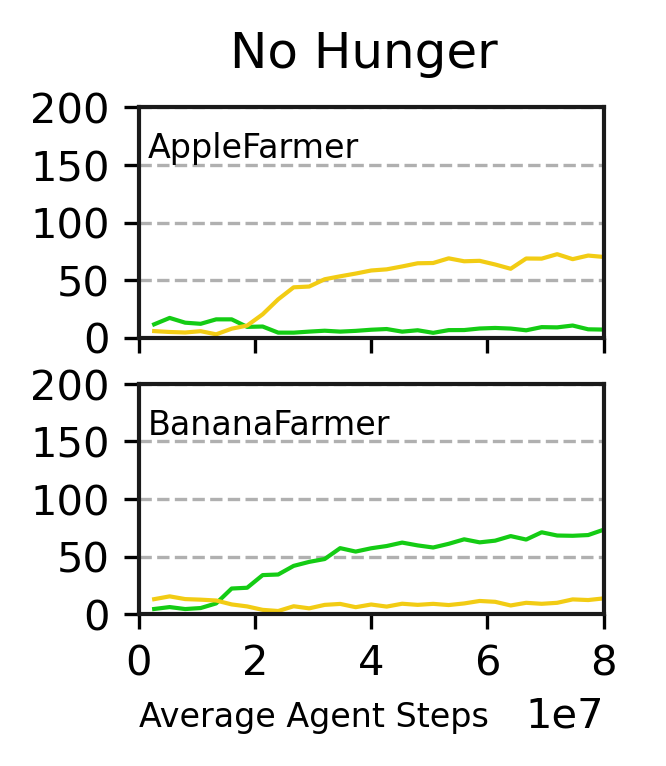}
        \caption{}
        \label{fig:ablation_hunger_prod_con:prod_no_hunger}
    \end{subfigure}%
    ~
    \begin{subfigure}{0.4\textwidth}
        \centering
        \includegraphics[height=2in]{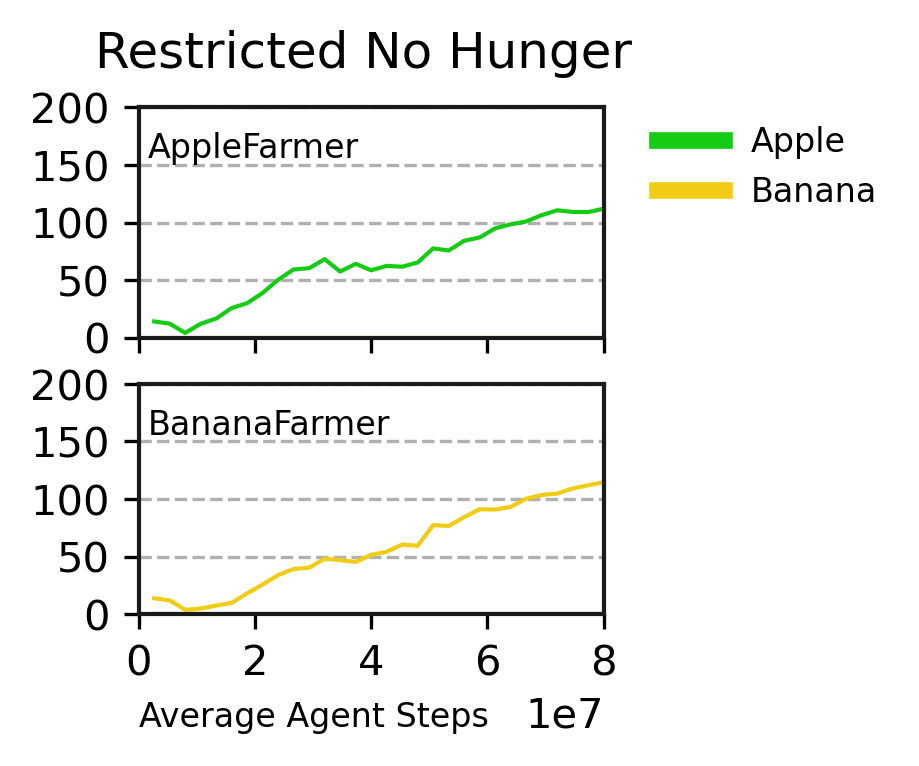}
        \caption{}
        \label{fig:ablation_hunger_prod_con:prod_restricted}
    \end{subfigure}
    
    \begin{subfigure}{0.3\textwidth}
        \centering
        \includegraphics[height=2in]{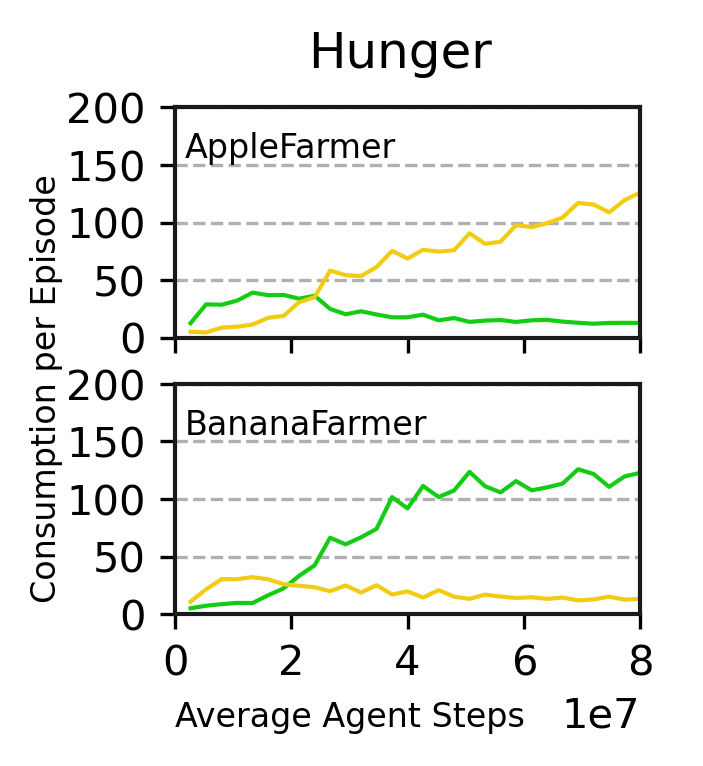}
        \caption{}
        \label{fig:ablation_hunger_prod_con:con_hunger}
    \end{subfigure}%
    ~
    \begin{subfigure}{0.3\textwidth}
        \centering
        \includegraphics[height=2in]{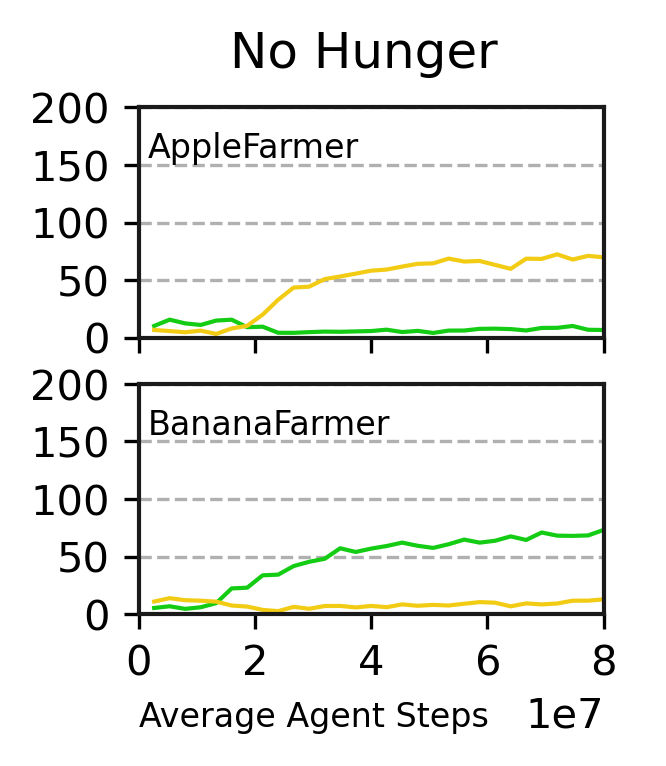}
        \caption{}
        \label{fig:ablation_hunger_prod_con:con_no_hunger}
    \end{subfigure}%
    ~
    \begin{subfigure}{0.4\textwidth}
        \centering
        \includegraphics[height=2in]{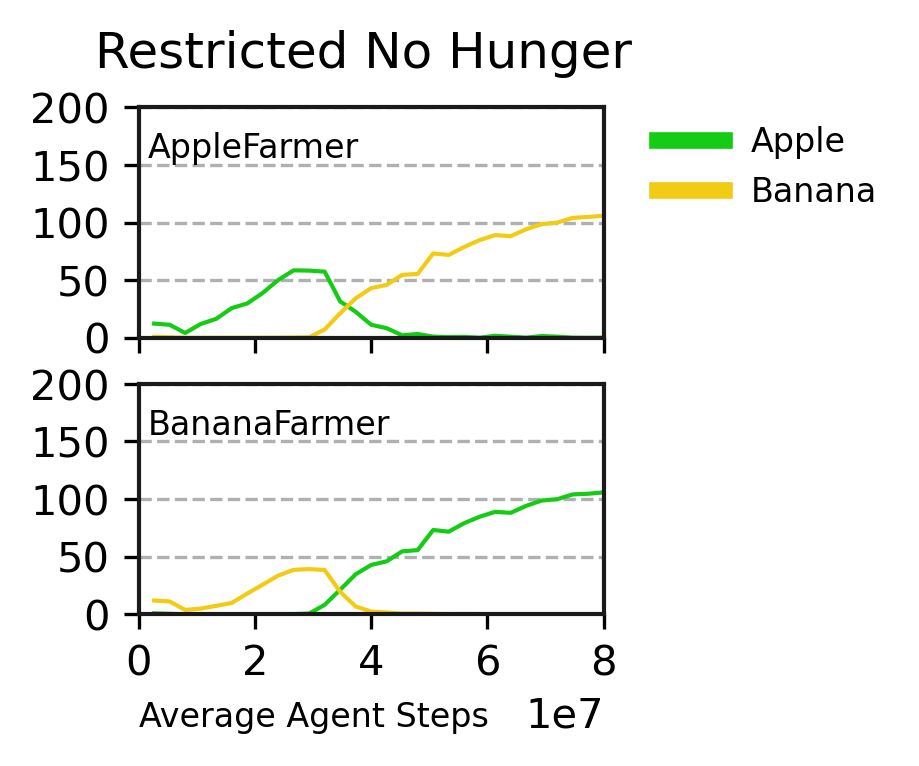}
        \caption{}
        \label{fig:ablation_hunger_prod_con:con_restricted}
    \end{subfigure}
    
    \caption{Agent production (a-c) and consumption (d-f) behaviour with and without the hunger penalty. Plots focus on the first 10\% of training. In the ``No Hunger'' case, agents predominantly produce the items they want to consume instead of the items that they can trade. In the other cases, agents produce the items their role is specialized towards, and consume the other item.}
    \label{fig:ablation_hunger_prod_con}
\end{figure}

However, there is also a second explanation to consider. Figure~\ref{fig:ablation_hunger_prod_con} shows the items produced and consumed by each role during the first 10\% of training. In the ``Hunger'' case, we see that Apple Farmers produce apples almost exclusively, and consume apples initially before switching to bananas, which were obtained through trade. In the ``No Hunger'' case, this is flipped: Apple Farmers produce a mix of mostly \textit{bananas}, and then consume that mix. Apple Farmers produce bananas inefficiently, with a 5\% probability of harvesting two bananas per timestep when standing on a tree with ripe bananas, as compared to a 100\% probability of harvesting apples. This presents a second hypothesis for why trade might fail to emerge in the ``No Hunger'' case: if agents learn to (almost) exclusively produce the goods they find most rewarding, and not the goods they can produce efficiently for trade, then they will have no tradable items with which to explore trading behaviour. The hunger penalty might also address this problem, independently of the inventory explanation, by encouraging Apple Farmers to split their time between reliably producing apples to stave off hunger, in addition to inefficiently producing bananas for reward. And then, since they have some apples \textit{at all} instead of only having bananas, they can explore their Offer actions and discover trade.

The ``Restricted No Hunger'' case lets us distinguish whether one of these two hypotheses, or a mix of both, is the best explanation. If the first hypothesis of ``Agents will eat all of their fruit and then have nothing left to explore trading with'' is true, then in the ``Restricted No Hunger'' case, we would expect the emergence of trading to be less likely. Apple Farmers would only produce apples, but without hunger to incentivise holding some in reserve, they could consume them quickly for one reward each. If the second hypothesis of ``Agents will primarily produce what they want to consume and then have no tradable goods to explore trading with'' is true, then in the ``Restricted No Hunger'' case, we would expect to see trade emerge. The restricted players could \textit{only} produce tradable goods, ensuring that they have some to explore trading, at least briefly before consuming them.

Both hypotheses appear to contribute to the end result. In the ``Restricted No Hunger'' case trade does emerge, and so the problem is not solely that agents learn to consume all of their produced goods and then have nothing to trade with. However, trade also takes longer to emerge than in the ``Hunger'' case, and the agents go through an initial period where they hold nearly zero items of each type on average. Figures~\ref{fig:ablation_hunger_prod_con:prod_restricted} and~\ref{fig:ablation_hunger_prod_con:con_restricted} show that they are producing many fruit of their specialized type, and so before trade emerges, they must be consuming them almost immediately. Thus, the first hypothesis appears to also play some role. 

Overall, the hunger penalty seems to give reinforcement learning agents a useful bit of initial reward shaping to help them discover trade. Our earlier results showed that the agents quickly learn to avoid it almost entirely, so this early experience affects their learning trajectory in a significant way, but does not directly encode the policy that they should use in general.

\subsection{Movement Penalty}
\label{sec:ablation:movement}

\begin{figure}
    \centering
    \begin{subfigure}{0.5\textwidth}
        \centering
        \includegraphics[height=2in]{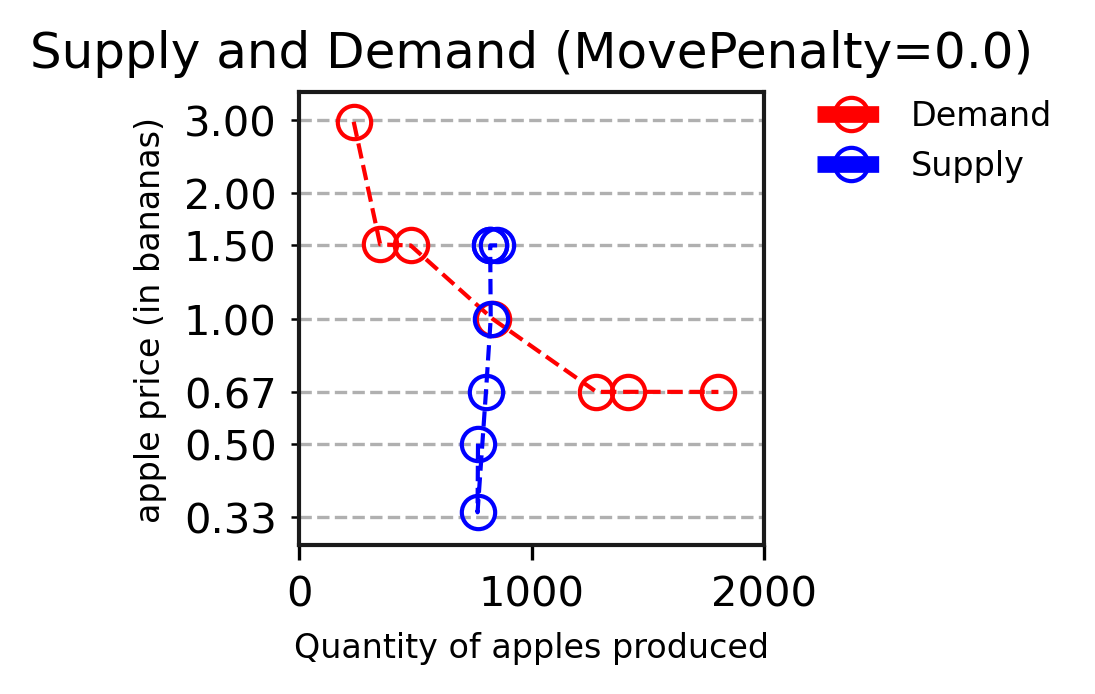}
        \caption{}
        \label{fig:ablation_movement:m0_produced}
    \end{subfigure}%
    ~
    \begin{subfigure}{0.5\textwidth}
        \centering
        \includegraphics[height=2in]{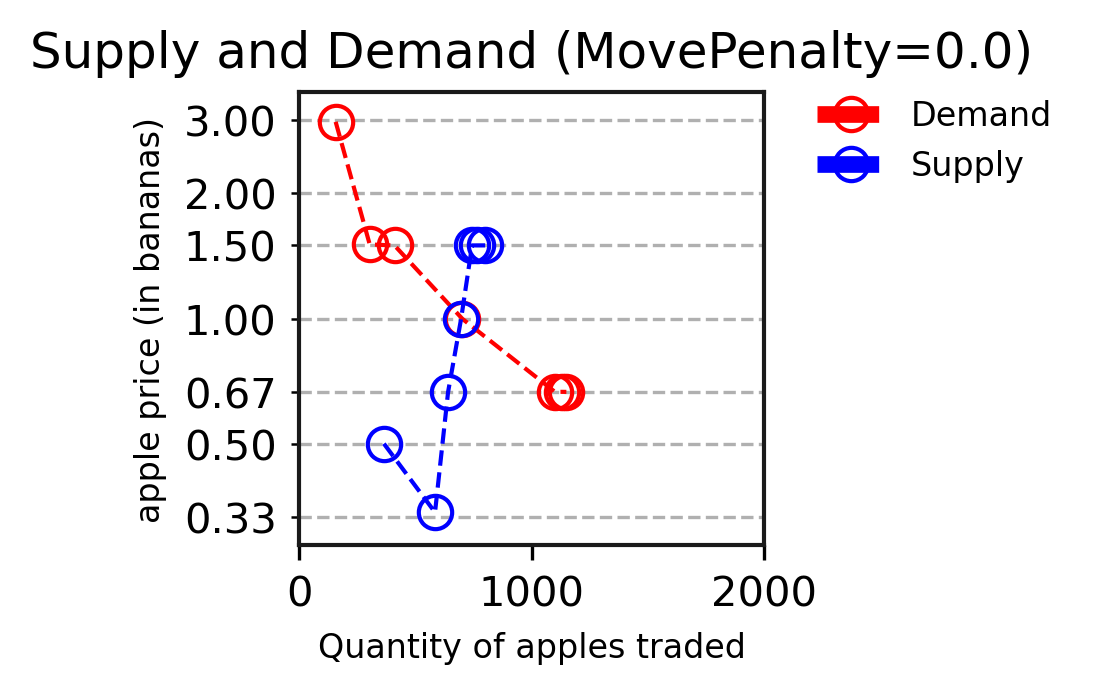}
        \caption{}
        \label{fig:ablation_movement:m0_traded}
    \end{subfigure}
    
    \begin{subfigure}{0.5\textwidth}
        \centering
        \includegraphics[height=2in]{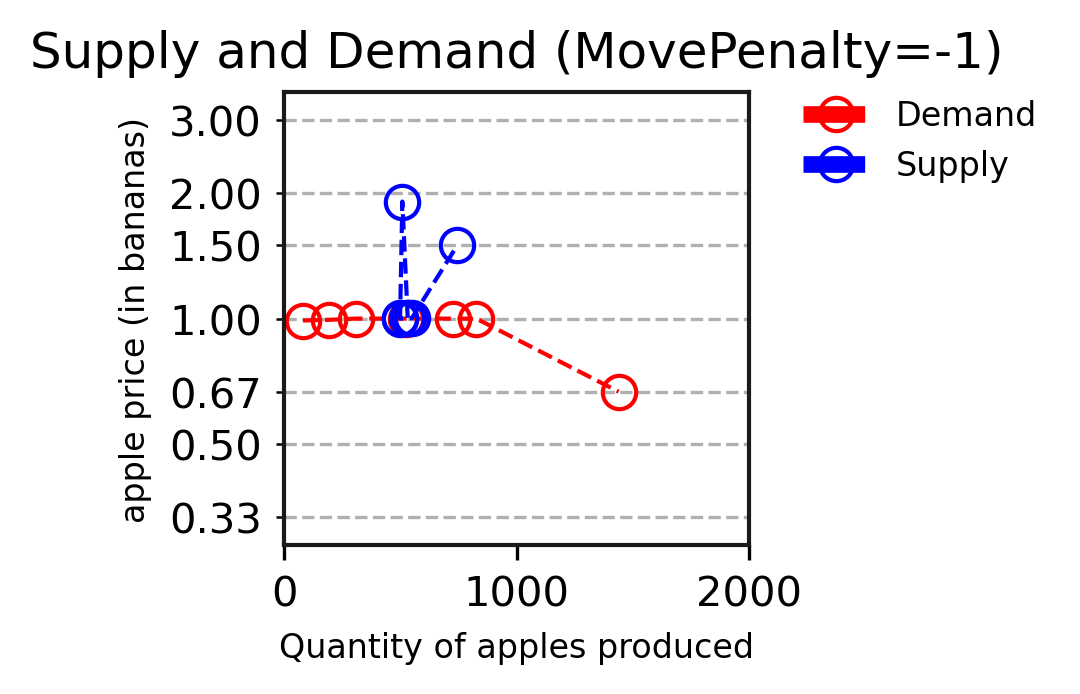}
        \caption{}
        \label{fig:ablation_movement:m1_produced}
    \end{subfigure}%
    ~
    \begin{subfigure}{0.5\textwidth}
        \centering
        \includegraphics[height=2in]{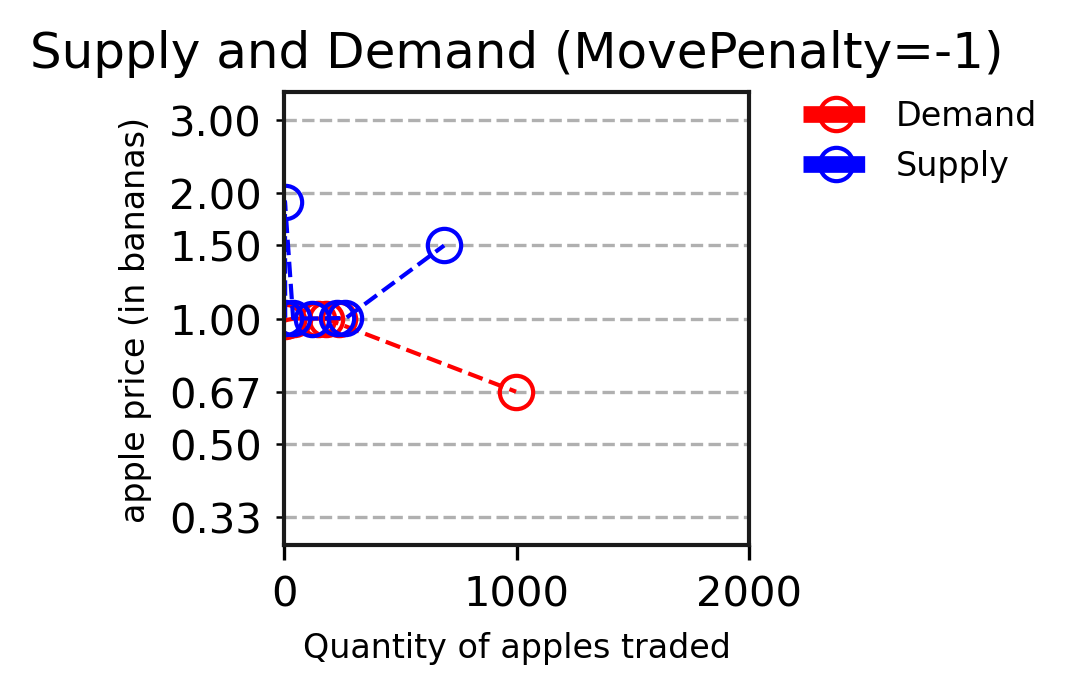}
        \caption{}
        \label{fig:ablation_movement:m1_traded}
    \end{subfigure}

    \caption{Comparison of supply and demand curves with a movement penalty of 0 reward/step in (a) and (c), and a movement penalty of -1 in (b) and (d). These graphs were produced by sweeping the spawn rate of apple and banana trees. See Figure~\ref{fig:sd-spawn-spawn} for comparison, which used our default movement penalty of -0.25 reward/step.}
    \label{fig:ablation_movement}
\end{figure}

The movement and water penalties described in Section~\ref{sec:environment} serve a different purpose from the hunger penalty. As we described in Sections~\ref{sec:environment:opportunity_costs} and~\ref{sec:experiments:marketplace}, these penalties create opportunity costs for the agents. Without them, an Apple Farmer's policy could be simple: produce as many apples as possible, trade them at any available price, and consume the resulting bananas. However, an increase in the population's price for apples would not result in any increase in apple production, because Apple Farmers would already be producing as many apples as possible at the lower price. By making travel costly (through the movement penalty) and regions farther from the center increasingly expensive to enter (through the water penalty), an Apple Farmer should instead consider the marginal cost of producing more apples, and if the available offers to buy apples justifies that cost. For example, if the price of apples is low, then nearby and convenient apples should be harvested and traded, but sitting idle (or inefficiently harvesting bananas) would be more rewarding than harvesting apples that are far away or across the water. If the population's price for apples increases, then harvesting the next closest apples becomes worthwhile, and only a truly high price should incentivise a player to walk all the way to the edge of the map, crossing many bodies of water.

Thus, changing the movement and water penalties allows us to control how agents should trade off between their available behaviours, including doing nothing. Of course, whether our agents will correctly learn how to adjust their behaviour in this way is an empirical question. In Figure~\ref{fig:ablation_movement}, we explore this by presenting Supply and Demand graphs in environments with movement penalties of 0 and -1. For comparison see our earlier presentation in Figure~\ref{fig:sd-spawn-spawn}, which used the default movement penalty of -0.25. All of these plots were generated by varying the spawn rate of apple and banana trees. 

Figures~\ref{fig:ablation_movement:m0_produced} and~\ref{fig:ablation_movement:m0_traded} used a movement penalty of 0 and present the same experiments, varying only by the x-axis measuring apples produced or net apples sold. In both plots but particularly in Figure~\ref{fig:ablation_movement:m0_produced} the supply curve is almost vertical, indicating that the agents only slightly vary their production in response to the price. In our earlier Figure~\ref{fig:sd-spawn-spawn} using a movement penalty of -0.25 per step, the Supply curve still had a steep slope, but less so than in these results, indicating that under the default movement penalty the agents slightly changed their production in response to the price.

However, setting the movement penalty to be too harsh also causes a problem, as is shown in Figures~\ref{fig:ablation_movement:m1_produced} and~\ref{fig:ablation_movement:m1_traded} which used a movement penalty of -1. In these cases, trading behaviour largely disappears; apples are still produced, but for the agent's own consumption and not for sale, as can be seen by comparing the plots. While it may be possible that a movement penalty of -1 is far too harsh and a smaller value such as -0.5 might have shown a more gradual curve, consider our earlier Figure~\ref{fig:marketplace:m1} which used a movement penalty of -1 and a marketplace. In those results, when the marketplace was available as a reliable trading partner throughout all of training --- always in the same location, always making the same offer --- the agents were quite consistent in learning to produce, trade, and consume goods, and the resulting supply and demand curves were smooth. The experiments presented here add the extra complication that an agent's trading partners are \textit{also} simultaneously learning how to trade, but in different locations and with varying offers. We believe that the extra difficulty of joint exploration is responsible for the sparse trading we see in Figure~\ref{fig:ablation_movement:m1_traded}.

Overall, the movement and water penalties were not added to encourage trade to emerge at all, as was true for the hunger penalty, but instead to force the agents to learn a richer set of behaviours: deciding when an apple is not worth harvesting. These particular penalties are of course not the only options. Instead of trying to tune this balance between the emergence of elastic supply while retaining the emergence of trading behaviour, it may be more fruitful to add other alternative sources of reward to the environment, or change from linear to diminishing rewards for repeated consumption of each type of fruit. Another alternative would be to reduce or remove each role's advantage in producing each good, so that Apple Farmers could more viably shift their production from apples to bananas if apple prices were too low. With our current production and consumption constants, trading apples for bananas with any offer makes apple production worthwhile, which gives Apple Farmers little incentive to switch to banana production if the apple price is too low.

\subsection{Agent Architecture}
\label{sec:ablation:agent}

\begin{figure}
    \centering
    \begin{subfigure}{0.45\textwidth}
        \centering
        \includegraphics[height=2in]{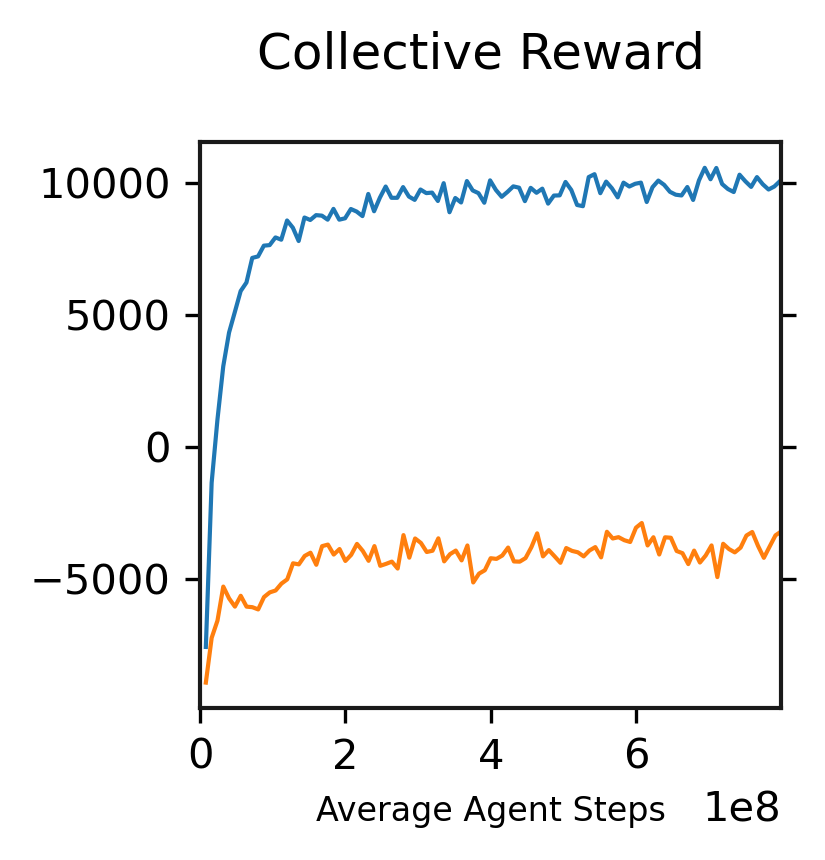}
        \caption{}
        \label{fig:ablation_agent:vmpo_a2c:collective_reward}
    \end{subfigure}%
    ~
    \begin{subfigure}{0.55\textwidth}
        \centering
        \includegraphics[height=2in]{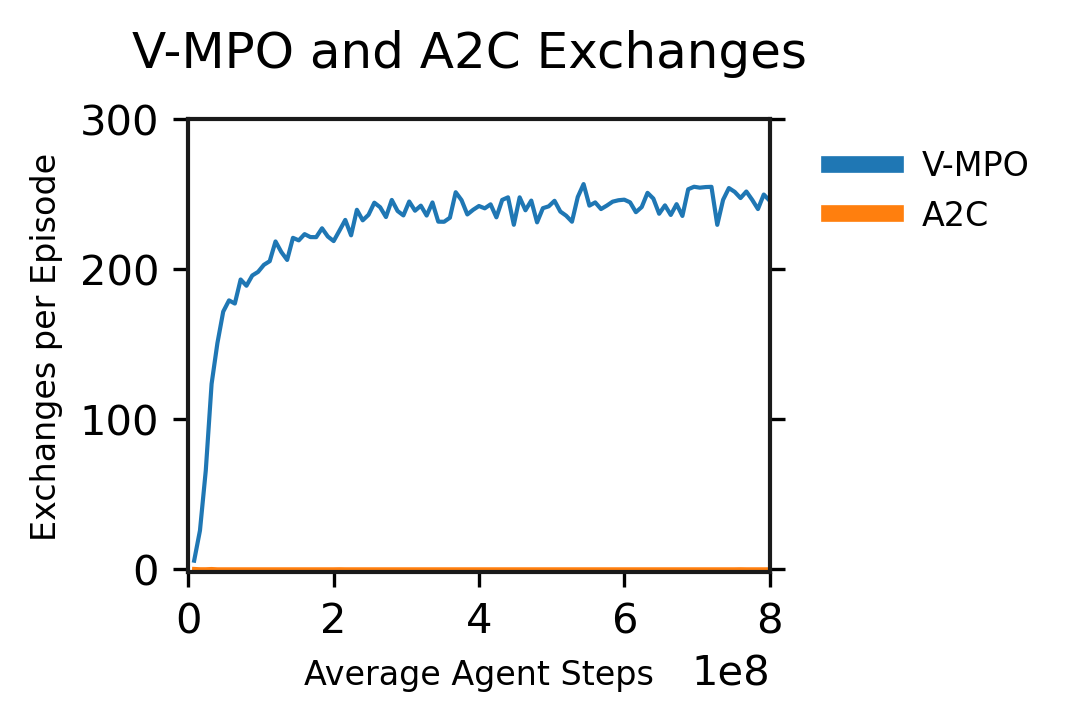}
        \caption{}
        \label{fig:ablation_agent:vmpo_a2c:exchanges}
    \end{subfigure}
    
    \caption{A comparison of V-MPO and A2C agents, measured by collective reward and quantity of exchanges.}
    \label{fig:ablation_agent:vmpo_a2c}
\end{figure}

\begin{figure}
    \centering
    \includegraphics{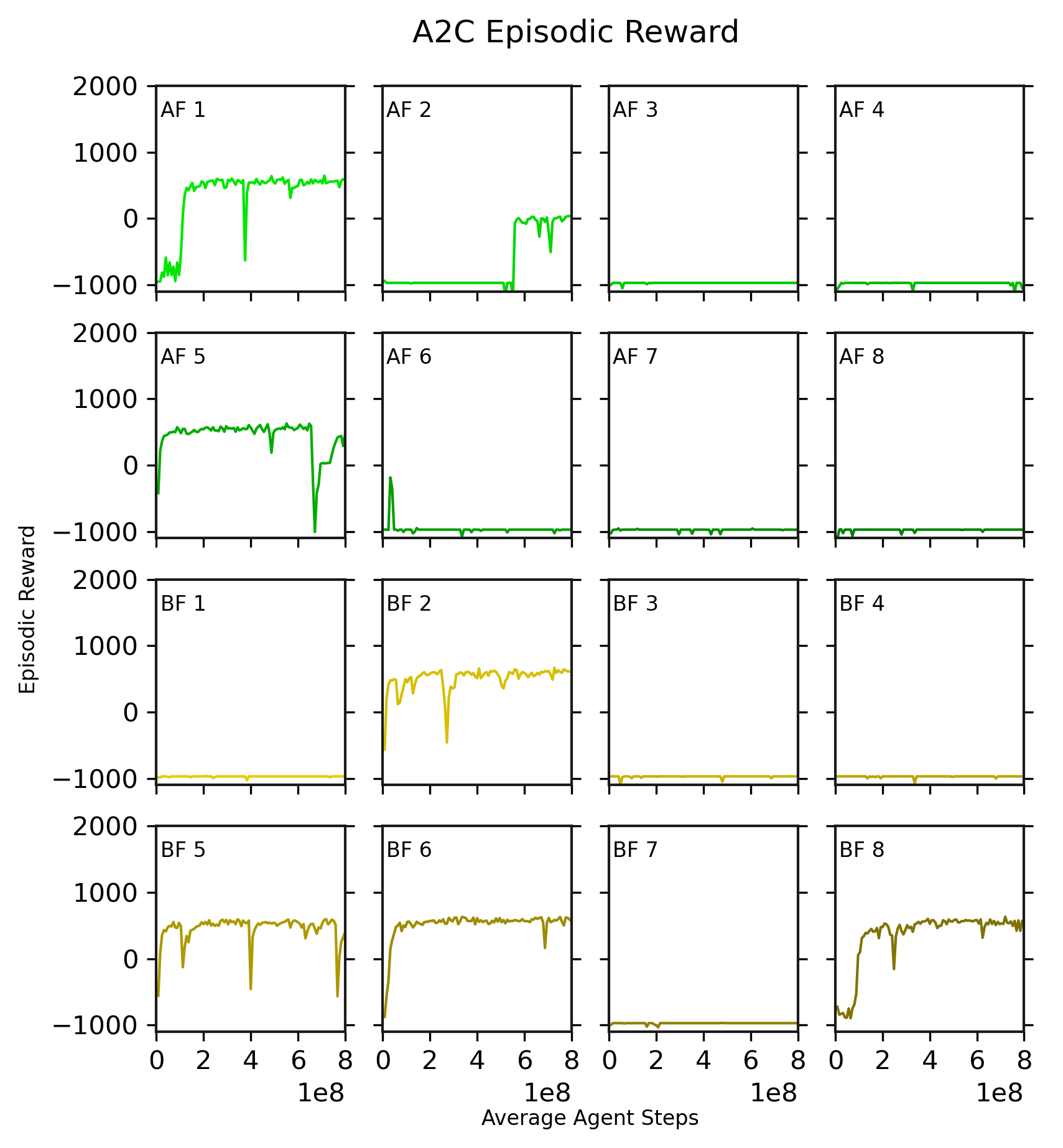}
    \caption{Average episodic reward for individual A2C agents. Note that only some agents learn to find reward, while most achieve almost -1000 reward per episode. An agent that never moves or consumes fruit will earn -970 reward per episode due to the hunger penalty.}
    \label{fig:ablation_agent:a2c_individual}
\end{figure}

In addition to the environmental conditions, the architecture of our reinforcement agents clearly also affects what they can learn. Throughout this paper, we have used the V-MPO architecture~\citep{song2019v}, which we have found to be remarkably consistent. For example, see Figure~\ref{fig:baseline_reward}, which shows every agent in the population learning with nearly the same curve of reward over training and reaching nearly the same long term performance. How would a slightly older agent architecture fare? We explore this in Figure~\ref{fig:ablation_agent:vmpo_a2c}, which compares V-MPO and A2C~\citep{mnih2016a3c} in terms of reward over time and the number of exchanges per episode. The difference is quite plain: A2C agents fail to learn to trade in our environment, and also obtain a negative collective reward.

To further investigate the A2C agents, in Figure~\ref{fig:ablation_agent:a2c_individual} we break out the performance of each individual agent over time. Out of sixteen agents, only eight ever achieve a reward greater than -970 per episode, and two of those eight are still well below zero reward on average. -970 is the reward obtained by an agent that never moves and never consumes fruit: it takes 30 timesteps for the hunger penalty to take effect, and then they suffer -1 reward for the next 970 timesteps. Any movement would further decrease that reward at a rate of -0.25 per tile moved. Thus, in this particular environment, A2C appears very unreliable in its ability to learn the basics of even producing and consuming fruit. It is possible that learning to trade is too difficult of a step beyond these fundamentals for even successful A2C agents. However, with less than half of the population even learning to produce fruit, the problem may be that encountering another potential trading partner is too rare for the agents to explore their offer actions and discover trade.

Despite this poor performance, A2C was used in much of the earlier Sequential Social Dilemma work~\citep{leibo2017ssd}. When our work on this project first began in 2018, we used the same codebase and A2C implementation that the SSD effort successfully used. Thus, we believe that the A2C implementation presented here is not simply faulty. In fact, in our early versions of Fruit Market from 2018 to 2019 which used smaller maps and a smaller set of Offer actions, we usually did observe \textit{some} trading occur between A2C agents. However, as with the individual agent results in Figure~\ref{fig:ablation_agent:a2c_individual}, we commonly observed up to half of the agents failing to learn any behaviour other than spinning in place. Further, even among successful agents that obtained more than zero reward, we frequently observed downward crashes in performance similar to those shown by agents AF1, AF5, BF2, or BF5. Even in experiments where trading behaviour emerged, it was intermittent and would often drop to zero exchanges per episode for spans of hundreds of episodes before recovering. Further, in Supply and Demand experiments, the A2C agents did not adjust their price as we varied the environmental conditions, and instead used only the 1a:1b offer. And as the above figures have shown, in the current environment presented throughout this paper where V-MPO is successful, using larger maps, more offers to choose from, and the movement penalty, our A2C agents fail to develop trading behaviour altogether.

In future work in this area, using whichever future agent architectures come after V-MPO, we anticipate further progress without having to adjust the environment. In particular, as we will discuss next, we hope that our agents can discover trading behaviour using even simpler actions than the Offer actions we have developed in this work.

\subsection{Trade Mechanics}
\label{sec:ablation:trade}

Finally, we will consider the actions that players use to express offers to each other, and the environmental mechanisms that pair those offers into exchanges. All of our results thus far have used the offer actions described in~\ref{sec:environment}, in which agents use actions that directly map to each possible offer expressed in quantities of apples and bananas. When two or more players are within trading range and are advertising compatible offers (\ie, each player will give at least as many items as the other requests), the environment automatically selects the most-generous pair of players, and swaps the items between the players' inventories in one step. This particular mechanism is just one option in the middle of a range options, with both simpler and more complex alternatives. We will now explore some of those alternatives, and justify our choice by demonstrating that it is a useful trade-off that provides agents with control over their trades, while still resulting in trading behavior emerging from current agents.

\subsubsection{Drop and Give Actions}
\label{sec:ablation:trade:drop_give}

We begin by exploring simpler actions that agents can use to exchange goods: simply dropping an apple or banana on the ground in front of them (two \textbf{Drop actions}) or giving an apple or banana to the closest player (two \textbf{Give actions}). These actions provide very simple ways for the agents to trade items: they do not require any abstract notion of offers, or observations to perceive others' offers, or for the environment to facilitate exchanges by deciding which pair of players should trade and swapping the items between their inventories. Instead, the Drop actions allow one player to approach another, drop an apple on the ground between them, wait for the other player to drop a banana, and then pick up each others' items. Likewise, a player could drop two apples to suggest a better offer, or wait for two bananas to be dropped before moving away from their apple. Similarly, the Give actions allow players to perform this exchange more quickly, requiring only one Give action each (possibly but not necessarily simultaneously) instead of a sequence of move and drop actions. 

While these examples are suggestive of bartering, the Drop and Give actions would also permit other, perhaps altruistic, uses such as simply gifting items to other players without immediately expecting an item in return, or communal production and pooling of goods. Or, perhaps agents could learn to remember who has given them items in the past so that they can repay the debt later on, thus stretching an exchange across time. In such cases, it may be difficult to call the behaviour ``trading'', or to identify which actions constituted ``an exchange'', or to determine the ratio of items that were exchanged. More likely, however, is the possibility of theft. Since the exchange requires a sequence of actions by both players, one player could take the dropped or given fruit without giving anything in return.

\begin{figure}
    \centering
    \includegraphics[height=2in]{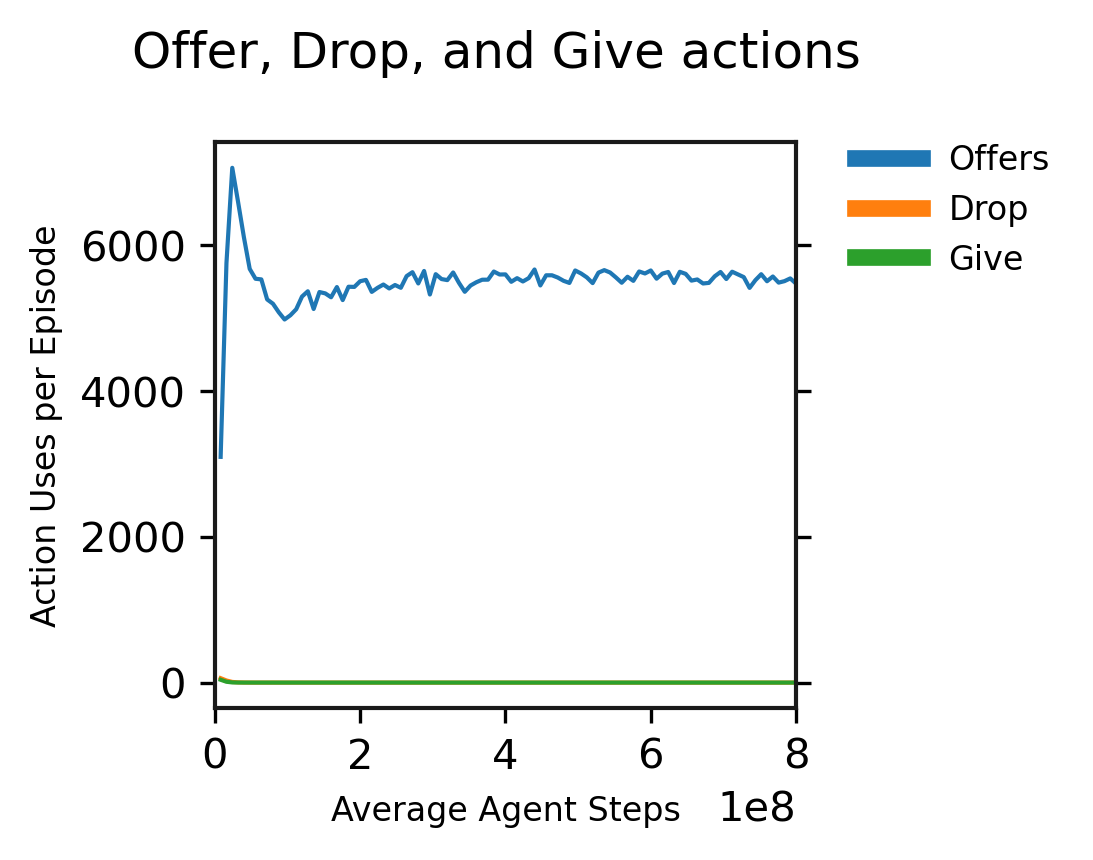}
    \caption{Frequency of use of the offer, drop, and give actions. Each line shows the collective use of each action type when only it, and not the alternatives, are provided. The 'Drop' and 'Give' lines overlap at zero use after a brief initial exploration.}
    \label{fig:ablation_drop_give}
\end{figure}

However, the possibility of such uses of the Drop and Give actions appears moot with current agents. In Figure~\ref{fig:ablation_drop_give}, we present experiments where agents had either the Offer actions, two Drop actions (``Drop Apple'' and ``Drop Banana''), or two Give actions (``Give Apple'' and ``Give Banana'') for giving an item to the nearest player. The environment is the uniform density $(a=1,b=1)$ setting used throughout Section~\ref{sec:experiments:baseline}. Each line shows the total usage of each type of action per episode, summed across all timesteps and players.\footnote{After an Offer action is taken, the offer may stay active for several timesteps until the player either finds a partner to trade with or uses the Cancel action. Thus, one use of an Offer action may result in several timesteps of offer usage in the graph. The Drop and Give actions have instantaneous effects, and so one use of the action counts as one activation in the graph.} While the Offer actions are quickly adopted as a method of exchange, the Drop and Give actions are both rarely explored initially, and almost immediately converge to zero use. 

The lack of use of the Drop and Give actions is perhaps not surprising: to a reinforcement learning agent that has just explored by dropping a rewarding item on the ground, the natural short-term reward maximizing actions would be to pick it up and then eat it. Exploring the drop action to discover a trading convention requires not only further exploration by resisting this temptation, but also requires another player do so simultaneously, and nearby, and with the desired other good, and then for both players to switch positions without one player picking up both items. The Give actions require just one action use by each player but are still difficult to explore, as a player who receives a gifted item is not required to reciprocate. By comparison, the Offer actions require a single action use by each player, stay active across time until fulfilled or cancelled, and theft is not possible because the environment simultaneously swaps their items.

Nonetheless, it would be very satisfying if our agents could learn to use very simple mechanics such as the four Drop and Give actions, without requiring the larger set of offer actions to be provided to the players, or for the environment to select partners and swap their items. These environmental mechanics encode a partial solution to the challenge of trading, instead of requiring our agents discover a solution for themselves. Unfortunately, while we believe groups of human players would learn to use actions like Drop or Give, these results suggest that current agents cannot. Thus, to make progress on eliciting trading behaviour, we arrived at the offer and exchange mechanism presented in this work which are learnable by current agents.

\subsubsection{Compatible versus Inverse Offer Resolution}
\label{sec:ablation:trade:inverse}

\begin{figure}
    \centering
    \includegraphics[height=3in]{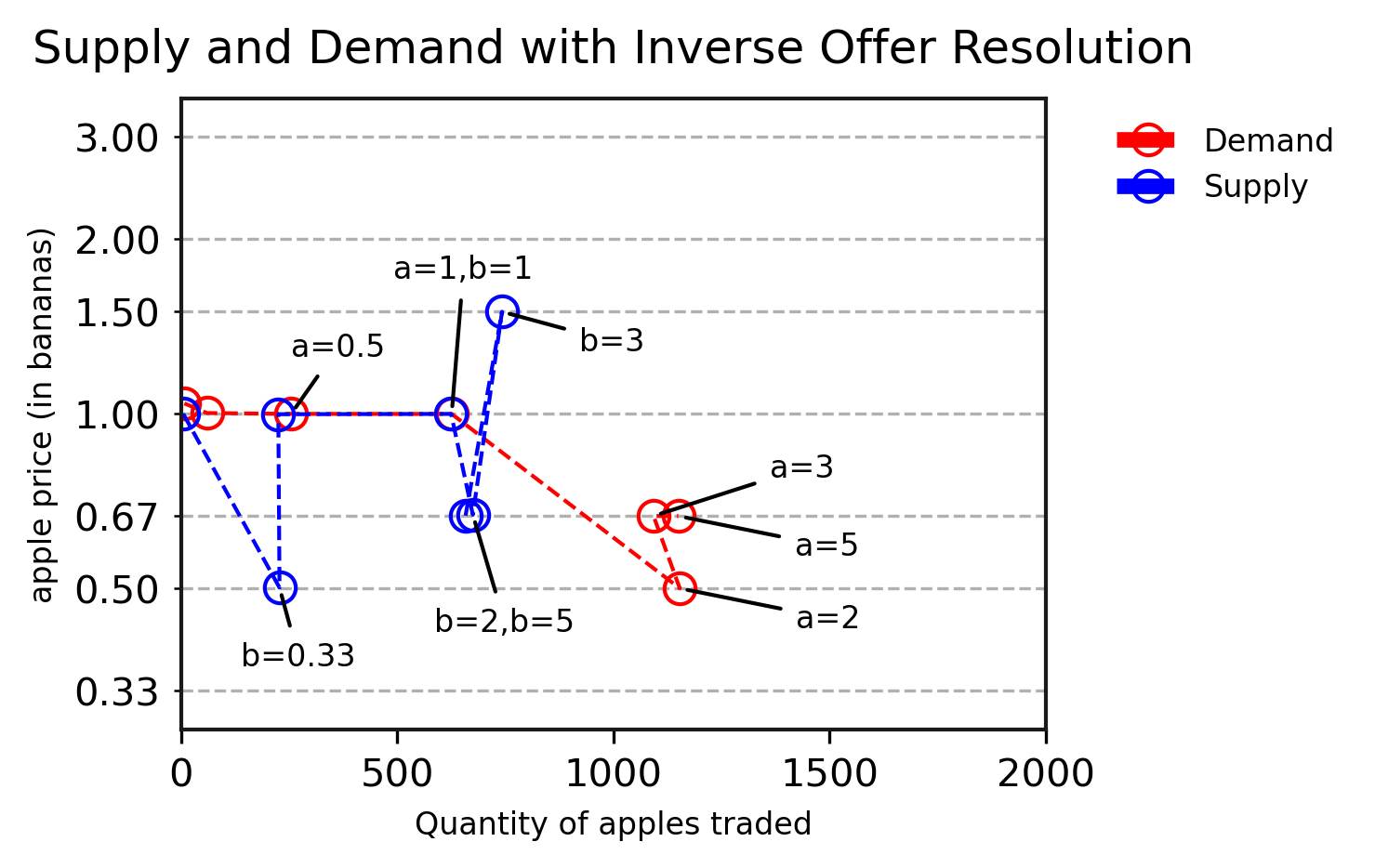}
    \caption{Supply and Demand experiment when only inverse offers are matched for exchanges, instead of the default mechanism of matching any compatible offers at the lowest price that satisfies both parties (see Figure~\ref{fig:sd-spawn-spawn:traded}). While trade largely does still emerge as a behavior, it is less frequent than in the default case, and the equilibrium prices are unpredictable.}
    \label{fig:ablation_inverse:sd}
\end{figure}

Next, we will explore removing the environment's bias towards higher offers when pairing players' offers into an exchange. As described in Section~\ref{sec:environment}, when two or more players are nearby and making compatible offers, the environment decides which pairs of players will exchange goods. Specifically, for each player, the environment finds the set of other players whose offers will provide the highest quantity of goods that the first player requests, while demanding no more than the first player is willing to give. The environment chooses a trade partner from that set, breaking ties by distance and then randomly, and then exchanges the goods at the lowest quantities that satisfy each player. Thus, a player can offer excess goods to prioritize their trade over any competitors, but may not have to pay those excess goods unless their partner's offer demands them.

This mechanism injects domain knowledge into the environment: first, about which offers are compatible with each other, and second, that excess goods should make an offer more attractive. We can remove this domain knowledge by exploring a different trade mechanic in the environment, where players only exchange goods if their offers are an exact inverse of each other. Ties can be broken by distance and then randomly, as before. This new mechanism would remove complexity from the environment, and also force agents to learn more about their observations in order to trade. If one player is offering 1 apple for 1 banana, and a nearby player is generously offering 2 bananas for 1 apple, the environment will no longer exchange their goods until one player changes their offer to the inverse of the other.

Figure~\ref{fig:ablation_inverse:sd} shows a Supply and Demand graph using this trade mechanism, which we call \textbf{inverse offer resolution}. The intervention on supply and demand is to vary the spawn rate of apple and banana trees, exactly as was performed in Figure~\ref{fig:sd-spawn-spawn:traded}, and these two figures should be compared to see the impact of this simpler offer resolution rule. While trading behavior does usually emerge with this alternate mechanism, the results are largely inconsistent as compared to our earlier results. Note the 4 out of 14 data points clustered at the left edge of the plot, indicating zero or near zero apples traded, where trading behaviour does not emerge at all with this mechanism. Further, when either item spawns with a multiplier under 1, the resulting price stays at 1-for-1 (with $b=0.33$ as the only exception), or trade does not emerge. When either item spawns with a multiplier over 1, the resulting price usually moves in the expected direction, such as a higher apple tree spawn rate resulting in a lower value for apples. However, this does not always happen: for example, the b=2 and b=5 cases where bananas are much more plentiful than apples and we would expect the apple price to go up, yet 0.67 bananas can buy one apple (\ie, 3 apples are worth 2 bananas).

We believe that this alternative mechanism, although simpler and encoding less domain knowledge, may introduce at least two problems. First, it may be more difficult for agents to learn how to trade at all, because two agents must jointly pick the exact inverse actions from the set of 18 possible offer actions to see any outcome, instead of only having to pick any two compatible offers. Thus, it may take longer for agents to discover trading behaviour, if at all. Second, it may be more difficult for agents to explore other offers beyond those currently used by the population, because it requires both participants to change their offers. For example, assume that the population is currently using only the ``Give 1 apple for 1 banana'' offer and its inverse ``Give 1 banana for 1 apple''. With our default mechanism, a banana selling agent may increase their offer to ``Give 2 bananas for 1 apple'' to prioritize their offer over competing lower offers, while still being matched for an exchange with banana buyers using the lower offer, who do not have to change their behaviour. With the exact inverse mechanism, a banana seller can increase their offer, but must wait for a banana buyer to notice a generous offer in their observation, and then change their own offer to the inverse of the banana seller's offer. While this would benefit the banana buyer, as the exchange would happen at the higher price (unlike our default mechanism), exploring this mechanic would be difficult because it would mean passing up offers from other players at the lower dominant price. Thus, whichever set of offers is first discovered by players may become difficult to shift away from.

\begin{figure}
    \centering
    \begin{subfigure}{0.35\textwidth}
        \includegraphics[height=2in]{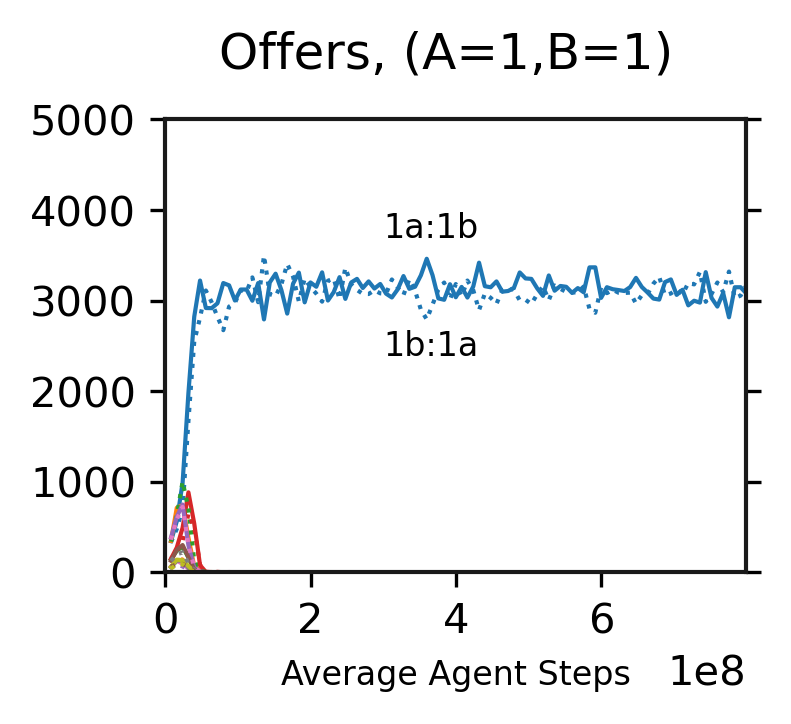}
        \caption{}
        \label{fig:ablation_inverse:offers}
    \end{subfigure}%
    ~
    \begin{subfigure}{0.65\textwidth}
        \includegraphics[height=2in]{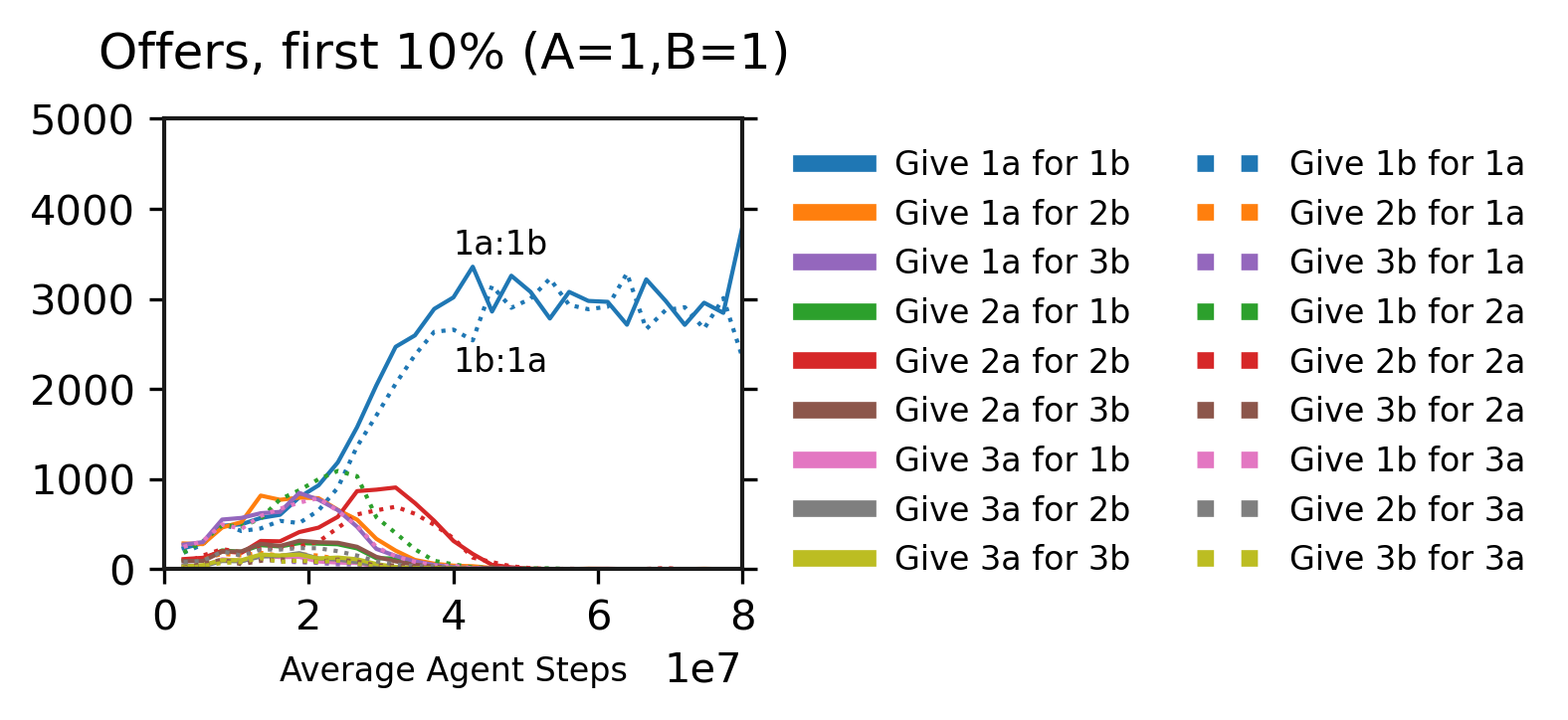}
        \caption{}
        \label{fig:ablation_inverse:offers_early}
    \end{subfigure}
    
    \begin{subfigure}{0.35\textwidth}
        \includegraphics[height=2in]{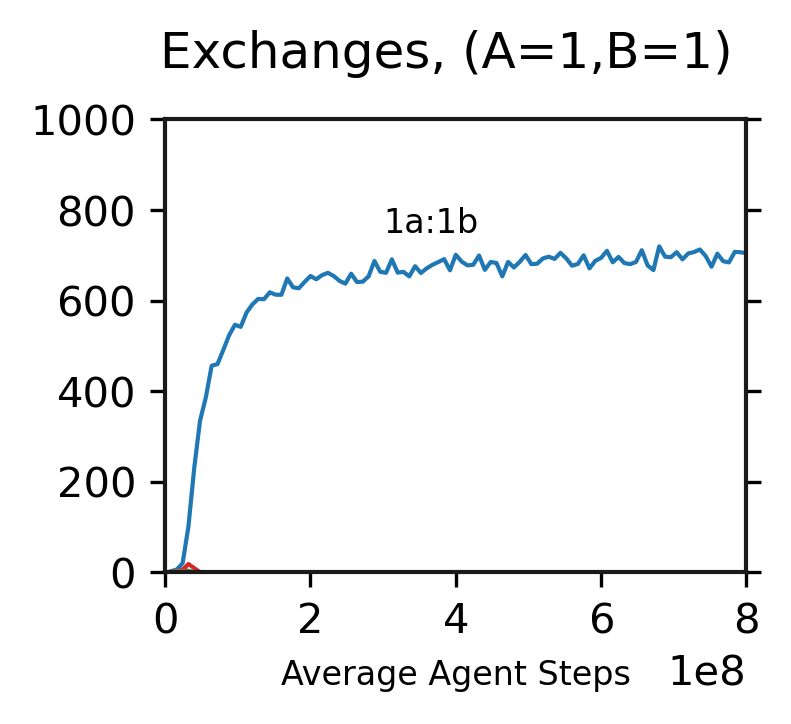}
        \caption{}
        \label{fig:ablation_inverse:exchanges}
    \end{subfigure}%
    ~
    \begin{subfigure}{0.65\textwidth}
        \includegraphics[height=2in]{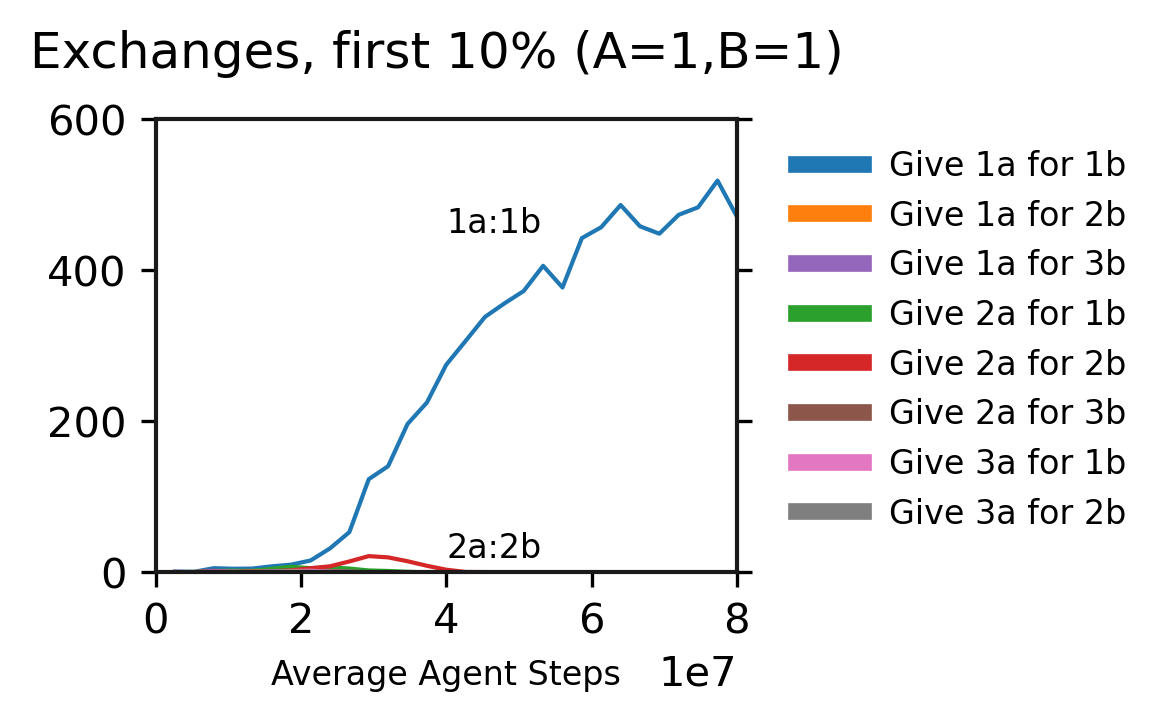}
        \caption{}
        \label{fig:ablation_inverse:exchanges_early}
    \end{subfigure}

    \caption{Offer and exchange frequency with the inverse offer resolution mechanism, in the $(a=1,b=1)$ setting. (b) and (d) zoom in on the first 10\% of the experiment.}
    \label{fig:ablation_inverse:offers_exchanges}
\end{figure}

In Figure~\ref{fig:ablation_inverse:offers_exchanges}, we examine the offers and exchanges made over time in the $(a=1,b=1)$ setting, mirroring our earlier analysis in Figure~\ref{fig:baseline_offers_a1} which used the default mechanism. With regard to the first potential problem, a difficulty to explore and discover trade as a joint behavior, there seems to only a small effect. Figures~\ref{fig:ablation_inverse:offers_early} and~\ref{fig:baseline_offers_a1:offers_early} focus on the first 10\% of the experiments, and in both cases, we see exploration over prices in the first $5e7$ timesteps, and consistent exchanges by the end of 1e8 timesteps. In fact, the alternate mechanism results in about 500 exchanges per episode after 1e8 timesteps compared to 200 exchanges using the default mechanism, although these exchanges are ``1 apple for 1 banana'', compared to ``3 apples for 3 bananas'' for the default case: the same ratio, but with less throughput in each exchange. The second potential problem appears more significant: the increased difficulty in exploring different prices. In Figure~\ref{fig:ablation_inverse:exchanges_early} we see that the ``1 apple for 1 banana'' exchange catches on early and remains dominant throughout the experiment, with only a brief and small use of ``2 apple for 2 banana'' exchanges. In comparison to the default mechanism in Figure~\ref{fig:baseline_offers_a1:exchanges_early}, where we saw an overlapping progression from exchanges at 1a:1b to 2a:2b to 3a:2b to 3a:3b, resulting in efficient throughput, the lack of visible movement to nearby prices is likely an issue. We believe the inconsistent Supply and Demand results presented in Figure~\ref{fig:ablation_inverse:sd} are likely related to this difficulty in moving away from whichever price is established first.

Overall, while the inverse offer resolution mechanism would also remove domain knowledge from the environment, the resulting behaviour does not respond to supply and demand changes, as shown in Figure~\ref{fig:ablation_inverse:sd}. Further, this mechanism would be even more difficult to learn if we increased the range of possible offers, as it would be even less likely for two nearby agents to simultaneously explore inverse offers. Thus, we decided to accept the inclusion of domain knowledge inherent in the compatible offer resolution mechanic.

\subsubsection{Accept Actions}
\label{sec:ablation:trade:accept}

Next, we will consider actions allowing players to directly accept offers proposed by other players. Recall from Section~\ref{sec:environment} that an exchange is performed automatically by the environment when two players are within each others' trade radius and are making compatible offers. While our results thus far have shown that this exchange mechanism can be learned by the players, it is unsatisfyingly artificial for the environment to have to select which pairs of players trade, and with what quantity of goods. In this section, we will explore the consequences of adding actions to let players directly and atomically accept an offer proposed by another player, without relying on the environment.

Specifically, we will add two new actions, ``Buy Apple'' and ``Buy Banana'', which are used in addition to the existing Offer actions. When a player uses a Buy action, the environment considers all affordable offers within their trade radius, and selects the offer with the highest ratio for the desired good. If such an offer exists, the player immediately exchanges goods with the player making the offer. Mechanically, this is handled exactly as if the player had chosen the inverse offer action, and the environment then selected that pair of players to exchange goods. Taking the Accept action simplifies the player's decision, as they do not have to learn which of their many offer actions is compatible with the offers around them, and their exchange is guaranteed to occur immediately without risk of another player being selected for the exchange instead. We could go even farther, by requiring the player to choose a particular other player's offer to accept instead of automatically selecting the ``best'' offer. But for now, introducing these two Accept actions strikes a balance between granting additional control to the agent while still being simple and easy to learn.

We can configure the environment to either only resolve trades through these Accept actions, or to use the Accept actions in addition to the existing compatible offer resolution mechanism. This gives us three settings to explore: ``Offer Resolution Only'', which is the default used in the earlier experiments, ``Offer Resolution \& Accepts', where exchanges are resolved either by both players making an offer and letting the environment facilitate the exchange or by one using an Accept action, and ``Accept Only'', where exchanges only occur when a player uses an Accept action. Note that in all three cases the players still have the Offer actions, and one player must use an Offer action in order for another to Accept that offer.

\begin{figure}
    \centering
    \centering
    \begin{subfigure}{0.5\textwidth}
        \centering
        \includegraphics[height=2in]{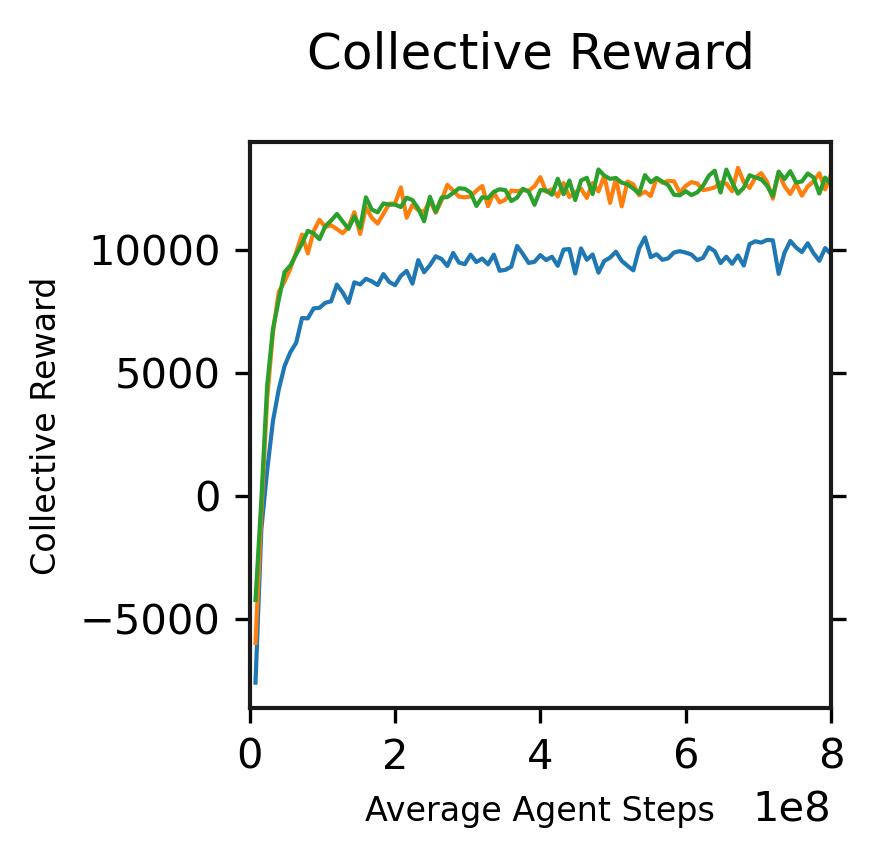}
        \caption{}
        \label{fig:ablation_accept:collective_reward}
    \end{subfigure}%
    ~
    \begin{subfigure}{0.5\textwidth}
        \centering
        \includegraphics[height=2in]{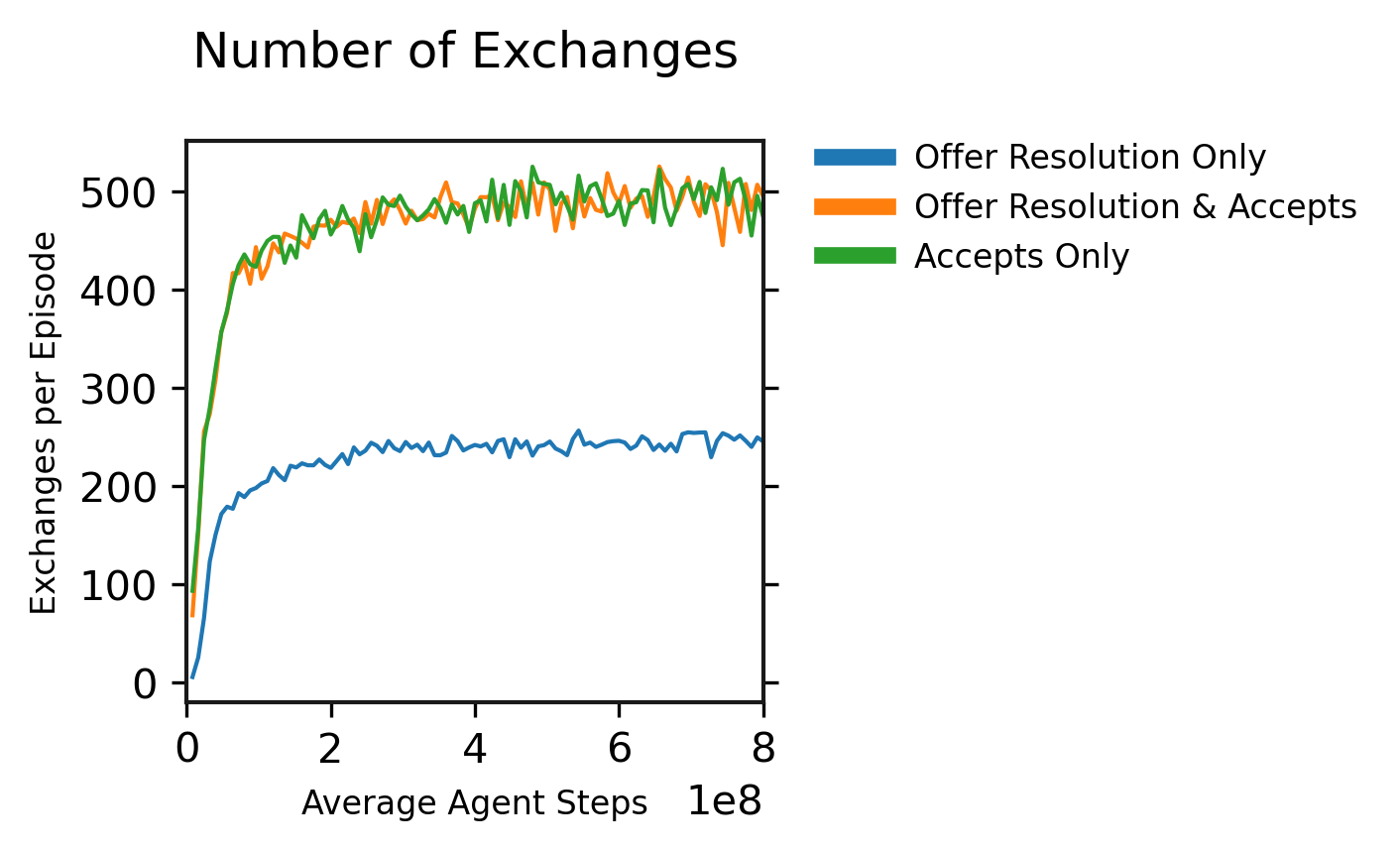}
        \caption{}
        \label{fig:ablation_accept:num_exchanges}
    \end{subfigure}
    \caption{Collective reward and trade frequency with and without Accept actions and automatic offer resolution. The presence of the Accept actions results in more exchanges and greater collective reward.}
    \label{fig:ablation_accept:reward_exchanges}
\end{figure}

Figure~\ref{fig:ablation_accept:reward_exchanges} plots the collective reward and number of exchanges over time in these three settings, using the $(a=1,b=1)$ environment used in Section~\ref{sec:experiments:baseline}. Both the collective reward and frequency of exchanges is increased when the Accept actions are available. This is perhaps unsurprising: an Accept action lets a player easily buy goods at the best nearby price instead of having to make an appropriate offer, and may thus be easier to learn and use.

\begin{figure}
    \centering
    \begin{subfigure}{0.3\textwidth}
    \centering
    \includegraphics[height=2in]{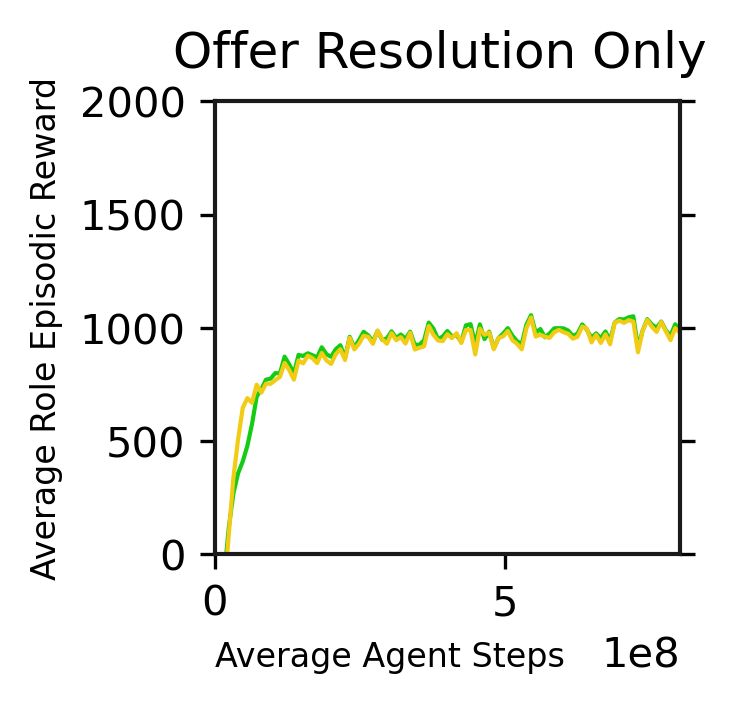}
    \caption{}
    \label{fig:ablation_accept:role_reward:offer_only}
    \end{subfigure}%
    ~
    \begin{subfigure}{0.3\textwidth}
    \includegraphics[height=2in]{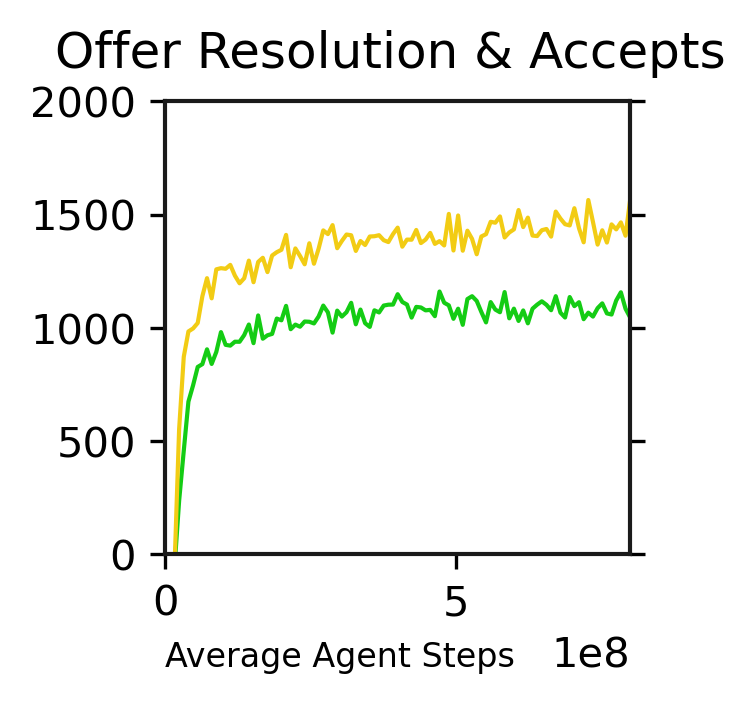}
    \caption{}
    \label{fig:ablation_accept:role_reward:offer_accept}
    \end{subfigure}%
    ~
    \begin{subfigure}{0.4\textwidth}
    \includegraphics[height=2in]{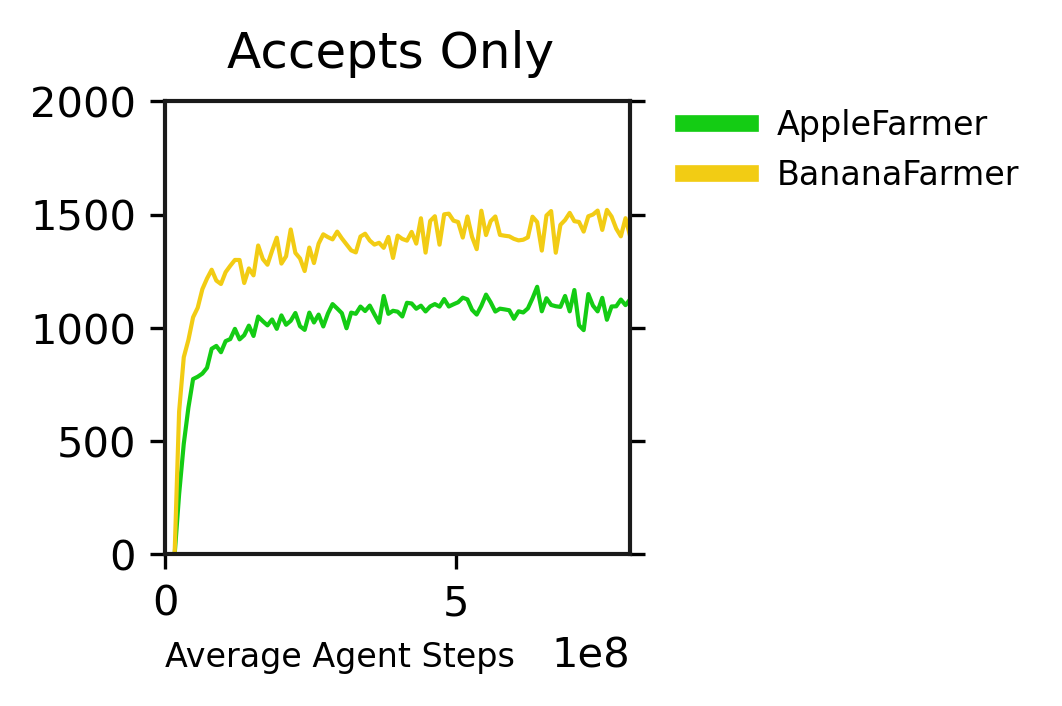}
    \caption{}
    \label{fig:ablation_accept:role_reward:accept_only}
    \end{subfigure}

    \caption{Average episodic reward for players of each role, with and without the Accept actions and automatic offer resolution. Note the higher reward for only one role when the Accept actions are introduced.}
    \label{fig:ablation_accept:role_reward}
\end{figure}

However, the Accept actions also introduce a problem, in that the agents' learned behaviour is no longer as rational. In Figure~\ref{fig:ablation_accept:role_reward}, we probe the collective reward results further by presenting the average reward for each role. Without the Accept action, in Figure~\ref{fig:ablation_accept:role_reward:offer_only}, the Apple Farmer and Banana Farmer roles perform nearly identically and earn about 1000 reward per episode on average. However, in both conditions where the Accept actions are introduced, we see an unexpected advantage for Banana Farmers, who earn almost 1500 reward per episode, while Apple Farmers earn just over 1000. This is quite strange, as there is no systematic advantage for either role in the environment or in the actions available; both roles can use the Offer actions and the Accept actions, and apple and banana trees are equally common, so why would the introduction of the Accept actions have this effect?

\begin{figure}
    \centering
    \includegraphics{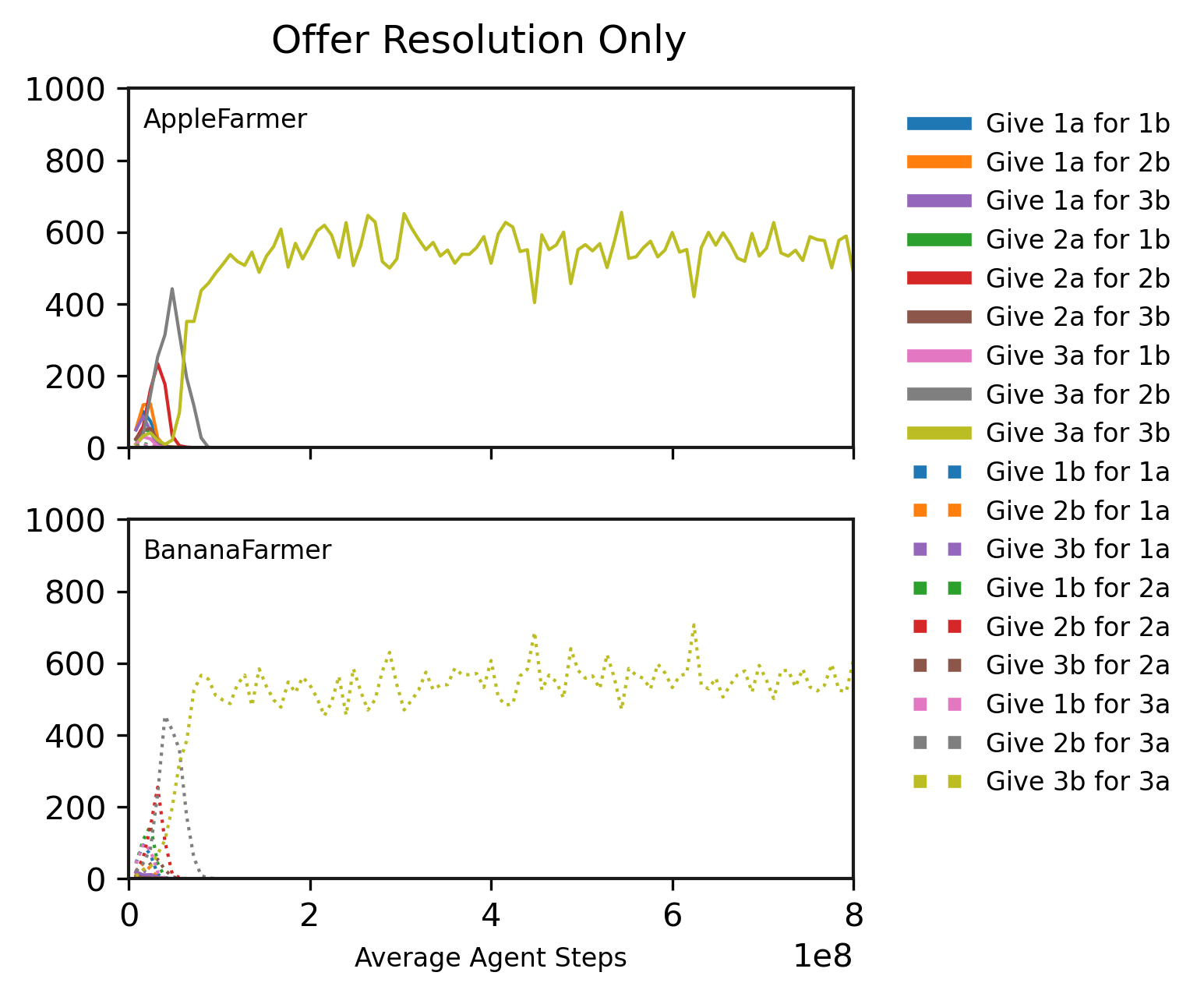}
    \caption{The average usage of each Offer per episode, separated by role, under the ``Offer Resolution Only'' mechanism.}
    \label{fig:ablation_accept:offers:offer_only}
\end{figure}

\begin{figure}
    \centering
    \includegraphics{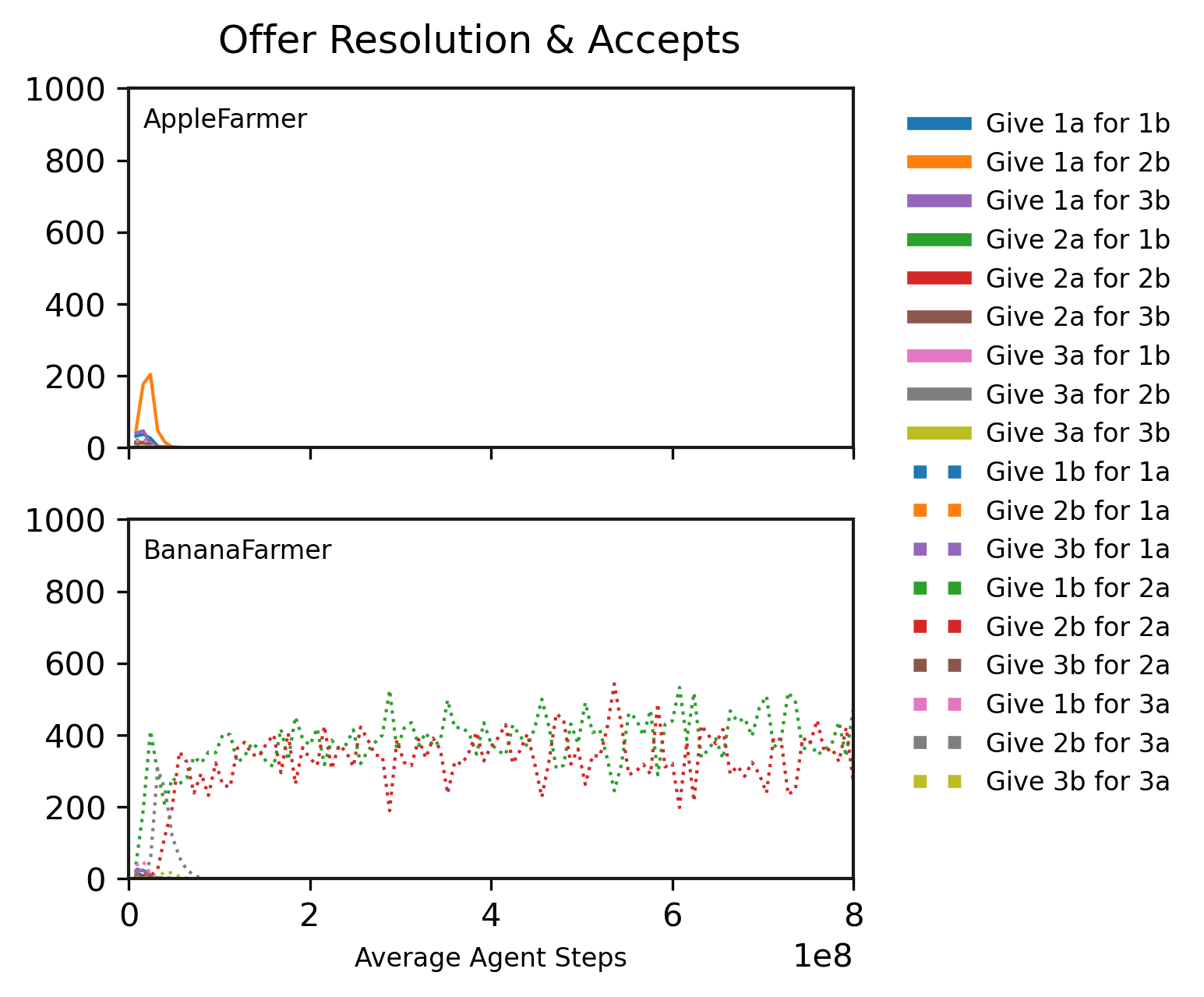}
    \caption{The average usage of each Offer per episode, separated by role, using the ``Offer Resolution and Accepts'' mechanisms. When Apple Farmers stop using the Offer actions, they are instead using the Accept actions to accept the Banana Farmers' offers.}
    \label{fig:ablation_accept:offers:offer_accept}
\end{figure}

\begin{figure}
    \centering
    \includegraphics{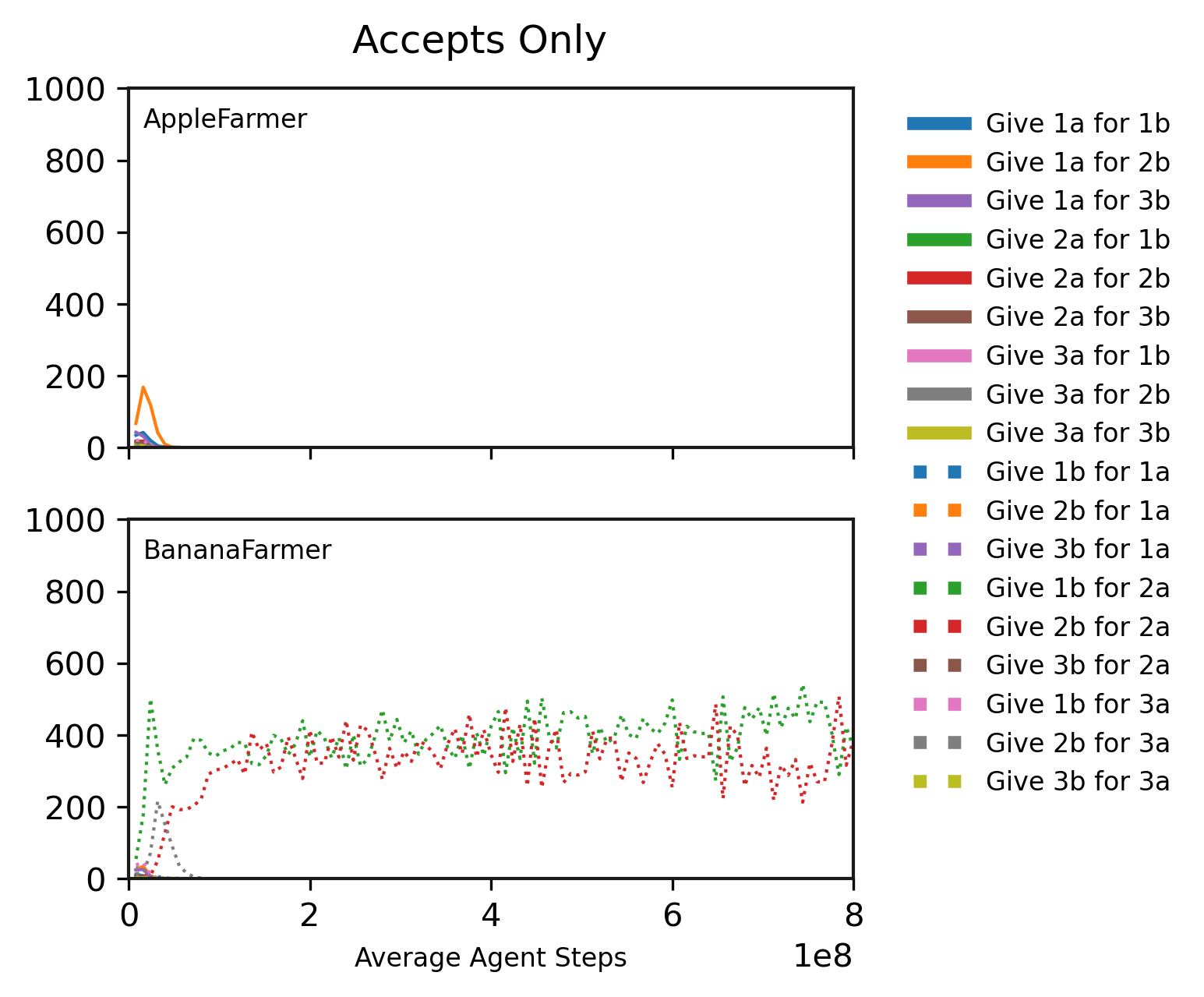}
    \caption{The average usage of each Offer per episode, separated by role, using the ``Accepts Only'' mechanism. When Apple Farmers stop using the Offer actions, they are instead using the Accept actions to accept the Banana Farmers' offers.}
    \label{fig:ablation_accept:offers:accept_only}
\end{figure}

\begin{figure}
    \centering
    \includegraphics[height=2in]{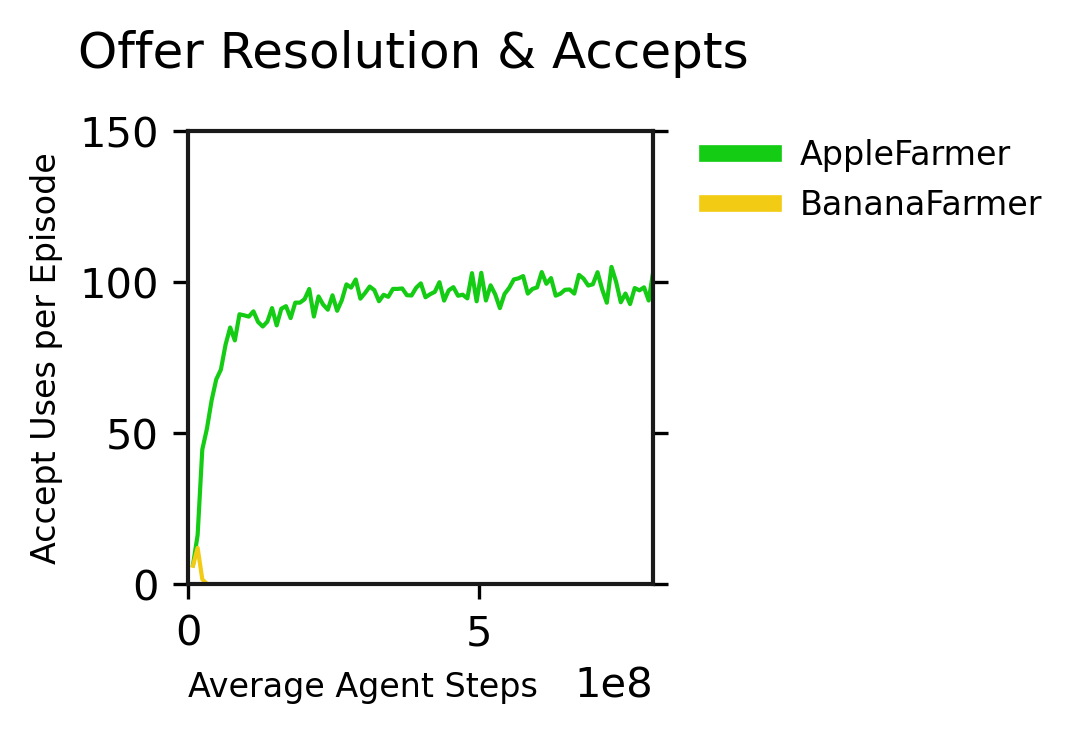}
    \caption{Usage of the Accept actions by role in the ``Offer Resolution \& Accepts'' setting. Apple Farmers quickly learn to use the Accept offer while Banana Farmers do not. Compare against Figure~\ref{fig:ablation_accept:offers:offer_accept}, which shows that Banana Farmers make offers, and Apple Farmers do not.}
    \label{fig:ablation_accept:num_accepts}
\end{figure}

We can further investigate this odd result by examining which offers are being made and by who. In Figures~\ref{fig:ablation_accept:offers:offer_only} through~\ref{fig:ablation_accept:offers:accept_only}, we plot the average usage of each offer per episode, separated by role. Recall from Section~\ref{sec:environment} that after a player uses an Offer action, the offer is active until the player trades, cancels it, or uses a different Offer action. In the default ``Offer Resolution Only'' case shown in Figure~\ref{fig:ablation_accept:offers:offer_only}, we see that both Apple and Banana Farmers make offers, and at about the same frequency of just under 600 timesteps of use per episode. The population briefly explores several lower offers (1a:2b, 2a:2b, 3a:2b) before settling on Apple Farmers offering 3a:3b and Banana Farmers offering 3b:3a. However, the behaviour is quite different when the Accept actions are available. In Figures~\ref{fig:ablation_accept:offers:offer_accept} and~\ref{fig:ablation_accept:offers:accept_only}, we see that Apple Farmers only briefly explore the offer actions before dropping to zero uses per episode, while the Banana Farmers consistently use both the 1 banana for 2 apples and 2 bananas for 2 apples offers. Since we know that exchanges happen (as shown in Figure~\ref{fig:ablation_accept:num_exchanges}) and only Banana Farmers make offers, the Apple Farmers must be using the Accept action to accept those offers. We confirm this in Figure~\ref{fig:ablation_accept:num_accepts}, which plots the usage of the Accept actions by role in the ``Offer Resolution \& Accepts'' case. There are two results here worth investigating: first, the behaviour where only one role's agents make offers which the other role's agents accept, and second, the Banana Farmers' use of two offers (``Give 1b for 2a'' and ``Give 2b for 2a'') that represent a lower value for apples than for bananas.

We will start with the behaviour where only one role makes offers. First, note that many behaviours are possible in the ``Offer Resolution \& Accepts'' case: agents of both roles could only make offers and not use the Accept actions at all, or agents of both roles could make offers and accept offers, or (as occurred here) agents of either role arbitrarily could learn to make offers which agents of the other role accept. Further, individual agents of the same role could learn different behaviours, such as some Apple Farmers making offers and other Apple Farmers accepting offers. Even if the rest of the population converged to the joint behaviour of Banana Farmers making offers and Apple Farmers accepting them, a lone Apple Farmer could make offers and trade successfully with Banana Farmers without the Banana Farmers having to change their behaviour.

In both of the cases with Accept actions in Figures~\ref{fig:ablation_accept:offers:offer_accept} and~\ref{fig:ablation_accept:offers:accept_only}, the population has arrived at a convention where Apple Farmers solely use the Accept actions to accept the Banana Farmers' two offers of 1b:2a and 2b:2a. There is no systemic environmental condition that would cause this particular convention to be adopted, and it is likely a coincidence that it was reached in both the ``Offer Resolution \& Accepts'' and ``Accepts Only'' cases, as we will demonstrate next. To investigate this further, using the ``Offer Resolution \& Accepts'' setting, we performed a parameter sweep of 14 experiments where we varied the spawn rate of either apple trees or banana trees in the range (0.2, 0.33, 0.5, 1.0, 2.0, 3.0, 5.0), as in our Supply and Demand experiments presented earlier in Figure~\ref{fig:sd-spawn-spawn}. With only one exception, in each experiment the agents converged to a behaviour where the agents producing the rarer good made the offers and agents producing the more common good accepted the offers. In the cases (a=0.2, a=0.33, b=2, b=3, b=5), Apple Farmers exclusively made the offers and Banana Farmers exclusively used the Accept actions. In the cases (b=0.2, b=0.33, b=0.5, a=2, a=3, a=5, and the exception, a=0.5), Banana Farmers exclusively made the offers and Apple Farmers exclusively used the Accept actions. In both of the a=1 and b=1 cases in the sweep, Banana Farmers made the offers and Apple Farmers accepted them. Thus, in all 14 of 14 cases, the agents reached an equilibrium where only one role made offers, even though any of the agents of any role, whether as a group or as individuals, could have learned to make offers as we found in the ``Offer Resolution Only'' case. We hypothesize that this behaviour where one role learns to always accept is an easier behaviour for the agents to converge towards than the behaviour where both roles make offers. This is perhaps not too surprising; once some other agents have learned to trade, an agent only has to learn to use one Accept action out of two options, instead of learning which of the 18 offer actions match the population's price.

\begin{figure}
    \centering
    \includegraphics[height=2in]{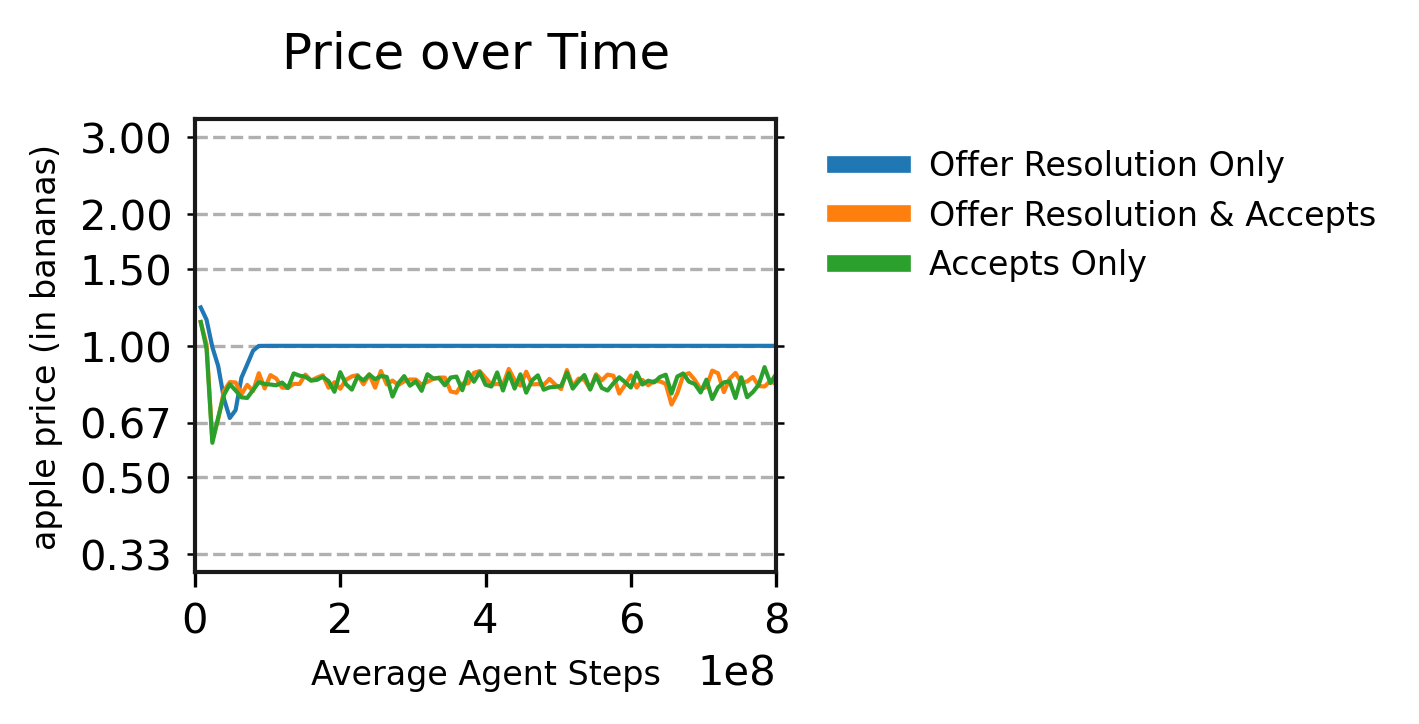}
    \caption{Average price of exchanges over time, with and without the Accept actions and automatic offer resolution. Note that the lower price arrived at when Accept actions are available is a result of the mixed offers observed in Figures~\ref{fig:ablation_accept:offers:offer_accept} and~\ref{fig:ablation_accept:offers:accept_only}, and this lower value for apples relative to bananas in turn explains the higher reward for Banana Farmers in Figure~\ref{fig:ablation_accept:role_reward}.}
    \label{fig:ablation_accept:prices}
\end{figure}

\begin{figure}
    \centering
    \includegraphics[width=\hsize]{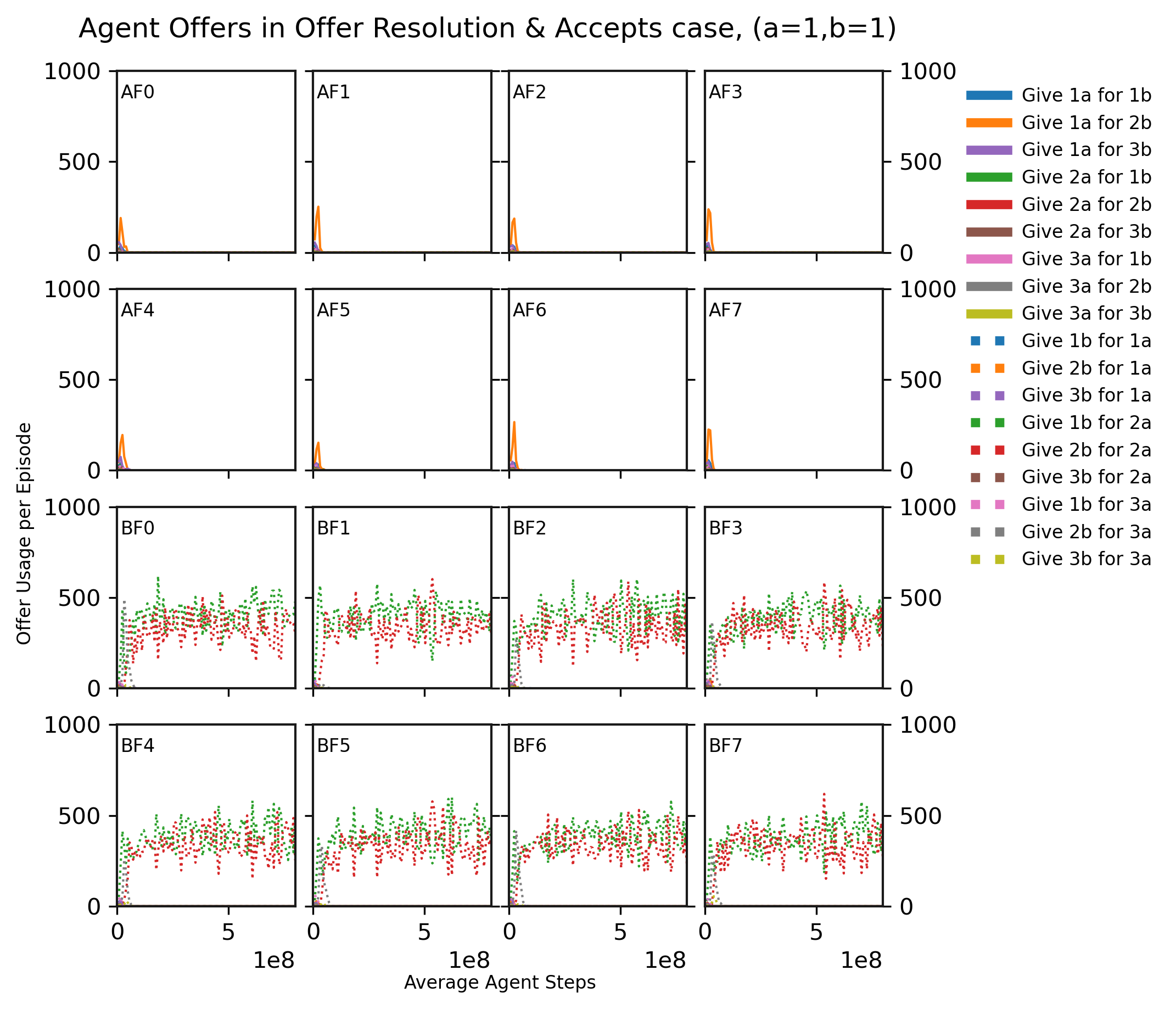}
    \caption{Offers made by agents over time, in the ``Offer Resolution \& Accepts'' case, with default spawn rates for apple and banana trees (a=1,b=1). This plot confirms that all Banana Farmers use both the ``Give 1b for 2a'' and ``Give 2b for 2a'' offers, in contrast to other possible behaviours such as half of the Banana Farmers using each offer.}
    \label{fig:ablation_accept:a1b1_agent_offers}
\end{figure}

Having established that this unexpected behaviour is consistently reached, we can now examine the effect that it has on price and the population-level supply and demand behaviour. In Figure~\ref{fig:ablation_accept:prices}, we plot the average price of exchanges over time in the a=1,b=1 setting for the ``Offer Resolution Only'', ``Offer Resolution \& Accepts'', and ``Accepts Only'' cases. In the default ``Offer Resolution Only'' case we see that exchanges happen at a price where one apple is worth 1 banana, which makes sense as Figure~\ref{fig:ablation_accept:offers:offer_only} showed that the agents converged to only using the ``3 apples for 3 bananas'' offer and its inverse. In the cases with Accept actions, we see that the agents converge to a lower price: on average, 0.83 bananas per apple. This result ties together the Banana Farmers' use of two offers that we observed in Figure~\ref{fig:ablation_accept:offers:offer_accept}, and the higher reward for Banana Farmers that we observed in Figure~\ref{fig:ablation_accept:role_reward}. By using both the ``1 banana for 2 apples'' and ``2 bananas for 2 apples'' offers, the Banana Farmers present a lower value for apples, which Apple Farmers accept. This grants the Banana Farmers more reward than the Apple Farmers, and also more than the Banana Farmers earned in the default ``Offer Resolution Only'' case. 

This use of two offers thus seems advantageous to the agents making offers, and we can probe deeper. While Figure~\ref{fig:ablation_accept:offers:offer_accept} showed that the Banana Farmer agents used two offers \textit{on average}, it does not distinguish whether every Banana Farmer used two offers, or if half offered ``1 banana for 2 apples'' and the other half offered ``2 bananas for 2 apples''. Figure~\ref{fig:ablation_accept:a1b1_agent_offers} presents each individual agent's offers, and confirms that every Banana Farmer agent used both offers. Our hypothesis is that the Banana Farmers were competing with each other by mixing between two offers: using the ``1 banana for 2 apples'' offer to obtain a better price, and the ``2 bananas for 2 apples'' offer to execute more trades by undercutting other Banana Farmers using the first offer, because Apple Farmers using the Accept action automatically select the offer with the best ratio of goods from their perspective\footnote{A more thorough analysis, which we have not performed, would investigate whether the usage of each offer depends on the presence and offers of other nearby Banana Farmers.}.

\begin{figure}
    \centering
    \begin{subfigure}{0.5\textwidth}
        \centering
        \includegraphics[height=2in]{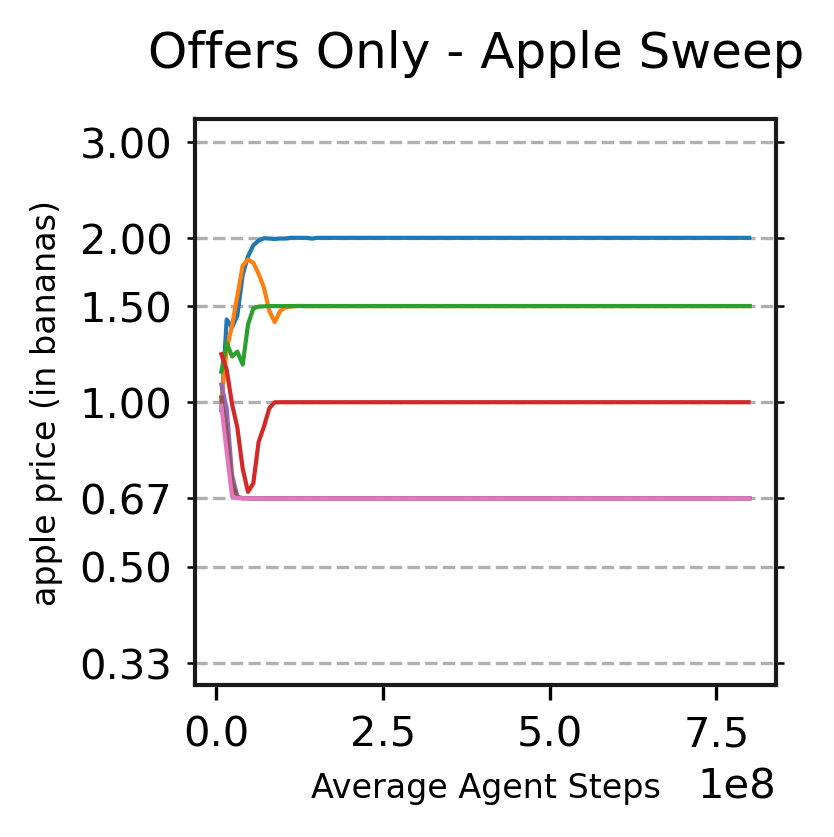}
        \caption{}
        \label{fig:ablation_accept:price_comparison:offer_only_prices_a}
    \end{subfigure}%
    ~
    \begin{subfigure}{0.5\textwidth}
        \centering
        \includegraphics[height=2in]{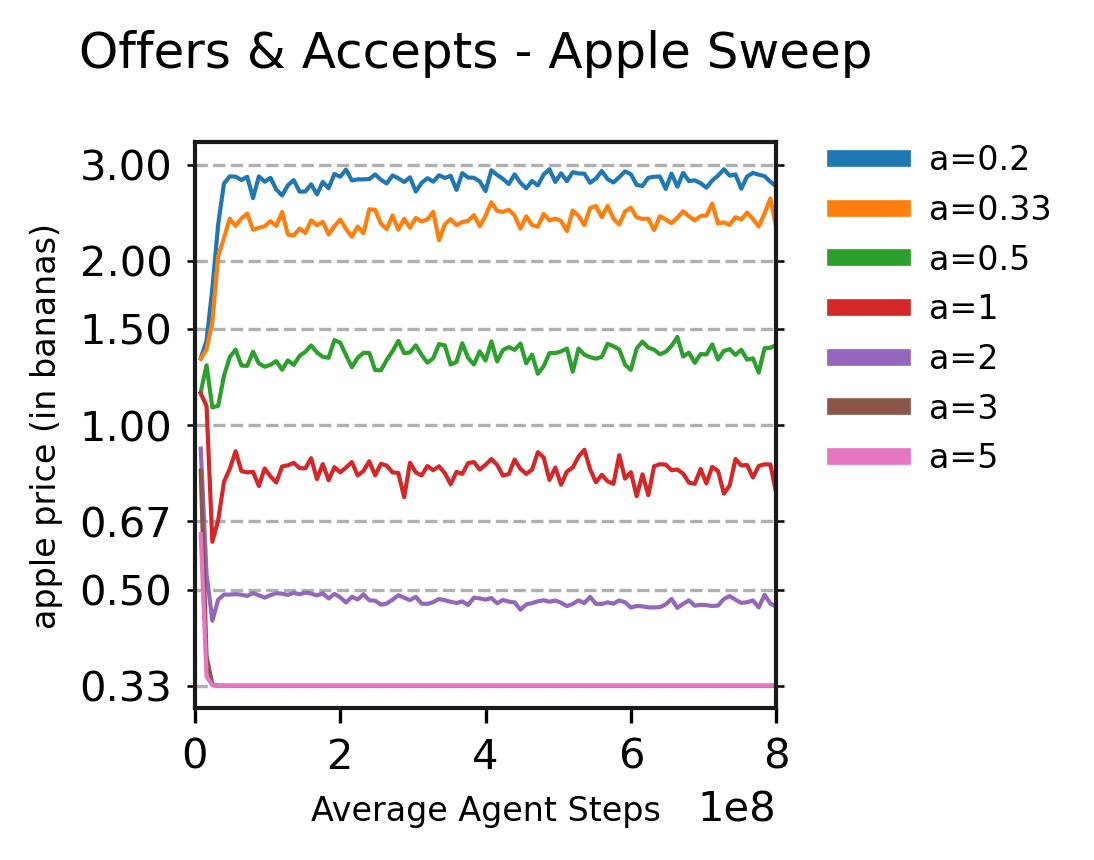}
        \caption{}
        \label{fig:ablation_accept:price_comparison:offer_accept_prices_a}
    \end{subfigure}
    
    \begin{subfigure}{0.5\textwidth}
        \centering
        \includegraphics[height=2in]{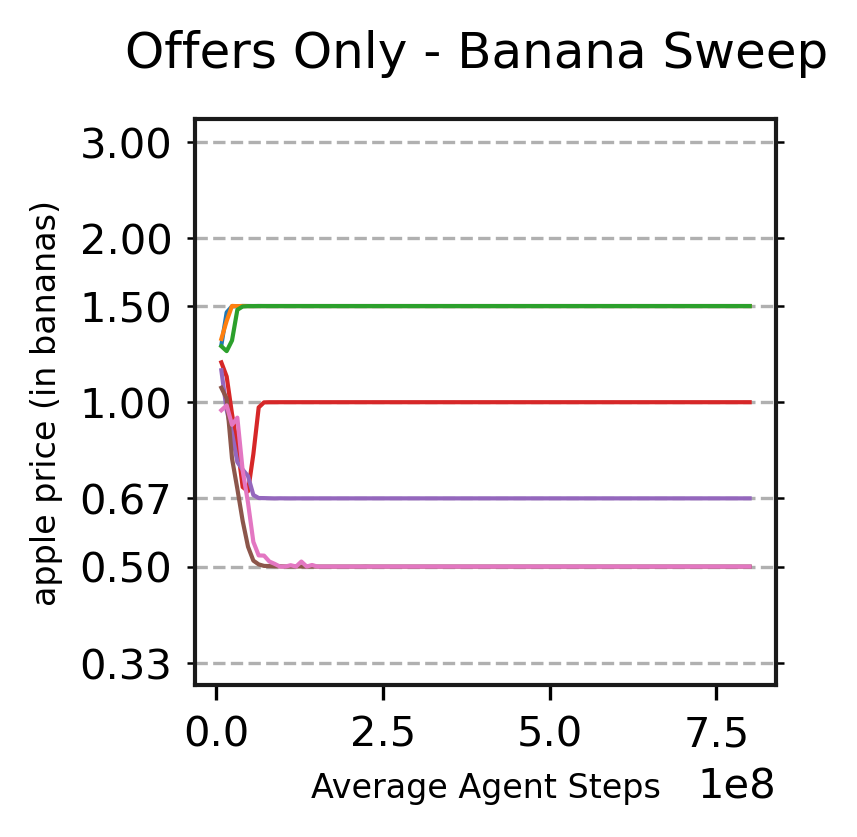}
        \caption{}
        \label{fig:ablation_accept:price_comparison:offer_only_prices_b}
    \end{subfigure}%
    ~
    \begin{subfigure}{0.5\textwidth}
        \centering
        \includegraphics[height=2in]{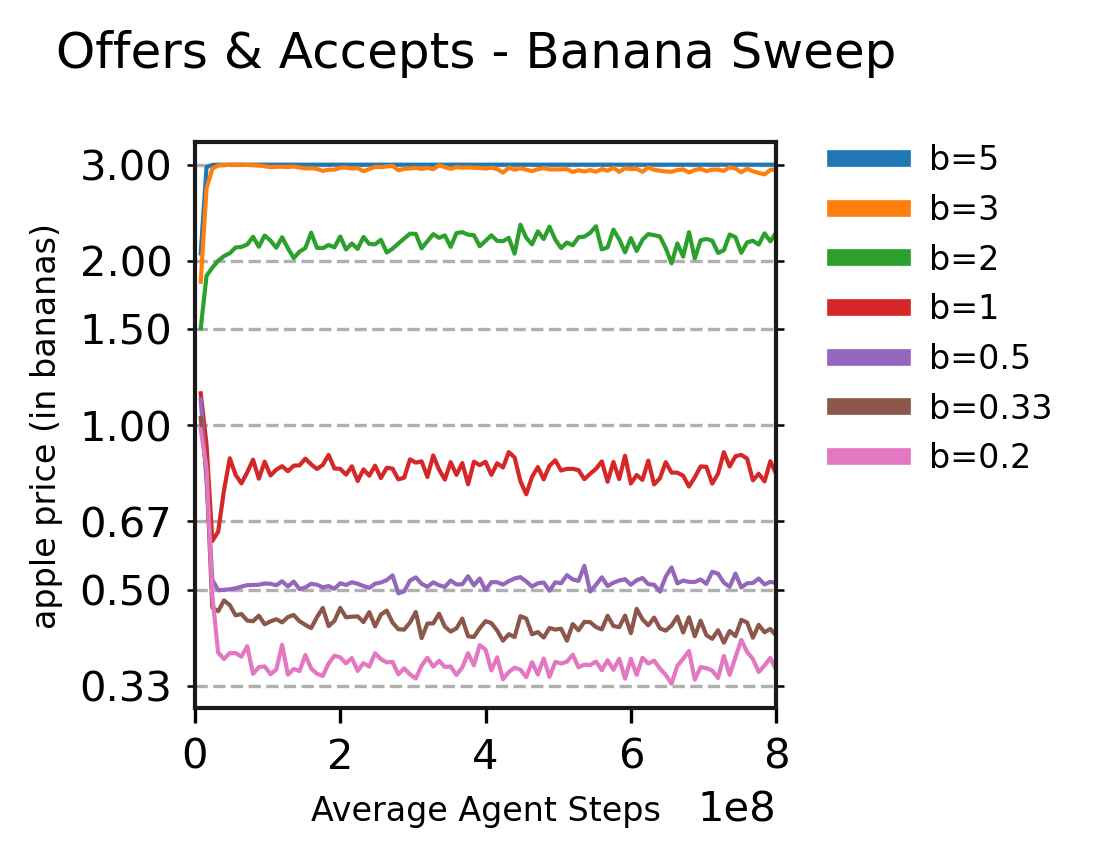}
        \caption{}
        \label{fig:ablation_accept:price_comparison:offer_accept_prices_b}
    \end{subfigure}
    
    \caption{Comparison of average exchange prices between the ``Offer Resolution Only'' and ``Offer Resolution \& Accepts'' cases, in a sweep over apple and banana tree spawn rates. In cases where one line is shadowed by another, note that the legend order matches the vertical order of the lines.}
    \label{fig:ablation_accept:price_comparison}
\end{figure}

Earlier, we described that in 14 of 14 experiments sweeping the tree spawn rates, all agents of one role learned to make offers and all agents of the other role learned to accept them. In 10 out of 14 of these experiments, the agents making offers converged to using either two or three offers. And in all 14 out of 14 experiments, the average price in exchanges was shifted in favour of the agents that made the offers: a higher price for apples if Apple Farmers made the offers, and a lower price for apples if Banana Farmers made the offers. Figure~\ref{fig:ablation_accept:price_comparison} demonstrates this by comparing average prices in the ``Offer Resolution Only'' and ``Offer Resolution \& Accepts'' cases. Recall that Apple Farmers made the offers in the a=0.33, a=0.2, b=2, b=3, and b=5 experiments; here, we see that the apple price was higher in those experiments in the right ``Offers \& Accepts'' column than in the left ``Offers Only'' column. Similarly, Banana Farmers made the offers in the a=0.5, a=1, a=2, a=3, a=5, b=0.2, b=0.33, b=0.5, and b=1 experiments, and the apple price was lower in those experiments in the right column than in the left. 

These 14 experiments each represent single runs, and the agents learn through a stochastic process. If the experiments were rerun, we would not be surprised if at least one experiment converged to a price one step higher or lower. However, that the price shifted in all 14 of 14 experiments in favour of the agents that learned to make offers, indicates a problem. Accepting offers is apparently easier for agents to learn than making offers, so the agents that learn to accept end up converging faster and are then exploited for having done so. This finding of exploitation of the faster-to-converge sub-population by the slower sub-population is reminiscent of results obtained in a similar---though not embodied---two player negotiation game~\citep{cao2018emergent}. \cite{noukhovitch2021emergent} found that when Accept actions are included, one agent tends to rapidly converge to always accepting all offers. This result is less surprising if you consider that the two player case resembles a temporally extended ultimatum game, \ie a game where it is rational to accept all non-zero offers. Thus a first-mover advantage benefits the sub-population who learn to make offers (ironically this is the slower-to-learn population), and they learn to take advantage of it. Likewise, in both our default ``Offers Only'' experiment and in the results of~\citet{noukhovitch2021emergent}, requiring both parties to make offers in order to exchange goods eliminated this first-mover advantage.

\begin{figure}
    \centering
    \includegraphics[height=3in]{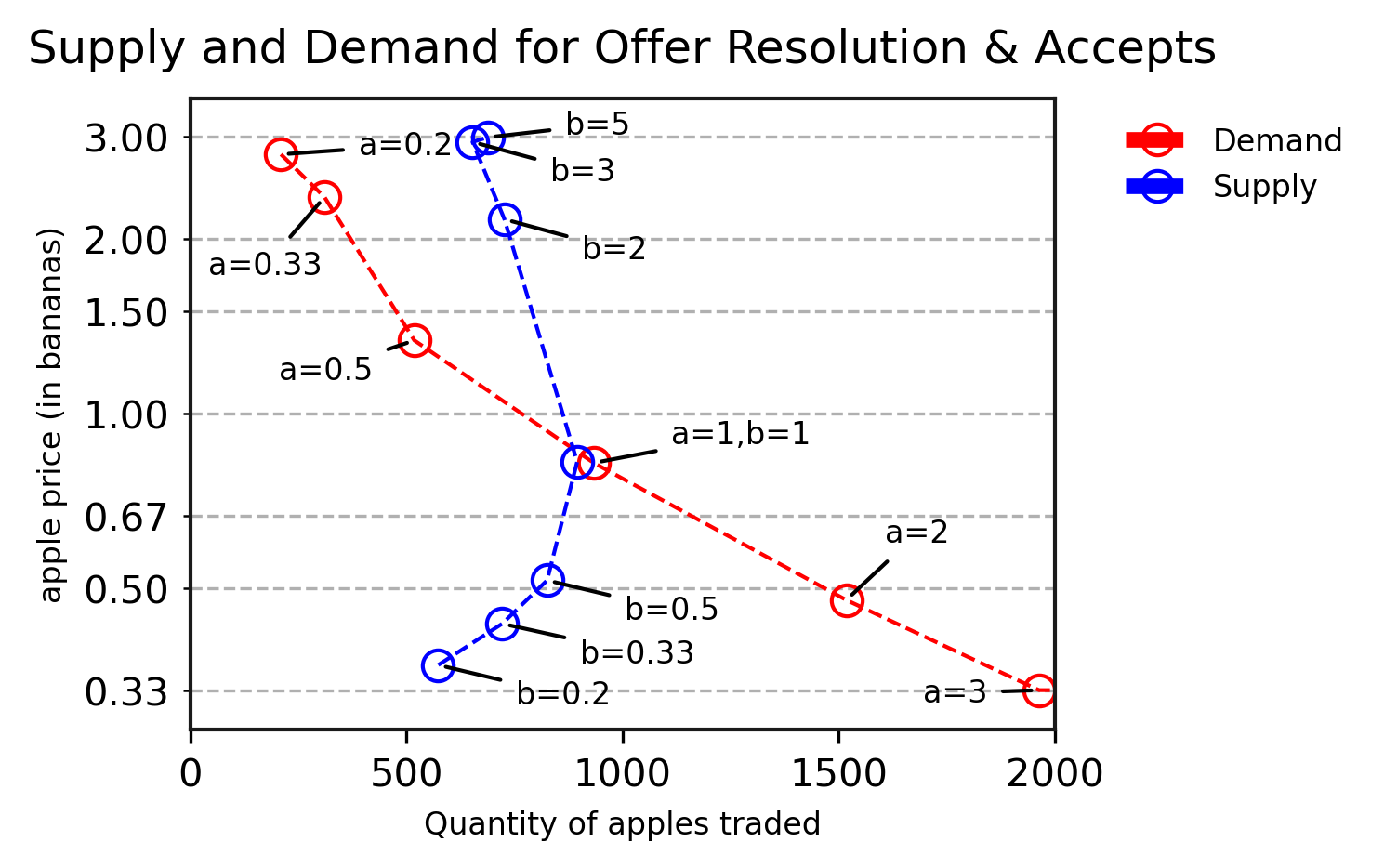}
    \caption{Supply and Demand graph using the ``Offer Resolution \& Accepts'' mechanisms. As in Figure~\ref{fig:sd-spawn-spawn}, the apple and banana tree spawn rates were varied to find the curves. Note that at high banana tree spawn rates we observe a higher price for apples than in the default condition, but also fewer apples being produced and traded.}
\label{fig:ablation_accept:sd}
\end{figure}

We highlight this difficulty in Figure~\ref{fig:ablation_accept:sd}, which presents a Supply and Demand graph using the parameter sweep that we described earlier. Comparing against our earlier Supply and Demand results in Figure~\ref{fig:sd-spawn-spawn}, we see that while both curves reach a wider range of prices and the Demand curve still appears reasonable, the Supply curve now bends back upon itself. Microeconomics predicts that a higher price should incentivise more production, but here we see the opposite, as the b=5, b=3, and b=2 datapoints indicate higher prices but fewer apples being produced and then sold. These three experiments are cases where Apple Farmers make offers and Banana Farmers accept them, suggesting that the Accept actions have unexpected effects beyond the effects on prices that we have been examining.

Overall, the related effects of the agents converging to asymmetric behaviours (some offering and some accepting, instead of all agents offering), the offering agents adjusting their offers to exploit the accepting agents, and the resulting supply and demand behaviour being less aligned with microeconomic predictions, dissuaded us from using the Accept actions in this work. This is somewhat unfortunate, as the simplicity of being able to directly accept another party's offer is attractive, particularly in the ``Accepts Only'' case which mostly removes the environment's role in facilitating exchanges. The environment still plays a (smaller) role in facilitating trade, since after taking an Accept action the environment automatically selects the best priced nearby offer. A stronger form of the Accept actions, perhaps worth exploring in future work, would be to accept only a specific partner's offer: perhaps the closest partner, or a partner standing directly in front of the player, or perhaps with one action per player. These would return even more control and also learning difficulty to the agents, while removing domain knowledge encoded in the environment.

\subsubsection{Dynamic Offer Actions}
\label{sec:ablation:trade:dynamic}

For our final experiment on alternative trade mechanics, we will consider a way to give agents more precise and consistent control over the offers they make. The offer mechanism described in Section~\ref{sec:environment} and enumerated in Table~\ref{tab:env:act_spec} gives the player one action for each possible offer, to set their offer vector to a specific set of values. Overall, to cover all offers up to a maximum quantity of 3, this requires 18 offer actions.

This mechanism has two problems. First, it does not scale efficiently as we increase either the maximum quantity in an offer (\eg moving from 3 to 4), or the types of items players may wish to trade (\eg, adding a Chocolate resource). Each of those changes would lead to an exponential increase in the number of actions required. Second, the agents learn about each action individually: actions that are semantically similar such as ``Give 1 apple for 1 banana'' and ``Give 2 apples for 1 banana'' are represented as discrete actions, with no suggestion to the agent that they are related and that knowledge about one can be transferred to the other. This makes it difficult for agents to explore different prices except through trial and error, as there is no simple action to ``Offer more'' or ``Offer less''.

But what if we used exactly those simple actions? To explore this, we implemented an alternative offer mechanism that we call \textbf{dynamic offer actions}. This mechanism replaces the 18 offer actions described previously with just two per item type: one action to increase the quantity of that item in the offer vector, and another to decrease it. Thus, in our environment, agents would have four dynamic offer actions: ``+ Apple'', ``- Apple'', ``+ Banana'', and ``- Banana'', in addition to the ``Cancel offer'' action that resets the offer vector to $[0, 0]$. For example, starting from the null offer vector of $[0, 0]$, taking the ``- Apple'' action changes the player's offer vector to $[-1, 0]$, and then taking the ``+ Banana'' action results in $[-1, 1]$: the same ``Give 1 apple for 1 banana'' offer that our earlier offer actions specify with one action. The player can then offer an additional apple with another ``- Apple'' action, changing their vector to $[-2, 1]$. From the agent's perspective, actions now have consistent meanings such as ``Offer more'' or ``Demand less'', and may be easier to learn about than having to explore an entirely new action.

Exchanges are handled exactly as before: when two nearby players are making compatible offers, the environment exchanges their goods and resets their offer vectors. Note that since offers require multiple actions to encode, some offer vectors would represent incomplete offers that do not both give and request an item, such as $[-1, 0]$ or $[0, 1]$. The environment does not consider such incomplete offers when pairing offers into exchanges.

The dynamic offer actions mechanic has advantages and disadvantages. One advantage is that it scales efficiently, unlike the default actions. Each additional item type (\eg, adding Chocolate) requires adding only two more actions to increment and decrement it, and no change is required to raise the maximum quantity of each item in an offer. Further, it may be easier for agents to learn how to adjust their offers in response to the community's prices. If an agent wants to try demanding another apple they only have to use the ``+ Apple'' action one extra time; similarly, if an agent's offer isn't competitive, they can use ``- Banana'' to make it more attractive. The disadvantage is that encoding an offer now requires a sequence of actions (\eg ``+ Apple'', ``+ Apple'', ``- Banana'') instead of only one (\eg ``Give 1 Banana for 2 Apples''). This sequence is both more difficult to learn through exploration, and even once learned, also gives trading a higher opportunity cost: every timestep spent adjusting the offer vector is a timestep not spent moving, harvesting, and consuming fruit.

A further challenge is that the order in which agents take these actions makes a difference. For example, the $[-1, 2]$ offer, or ``Give 1 Apple for 2 Bananas'', could be reached through the sequences (``+ Banana'', ``+ Banana'', ``- Apple'') or (``- Apple'', ``+ Banana'', ``+ Banana''). However, the first (starting with the requested item) is better, because the second sequence will briefly encode the lower-than-intended offer of ``Give 1 Apple for 1 Banana'' and might result in a trade. The first sequence is not a complete offer until the final action, and so cannot trade at an unintended price. For more complex offers like $[-2, 3]$, encoding the requested items results in the offer vector moving from a high price down to the target price, and any unintended exchanges would benefit the agent.

\begin{figure}
    \centering
    \begin{subfigure}{0.4\textwidth}
        \centering
        \includegraphics[height=2.5in]{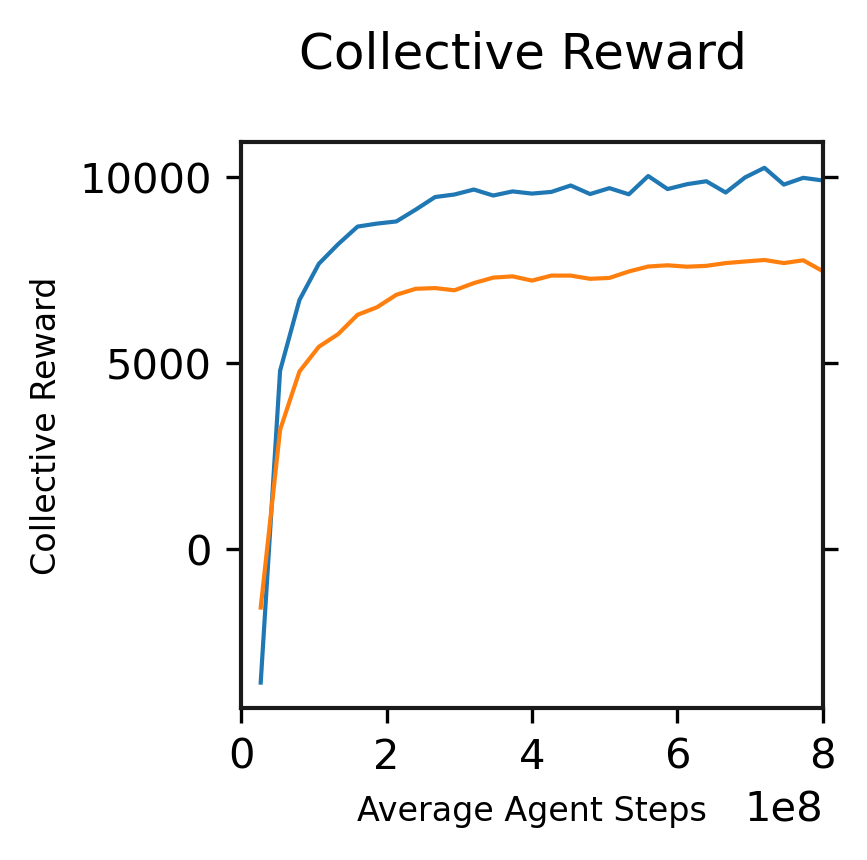}
        \caption{Collective reward}
        \label{fig:ablation_dynamic:comparison:collective_reward}
    \end{subfigure}%
    ~
    \begin{subfigure}{0.6\textwidth}
        \includegraphics[height=2.5in]{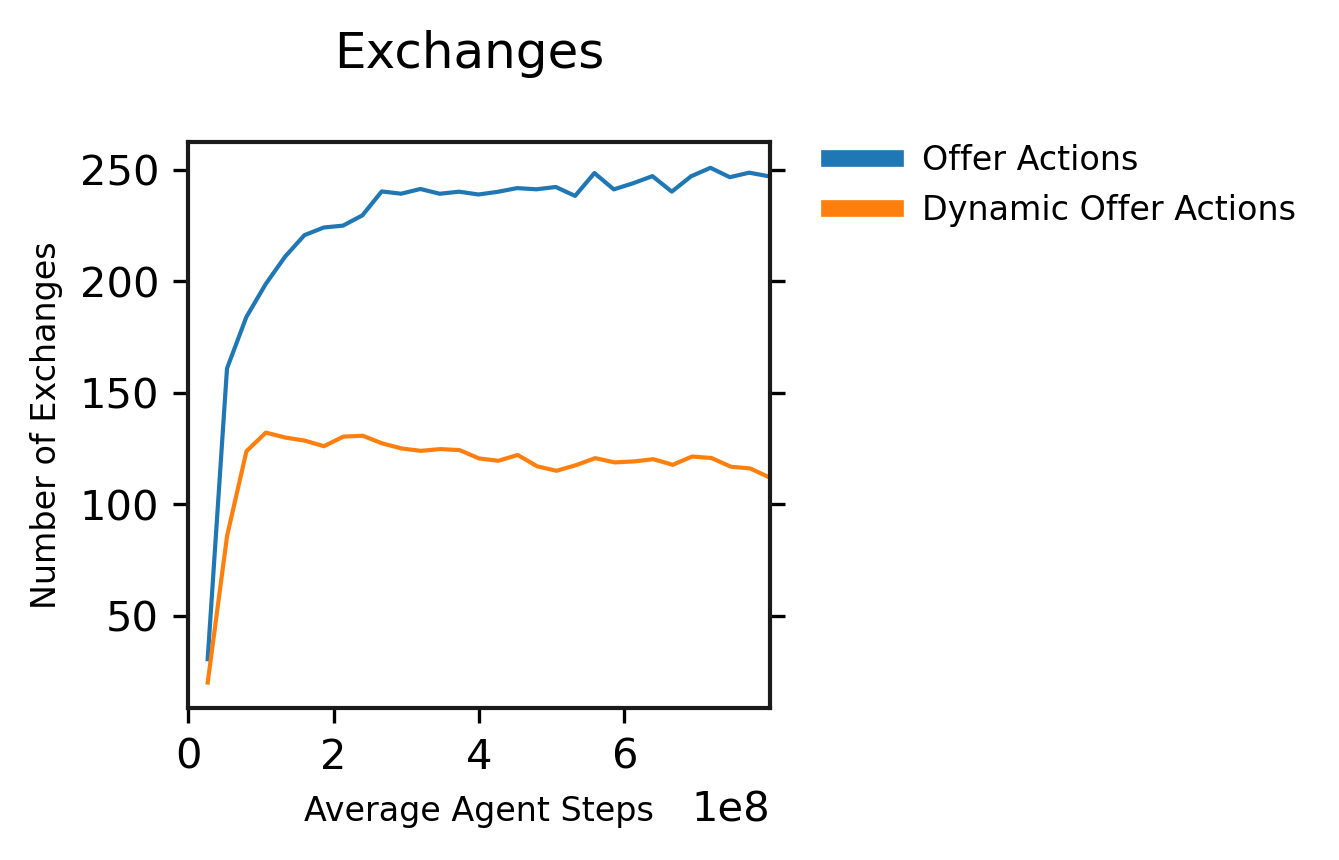}
        \caption{Exchanges per episode}
        \label{fig:ablation_dynamic:comparison:exchanges}
    \end{subfigure}
    \caption{Comparison of Offer actions and Dynamic Offer actions in the a=1,b=1 setting.}
    \label{fig:ablation_dynamic:comparison}
\end{figure}

\begin{figure}
    \centering
    \begin{subfigure}{0.45\textwidth}
        \centering
        \includegraphics[height=4in]{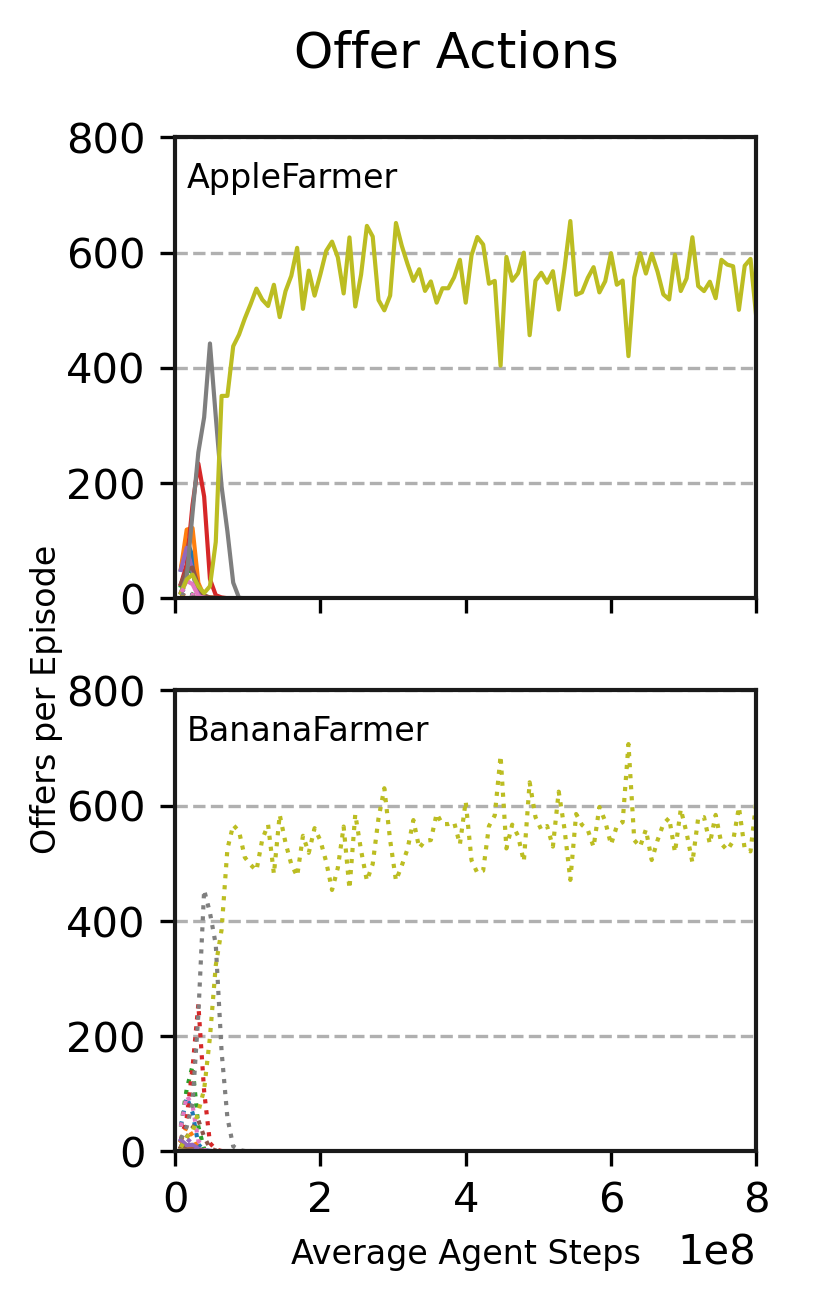}
        \caption{Offer Actions.}
        \label{fig:ablation_dynamic:offers:baseline}
    \end{subfigure}%
    ~
    \begin{subfigure}{0.55\textwidth}
        \centering
        \includegraphics[height=4in]{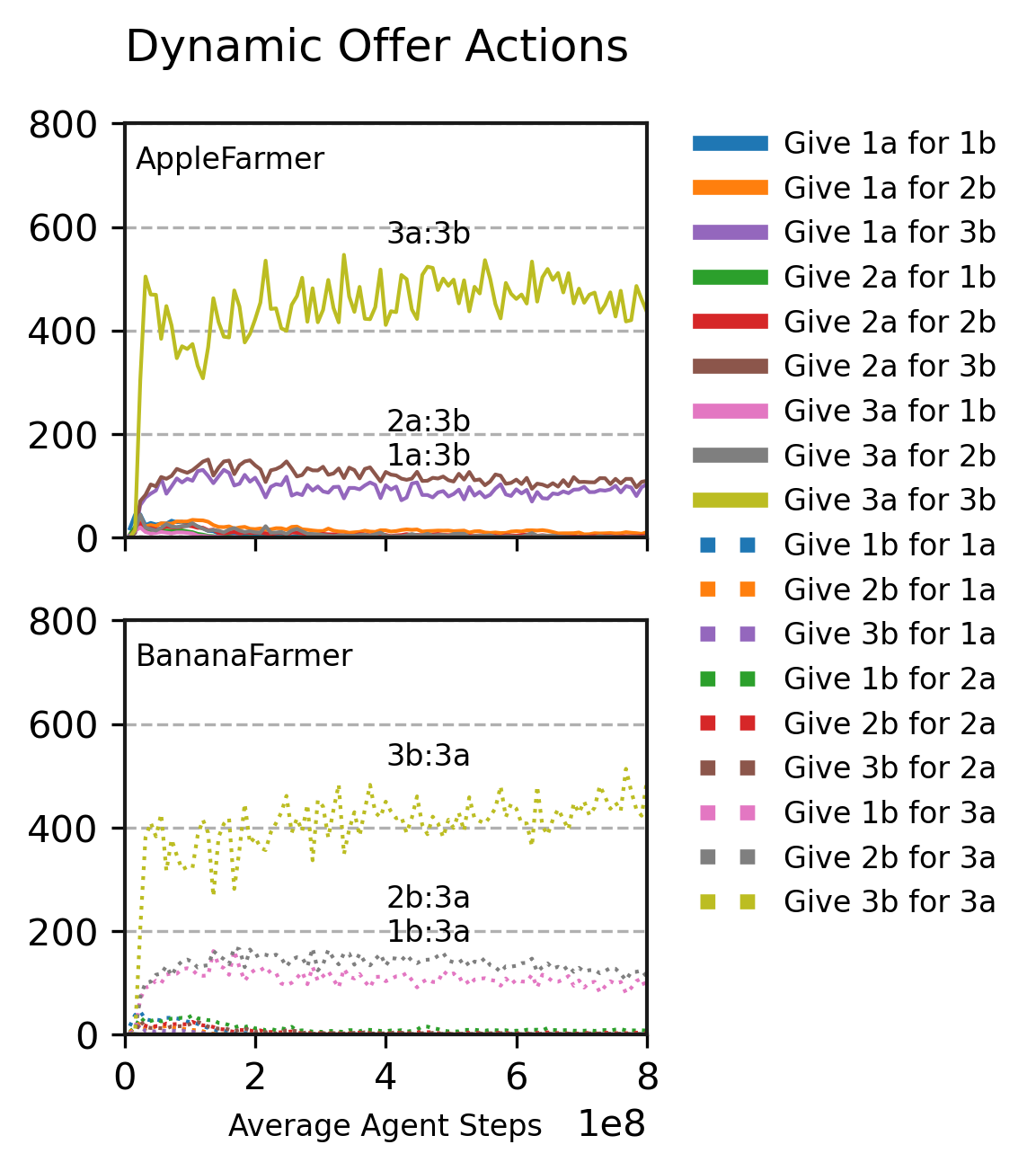}
        \caption{Dynamic Offer Actions.}
        \label{fig:ablation_dynamic:offers:dynamic}
    \end{subfigure}

    \caption{Offer usage per episode by role in the a=1,b=1 setting, using (a) Offer actions and (b) Dynamic Offer actions.}
    \label{fig:ablation_dynamic:offers}
\end{figure}

Figure~\ref{fig:ablation_dynamic:comparison} compares populations using the default and dynamic offer actions, by measuring collective reward and total exchanges per episode. The dynamic offer agents do still learn to trade, although with half as many exchanges per episode and three quarters the collective reward. Figure~\ref{fig:ablation_dynamic:offers} presents the offers made by each role over time. We observe that the dynamic offer agents also learn the correct order of actions to encode their offers, as we described above. For example, in the Apple Farmer plot of Figure~\ref{fig:ablation_dynamic:offers:dynamic} we observe the 1a:3b (or $[-1, 3]$) and 2a:3b (or $[-2, 3]$) offers are active for approximately 100 timesteps per episode, and the 3a:3b (or $[-3, 3]$) offer active for over 400 timesteps per episode. There is no significant use of the 3a:1b or 3a:2b offers. This suggests that the agents encode their requests first and then what is given (\eg, using ``+ Banana'' three times and then ``- Apple'' three times). Banana Farmers also encode offers in the order that is best for them, by requesting apples first and then offering bananas. Together, these results show that the dynamic offer actions are still learnable by agents in the $(a=1,b=1)$ setting, including the additional challenge of ordering the actions. However, as the agents trade less often and earn less reward, the simplicity and flexibility of this mechanic does come at a cost.

\begin{figure}
    \centering
    \includegraphics{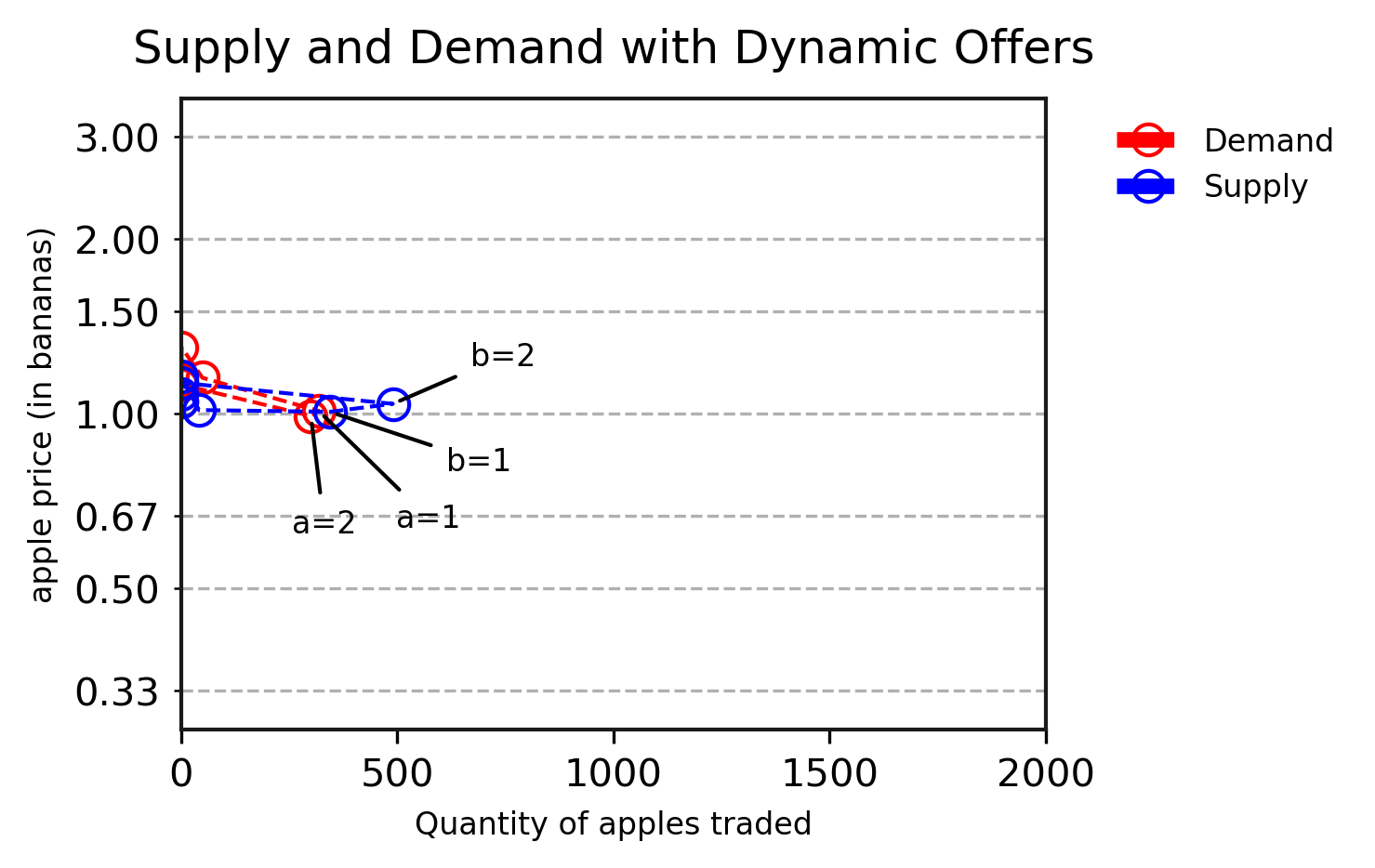}
    \caption{Supply and Demand graph with Dynamic Offer Actions, produced by sweeping apple and banana tree spawn rates.}
    \label{fig:ablation_dynamic:sd}
\end{figure}

However, the results are less promising outside of the $(a=1,b=1)$ case. Figure~\ref{fig:ablation_dynamic:sd} presents a Supply and Demand graph using the dynamic offer actions. As in Figure~\ref{fig:sd-spawn-spawn}, this experiment swept supply and demand by varying the spawn rate of apple and banana trees respectively. Unfortunately, out of the 14 experiments in the sweep, trading behaviour only emerged in the four runs where both items were similarly plentiful: a=1, a=2, b=1, and b=2. In all other experiments, either zero or nearly zero apples were produced and then traded per episode. Further, even when trading emerged, the average price was near 1.0; in Figure~\ref{fig:ablation_dynamic:offers}, we saw the agents use the ``3 Apples for 3 Bananas'' offer to obtain this ratio. Thus, in the a=2 and b=2 runs, changes in relative scarcity did not affect the offers chosen by the players. In comparison, using the default offer actions in Figure~\ref{fig:sd-spawn-spawn:traded}, the a=2 and b=2 runs resulted in prices of 1.5 and 0.67 respectively, thus moving as expected to devalue the more plentiful item. Further, since trade does not emerge in the abundant settings of a=3, a=5, b=3, or b=5 settings, the problem is not simply one of scarcity making the actions more difficult to learn.

Thus, we conclude that our current agents only \textit{sometimes} learn to use the dynamic offer actions, and even then less efficiently and rationally than with the default offer actions. Although the dynamic offer actions are promising for their simplicity and scaling properties, we may require more effective agents, or adjustments to the mechanism to make offers easier to encode, in order to adopt this mechanism. For example, it might help to add an action to reset the vector to its most recent value before the last trade (thus reducing the cost for encoding the same offer repeatedly or adjusting it to a nearby offer), add actions like ``+3 Apples'' in addition to ``+1 Apple'' to let agents more quickly encode an offer, or let agents directly output an offer vector on each timestep instead of encoding it through a series of discrete environmental actions.

\section{Future Work}
\label{sec:future}

In this work we have presented our environment and agents, discussed the situations in which the agents succeed in learning a range of microeconomic behaviours, and examined many of the design choices that enable that learning. Looking ahead, we are excited by many paths to extend this work.

\begin{itemize}
    \item \textbf{More and varied resources}. Throughout our experiments, our agents traded only two similar goods: (A)pples and (B)ananas. One straightforward direction would be to add further goods: perhaps (C)hocolate, (D)urian, and so on; this would provide a further challenge in finding and negotiating offers, and would likely require agents to learn the dynamic offer actions in order to scale. However, we believe a more exciting direction would be to explore different \textit{types} of resources to produce and trade. For example: both durable and non-durable goods that decay over time (as apples and bananas should), processed goods (\eg baked apples or applesauce) that are more rewarding but require other goods as inputs and provide another potential for agent specialization, finite goods that can be traded or picked up but not produced, or tools such as stone hand-axes that require effort to produce (\eg by knapping flint) and make agents carrying one more efficient at producing other goods. Will agents learn to trade apples for a stone hand axe, and at what prices? Will agents learn to specialize in tool production? In particular, we would like to discover whether agents learn to barter arbitrary goods for each other, or if one good emerges as a currency that is predominantly used on one side of exchanges. If agents do arrive at such a convention through their own experience, we could then explore which properties influence a good's adoption as the numeraire good: durability, finite quantity, fungibility, universal appeal, and so on. This work could examine hypotheses motivated by theories of the origin of money~\citep{smit2011money}.
    
    \item \textbf{Further removal of environmental knowledge}. As we have discussed throughout this work, the environment facilitates exchanges between players by detecting pairs of compatible offers and then atomically exchanging goods between parties. While our agents still have control over what offers they make and where they make them, we would prefer for agents to have to learn how to trade without this assistance. In real-life multi-agent robotics tasks or in high-fidelity simulations, for example, no such environmental assistance will be available, and agents will have to learn to explore these interactions on their own. In future work, we are excited to revisit the Drop and Give actions discussed in Section~\ref{sec:ablation:trade:drop_give}, to discover if conditions exist where current agents can learn to use them to discover trade entirely from scratch.
    
    \item \textbf{Non-stationary environments}. From an agent's point of view, its environment is non-stationary because the rest of the population is learning. This may continually produce new niches and make others obsolete. However, throughout our experiments (and particularly in our Supply and Demand experiments) we used the approach of comparative statics, where we trained a new population of agents for each environmental condition. We did not study how a population might move from one equilibrium to another following a gradual or sudden environmental shift (\eg, by scheduling ahead of time a continuous or discontinuous change in the spawn rates of trees, perhaps with a natural interpretation such as different growing seasons). However, the robustness of the population's behaviour during that transition would be interesting to study: we would prefer to have agents that can smoothly adjust to new conditions, as opposed to agents that are overfit to one environmental condition and have to slowly and painfully unlearn and then relearn their behaviour if those conditions change. Ideally, the agents would be able to recognize their environmental conditions and adjust their behaviour within one or a small number of episodes, such that one population could be trained and reused across many experiments.
    
\end{itemize}

\section{Conclusion}
\label{sec:conclusion}

Multi-agent settings are a key element of reinforcement learning research: the real world is multi-agent, and the potential to cooperate or compete with a population of other agents provides an ongoing curriculum for agents to learn from. In this work we have investigated the emergence of microeconomic behaviour---production, consumption, and trading---in populations of agents, from the perspective of multi-agent reinforcement learning. Our contributions have touched on both reinforcement learning and agent-based microeconomics.

From the reinforcement learning side, we see four contributions. First, our original motivation for this work was to investigate a structured and grounded language for communication between agents: bridging the gap between cooperative sequential social dilemma environments without a dedicated communications channel, and emergent communication environments that provide a ``cheap talk'' communication channel for arbitrary use. Trade offers in Fruit Market provide a binding and grounded form of communication. The reward implications of trade provide agents with reasons to use offers for negotiation. Second, our results highlight an obstacle for learning agents that may not be commonly known: once an agent learns one rewarding behaviour, such as eating an apple, it is difficult for them to learn other things that the behaviour precludes, such as selling the apple to get a more rewarding banana. Once they have eaten all of their apples, their offer actions have no effect even when the agent explores them; holding an apple is required to discover the benefits of trade. Even though our environment was designed to make trading behaviour unambiguously rewarding for both parties since half the agents are good at producing apples but prefer bananas while the other half are good at producing bananas but prefer apples, we found in our ``hunger penalty'' ablation experiments that our agents could still fail to discover trade. Agents can't ``have their apple and eat it too''. Third, our experiments in Section~\ref{sec:ablation:trade:drop_give} showed that our current agents do not learn to trade with the Give and Drop actions, presumably because the joint exploration task is too difficult: an agent exploring by giving an item away is unlikely to receive one in return from an agent that does not know how to trade, and so the behaviour is not rewarded. In the real world, however, humans have conventions, norms, and institutions to teach---and enforce!---this behaviour, and we have social and evolutionary motivations---not just economic motivations---for making sure that our relatives and colleagues are happy and healthy. Thus, the frontier of research in this area involves discovering which of these structures and motivations are necessary for our agents to learn the foundational social behaviour that trading behaviour can be built on top of. Fourth, we think that all these results taken together provide strong evidence that economic behaviours such as production, consumption, trade, arbitrage, and so on, are   natural frontiers of multi-agent social interaction that we should endeavor to get our artificial agents to learn about.

From the Microeconomics side, our work fits into the existing literature on agent-based computational economics. We have presented experiments showing that state-of-the-art deep reinforcement learning agents such as V-MPO are capable of learning microeconomic behaviour through their own experience, starting from a random initialization and with no domain-specific code or knowledge added to the agents. This learned behaviour includes agents discovering their preferences for goods, learning what items to produce for later consumption, the emergence of trade between pairs of agents, adjustments in their prices and production and consumption quantities in response to supply and demand shifts, and the emergence of local prices and then arbitrage as agents learn to specialize in resource transport instead of resource production. The agents are also general: with no modifications, they can also succeed in the variety of domains contained in the Melting Pot task suite. The agent-based computational economics community has a vast literature investigating microeconomic behaviour of agents, including reinforcement learning agents. However, aside from the recent AI Economist work~\citep{zheng2020ai}, we are unaware of an application of state-of-the-art deep reinforcement learning agents from the multi-agent reinforcement learning community to this area. In particular, we hope that our demonstration of the flexibility of deep reinforcement learning agents will help to address the problems frequently highlighted in that literature around agents being difficult to write, tune, and reuse across projects. We also hope that the upcoming release of Fruit Market as part of the open-source Melting Pot framework will be of interest both to AI researchers and to the agent-based computational economics community, and we look forward to collaborating with practicing economists in the future.

If we aim to build human-like AGI using MARL, this research program must eventually come to encompass all the critical domains of social intelligence. However, until now this line of work has not incorporated traditional economic phenomena such as trade, bargaining, specialisation, consumption, and production. This paper fills that gap and, we hope, provides a useful platform for further research.

\section*{Acknowledgements}
We would like to thank Gillian Hadfield for a very helpful discussion with us on an early version of this work. We would also like to thank many of our colleagues at DeepMind for the helpful conversations that have guided this work: Angeliki Lazaridou, Yoram Bachrach, Richard Everett, Edgar Du\'{e}\~{n}ez-Guzm\'{a}n, Chris Summerfield, Andrew Butcher, Michael Bowling, Patrick Pilarski, Leslie Acker, Nolan Bard, Josh Davidson, Neil Burch, Anna Koop, Oliver Smith, Thore Graepel, Sasha Vezhnevets, John P. Agapiou, Peter Sunehag, Raphael Koster, Jayd Matyas, Mina Khan, and Yiran Mao.

\bibliographystyle{abbrvnat}
\setlength{\bibsep}{5pt}
\setlength{\bibhang}{0pt}
\bibliography{bibliography}

\clearpage
\appendix
\appendixpage
\addappheadtotoc

\section{Agent Architecture}
\label{app:agent}

Tables~\ref{tab:agent_architecture}, \ref{tab:agent_hyperparameters}, and \ref{tab:training_parameters} list the neural net architecture used for our V-MPO agent and associated hyperparameters for the agent and training procedure.

\begin{table}[H]
    \centering
    \begin{tabular}{|l|l|l|}
        \hline
        \multicolumn{2}{|c|}{Layer} & Parameters or Description\\
        \hline
        \multirow{2}{*}{\makecell[l]{Visual\\ Processing}} & Convolution & channels: (24,), kernel: (1,), strides: (1,) \\ 
        & MLP & size: (256,) \\
        \hline
        \multirow{3}{*}{Torso} & Flatten & Flatten nonvisual observations to a vector \\
        & Concat & Concatenate visual and nonvisual vectors \\
        & LSTM & size: (128,) \\
        \hline
        \multirow{2}{*}{Policy Head} & MLP & size: (64, 64) \\
        & Policy Output & size: (28,) \\
        \hline
        \multirow{3}{*}{Value Head} & MLP & size: (64, 64) \\
        & PopArt Normalization & \makecell[l]{output size: (1,), step size: \num{1e-3}, scale lower bound: \num{1e-2},\\ scale upper bound: \num{1e6}} \\
        \hline
    \end{tabular}
    \caption{V-MPO neural network layers, parameters, and description, for the current 'single pixel per tile' version of Fruit Market; see Figure~\ref{fig:map:minilab}.}
    \label{tab:agent_architecture}
\end{table}

\begin{table}[H]
    \centering
    \begin{tabular}{|l|l|l|}
        \hline
        Hyperparameter & Value \\
        \hline
        Discount Factor & 0.99 \\
        Optimizer & Adam, learning rate \num{1e-4} \\
        Target Update Period & 10 \\
        MPO Epsilon Temperature & \num{1e-1} \\
        \hline
    \end{tabular}
    \caption{V-MPO Agent Hyperparameters.}
    \label{tab:agent_hyperparameters}
\end{table}

\begin{table}[H]
    \centering
    \begin{tabular}{|l|l|}
        \hline
        Parameter & Value \\
        \hline
        Number of agents & 16 \\
        Players per episode & 10 \\
        Number of episodes run in parallel & 800 \\
        Episode Length & 1000 timesteps \\
        \hline
    \end{tabular}
    \caption{Training Parameters.}
    \label{tab:training_parameters}
\end{table}

\end{document}